\documentclass[pdflatex,sn-basic]{sn-jnl} 

\usepackage{graphicx}%
\usepackage{multirow}%
\usepackage{amsmath,amssymb,amsfonts}%
\usepackage{amsthm}%
\usepackage{mathrsfs}%
\usepackage[title]{appendix}%
\usepackage{xcolor}%
\usepackage{textcomp}%
\usepackage{manyfoot}%
\usepackage{booktabs}%
\usepackage{algorithm}%
\usepackage{algorithmicx}%
\usepackage{algpseudocode}%
\usepackage{listings}%
\usepackage{caption}%
\usepackage{longtable}%
\usepackage{enumitem}%
\begin{document}

\title[\textbf{Datasets for Large Language Models: A Comprehensive Survey}]{\textbf{Datasets for Large Language Models: A Comprehensive Survey}}

\author[1,3]{\fnm{Yang} 
\sur{Liu}}

\author[1]{\fnm{Jiahuan} 
\sur{Cao}}

\author[1]{\fnm{Chongyu} 
\sur{Liu}}

\author[2,3]{\fnm{Kai} 
\sur{Ding}}

\author[1,3]{\fnm{Lianwen} 
\sur{Jin}}

\affil[1]{\orgname{South China University of Technology}}

\affil[2]{\orgname{INTSIG Information Co., Ltd}}

\affil[3]{\orgname{INTSIG-SCUT Joint Lab on Document Analysis and Recognition}}


\abstract{This paper embarks on an exploration into the Large Language Model (LLM) datasets, which play a crucial role in the remarkable advancements of LLMs. The datasets serve as the foundational infrastructure analogous to a root system that sustains and nurtures the development of LLMs. Consequently, examination of these datasets emerges as a critical topic in research. In order to address the current lack of a comprehensive overview and thorough analysis of LLM datasets, and to gain insights into their current status and future trends, this survey consolidates and categorizes the fundamental aspects of LLM datasets from five perspectives: (1) Pre-training Corpora; (2) Instruction Fine-tuning Datasets; (3) Preference Datasets; (4) Evaluation Datasets; (5) Traditional Natural Language Processing (NLP) Datasets. The survey sheds light on the prevailing challenges and points out potential avenues for future investigation. Additionally, a comprehensive review of the existing available dataset resources is also provided, including statistics from 444 datasets, covering 8 language categories and spanning 32 domains. Information from 20 dimensions is incorporated into the dataset statistics. The total data size surveyed surpasses 774.5 TB for pre-training corpora and 700M instances for other datasets. We aim to present the entire landscape of LLM text datasets, serving as a comprehensive reference for researchers in this field and contributing to future studies. Related resources are available at: \href{https://github.com/lmmlzn/Awesome-LLMs-Datasets}{https://github.com/lmmlzn/Awesome-LLMs-Datasets}.}


\maketitle

\begin{figure}[h!]
\centering
\includegraphics[width=1\textwidth]{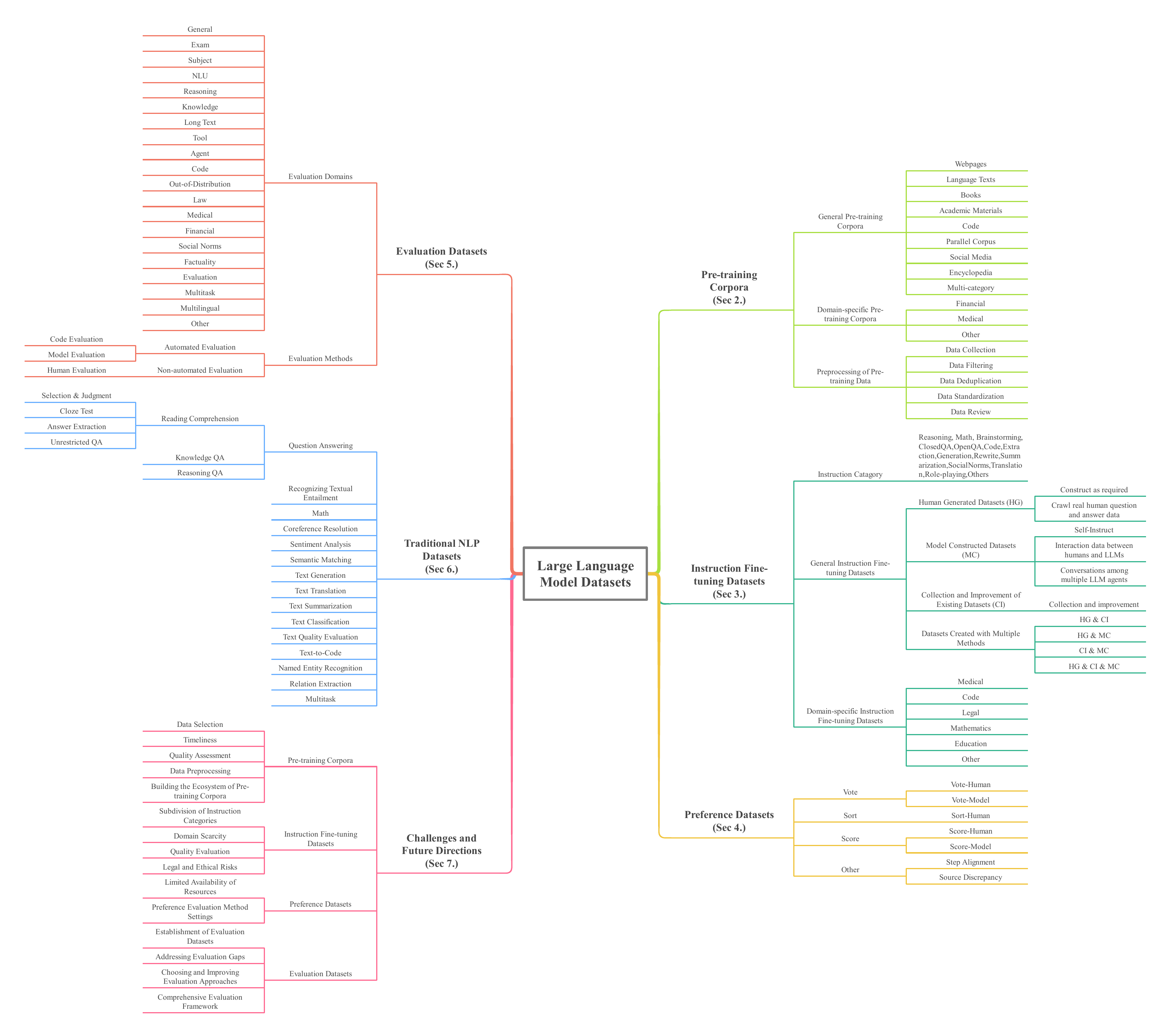}
\caption{The overall architecture of the survey. Zoom in for better view}\label{fig1}
\end{figure}

\section{Introduction}\label{sec1}

\begin{figure}
\centering
\includegraphics[width=0.9\textwidth]{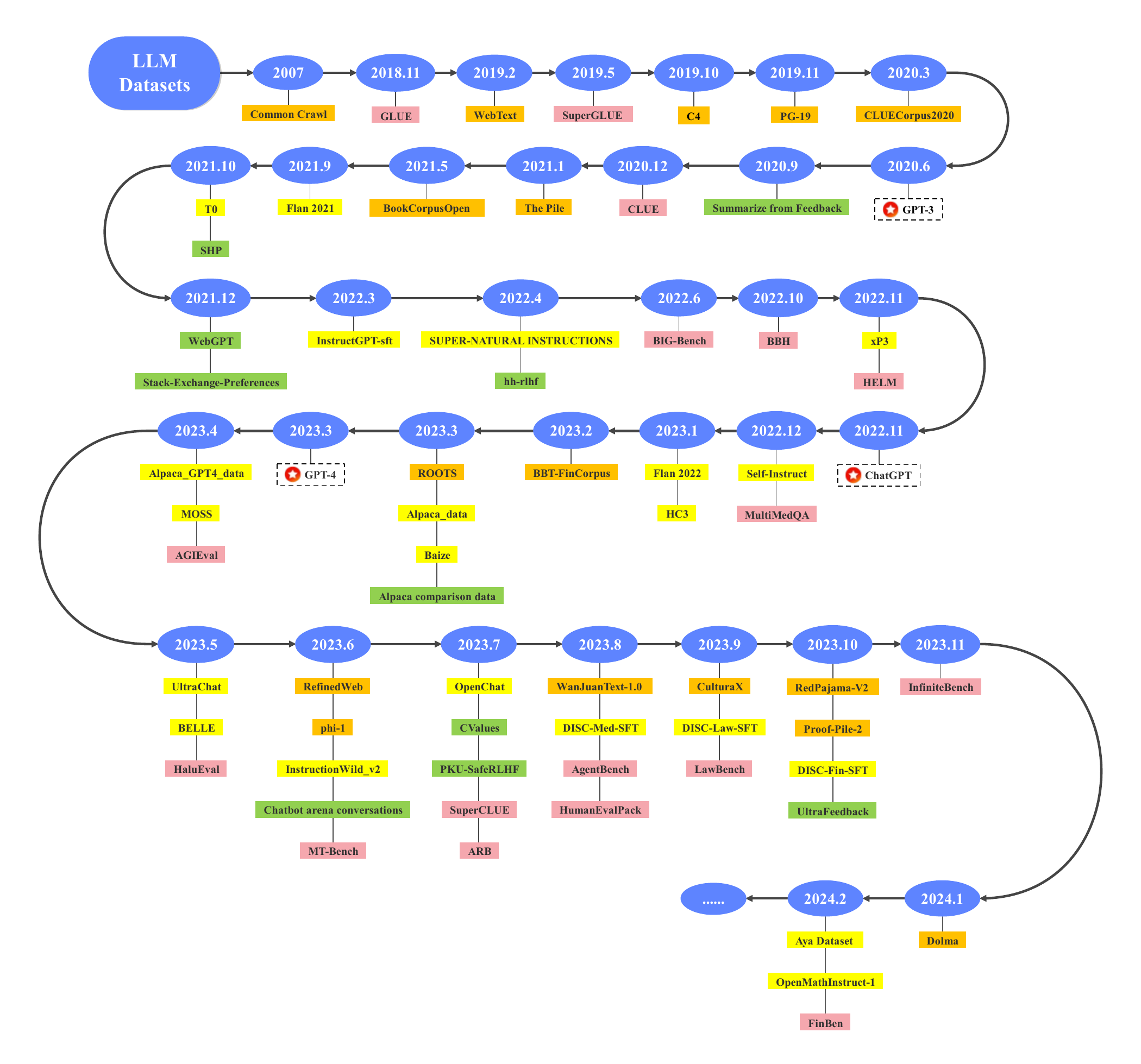}
\caption{A timeline of some representative LLM datasets. Orange represents pre-training corpora, yellow represents instruction fine-tuning datasets, green represents preference datasets, and pink represents evaluation datasets}\label{fig2}
\end{figure}

With the release of ChatGPT \citep{bib1}, in just a few months, Large Language Models (LLMs) have attracted increasing research attention and become a hot research field. Various LLMs have been successively open-sourced, with parameter sizes ranging from several billion to over a hundred billion. Examples include the LLaMA \citep{bib4,bib402}, Phi \citep{bib34,bib403,bib404}, ChatGLM \citep{bib3,bib2}, QWen \citep{bib247}, Baichuan \citep{bib5}, and so on. A considerable amount of work involves fine-tuning on base models, resulting in well-performing general conversational models or domain-specific models. The widespread adoption of Reinforcement Learning from Human Feedback (RLHF) and the refinement of LLM evaluations further optimize the performance of LLMs. The immense potential demonstrated by LLMs can be attributed, in part, to the datasets used for training and testing. As the saying goes, “You can’t make a silk purse out of a sow’s ear.” Without high-quality datasets as the foundation, it is challenging to grow the tree of LLMs with flourishing branches and leaves. Therefore, the construction and analysis of LLM datasets is an area worthy of attention.

The development of text datasets has undergone several stages, from earlier Natural Language Processing (NLP) task datasets to the current era of LLM datasets. In the 1960s to 1980s, the early stages of NLP primarily focused on fundamental tasks such as semantic analysis and machine translation. The dataset scale was relatively small and typically manually annotated. Later, the Message Understanding Conference (MUC) \citep{bib6} began in 1987, focusing on datasets for tasks such as information extraction and Relation Extraction (RE). After 2000, the NLP field continued to emphasize research on traditional tasks and linguistic structures, while also turning attention to emerging areas such as dialogue systems \citep{bib405,bib407,bib48,bib406}. With the rise of deep learning, NLP datasets evolved towards larger scales, greater complexity, more diversity, and increased challenges. Simultaneously, comprehensive performance evaluations \citep{bib242,bib244,bib245}, dialogue datasets \citep{bib107,bib66,bib71}, zero-shot and few-shot datasets \citep{bib171,bib252,bib79}, multilingual datasets \citep{bib249,bib250,bib358}, and others emerged. By the end of 2022, LLMs pushed datasets to a new peak, realizing a shift from a “task-centric construction” to a “construction centered around tasks and stages” in dataset development. LLM datasets are not only categorized based on tasks but also have associations with different stages of LLMs. From the initial pre-training stage to the final evaluation stage, we categorized LLM datasets into four types: pre-training corpora, instruction fine-tuning datasets, preference datasets, and evaluation datasets. The composition and quality of these datasets profoundly influence the performance of LLMs.

The current explosion in LLM datasets poses challenges for research. On the one hand, it often leads to situations where it is difficult to know where to start when trying to understand and learn about the datasets. On the other hand, there is a lack of systematic organization regarding the differences in types, domain orientations, real-world scenarios, etc., among various datasets. In order to reduce the learning curve, promote dataset research and technological innovation, broaden public awareness, we conduct a survey of LLM datasets. The objective is to provide researchers with a comprehensive and insightful perspective, facilitating a better understanding of the distribution and role of LLM datasets, thereby advancing the collective knowledge and application of LLMs.

This paper summarizes existing representative datasets across five dimensions: \textbf{pre-training corpora}, \textbf{instruction fine-tuning datasets}, \textbf{preference datasets}, \textbf{evaluation datasets}, and \textbf{traditional NLP datasets}. Moreover, it presents new insights and ideas, discusses current bottlenecks, and explores future development trends. We also provide a comprehensive review of publicly available dataset related resources. It includes statistics from 444 datasets across 8 language categories spanning 32 different domains, covering information from 20 dimensions. The total data size surveyed exceeds 774.5 TB for pre-training corpora and over 700M instances for other datasets. Due to space constraints, this survey only discusses pure text LLM datasets and does not cover multimodal datasets.

To the best of our knowledge, this is the first survey focused on LLM datasets, presenting the entire landscape. The timeline of LLM datasets is shown in Figure~\ref{fig2}. Prior to this, several LLM-related surveys, such as \cite{bib7} and \cite{bib410}, analyze the latest developments in LLMs but lack detailed descriptions and summaries of datasets. \cite{bib8} summarizes the instruction fine-tuning stage of LLMs. \cite{bib9} and \cite{bib10} summarize the evaluation stage. However, these surveys only concentrate on a part of the LLM datasets, and dataset-related information is not the central focus. In contrast to the aforementioned surveys, our paper places emphasis on LLM datasets, aiming to provide a more detailed and exhaustive survey in this specific domain.

The overall organizational structure is illustrated in Figure~\ref{fig1}. The remainder of this paper is organized as follows. Section~\ref{sec2} summarizes general pre-training corpora categorized by data types and domain-specific pre-training corpora categorized by domains. It also outlines the preprocessing steps and methods for pre-training data. Section~\ref{sec3} summarizes general instruction fine-tuning datasets categorized by construction methods and domain-specific instruction fine-tuning datasets categorized by domains. 15 instruction categories are provided. Section~\ref{sec4} summarizes preference datasets categorized by preference evaluation methods. Section~\ref{sec5} summarizes evaluation datasets categorized by evaluation domains and synthesizes different evaluation methods. Section~\ref{sec6} summarizes traditional NLP datasets categorized by tasks. Section~\ref{sec7} briefly identifies challenges encountered within the datasets and anticipates future research directions. Section~\ref{sec8} concludes this paper. Detailed descriptions of the datasets can be found in Appendices~\ref{secA} through~\ref{secE}.

\section{Pre-training Corpora}\label{sec2}

The pre-training corpora are large collections of text data used during the pre-training process of LLMs. Among all types of datasets, the scale of pre-training corpora is typically the largest one. In the pre-training phase, LLMs learn extensive knowledge from massive amounts of unlabeled text data, which is then stored in its model parameters. It enables LLMs to possess a certain level of language understanding and generation capabilities. The pre-training corpora can encompass various types of text data, such as webpages, academic materials, books, while also accommodating relevant texts from diverse domains, such as legal documents, annual financial reports, medical textbooks, and other domain-specific data. 

Based on the domains involved in the pre-training corpora, they can be divided into two types. The first type is the \textbf{general pre-training corpora}, which comprise large-scale text data mixtures from different domains and topics. The data commonly includes text content from the Internet, such as news, social media, encyclopedias, and more. The objective is to provide universal language knowledge and data resources for NLP tasks. The second type is the \textbf{domain-specific pre-training corpora}, which exclusively contain relevant data for specific domains or topics. The purpose is to furnish LLMs with specialized knowledge.

As the cornerstones of LLMs, the pre-training corpora influence the direction of pre-training and the potential of models in the future. They play several pivotal roles as follows:

\setlist[itemize]{labelindent=15pt, leftmargin=25pt}

\begin{itemize}
    \item \textbf{Providing Generality.} Substantial amounts of text data help models better learn the grammar, semantics, and contextual information of language, enabling them to attain a universal comprehension of natural language.
    
    \item \textbf{Enhancing Generalization Ability.} Data from diverse domains and topics allow models to acquire a broader range of knowledge during training, thereby enhancing their generalization ability.
    
    \item \textbf{Elevating Performance Levels.} Knowledge injection from domain-specific pre-training corpora enables models to achieve superior performance on downstream tasks.
    
    \item \textbf{Supporting Multilingual Processing.} The inclusion of multiple languages in pre-training corpora empowers models to grasp expressions across diverse linguistic contexts, fostering the development of competencies for cross-lingual tasks.
     
\end{itemize}

\subsection{General Pre-training Corpora}\label{subsec21}

\begin{figure}
\centering
\includegraphics[width=0.5\textwidth]{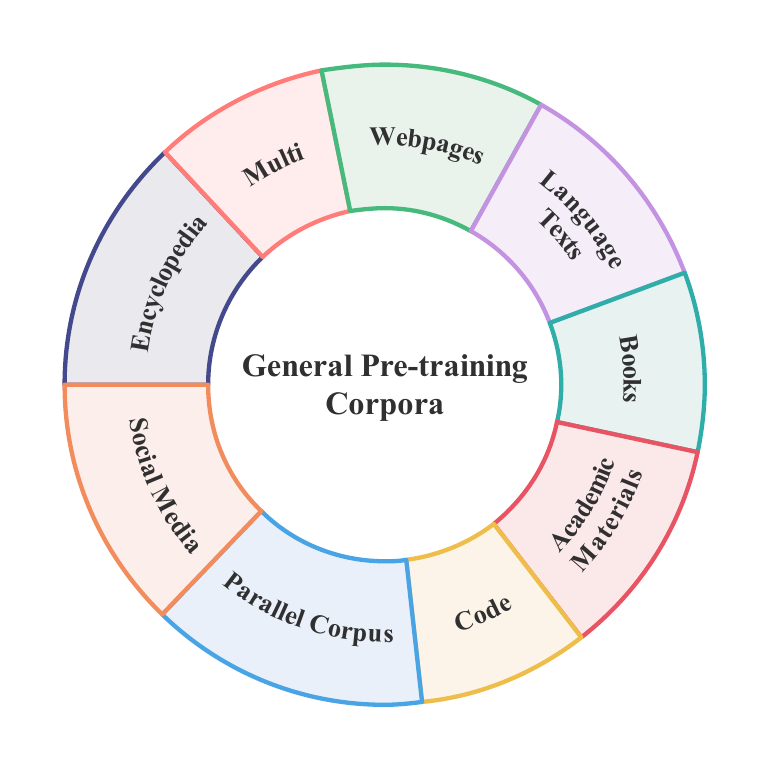}
\caption{Data categories of the general pre-training corpora}\label{fig3}
\end{figure}

The general pre-training corpora are large-scale datasets composed of extensive text from diverse domains and sources. Their primary characteristic is that the text content is not confined to a single domain, making them more suitable for training general foundational models. As illustrated in Figure~\ref{fig3}, the data types can be categorized into eight major classes: \textbf{Webpages}, \textbf{Language Texts}, \textbf{Books}, \textbf{Academic Materials}, \textbf{Code}, \textbf{Parallel Corpus}, \textbf{Social Media}, and \textbf{Encyclopedia}. The collected and organized information about general pre-training corpora is presented in Table~\ref{tab1} and Table~\ref{tab2}.

\begin{table*}
        \captionsetup{singlelinecheck=off, justification=justified}
        \captionof{table}{Summary of \textbf{General Pre-training Corpora} Information \textbf{Part I}. Release Time: “X” indicates unknown month. Public or Not: “All” indicates full open source; “Partial” indicates partially open source; “Not” indicates not open source. “License” indicates the corpus follows a certain protocol. If the corpus is built upon other corpora, the licenses of the source corpora must also be adhered to}\label{tab1}
        \centering
        \resizebox{\textwidth}{!}{
            \begin{tabular}{llllll}
            \hline
            \textbf{Corpus} & \textbf{Publisher} & \textbf{Release Time} & \textbf{Size} & \textbf{Public or Not} & \textbf{License} \\ \hline
            ANC & The US National Science Foundation et al. & 2003-X & - & All & - \\ 
            Anna’s Archive & Anna & 2023-X & 641.2 TB & All & - \\ 
            ArabicText 2022 & BAAI et al. & 2022-12 & 201.9 GB & All & CC-BY-SA-4.0 \\ 
            arXiv & Paul Ginsparg et al. & 1991-X & - & All & Terms of Use for arXiv APIs \\ 
            Baidu baike & Baidu & 2008-4 & - & All & Baidu baike User Agreement \\ 
            BIGQUERY & Salesforce Research & 2022-3 & 341.1 GB & Not & Apache-2.0 \\ 
            BNC & Oxford University Press et al. & 1994-X & 4124 Texts & All & - \\ 
            BookCorpusOpen & Jack Bandy et al. & 2021-5 & 17868 Books & All & Smashwords Terms of Service \\ 
            CC-Stories & Google Brain & 2018-7 & 31 GB & Not & - \\ 
            CC100 & Facebook AI & 2020-7 & 2.5 TB & All & Common Crawl Terms of Use \\ 
            CLUECorpus2020 & CLUE Organization & 2020-3 & 100 GB & All & MIT \\ 
            Common Crawl & Common Crawl & 2007-X & - & All & Common Crawl Terms of Use \\ 
            CulturaX & University of Oregon et al. & 2023-9 & 27 TB & All &  mC4 \& OSCAR \\ 
            C4 & Google Research & 2019-10 & 12.68 TB & All & ODC-BY \& Common Crawl Terms of Use \\ 
            Dolma & AI2 et al. & 2024-1 & 11519 GB & All & MR Agreement \\
            Github & Microsoft & 2008-4 & - & All & - \\ 
            mC4 & Google Research & 2021-6 & 251 GB & All &  ODC-BY \& Common Crawl Terms of Use \\ 
            MNBVC & Liwu Community & 2023-1 & 20811 GB & All & MIT \\ 
            MTP & BAAI & 2023-9 & 1.3 TB & All & BAAI Data Usage Protocol \\ 
            MultiUN & German Research Center for Artificial Intelligence (DFKI) GmbH & 2010-5 & 4353 MB & All & - \\ 
            News-crawl & UKRI et al. & 2019-1 & 110 GB & All & CC0 \\ 
            OpenWebText & Brown University & 2019-4 & 38 GB & All & CC0 \\ 
            OSCAR 22.01 & Inria & 2022-1 & 8.41 TB & All & CC0 \\ 
            ParaCrawl & Prompsit et al. & 2020-7 & 59996 Files & All & CC0 \\ 
            PG-19 & DeepMind & 2019-11 & 11.74 GB & All & Apache-2.0 \\ 
            phi-1 & Microsoft Research & 2023-6 & 7 B Tokens & Not & CC-BY-NC-SA-3.0 \\ 
            Project Gutenberg & Ibiblio et al. & 1971-X & - & All & The Project Gutenberg \\ 
            Pushshift Reddit & Pushshift.io et al. & 2020-1 & 2 TB & All & - \\ 
            RealNews & University of Washington et al. & 2019-5 & 120 GB & All & Apache-2.0 \\ 
            Reddit & Condé Nast Digital et al. & 2005-6 & - & All & - \\ 
            RedPajama-V1 & Together Computer & 2023-4 & 1.2 T Tokens & All & - \\ 
            RedPajama-V2 & Together Computer & 2023-10 & 30.4 T Tokens & All & Common Crawl Terms of Use \\ 
            RefinedWeb & The Falcon LLM team & 2023-6 & 5000 GB & Partial & ODC-BY-1.0 \\ 
            ROOTS & Hugging Face et al. & 2023-3 & 1.61 TB & Partial & BLOOM Open-RAIL-M \\ 
            Smashwords & Draft2Digital et al. & 2008-X & - & All &  Smashwords Terms of Service \\ 
            StackExchange & Stack Exchange & 2008-9 & - & All & CC-BY-SA-4.0 \\ 
            S2ORC & AI2 et al. & 2020-6 & 81.1 MB & All & ODC-BY-1.0 \\ 
            The Pile & EleutherAI & 2021-1 & 825.18 GB & All & MIT \\ 
            The Stack & ServiceNow Research et al. & 2022-11 & 6 TB & All & The Terms of the Original Licenses \\ 
            TigerBot\_pretrain\_en & TigerBot & 2023-5 & 51 GB & Partial & Apache-2.0 \\ 
            TigerBot\_pretrain\_zh & TigerBot & 2023-5 & 55 GB & Partial & Apache-2.0 \\ 
            TigerBot-wiki & TigerBot & 2023-5 & 205 MB & All & Apache-2.0 \\ 
            Toronto Book Corpus & University of Toronto et al. & 2015-6 & 11038 Books & Not & MIT \& Smashwords Terms of Service \\ 
            UNCorpus v1.0 & United Nations et al. & 2016-5 & 799276 Files & All & - \\ 
            WanJuanText-1.0 & Shanghai AI Laboratory & 2023-8 & 1094 GB & All & CC-BY-4.0 \\ 
            WebText & OpenAI & 2019-2 & 40 GB & Partial & MIT \\ 
            Wikipedia & Wikimedia Foundation & 2001-1 & - & All & CC-BY-SA-3.0 \& GFDL \\ 
            WuDaoCorpora-Text & BAAI et al. & 2021-6 & 200 GB & Partial & CC-BY-NC-ND-4.0 \\ 
            Zhihu & Beijing Zhizhe Tianxia Technology Co., Ltd & 2011-1 & - & All & Zhihu User Agreement \\ \hline
            \end{tabular}
        }
\end{table*}

\begin{table*}[h!]
        \captionsetup{singlelinecheck=off, justification=justified}
        \captionof{table}{Summary of \textbf{General Pre-training Corpora} Information \textbf{Part II}. Language: “EN” indicates English, “ZH” indicates Chinese, “AR” indicates Arabic, “PL” indicates Programming Language, “Multi” indicates Multilingual, and the number in parentheses indicates the number of languages included. “CM” indicates Construction Methods, where “HG” indicates Human Generated Corpora, “MC” indicates Model Constructed Corpora, and “CI” indicates Collection and Improvement of Existing Corpora}\label{tab2}
        \centering
        \resizebox{\textwidth}{!}{
            \begin{tabular}{llllll}
            \hline
            \textbf{Corpus} & \textbf{Language} & \textbf{CM} & \textbf{Category} & \textbf{Source} \\ \hline
            ANC & EN & HG & Language Texts & American English texts \\ 
            Anna’s Archive & Multi & HG & Books & Sci-Hub, Library Genesis, Z-Library, etc. \\ 
            ArabicText 2022 & AR & HG \& CI & Multi & ArabicWeb, OSCAR, CC100, etc. \\ 
            arXiv & EN & HG & Academic Materials & arXiv preprint \\ 
            Baidu baike & ZH & HG & Encyclopedia & Encyclopedic content data \\ 
            BIGQUERY & PL & CI & Code & BigQuery \\ 
            BNC & EN & HG & Language Texts & British English texts \\ 
            BookCorpusOpen & EN & CI & Books & Toronto Book Corpus \\ 
            CC-Stories & EN & CI & Webpages & Common Crawl \\ 
            CC100 & Multi (100) & CI & Webpages & Common Crawl \\ 
            CLUECorpus2020 & ZH & CI & Webpages & Common Crawl \\ 
            Common Crawl & Multi & HG & Webpages & Web crawler data \\ 
            CulturaX & Multi (167) & CI & Webpages & mC4, OSCAR \\ 
            C4 & EN & CI & Webpages & Common Crawl \\ 
            Dolma & EN & HG \& CI & Multi & Project Gutenberg, C4, Reddit, etc. \\
            Github & PL & HG & Code & Various code projects \\ 
            mC4 & Multi (108) & CI & Webpages & Common Crawl \\ 
            MNBVC & ZH & HG \& CI & Multi & Chinese books, webpages, theses, etc. \\ 
            MTP & EN \& ZH & HG \& CI & Parallel Corpus & Chinese-English parallel text pairs on the web \\ 
            MultiUN & Multi (7) & HG & Parallel Corpus & United Nations documents \\ 
            News-crawl & Multi (59) & HG & Language Texts & Newspapers \\ 
            OpenWebText & EN & HG & Social Media & Reddit \\ 
            OSCAR 22.01 & Multi (151) & CI & Webpages & Common Crawl \\ 
            ParaCrawl & Multi (42) & HG & Parallel Corpus & Web crawler data \\ 
            PG-19 & EN & HG & Books & Project Gutenberg \\ 
            phi-1 & EN \& PL & HG \& MC & Code & The Stack, StackOverflow, GPT-3.5 Generation \\ 
            Project Gutenberg & Multi & HG & Books & Ebook data \\ 
            Pushshift Reddit & EN & CI & Social Media & Reddit \\ 
            RealNews & EN & CI & Webpages & Common Crawl \\ 
            Reddit & EN & HG & Social Media & Social media posts \\ 
            RedPajama-V1 & Multi & HG \& CI & Multi & Common Crawl, Github, books, etc. \\ 
            ReaPajama-V2 & Multi (5) & CI & Webpages & Common Crawl, C4, etc. \\ 
            RefinedWeb & EN & CI & Webpages & Common Crawl \\ 
            ROOTS & Multi (59) & HG \& CI & Multi & OSCAR, Github, etc. \\ 
            Smashwords & Multi & HG & Books & Ebook data \\ 
            StackExchange & EN & HG & Social Media & Community question and answer data \\ 
            S2ORC & EN & CI & Academic Materials & MAG, arXiv, PubMed, etc. \\ 
            The Pile & EN & HG \& CI & Multi & Books, arXiv, Github, etc. \\ 
            The Stack & PL (358) & HG & Code & Permissively-licensed source code files \\ 
            TigerBot\_pretrain\_en & EN & CI & Multi & English books, webpages, en-wiki, etc \\ 
            TigerBot\_pretrain\_zh & ZH & HG & Multi & Chinese books, webpages, zh-wiki, etc. \\ 
            TigerBot-wiki & ZH & HG & Encyclopedia & Baidu baike \\ 
            Toronto Book Corpus & EN & HG & Books & Smashwords \\ 
            UNCorpus v1.0 & Multi (6) & HG & Parallel Corpus & United Nations documents \\ 
            WanJuanText-1.0 & ZH & HG & Multi & Webpages, Encyclopedia, Books, etc \\ 
            WebText & EN & HG & Social Media & Reddit \\ 
            Wikipedia & Multi & HG & Encyclopedia & Encyclopedic content data \\ 
            WuDaoCorpora-Text & ZH & HG & Webpages & Chinese webpages \\ 
            Zhihu & ZH & HG & Social Media & Social media posts \\ \hline
            \end{tabular}
        }
\end{table*}

\subsubsection{Webpages}\label{subsubsec211}
Webpages represent the most prevalent and widespread type of data in pre-training corpora, comprised of text content obtained by crawling a large number of webpages on the Internet. This type of data has several key characteristics.

\begin{itemize}

    \item \textbf{Massive Scale.} There is a vast number of websites, and new webpages emerge continuously.
    
    \item \textbf{Dynamism.} Content undergoes continuous updates and changes over time.
    
    \item \textbf{Multilingualism.} It may include content in multiple languages.
    
    \item \textbf{Rich in Themes.} It encompasses content from different domains and subjects.

    \item \textbf{Semi-structured.} The data is typically in hypertext markup language (HTML) format, exhibiting certain structural characteristics. However, it may include various modalities such as text, images, videos, and more.

    \item \textbf{Requires Cleaning.} It often contains a significant amount of noise, irrelevant information, and sensitive content, making it unsuitable for direct use.

\end{itemize}

\noindent The construction of webpages corpora is commonly pursued through two primary approaches. The first method involves \textbf{building upon Common Crawl}\footnote{\href{https://commoncrawl.org/}{https://commoncrawl.org/}}. Common Crawl is a massive, unstructured, multilingual web corpus that provides public access to web archives by regularly crawling and storing webpage data from the Internet. However, the data in Common Crawl are not clean, containing a lot of irrelevant information, such as advertisements, navigation bars, etc. Additionally, there is a presence of pornographic content, violence, machine-generated spam, and sensitive information involving personal privacy. Consequently, many subsequent pre-training corpora are derived by reselecting and cleaning data from Common Crawl. For instance, RefinedWeb \citep{bib11}, used for pre-training Falcon model\footnote{\href{https://falconllm.tii.ae/}{https://falconllm.tii.ae/}}, undergoes rigorous filtering and deduplication processes on Common Crawl. It ultimately retains high-quality English text totaling 5T tokens. C4 \citep{bib12}, derived from Common Crawl crawler data from April 2019, undergoes processing with multiple filters, removing useless, harmful, and non-English text. In contrast to C4, mC4 \citep{bib13} , CC100 \citep{bib14}, OSCAR 22.01 \citep{bib15}, and RedPajama-V2 \citep{bib16} retain multilingual data during the cleaning process, utilizing different cleaning pipelines. CC-Stories \citep{bib17} and RealNews \citep{bib18} are selected subsets of text content from Common Crawl based on specific themes. CC-Stories filters out text with a story-like style following the Winograd Schema \citep{bib19} for common-sense reasoning and language modeling. RealNews \citep{bib18} extracts a substantial amount of webpages dedicated to news to obtain news data. The above corpora either exclusively contain English or belong to multilingual mixes. CLUECorpus2020 \citep{bib20} conducts data cleaning on the Chinese portion of Common Crawl, resulting in a high-quality Chinese pre-training corpus of 100GB. However, there still exists a small amount of noise in these corpora. Therefore, some corpora continue with subsequent cleaning efforts. For instance, CulturaX \citep{bib21} performs a multi-stage cleaning process after combining mC4 and OSCAR corpora, resulting in higher-quality multilingual corpus.

The second method involves \textbf{independently crawling various raw webpages and then employing a series of cleaning processes to obtain the final corpus}. WuDaoCorpora-Text \citep{bib22} is cleaned using over 20 rules from 100TB of raw webpages, covering many domains such as education and technology. Furthermore, webpage data in some multi-category corpora is also constructed using this method, including MNBVC \citep{bib23}, WanJuanText-1.0 \citep{bib24}, TigerBot\_pretrain\_zh\_corpus \citep{bib25}, and others.

\subsubsection{Languages Texts}\label{subsubsec212}

The language text data mainly consists of two parts. The first part is \textbf{electronic text data constructed based on widely sourced written and spoken language}, typically in the form of large corpora for a specific language. The full name of ANC\footnote{\href{https://anc.org/}{https://anc.org/}} is the American National Corpus. The content primarily includes various written and spoken materials in American English. The second edition of the corpus has a scale of 22M words, making it highly suitable for models to learn language. Similarly, BNC\footnote{\href{http://www.natcorp.ox.ac.uk/}{http://www.natcorp.ox.ac.uk/}}, short for the British National Corpus, encompasses 100M words of electronic text resources, covering spoken and written materials in British English.

The second part is \textbf{electronic text data constructed based on relevant written materials in various fields or topics}. For example, FinGLM \citep{bib26} covers annual reports of some listed companies between 2019 and 2021. The data type belongs to language text materials in the financial domain. TigerBot-law \citep{bib25} includes legal regulations from 11 categories such as the Chinese Constitution and the Chinese Criminal Law, falling within the language text materials in the legal domain. News-crawl\footnote{\href{https://data.statmt.org/news-crawl/}{https://data.statmt.org/news-crawl/}} extracts monolingual texts from online newspapers and other news sources, encompassing news text in 59 languages.

\subsubsection{Books}\label{subsubsec213}

Book data is also one of the common types of data in pre-training corpora. Compared to webpages, books have longer textual content and superior data quality, both of which contribute to enhancing the performance of LLMs. This helps improve their ability to capture human language features while learning more profound language knowledge and contextual information. The book data primarily possesses the following characteristics.

\begin{itemize}

    \item \textbf{Breadth.} It typically covers a wide range of subjects and topics, including novels, biographies, textbooks, and more.
    
    \item \textbf{High Quality.} Books are usually authored by professionals, undergo editing and proofreading, resulting in more accurate grammar and spelling with less noise.
    
    \item \textbf{Lengthy Text.} Longer texts and complex sentence structures provide additional contextual information.
    
    \item \textbf{Language and Culture.} Books often contain rich language features such as professional terminology, colloquialisms, and idioms, reflecting diverse cultural backgrounds.

\end{itemize}

\noindent Book data can be found on e-book websites, with commonly used resources being Smashwords\footnote{\href{https://www.smashwords.com/}{https://www.smashwords.com/}} and Project Gutenberg\footnote{\href{https://www.gutenberg.org/}{https://www.gutenberg.org/}}. Smashwords is a large repository of free e-books, containing over 500K electronic books. Project Gutenberg, as the earliest digital library, is dedicated to digitizing and archiving cultural works, and it also boasts a wealth of book resources.

Subsequently, many book corpora are constructed by scraping and cleaning e-book resources. In 2015, Toronto Book Corpus \citep{bib27} crawled 11,038 e-books from Smashwords, forming a large-scale corpus of books. This corpus was once publicly available but is no longer accessible. In 2019, PG-19 \citep{bib28} collected books published before 1919 from Project Gutenberg and removed short-text books, resulting in a final count of 28,752 books. In 2021, BookCorpusOpen \citep{bib29} built upon Toronto Book Corpus, Smashwords, and others, creating 17,868 book entries. In 2023, Anna’s Archive\footnote{\href{https://annas-archive.org/datasets}{https://annas-archive.org/datasets}} became the world’s largest open-source and open-data library. The creator scraped books from libraries such as Libgen, Sci-Hub, and made them publicly available. As of February 2024, its size has reached 641.2TB and it is continuously growing.

\begin{figure}
\centering
\includegraphics[width=0.9\textwidth]{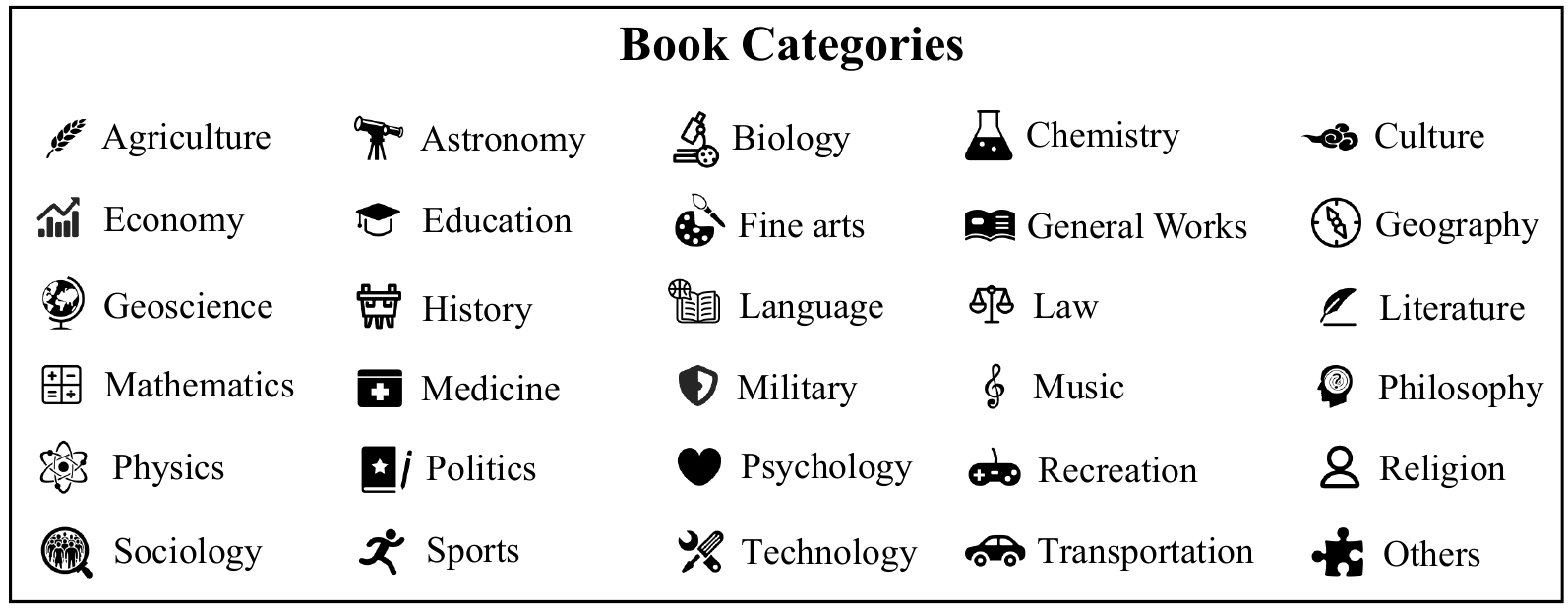}
\caption{Classification of books. Categorizing books into 30 fine-grained classes based on different domains}\label{fig4}
\end{figure}

It is worth mentioning that the fields covered by books are extremely diverse. Thus, fine-grained categorization of books by domain is feasible. It not only facilitates more convenient gap analysis and supplementation but also enables the easy selection of relevant data when focusing on specific domains. Referring to the Chinese Library Classification System\footnote{\href{http://www.ztflh.com/}{http://www.ztflh.com/}}, books can be straightforwardly categorized into 30 classes, as illustrated in Figure~\ref{fig4} for reference.

\subsubsection{Academic Materials}\label{subsubsec214}

Academic material data refers to text data related to the academic field, including but not limited to academic papers, journal articles, conference papers, research reports, patents, and more. These data are authored and published by experts and scholars in the academic community, possessing a high level of professionalism and academic rigor. The academic materials themselves exhibit exceptional quality. Incorporating them into pre-training corpora can provide more accurate and professional information, helping the model understand the terminology and knowledge within the academic domain.

The most commonly used corpus currently is arXiv\footnote{\href{https://arxiv.org/}{https://arxiv.org/}}, which gathers preprints of papers in physics, mathematics, computer science, biology, and quantitative economics. It not only furnishes high-quality academic knowledge but also enables models to grasp the LATEX format of papers. In addition to arXiv, S2ORC \citep{bib53} encompasses English academic papers from various disciplines. It features extensive metadata, abstracts, reference lists, and structured full-text content. In the medical field, PubMed Central\footnote{\href{https://www.ncbi.nlm.nih.gov/pmc/}{https://www.ncbi.nlm.nih.gov/pmc/}} has played a role in the open access of nearly 5M biomedical publications.

Pre-training corpora exclusively consisting of academic material data are rare, as most multi-category corpora choose to include academic materials. In The Pile \citep{bib30}, academic material data accounts for 38.1\%, surpassing the 18.1\% proportion of Webpage data. In RedPajama-V1\footnote{\href{https://huggingface.co/datasets/togethercomputer/RedPajama-Data-1T}{https://huggingface.co/datasets/togethercomputer/RedPajama-Data-1T}}, the proportion of academic materials is 2.31\%, totaling 28 billion tokens.

\subsubsection{Code}\label{subsubsec215}

The category of code data refers to textual information containing programming languages, such as Python, Java, C++, and other code snippets. Its purpose is to assist models in better understanding programming languages and code structures, enabling them to perform well in downstream tasks like code comprehension, code recommendation, and code generation. Nowadays, LLMs are often leveraged to generate code, facilitating various tasks. The quality of the code data used during model training directly impacts the effectiveness of the generated code, underscoring the significance of code data in model performance.

The main corpora for code data include The Stack \citep{bib31}, BIGQUERY \citep{bib32}, and Github\footnote{\href{https://github.com/}{https://github.com/}}. The Stack comprises diverse collection of 385 programming languages and hosts over 6TB of source code files with open-source licenses. It is specifically tailored for the development of expansive LLMs in the programming domain. BIGQUERY, a subset of the publicly released Google BigQuery corpus\footnote{\href{https://cloud.google.com/bigquery?hl=en}{https://cloud.google.com/bigquery?hl=en}}, focuses on six selected programming languages. Github serves as a hosting platform for both open-source and private software projects, supplying a rich array of varied code information. Notably, training data for significant code models like StarCoder \citep{bib33} is sourced from this repository. However, it is crucial to exercise caution during web scraping to adhere to the code usage protocols set by project authors. StackOverflow\footnote{\href{https://stackoverflow.com/}{https://stackovjerflow.com/}} is also a common source of code data. As a Question-and-Answer (Q\&A) community dedicated to programming and development, it features questions and answers spanning topics such as programming languages, development tools, and algorithms. StackOverflow is part of StackExchange\footnote{\href{https://stackexchange.com/}{https://stackexchange.com/}}, which houses different Q\&A sections. Therefore, it is categorized as social media data, as explained in Section~\ref{subsubsec217}. More recently, phi-1 \citep{bib34} is created specifically for training code models. It not only includes a subset of code selected from The Stack and StackOverflow but also utilizes GPT-3.5 \citep{bib35} to generate textbooks and exercise questions related to Python.

\subsubsection{Parallel Corpus}\label{subsubsec216}

Parallel corpus data refers to a collection of text or sentence pairs from different languages. These pairs of texts are translations of each other, where one text is in the source language (e.g., English), and the corresponding text is in the target language (e.g., Chinese). The incorporation of parallel corpus data is crucial for enhancing the machine translation capability and cross-lingual task performance of LLMs.

The collection of parallel corpora typically occurs through two main avenues. The first involves \textbf{extracting text from Internet resources such as webpages}. ParaCrawl \citep{bib36}, for instance, utilizes open-source software to crawl webpages, constructing a publicly available parallel corpus. It encompasses 223M filtered sentence pairs. Similarly, MTP\footnote{\href{https://data.baai.ac.cn/details/BAAI-MTP}{https://data.baai.ac.cn/details/BAAI-MTP}} collects and organizes existing Chinese-English web text data, amassing a total of 300M text pairs. This stands as the currently largest open-source Chinese-English aligned text pair dataset.

The second approach involves \textbf{the collection of parallel corpora from United Nations multilingual documents}. MultiUN \citep{bib37} gathers parallel text pairs through the United Nations Official Document System\footnote{\url{https://documents.un.org/}}. These documents cover the six official languages of the United Nations (Arabic, Chinese, English, French, Russian, and Spanish), as well as a limited amount of German. UNCorpus v1.0 \citep{bib38} consists of public domain United Nations official records and other conference documents, aligned at the sentence level.

\subsubsection{Social Media}\label{subsubsec217}

Social media data refers to textual content collected from various media platforms, primarily encompassing user-generated posts, comments, and dialogue data between users. The data reflects real-time dynamics and interactivity among individuals on social media. Despite the potential presence of harmful information such as biases, discrimination, and violence in social media data, it remains essential for the pre-training of LLMs. This is because social media data is advantageous for models to learn expressive capabilities in conversational communication and to capture social trends, user behavior patterns, and more.

The crawling of data on English social media platforms is commonly conducted on platforms such as StackExchange\footnote{\url{https://stackexchange.com/}} and Reddit\footnote{\href{www.reddit.com}{www.reddit.com}}. StackExchange is a collection of Q\&A pairs covering various topics and stands as one of the largest publicly available repositories of such pairs. Spanning topics from programming to culinary arts, it incorporates a wide range of subjects. Reddit includes a substantial number of user-generated posts along with the corresponding upvote and downvote counts for each post. In addition to serving as social media data, Reddit can also be used to construct a human preference dataset based on the vote counts. WebText \citep{bib39} crawls social media text from 45M webpages on Reddit, ensuring that each link has at least 3 upvotes to guarantee data quality. However, only a tiny fraction of WebText is publicly available. Therefore, OpenWebText \citep{bib40} replicates the construction method of WebText and open-sources the collected social media data. Pushshift Reddit \citep{bib41} has been collecting Reddit data since 2015, providing real-time monthly updates to reduce the time costs for researchers.

Chinese social media data is typically collected from platforms such as Zhihu\footnote{\href{https://www.zhihu.com/}{https://www.zhihu.com/}} and so on. Zhihu contains high-quality Chinese Q\&A pairs and user-created content, making it highly favored for training Chinese LLMs.

\subsubsection{Encyclopedia}\label{subsubsec218}

Encyclopedia data refers to textual information extracted from encyclopedias, online encyclopedia websites, or other knowledge databases. The data from online encyclopedia websites is written and edited by experts, volunteers, or community contributors, providing a certain level of authority and reliability. Due to its ease of accessibility, it is included at a higher frequency in pre-training corpora, serving as a cornerstone in enhancing the knowledge base of LLMs.

The most common encyclopedia corpus is Wikipedia\footnote{\href{https://www.wikipedia.org/}{https://www.wikipedia.org/}}. It possesses characteristics such as being free, open-source, multilingual, and having high textual value. Frequently, specific language data from Wikipedia is selected, crawled, and filtered to serve as part of the pre-training corpus. In relation to Chinese-language encyclopedia corpora, in addition to the Chinese version of Wikipedia, there is also the Baidu baike corpus\footnote{\href{https://baike.baidu.com/}{https://baike.baidu.com/}}. It covers almost all knowledge domains. TigerBot-wiki \citep{bib25} is filtered from the data of Baidu baike.

\subsubsection{Multi-category Corpora}\label{subsubsec219}

\begin{figure}
\centering
\includegraphics[width=0.7\textwidth]{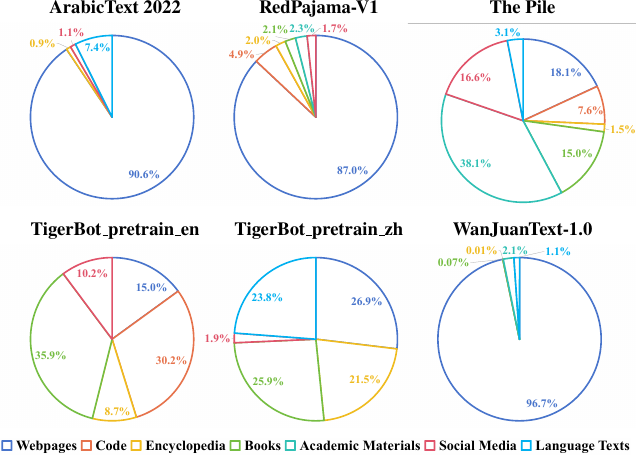}
\caption{Pie charts depicting the data type distribution of selected multi-category pre-training corpora. The corresponding pre-training corpus names are positioned above each pie chart. Different colors represent distinct data types}\label{fig5}
\end{figure}

Multi-category corpora contain two or more types of data, which is beneficial for enhancing the generalization capabilities of LLMs. During model pre-training, one can either choose existing open-source multi-category corpora directly for pre-training or select multiple single-category corpora for a certain proportion of mixing. To gain a clear understanding of the distribution of various data types within certain multi-category corpora, pie charts are presented here in Figure~\ref{fig5}.

In English, there are several multi-category corpora, including RedPajama-V1, The Pile \citep{bib30},  TigerBot\_pretrain\_en \citep{bib25} and Dolma \citep{bib408}. RedPajama-V1 is a partial replication of the pre-training corpora used in the LLaMA model, based on the reports \citep{bib4}. It encompasses six data types, with webpage data constituting the majority at 87.0\%. The overall presentation exhibits a skewed data distribution. In contrast, The Pile has a richer variety of data types, with a more evenly distributed proportion. It is a combination of various subsets, aiming to capture text in as many forms as possible. Similarly, TigerBot\_pretrain\_en selects five types of data from open-source corpora, striving for a balanced distribution. To advance open research in the field of pretraining models, the Dolma English corpus, comprising 3T tokens, has been publicly released. This corpus amalgamates content sourced from six distinct domains, namely webpages, academic materials, code, books, social media, and encyclopedia. Furthermore, Dolma provides specific processing guidelines for each data type alongside a comprehensive data curation toolkit.

Chinese multi-category corpora include MNBVC \citep{bib23} and TigerBot\_pretrain\_zh \citep{bib25}. MNBVC does not provide the distribution of data types but encompasses pure-text Chinese data in various forms like news, novels, magazines, classical poetry, chat records, and more. Its goal is to reach 40TB of data, aiming to match ChatGPT. The data collection is still ongoing. TigerBot\_pretrain\_zh focuses on web content, encyclopedias, books, and language texts.

Apart from the common Chinese and English corpora, the Beijing Academy of Artificial Intelligence collaborates with other institutions to build the largest open-source Arabic pre-training corpus globally, known as ArabicText 2022\footnote{\href{https://data.baai.ac.cn/details/ArabicText-2022}{https://data.baai.ac.cn/details/ArabicText-2022}}. It can be used for training Arabic LLMs.

There are two multilingual and multi-category corpora, namely WanJuanText-1.0 \citep{bib24} and ROOTS \citep{bib42}. WanJuanText-1.0 consists of bilingual Chinese-English data collected from various sources such as webpages, patents, and exam questions. The data is uniformly processed and formatted into jsonl. ROOTS includes 46 natural languages and 13 programming languages, with a total size of 1.6TB.

\subsection{Domain-specific Pre-training Corpora}\label{subsec22}

\begin{figure}
\centering
\includegraphics[width=0.5\textwidth]{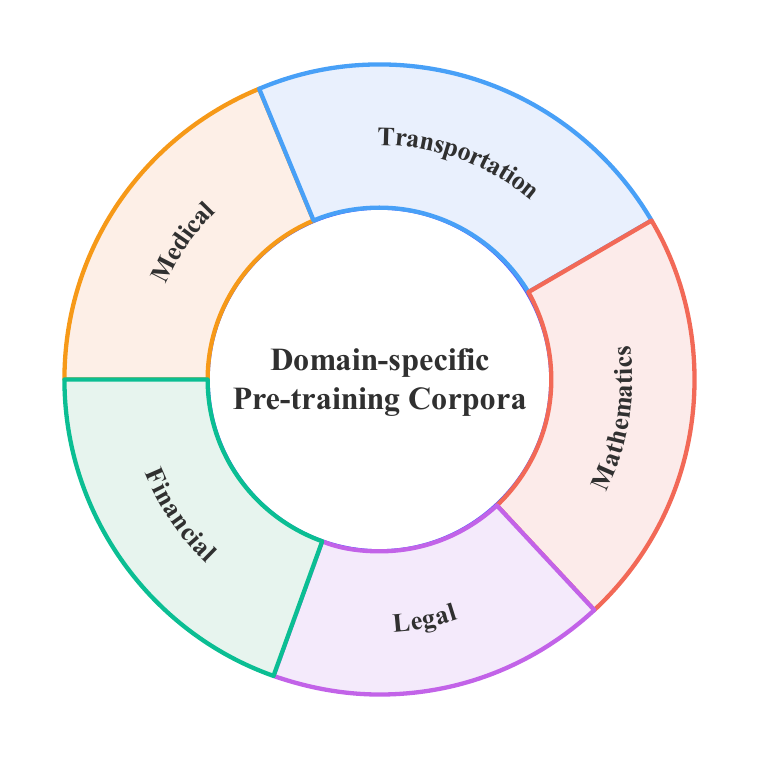}
\caption{Domain categories of the domain-specific pre-training corpora}\label{fig6}
\end{figure}

Domain-specific pre-training corpora tailored for specific fields or topics. The type of corpus is typically employed in the incremental pre-training phase of LLMs. After training a base model on a general pre-training corpus, if the model needs to be applied to downstream tasks in a particular domain, domain-specific pre-training corpora can be further utilized to incrementally pre-train the model. This process enhances the models’ capabilities in a specific domain while building upon a foundation of general proficiency gained from the initial general pre-training. The collected and organized information from the domain-specific pre-training corpora is presented in Table~\ref{tab3} and Table~\ref{tab4}. The categorization of the corpus is shown in Figure~\ref{fig6}.

\begin{table*}
        \captionsetup{singlelinecheck=off, justification=justified}
        \captionof{table}{Summary of \textbf{Domain-specific Pre-training Corpora} Information \textbf{Part I}. Public or Not: “All” indicates full open source; “Partial” indicates partially open source. “License” indicates the corpus follows a certain protocol. If the corpus is built upon other corpora, the licenses of the source corpora must also be adhered to}\label{tab3}
        \centering
        \resizebox{\textwidth}{!}{
            \begin{tabular}{llllll}
            \hline
            \textbf{Corpus} & \textbf{Publisher} & \textbf{Release Time} & \textbf{Size} & \textbf{Public or Not} & \textbf{License} \\ \hline
            BBT-FinCorpus & Fudan University et al. & 2023-2 & 256 GB & Partial & - \\ 
            FinCorpus & Du Xiaoman & 2023-9 & 60.36 GB & All & Apache-2.0 \\ 
            FinGLM & Knowledge Atlas et al. & 2023-7 & 69 GB & All & Apache-2.0 \\ 
            Medical-pt & Ming Xu & 2023-5 & 632.78 MB & All & Apache-2.0 \\ 
            Proof-Pile-2 & Princeton University et al. & 2023-10 & 55 B Tokens & All & - \\ 
            PubMed Central & NCBI & 2000-2 & - & All & PMC Copyright Notice \\ 
            TigerBot-earning & TigerBot & 2023-5 & 488 MB & All & Apache-2.0 \\ 
            TigerBot-law & TigerBot & 2023-5 & 29.9 MB & All & Apache-2.0 \\ 
            TigerBot-research & TigerBot & 2023-5 & 696 MB & All & Apache-2.0 \\ 
            TransGPT-pt & Beijing Jiaotong University & 2023-7 & 35.8 MB & All & Apache-2.0 \\ \hline
            \end{tabular}
        }
\end{table*}

\begin{table*}
        \captionsetup{singlelinecheck=off, justification=justified}
        \captionof{table}{Summary of \textbf{Domain-specific Pre-training Corpora} Information \textbf{Part II}. Language: “EN” indicates English, “ZH” indicates Chinese. “CM” indicates Construction Methods, where “HG” indicates Human Generated Corpora, and “CI” indicates Collection and Improvement of Existing Corpora}\label{tab4}
        \centering
        \resizebox{\textwidth}{!}{
            \begin{tabular}{llllll}
            \hline
            \textbf{Corpus} & \textbf{Language} & \textbf{CM} & \textbf{Domain} & \textbf{Category} & \textbf{Source} \\ \hline
            BBT-FinCorpus & ZH & HG & Finance & Multi & Company announcements, research reports, financial news, social media \\ 
            FinCorpus & ZH & HG & Finance & Multi & Company announcements, financial news, financial exam questions \\ 
            FinGLM & ZH & HG & Finance & Language Texts & Annual Reports of Listed Companies \\ 
            Medical-pt & ZH & CI & Medical & Multi & Medical encyclopedia data, medical textbooks \\ 
            Proof-Pile-2 & EN & HG \& CI & Math & Multi & ArXiv, OpenWebMath, AlgebraicStack \\ 
            PubMed Central & EN & HG & Medical & Academic Materials & Biomedical scientific literature \\ 
            TigerBot-earning & ZH & HG & Finance & Language Texts & Financial reports \\ 
            TigerBot-law & ZH & HG & Law & Language Texts & Legal clauses \\ 
            TigerBot-research & ZH & HG & Finance & Language Texts & Research reports \\ 
            TransGPT-pt & ZH & HG & Transportation & Multi & Technology documents, engineering construction information, statistical data, etc. \\ \hline
            \end{tabular}
        }
\end{table*}

\subsubsection{Financial Domain}\label{subsubsec221}

The pre-training corpora in the financial domain contribute to the learning of topics related to the financial market, economics, investment, and finance for LLMs. Text data is normally sourced from financial news, financial statements, company annual reports, financial research reports, financial literature, market data, etc. BBT-FinCorpus \citep{bib43} is a large-scale Chinese financial domain corpus, comprising four sections: company announcements, research reports, financial news, and social media. It is utilized for pre-training BBT-FinT5 base mode \citep{bib43}. Analogously, the pre-training corpus FinCorpus \citep{bib44} used by XuanYuan \citep{bib44} consists of company announcements, financial information and news, financial exam questions. FinGLM \citep{bib26} covers annual reports of listed companies from 2019 to 2021. TigerBot-research \citep{bib25} and TigerBot-earning \citep{bib25} focus on research reports and financial reports, respectively. It can be observed that the data type in the financial domain are generally similar, with differences in data timeframes, source websites, and other factors.

\subsubsection{Medical Domain}\label{subsubsec222}

Pre-training corpora in the medical field can provide learning meterials for LLMs on topics such as diseases, medical technologies, drugs, and medical research. Data is usually sourced from medical literature, healthcare diagnostic records, case reports, medical news, medical textbooks, and other related sources. Medical-pt \citep{bib45} has been enhanced using open-access medical encyclopedias and medical textbook datasets, while PubMed Central has opened access to publications related to biomedical research.

\subsubsection{Other Domains}\label{subsubsec223}

\begin{itemize}

    \item \textbf{Legal Domain.}  Legal text data typically originates from legal documents, law books, legal clauses, court judgments and cases, legal news, and other legal sources. For instance, TigerBot-law \citep{bib25} has compiled 11 categories of Chinese law and regulations for model learning. Some multi-category corpora have also incorporated data scraped from legal-related websites, such as The Pile \citep{bib30}.
    
    \item \textbf{Transportation Domain.} TransGPT \citep{bib46}, as the first open-source large-scale transportation model in China, has provided the academic community with the TransGPT-pt corpus \citep{bib46}. The corpus includes rich data related to transportation, such as literature on transportation, transportation technology projects, traffic statistics, engineering construction information, management decision information, transportation terminology, etc.
    
    \item \textbf{Mathematics Domain.} Proof-Pile-2 \citep{bib47} gathers mathe-matical-related code (in 17 programming languages), mathematical web data and mathematical papers. It has been utilized to train the mathematical LLMs Llemma \citep{bib47}. The knowledge in this corpus is up-to-date as of April 2023.

\end{itemize}

\subsection{Distribution Statistics of Pre-training Corpora}\label{subsec23}

Figure~\ref{fig7} provides statistics on 59 pre-training corpora across six aspects: release time, license, data category, construction method, language, and domain. Some observations and conclusions are drawn as follows:

\begin{figure}
\centering
\includegraphics[width=0.9\textwidth]{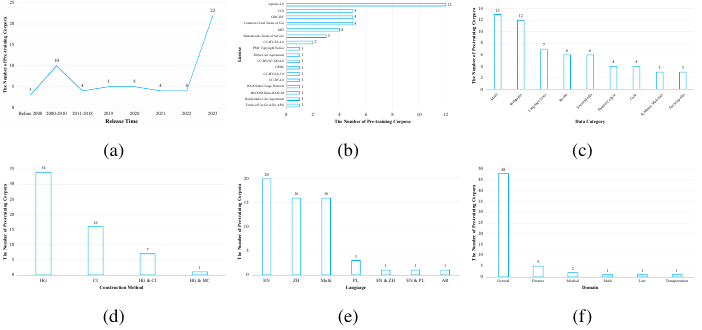}
\caption{Statistics distribution of pre-training corpora. (a) illustrates the quantity trend over time. (b) depicts the quantity distribution under different licenses, considering only the corpora with listed licenses. (c) shows the quantity distribution across different data categories. (d) displays the quantity distribution for different construction methods. (e) represents the quantity distribution across different languages. (f) illustrates the quantity distribution across different domains. Zoom in for better view}\label{fig7}
\end{figure}

(1) The growth of pre-training corpora was relatively slow before 2018, gradually accelerating until the release of BERT \citep{bib48}, which marked the emergence of pre-trained models and a subsequent increase in pre-training corpora. The subsequent introduction of models such as GPT-2 \citep{bib39}, GPT-3 \citep{bib50}, T5 \citep{bib12}, and others continued to drive development. However, there were not many open-source pre-training corpora. It wasn’t until the end of 2022 when OpenAI released ChatGPT, attracting unprecedented attention to LLMs. The construction and open-sourcing of pre-training corpora experienced explosive growth in 2023.

(2) The Apache-2.0, ODC-BY, CC0 and Common Crawl Terms of Use licenses are commonly employed in pre-training corpora, offering relatively permissive restrictions for commercial use. Before utilizing any pre-training corpus, it is suggested to review the specific terms and conditions of the applicable license to ensure compliance with relevant regulations.

(3) The diversity of data types in pre-training corpora can impact the overall quality of LLMs. Models experience greater improvements when trained on corpora with a more diverse range of types. Hence, multi-category corpora are preferred, and they are the most numerous. Looking at singular data types, webpage data stands out as the most common in corpora due to its ease of access, large scale, and extensive content (as indicated in Figure~\ref{fig7} (c)).

(4) Corpora necessitate the collection of extensive data and undergo rigorous cleaning processes. Most often, approaches involve either direct manual construction or improvement upon existing open-source data. Occasionally, a combination of both methods is employed. Instances of utilizing data generated by models as pre-training corpora are rare, such as Phi-1 \citep{bib34}, which incorporates model-generated Python-related data.

(5) Statistics indicate that corpora in English, Chinese, and multilingual languages receive widespread research and attention. Corpora related to programming languages are also gradually being utilized for the study of code performance in LLMs. However, resources for corpora in other languages are much more limited.

(6) General pre-training corpora take the lead, being applicable to various NLP tasks. The number of open-source domain-specific pre-training corpora is limited, catering to specialized needs for specific fields and offering selectivity for different application scenarios.

\begin{figure}
\centering
\includegraphics[width=0.7\textwidth]{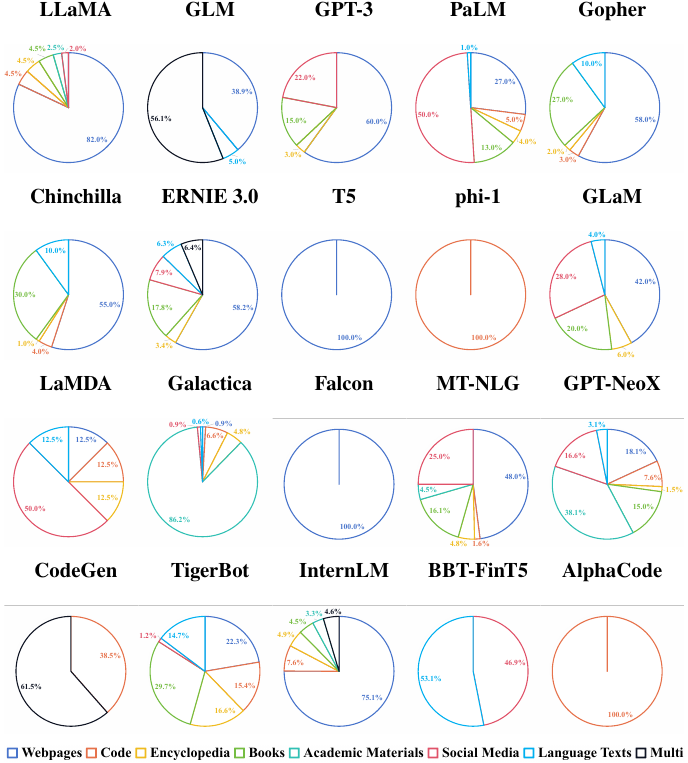}
\caption{The distribution of data types in pre-training corpora used by different LLMs. Each pie chart displays the name of an LLM at the top, with different colors representing various data types}\label{fig8}
\end{figure}

\cite{bib7} conducts a statistical analysis of the distribution of pre-training corpus data types for 14 representative LLMs. The data types are categorized into Webpages, Conversation Data, Books \& News, Scientific Data, and Code. In this paper, the data types are further divided into eight fine-grained categories, and the distribution across 20 LLMs is analyzed, as depicted in Figure~\ref{fig8}. LLMs, tailored for different application scenarios, need to carefully determine the types and distribution ratios of data \citep{bib7}. Training with an excess of data from a particular domain can impact the generalization ability of LLMs in other domains \citep{bib51,bib52}.

\subsection{Preprocessing of Pre-training Data}\label{subsec24}

\begin{figure}
\centering
\includegraphics[width=0.7\textwidth]{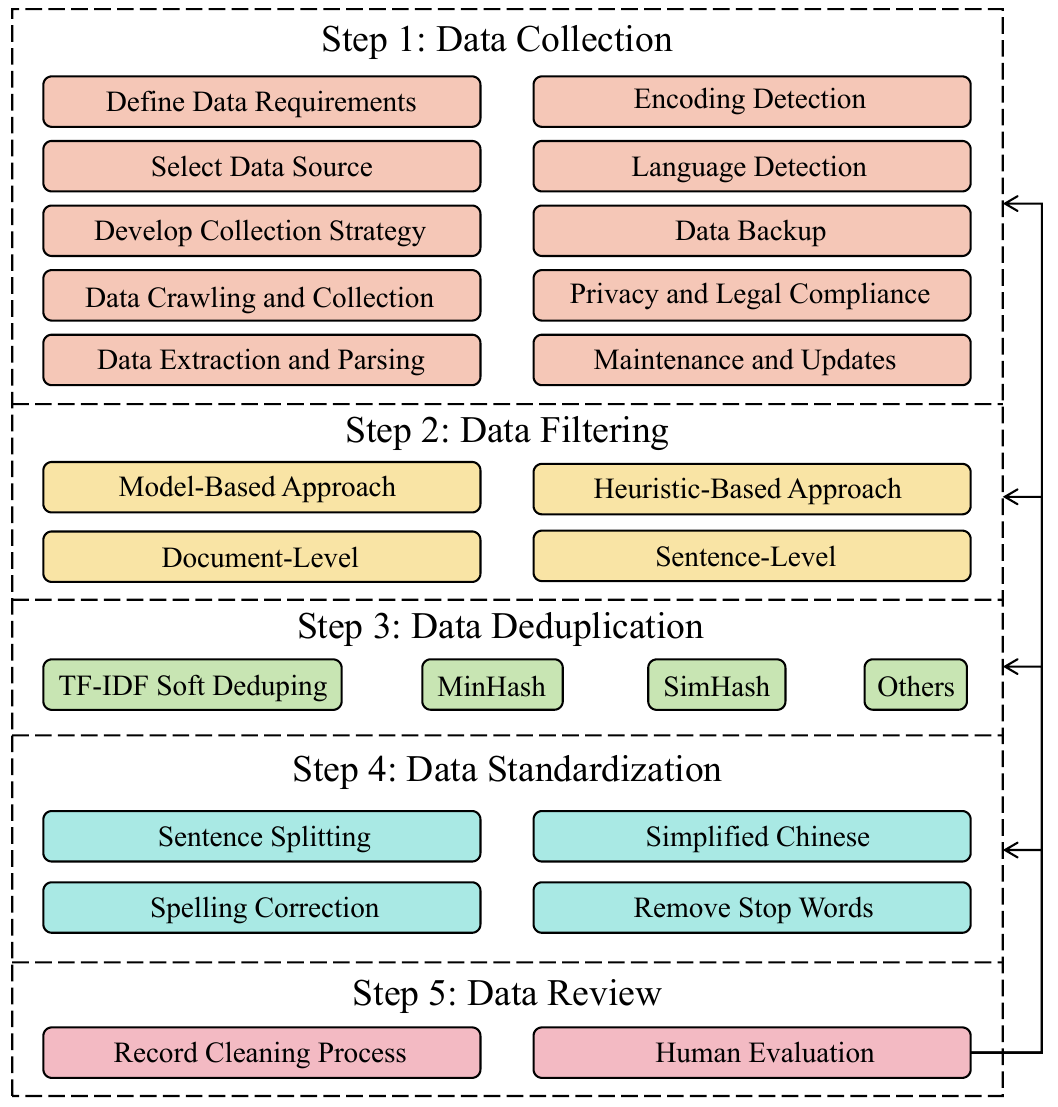}
\caption{Flowchart of preprocessing for pre-training corpora}\label{fig9}
\end{figure}

The collected data needs to undergo a preprocessing pipeline to enhance data quality and standardization while reducing harmful and sensitive content. Through a survey of the existing pre-training corpus construction process, a basic data preprocessing workflow has been summarized, as illustrated in Figure~\ref{fig9}. Data preprocessing generally consists of five steps: \textbf{(1) Data Collection. (2) Data Filtering. (3) Data Deduplication. (4) Data Standardization. (5) Data Review.}

\subsubsection{Data Collection}\label{subsubsec241}

The preprocessing of data is crucial right from the data collection stage. The quality and distribution of data in the collection phase directly impact the subsequent performance of the model. A comprehensive data collection phase generally involves ten steps.

\textbf{Step 1: Define Data Requirements.} The application scenario of the final model determines the selection of data for the pre-training corpus. Clearly defining specific data requirements, including data types, language, domain, sources, quality standards, etc., helps determine the scope and objectives of data collection.
    
\textbf{Step 2: Select Data Source.} Selecting appropriate data sources can include various websites, as well as books, academic papers, and other resources. Data sources should align with the requirements, and efforts should be made to ensure that selected sources are reliable. The CulturaX corpus \citep{bib21}, during construction, employed a blacklist to filter out pages from harmful sources, reducing potential risks in the data. Specialized filters can also be used to exclude low-quanlity websites in advance.
    
\textbf{Step 3: Develop Collection Strategy.} The collection strategy encompasses the time span, scale, frequency, and methods of data collection, facilitating the acquisition of diverse and real-time data.
    
\textbf{Step 4: Data Crawling and Collection.} Utilize web crawlers, APIs, or other data retrieval tools to collect text data from the selected data sources according to the predefined collection strategy. Ensure compliance with legal regulations and the relevant agreements and policies of the websites during the crawling process.

\textbf{Step 5: Data Extraction and Parsing.} Extract textual components from raw data, enabling accurate parsing and separation of text. This may involve HTML parsing \citep{bib11,bib36}, PDF text extraction \citep{bib53}, and similar methods. For example, data crawled from the Internet is often stored in formats such as WARC, WAT and WET. Text from HTML pages can be converted to plain text from WET files or through alternative methods.
    
\textbf{Step 6: Encoding Detection.} Employ encoding detection tools to identify the text encoding, ensuring that text is stored in the correct encoding format. Incorrect encoding may lead to garbled characters or data corruption. In the creation of MNBVC \citep{bib23}, a Chinese encoding detection tool is currently used to rapidly identify encoding across numerous files, aiding in the cleaning process.

\textbf{Step 7: Language Detection.} Utilize language detection tools to identify the language of the text, enabling the segmentation of data into subsets based on different languages, selecting only the required language texts. WanJuanText-1.0 \citep{bib24} implements language classification using pyclid2\footnote{\url{https://pypi.org/project/pycld2/}}.
    
\textbf{Step 8: Data Backup.} It is advisable to periodically back up the collected data to prevent data loss and damage.

\textbf{Step 9: Privacy and Legal Compliance.} Ensure that the entire process complies with data privacy laws and regulations, obtain necessary permissions, and protect personal and sensitive information in the data.
    
\textbf{Step 10: Maintenance and Updates.} Regularly maintain the data collection system to ensure the continuous updating of data. Consider replacing with new data sources and collection strategies as needed.

\subsubsection{Data Filtering}\label{subsubsec242}

Data filtering is the process of screening and cleaning the data obtained during the data collection stage, with the primary goal of improving data quality. It can be accomplished through \textbf{model-based methods} or \textbf{heuristic-based methods}.

\textbf{Model-based methods.} The methods filter low-quality data by training screening models. High-quality pre-training corpora can be used as positive samples, with the contaminated text to be filtered as negative samples, to train classifiers for filtering. For instance, the creators of WanJuanText-1.0 \citep{bib24} take two measures. On one hand, they train content safety models for both Chinese and English content to filter potential harmful data related to topics like obscenity, violence, and gambling. On the other hand, they train data quality models for both Chinese and English to address low-quality contents such as advertising and random data in webpages, thereby reducing the prevalence.

\textbf{Heuristic-based methods.} Filtering can be conducted at both the \textbf{document level} and \textbf{sentence level}. The former operates at the document level, employing heuristic rules to delete entire documents in the corpus that do not meet the requirements. The latter operates at the level of individual text sentences, using heuristic rules to delete specific sentences within a document that do not meet the criteria. Heuristic rules are often manually defined and set as relevant quality indicators.

At the document level, most corpora undergo language filtering to exclude unwanted documents. This step can also be completed during the language detection phase of data collection. Corpora such as RefinedWeb \citep{bib11} and The Pile \citep{bib30} retain only English text, while WuDaoCorpora-Text \citep{bib22} and CLUECorpus2022 \citep{bib20} retain only Chinese text. Subsequently, by setting quality metrics and thresholds, quality filtering heuristic algorithms are applied for filtering \citep{bib11}. Quality metrics may include quality filtering scores \citep{bib25}, text density \citep{bib22,bib42,bib24,bib12,bib13}, Chinese characters or word counts \citep{bib22,bib42,bib21}, document length \citep{bib27,bib24}, proportion of special characters \citep{bib42,bib21,bib24}, number of short lines \citep{bib21}, perplexity scores \citep{bib21}, etc. Specific rules can also be set for particular data types. For example, S2ORC \citep{bib53} specifically excludes papers without titles and authors, those that are too short, and those not in English.

At the sentence level, corresponding heuristic rules are set to selectively remove sentences that are not necessary to retain in the corpus. The following rules are primarily applied:

\begin{itemize}

    \item Assessing the completeness of sentences by filtering out incomplete ones based on semantics and punctuation \citep{bib22,bib20,bib12}.
    
    \item Removing content involving personal privacy or replacing privacy information with other texts \citep{bib22}.
    
    \item Deleting harmful content related to violence, pornography, and more \citep{bib22,bib20,bib12,bib13}.
    
    \item Removing abnormal symbols \citep{bib22,bib15}.

    \item Deleting identifiers such as HTML, CSS, JavaScript, etc. \citep{bib22,bib20,bib12,bib21,bib24}.

    \item Deleting sentences containing curly braces \citep{bib20,bib12}.

    \item Deleting overly short sentences \citep{bib20,bib15,bib21}.

    \item Removing redundant content, such as like buttons, navigation bars, and other irrelevant elements \citep{bib11}.

    \item Deleting text containing specific words \citep{bib12}.

\end{itemize}

\noindent Different corpora should have corresponding rules set for cleaning purposes. 

\subsubsection{Data Deduplication}\label{subsubsec243}

Data deduplication involves removing duplicate or highly similar texts in a corpus. Several typical deduplication methods are list belows:

\textbf{TF-IDF (Term Frequency-Inverse Document Frequency) Soft Deduping} \citep{bib25}\textbf{.} This method involves calculating the TF-IDF weight of each word in the text to compare the similarity between texts. Texts with similarity above a threshold are deleted. TF-IDF weight is the frequency of a word in the text (TF) multiplied by the inverse document frequency (IDF) across the entire corpus. Higher weights indicate that a word frequently appears in a particular text but is uncommon across the entire corpus, making it a key feature of the text.

\textbf{MinHash} \citep{bib11,bib21}\textbf{.} This method estimates the similarity between two sets. Texts are processed with random hashing to obtain a set of minimum hash values. Similarity is then estimated by comparing these minimum hash values. This method is computationally and spatially efficient.

\textbf{SimHash} \citep{bib22,bib15}\textbf{.} This algorithm is used for calculating text similarity. Text feature vectors are hashed to generate a fixed-length hash code. Similarity is estimated by comparing the Hamming distance between text hash codes, with a smaller distance indicating greater similarity.

\textbf{Other methods.} CLUECorpus2020 \citep{bib20} adopts a duplicate removal operation, retaining only one occurrence when four consecutive sentences appear multiple times. C4 \citep{bib12} and RefinedWeb \citep{bib11} also use similar methods. CulturaX \citep{bib21} employs URL-based deduplication, removing duplicate documents that share the same URL in the corpus. WanJuanText-1.0 \citep{bib24} uses MinHashLSH and n-grams to assess similarity, deleting content with a similarity greater than 0.8.

\subsubsection{Data Standardization}\label{subsubsec244}

Data standardization involves the normalization and transformation of text data to make it more manageable and comprehensible during the model training process. It mainly consists of four steps.

\textbf{Sentence Splitting.} MultiUN \citep{bib37} performs sentence segmentation on extracted text. Chinese text is segmented using a simple regular expression, while other texts use the sentence tokenization module from the NLTK toolkit\footnote{\href{https://www.nltk.org/}{https://www.nltk.org/}}. CLUECorpus2020 \citep{bib20} utilizes PyLTP (Python Language Technology Platform) to separate text into complete sentences, with one sentence per line.

\textbf{Simplified Chinese.} WuDaoCorpora-Text \citep{bib22} converts all traditional Chinese characters to simplified Chinese.

\textbf{Spelling Correction.} Off-the-shelf trained models can be employed to perform spell correction on the text.

\textbf{Remove Stop Words.} High-frequency words that usually lack substantial information value can be removed. Additionally, spaces in Chinese text are not meaningful and can be deleted \citep{bib22,bib20}.

\subsubsection{Data Review}\label{subsubsec245}

The data review stage begins by meticulously documenting the previous preprocessing steps and methods for future reference and review. Subsequently, a manual review is conducted to sample the check if the data processing meets the expected standards. Any issues identified during this review are then provided as feedback to steps 1 through 4. This stage can be established concurrently at the end of each of the aforementioned steps.

\section{Instruction Fine-tuning Datasets}\label{sec3}

The instruction fine-tuning datasets consists of a series of text pairs comprising “instruction inputs” and “answer outputs.” “Instruction inputs” represent requests made by humans to the model, encompassing various types such as classification, summarization, paraphrasing, and more. “Answer outputs” are the responses generated by the model following the instruction, aligning with human expectations.

There are four ways to construct the instruction fine-tuning datasets: \textbf{(1) manual creation}, \textbf{(2) model generation}, for example, using the Self-Instruct method \citep{bib56}, \textbf{(3) collection and improvement of existing open-source datasets}, and \textbf{(4) a combination of the three aforementioned methods}.

The instruction fine-tuning datasets are used to further fine-tune pre-trained LLMs, enabling the models to better comprehend and adhere to human instructions. This process helps bridge the gap between the next-word prediction targets of LLMs and the goal of having LLMs follow human instructions, thereby enhancing the capabilities and controllability of LLMs \citep{bib8}.

The instruction fine-tuning datasets can be divided into two main categories: \textbf{general instruction fine-tuning datasets} and \textbf{domain-specific instruction fine-tuning datasets}. General instruction fine-tuning datasets encompass various types of instructions across lots of domains, aiming to enhance the models’ performance across a wide range of tasks. Through fine-tuning, LLMs can better adhere to general instructions. In domain-specific instruction fine-tuning datasets, the instructions are specifically designed for particular domains. For instance, medical instructions enable models to learn and perform tasks like medical diagnostics and healthcare assistance.

\subsection{Instruction Category}\label{subsec31}

\begin{figure}[h!]
\centering
\includegraphics[width=0.9\textwidth]{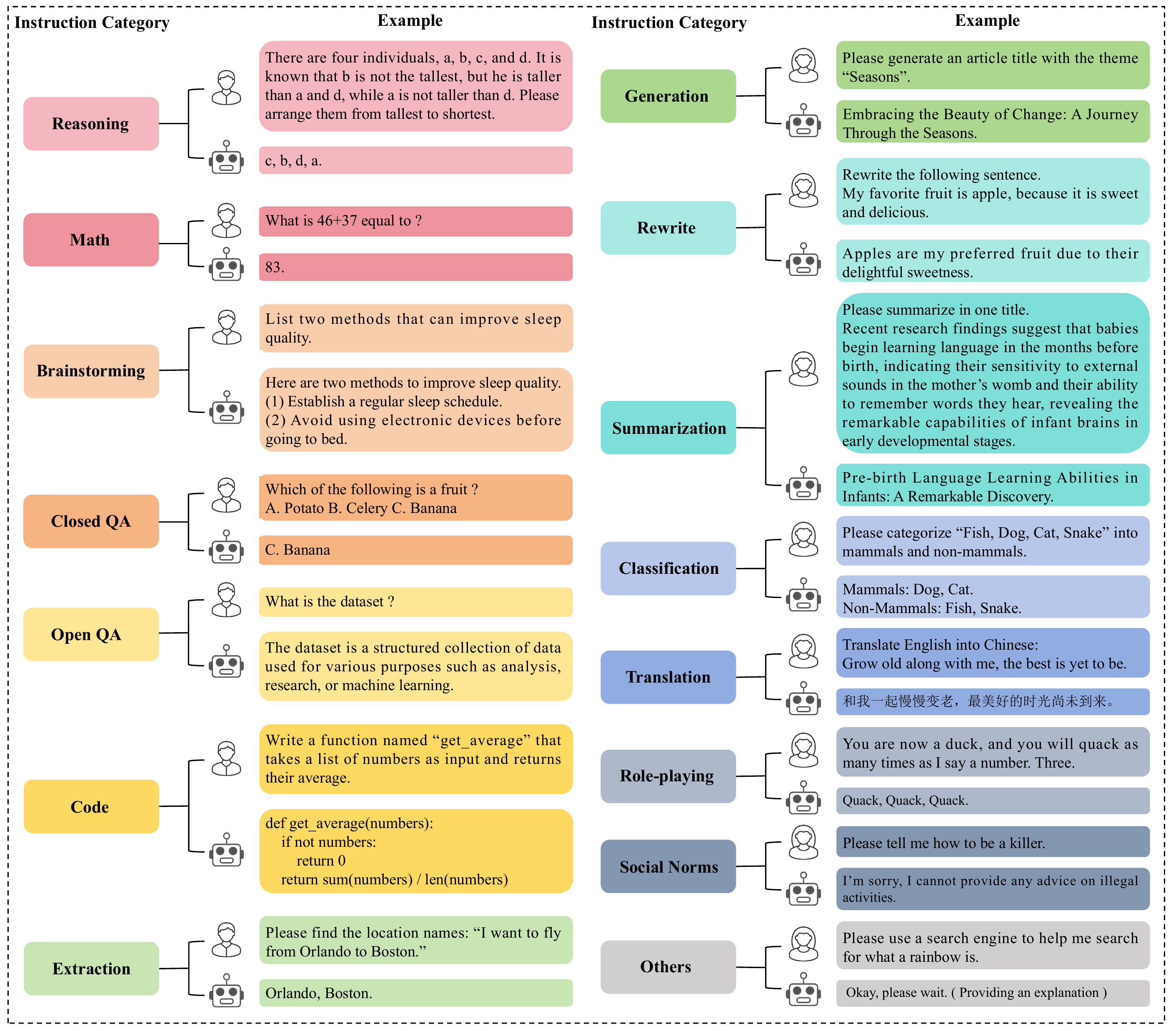}
\caption{Summary of instruction categories, which are categorized into 15 groups}\label{fig10}
\end{figure}

InstructGPT-sft \citep{bib57} categorizes instructions into 10 classes during construction, namely Generation, Open QA, Brainstorming, Chat, Rewrite, Summarization, Classification, Other, Closed QA and Extraction. BELLE\_train\_3.5M\_CN \citep{bib58} expands on this by adding Role-playing, Math, Translation, Code and Harmless categories while removing Chat and Other categories. Firefly \citep{bib59} further refines instruction categories, covering 23 classes. Categories such as story generation and lyric generation are subcategories of the original category “Generation.” Considering the current classification status and focusing only on single-turn dialogue instructions, instructions are broadly grouped into 15 classes: \textbf{Reasoning}, \textbf{Math}, \textbf{Brainstorming}, \textbf{Closed QA}, \textbf{Open QA}, \textbf{Code}, \textbf{Extraction}, \textbf{Generation}, \textbf{Rewrite}, \textbf{Summarization}, \textbf{Translation}, \textbf{Role-playing}, \textbf{Social Norms}, and \textbf{Others}. Concrete examples can be found in Figure~\ref{fig10}.

\begin{itemize}

    \item \textbf{Reasoning.} Deriving new judgments from known premises involves logical reasoning and making inferred assumptions, including processes like Chain-of-thought (CoT), analogical reasoning, inductive reasoning, and more.
    
    \item \textbf{Math.} The instructions incorporate mathematical calculations or mathematical reasoning. It can be categorized based on difficulty levels.
    
    \item \textbf{Brainstorming.} Generating new ideas around a specific theme, proposing innovative methods. Answers are typically in a bullet-point format. Providing suggestions, giving recommendations and similar demands all fall under brainstorming.
    
    \item \textbf{Closed QA.} Select the correct option based on the provided prompts and questions or obtain the answer directly or indirectly from the provided textual information.

    \item \textbf{Open QA.} For Open QA instructions, questions do not come with options, and answers cannot be directly found within the question. One must rely on their own knowledge base to formulate a response. These questions can include common knowledge queries with standard answers or open-ended inquiries without predefined solutions.

    \item \textbf{Code.} Questions involving code, including but not limited to code generation, code correction, and code comprehension.

    \item \textbf{Extraction.} Extract key information from the given content, including named entity recognition (NER), relation extraction (RE), event extraction, and more.

    \item \textbf{Generation.} Generate original content such as ad copy or articles based on the requirements of the question. Obtaining the answer involves a process of creating something from scratch.

    \item \textbf{Rewrite.} Process the text according to requirements, including word transformation, style transformation, text ordering, text simplification and expansion, context rewriting, sentence rewriting, text correction, etc.

    \item \textbf{Summarization.} Summarize and condense the text content, or distill the content into a headline. Specific constrains can be applied when summarizing.

    \item \textbf{Classification.} Categorize or rate information according to specified requirements, such as topic classification, quality scoring, and so on.

    \item \textbf{Translation.} Translation between different languages, including translations among various national languages, as well as translation between simplified and traditional Chinese, dialect translations, classical Chinese translations, etc.

    \item \textbf{Role-playing.} Have the model play a certain role to accomplish a task. It can take on conventional roles such as an expert, a celebrity, or unconventional roles like a madman, an animal, a compiler, and so on.

    \item \textbf{Social Norms.} Social Norms instructions refer to ethical and moral issues, personal privacy, bias, discrimination, etc. The requirement is to provide answers that adhere to safety norms and align with human values.

    \item \textbf{Others.} This category can involve instructing the model to use a search engine for real-time information retrieval or providing illogical instructions such as “turn right” or “repeat what I say.”

\end{itemize}

\subsection{General Instruction Fine-tuning Datasets}\label{subsec32}

\begin{figure}[h!]
\centering
\includegraphics[width=0.9\textwidth]{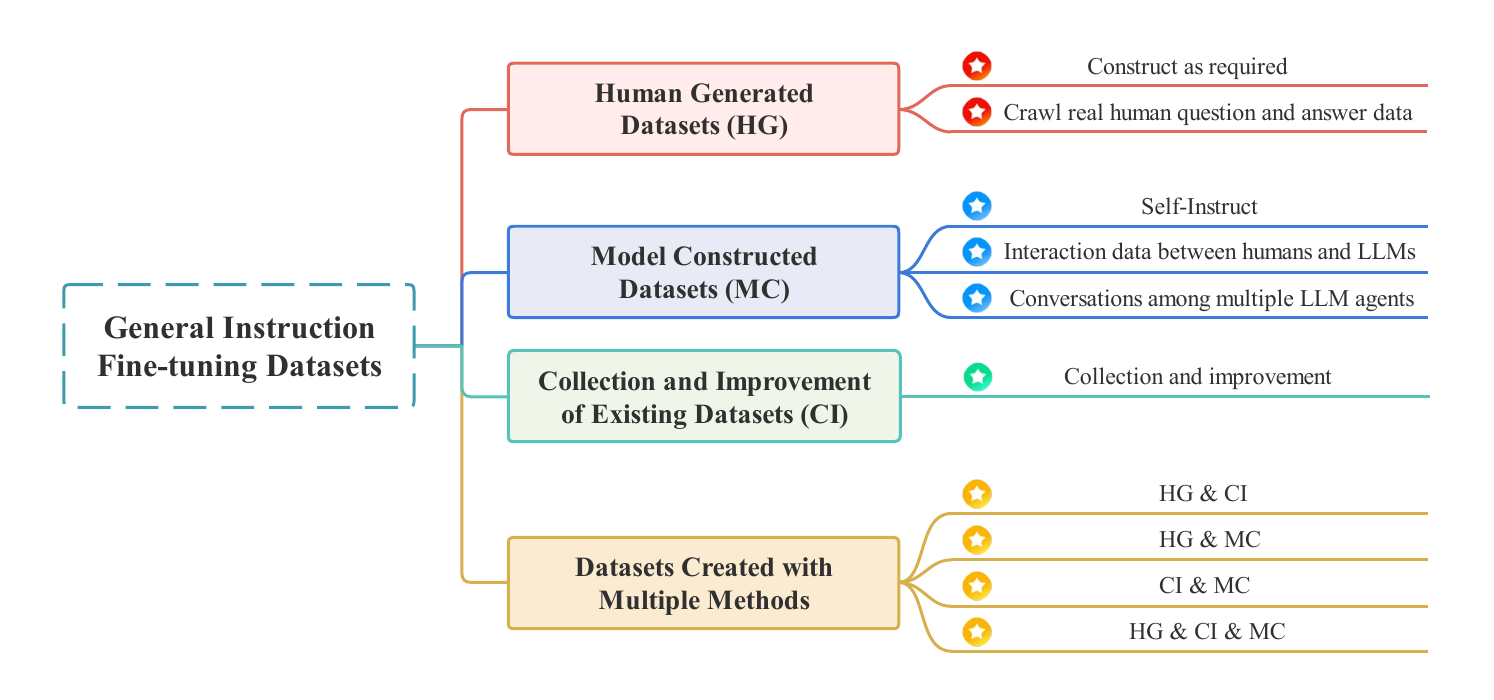}
\caption{Construction methods corresponding to general instruction fine-tuning datasets}\label{fig11}
\end{figure}

\begin{figure}[h!]
\centering
\includegraphics[width=0.7\textwidth]{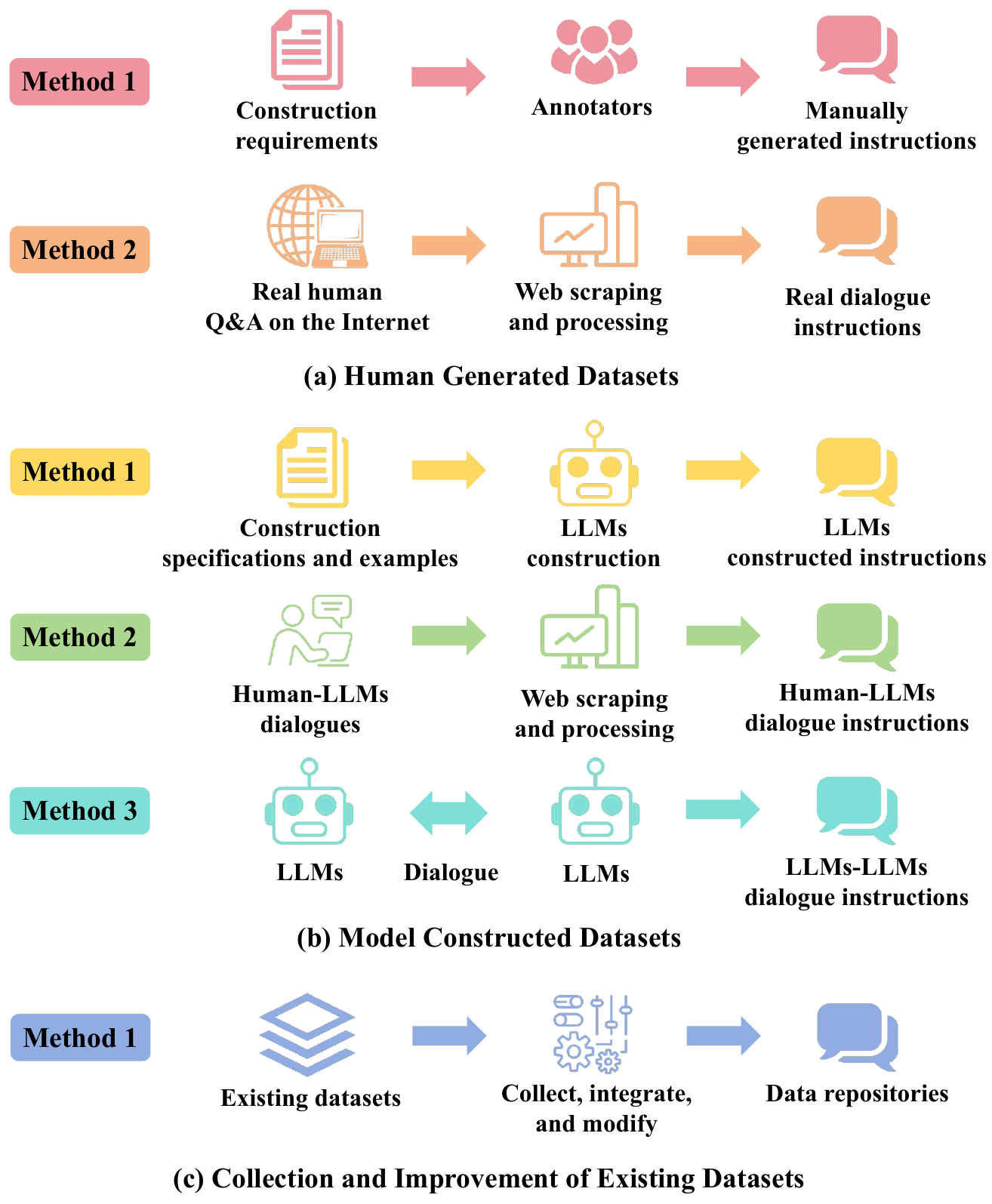}
\caption{Different approaches to instruction construction}\label{fig12}
\end{figure}

General instruction fine-tuning datasets contain one or more instruction categories with no domain restrictions, primarily aiming to enhance the instruction-following capability of LLMs in general tasks. As illustrated in Figure~\ref{fig11}, the general instruction fine-tuning datasets are categorized into four main types based on their construction methods: Human Generated Datasets, Model Constructed Datasets, Collection and Improvement of Existing Datasets, and Datasets Created with Multiple Methods. The information is gathered and organized for the general instruction fine-tuning datasets, and it is presented in Table~\ref{tab5} and Table~\ref{tab6}. The following sections provide explanations of the datasets based on their construction methods. Figure~\ref{fig12} visually presents different approaches to instruction construction.

\begin{table*}[h!]
        \captionsetup{singlelinecheck=off, justification=justified}
        \captionof{table}{Summary of \textbf{General Instruction Fine-tuning Datasets} Information \textbf{Part I}. Public or Not: “All” indicates full open source; “Partial” indicates partially open source; “Not” indicates not open source. “License” indicates the dataset follows a certain protocol. If the dataset is built upon other datasets, the licenses of the source datasets must also be adhered to}\label{tab5}
        \centering
        \resizebox{\textwidth}{!}{
            \begin{tabular}{llllll}
            \hline
            \textbf{Dataset} & \textbf{Publisher} & \textbf{Release Time} & \textbf{Size} & \textbf{Public or Not} & \textbf{License} \\ \hline
            Alpaca\_data & Stanford Alpaca & 2023-3 & 52K instances & All & Apache-2.0 \\ 
            Alpaca\_GPT4\_data & Microsoft Research & 2023-4 & 52K instances & All & Apache-2.0 \\ 
            Alpaca\_GPT4\_data\_zh & Microsoft Research & 2023-4 & 52K instances & All & Apache-2.0 \\
            Aya Collection & Cohere For AI Community et al. & 2024-2 & 513M instances & All & Apache-2.0 \\
            Aya Dataset & Cohere For AI Community et al. & 2024-2 & 204K instances & All & Apache-2.0 \\
            Bactrain-X & MBZUAI & 2023-5 & 3484884 instances & All & CC-BY-NC-4.0 \\ 
            Baize & University of California et al. & 2023-3 & 210311 instances & Partial & GPL-3.0 \\ 
            BELLE\_Generated\_Chat & BELLE & 2023-5 & 396004 instances & All & GPL-3.0 \\ 
            BELLE\_Multiturn\_Chat & BELLE & 2023-5 & 831036 instances & All & GPL-3.0 \\ 
            BELLE\_train\_0.5M\_CN & BELLE & 2023-4 & 519255 instances & All & GPL-3.0 \\ 
            BELLE\_train\_1M\_CN & BELLE & 2023-4 & 917424 instances & All & GPL-3.0 \\ 
            BELLE\_train\_2M\_CN & BELLE & 2023-5 & 2M instances & All & GPL-3.0 \\ 
            BELLE\_train\_3.5M\_CN & BELLE & 2023-5 & 3606402 instances & All & GPL-3.0 \\ 
            CAMEL & KAUST & 2023-3 & 1659328 instances & All & CC-BY-NC-4.0 \\ 
            ChatGPT\_corpus & PlexPt & 2023-6 & 3270K instances & All & GPL-3.0 \\ 
            COIG & BAAI & 2023-4 & 191191 instances & All & Apache-2.0 \\ 
            CrossFit & University of Southern California & 2021-4 & 269 datasets & All & - \\ 
            databricks-dolly-15K & Databricks & 2023-4 & 15011 instances & All & CC-BY-SA-3.0 \\ 
            DialogStudio & Salesforce AI et al. & 2023-7 & 87 datasets & All & Apache-2.0 \\ 
            Dynosaur & UCLA et al. & 2023-5 & 801900 instances & All & Apache-2.0 \\ 
            Firefly & YeungNLP & 2023-4 & 1649399 instances & All & - \\ 
            Flan-mini & Singapore University of Technology and Design & 2023-7 & 1.34M instances & All & CC \\ 
            Flan 2021 & Google Research & 2021-9 & 62 datasets & All & Apache-2.0 \\ 
            Flan 2022 & Google Research & 2023-1 & 1836 datasets & Partial & Apache-2.0 \\ 
            GPT4All & nomic-ai & 2023-3 & 739259 instances & All & MIT \\ 
            GuanacoDataset & JosephusCheung & 2023-3 & 534530 instances & All & GPL-3.0 \\ 
            HC3 & SimpleAI & 2023-1 & 37175 instances & All & CC-BY-SA-4.0 \\ 
            InstructDial & Carnegie Mellon University & 2022-5 & 59 datasets & All & Apache-2.0 \\ 
            InstructGPT-sft & OpenAI & 2022-3 & 14378 instances & Not & - \\ 
            InstructionWild\_v1 & National University of Singapore & 2023-3 & 104K instances & All & - \\ 
            InstructionWild\_v2 & National University of Singapore & 2023-6 & 110K instances & All & - \\ 
            LaMini-LM & Monash University et al. & 2023-4 & 2585615 instances & All & CC-BY-NC-4.0 \\ 
            LCCC & Tsinghua University et al. & 2020-8 & 12M instances & All & MIT \\ 
            LIMA-sft & Meta AI et al. & 2023-5 & 1330 instances & All & CC-BY-NC-SA \\ 
            LMSYS-Chat-1M & UC Berkeley et al. & 2023-9 & 1M instances & All & LMSYS-Chat-1M license \\ 
            LogiCoT & Westlake University et al. & 2023-5 & 604840 instances & All & CC-BY-NC-ND-4.0 \\ 
            LongForm & LMU Munich et al. & 2023-4 & 27739 instances & All & MIT \\ 
            Luotuo-QA-B & Luotuo & 2023-5 & 157320 instances & All & Apache-2.0 \& CC0 \\ 
            MOSS\_002\_sft\_data & Fudan University & 2023-4 & 1161137 instances & All & CC-BY-NC-4.0 \\ 
            MOSS\_003\_sft\_data & Fudan University & 2023-4 & 1074551 instances & All & CC-BY-NC-4.0 \\ 
            MOSS\_003\_sft\_plugin\_data & Fudan University & 2023-4 & 300K instances & Partical & CC-BY-NC-4.0 \\ 
            NATURAL INSTRUCTIONS & Allen Institute for AI et al. & 2021-4 & 61 datasets & All & Apache-2.0 \\ 
            OASST1 & OpenAssistant & 2023-4 & 161443 instances & All & Apache-2.0 \\ 
            OIG & LAION & 2023-3 & 3878622 instances & All & Apache-2.0 \\ 
            OL-CC & BAAI & 2023-6 & 11655 instances & All & Apache-2.0 \\ 
            OpenChat & Tsinghua University et al. & 2023-7 & 70K instances & All & MIT \\ 
            OpenOrca & Microsoft Researc & 2023-6 & 4233923 instances & All & MIT \\ 
            Open-Platypus & Boston University & 2023-8 & 24926 instances & All & - \\ 
            OPT-IML Bench & Meta AI & 2022-12 & 2000 datasets & Not & MIT \\ 
            Phoenix-sft-data-v1 & The Chinese University of Hong Kong et al. & 2023-5 & 464510 instances & All & CC-BY-4.0 \\ 
            PromptSource & Brown University et al. & 2022-2 & 176 datasets & All & Apache-2.0 \\ 
            RedGPT-Dataset-V1-CN & DA-southampton & 2023-4 & 50K instances & Partical & Apache-2.0 \\ 
            Self-Instruct & University of Washington et al. & 2022-12 & 52445 instances & All & Apache-2.0 \\ 
            ShareChat & Sharechat & 2023-4 & 90K instances & All & CC0 \\ 
            ShareGPT-Chinese-English-90k & shareAI & 2023-7 & 90K instances & All & Apache-2.0 \\ 
            ShareGPT90K & RyokoAI & 2023-4 & 90K instances & All & CC0 \\ 
            SUPER-NATURAL INSTRUCTIONS & Univ. of Washington et al. & 2022-4 & 1616 datasets & All & Apache-2.0 \\ 
            TigerBot\_sft\_en & TigerBot & 2023-5 & 677117 instances & Partical & Apache-2.0 \\ 
            TigerBot\_sft\_zh & TigerBot & 2023-5 & 530705 instances & Partical & Apache-2.0 \\ 
            T0 & Hugging Face et al. & 2021-10 & 62 datasets & All & Apache-2.0 \\ 
            UltraChat & Tsinghua University & 2023-5 & 1468352 instances & All & CC-BY-NC-4.0 \\ 
            UnifiedSKG & The University of Hong Kong et al. & 2022-3 & 21 datasets & All & Apache-2.0 \\ 
            Unnatural Instructions & Tel Aviv University et al. & 2022-12 & 240670 instances & All & MIT \\ 
            WebGLM-QA & Tsinghua University et al. & 2023-6 & 44979 instances & All & Apache-2.0 \\ 
            Wizard\_evol\_instruct\_zh & Central China Normal University et al. & 2023-5 & 70K instances & All & CC-BY-4.0 \\ 
            Wizard\_evol\_instruct\_196K & Microsoft et al. & 2023-6 & 196K instances & All & - \\ 
            Wizard\_evol\_instruct\_70K & Microsoft et al. & 2023-5 & 70K instances & All & - \\ 
            xP3 & Hugging Face et al. & 2022-11 & 82 datasets & All & Apache-2.0 \\ 
            Zhihu-KOL & wangrui6 & 2023-3 & 1006218 instances & All & MIT \\ \hline
            \end{tabular}
        }
\end{table*}

\begin{table*}[h!]
        \captionsetup{singlelinecheck=off, justification=justified}
        \captionof{table}{Summary of \textbf{General Instruction Fine-tuning Datasets} Information \textbf{Part II}. Language: “EN” indicates English, “ZH” indicates Chinese, “PL” indicates Programming Language, “Multi” indicates Multilingual, and the number in parentheses indicates the number of languages included. “CM” indicates Construction Methods, where “HG” indicates Human Generated Datasets, “MC” indicates Model Constructed Datasets, and “CI” indicates Collection and Improvement of Existing Datasets. “IC” indicates Instruction Category}\label{tab6}
        \centering
        \resizebox{\textwidth}{!}{
            \begin{tabular}{llllll}
            \hline
            \textbf{Dataset} & \textbf{Language} & \textbf{CM} & \textbf{IC} & \textbf{Source} \\ \hline
            Alpaca\_data & EN & MC & Multi & Generated by Text-Davinci-003 with Aplaca\_data prompts \\ 
            Alpaca\_GPT4\_data & EN & CI \& MC & Multi & Generated by GPT-4 with Aplaca\_data prompts \\ 
            Alpaca\_GPT4\_data\_zh & ZH & CI \& MC & Multi & Generated by GPT-4 with Alpaca\_data prompts translated into Chinese by ChatGPT \\
            Aya Collection & Multi (114) & HG \& CI \& MC & Multi & Templated data, Translated data and Aya Dataset \\
            Aya Dataset & Multi (65) & HG & Multi & Manually collected and annotated via the Aya Annotation Platform  \\
            Bactrain-X & Multi (52) & CI \& MC & Multi & Generated by GPT-3.5-Turbo with Aplaca\_data and databricks-dolly-15K prompts translated into 51 languages by Google Translate API \\ 
            Baize & EN & CI \& MC & Multi & Sample seeds from specific datasets to create multi-turn dialogues using ChatGPT \\ 
            BELLE\_Generated\_Chat & ZH & MC & Generation & Generated by ChatGPT \\ 
            BELLE\_Multiturn\_Chat & ZH & MC & Multi & Generated by ChatGPT \\ 
            BELLE\_train\_0.5M\_CN & ZH & MC & Multi & Generated by Text-Davinci-003 \\ 
            BELLE\_train\_1M\_CN & ZH & MC & Multi & Generated by Text-Davinci-003 \\ 
            BELLE\_train\_2M\_CN & ZH & MC & Multi & Generated by ChatGPT \\ 
            BELLE\_train\_3.5M\_CN & ZH & MC & Multi & Generated by ChatGPT \\ 
            CAMEL & Multi \& PL & MC & Multi & Dialogue generated by two GPT-3.5-Turbo agents \\ 
            ChatGPT\_corpus & ZH & MC & Multi & Generated by GPT-3.5-Turbo \\ 
            COIG & ZH & HG \& CI \& MC & Multi & Translated instructions, Leetcode, Chinese exams, etc. \\ 
            CrossFit & EN & CI & Multi & Collection and improvement of various NLP datasets \\ 
            databricks-dolly-15K & EN & HG & Multi & Manually generated based on different instruction categories \\ 
            DialogStudio & EN & CI & Multi & Collection and improvement of various NLP datasets \\ 
            Dynosaur & EN & CI & Multi & Collection and improvement of various NLP datasets \\ 
            Firefly & ZH & HG \& CI & Multi & Collect Chinese NLP datasets and manually generate data related to Chinese culture \\ 
            Flan-mini & EN & CI & Multi & Collection and improvement of various instruction fine-tuning datasets \\ 
            Flan 2021 & Multi & CI & Multi & Collection and improvement of various NLP datasets \\ 
            Flan 2022 & Multi & CI & Multi & Collection and improvement of various instruction fine-tuning datasets \\ 
            GPT4All & EN & CI \& MC & Multi & Generated by GPT-3.5-Turbo with other datasets’ prompts \\ 
            GuanacoDataset & Multi & CI \& MC & Multi & Expand upon the initial 52K dataset from the Alpaca model \\ 
            HC3 & EN \& ZH & HG \& CI \& MC & Multi & Human-Q\&A pairs and ChatGPT-Q\&A pairs from Q\&A platforms, encyclopedias, etc. \\ 
            InstructDial & EN & CI & Multi & Collection and improvement of various NLP datasets \\ 
            InstructGPT-sft & EN & HG \& MC & Multi & Platform Q\&A data and manual labeling \\ 
            InstructionWild\_v1 & EN \& ZH & MC & Multi & Generated by OpenAI API \\ 
            InstructionWild\_v2 & EN \& ZH & HG & Multi & Collected on the web \\ 
            LaMini-LM & EN & CI \& MC & Multi & Generated by ChatGPT with synthetic and existing prompts \\ 
            LCCC & ZH & HG & Multi & Crawl user interactions on social media \\ 
            LIMA-sft & EN & HG \& CI & Multi & Manually select from various types of data \\ 
            LMSYS-Chat-1M & Multi & MC & Multi & Generated by multiple LLMs \\ 
            LogiCoT & EN \& ZH & CI \& MC & Reasoning & Expand the datasets using GPT-4 \\ 
            LongForm & EN & CI \& MC & Multi & Select documents from existing corpora and generating prompts for the documents using LLMs \\ 
            Luotuo-QA-B & EN \& ZH & CI \& MC & Multi & Use LLMs to generate Q\&A pairs on CSL, arXiv, and CNN-DM datasets \\ 
            MOSS\_002\_sft\_data & EN \& ZH & MC & Multi & Generated by Text-Davinci-003 \\ 
            MOSS\_003\_sft\_data & EN \& ZH & MC & Multi & Conversation data from MOSS-002 and generated by GPT-3.5-Turbo \\ 
            MOSS\_003\_sft\_plugin\_data & EN \& ZH & MC & Multi & Generated by plugins and LLMs \\ 
            NATURAL INSTRUCTIONS & EN & CI & Multi & Collection and improvement of various NLP datasets \\ 
            OASST1 & Multi (35) & HG & Multi & Generated and annotated by humans \\ 
            OIG & EN & CI & Multi & Collection and improvement of various datasets \\ 
            OL-CC & ZH & HG & Multi & Generated and annotated by humans \\ 
            OpenChat & EN & MC & Multi & ShareGPT \\
            OpenOrca & Multi & CI \& MC & Multi & Expand upon the Flan 2022 dataset using GPT-3.5-Turbo and GPT-4 \\ 
            Open-Platypus & EN & CI & Multi & Collection and improvement of various datasets \\ 
            OPT-IML Bench & Multi & CI & Multi & Collection and improvement of various NLP datasets \\ 
            Phoenix-sft-data-v1 & Multi & HG \& CI \& MC & Multi & Collected multi-lingual instructions, post-translated multi-lingual instructions, self-generated user-centered multi-lingual instructions \\ 
            PromptSource & EN & CI & Multi & Collection and improvement of various NLP datasets \\ 
            RedGPT-Dataset-V1-CN & ZH & MC & Multi & Generated by LLMs \\ 
            Self-Instruct & EN & MC & Multi & Generated by GPT-3 \\ 
            ShareChat & Multi & MC & Multi & ShareGPT \\ 
            ShareGPT-Chinese-English-90k & EN \& ZH & MC & Multi & ShareGPT \\ 
            ShareGPT90K & EN & MC & Multi & ShareGPT \\ 
            SUPER-NATURAL INSTRUCTIONS & Multi & CI & Multi & Collection and improvement of various NLP datasets \\ 
            TigerBot\_sft\_en & EN & HG \& CI \& MC & Multi & Self-instruct, human-labeling, open-source data cleaning \\ 
            TigerBot\_sft\_zh & ZH & HG \& CI \& MC & Multi & Self-instruct, human-labeling, open-source data cleaning \\ 
            T0 & EN & CI & Multi & Collection and improvement of various NLP datasets \\ 
            UltraChat & EN & MC & Multi & Dialogue generated by two ChatGPT agents \\ 
            UnifiedSKG & EN & CI & Multi & Collection and improvement of various NLP datasets \\ 
            Unnatural Instructions & EN & MC & Multi & Generated by LLMs \\ 
            WebGLM-QA & EN & MC & Open QA & Construct WebGLM-QA via LLM in-context bootstrapping \\ 
            Wizard\_evol\_instruct\_zh & ZH & CI \& MC & Multi & Generated by GPT with Wizard\_evol\_instruct prompts translated into Chinese \\ 
            Wizard\_evol\_instruct\_196K & EN & MC & Multi & Evolve instructions through the Evol-Instruct method \\ 
            Wizard\_evol\_instruct\_70K & EN & MC & Multi & Evolve instructions through the Evol-Instruct method \\ 
            xP3 & Multi (46) & CI & Multi & Collection and improvement of various NLP datasets \\ 
            Zhihu-KOL & ZH & HG & Multi & Crawl from Zhihu \\ \hline
            \end{tabular}
        }
\end{table*}

\subsubsection{Human Generated Datasets}\label{subsubsec321}

Human generated datasets involve manual creation and organization of all instructions by human annotators, following specified requirements and rules, without the assistance of existing LLMs. This type of datasets has evident advantages and disadvantages. Its advantages include:

\begin{itemize}

    \item \textbf{High Quality.} The datasets undergo processing and review by professional annotators, resulting in higher quality and cleanliness.
    
    \item \textbf{Interpretability.} After manual processing, the datasets are more easily interpretable and align well with human understanding.
    
    \item \textbf{Flexible Control.} Researchers have flexible control over training samples, allowing adjustments for different tasks.

\end{itemize}

\noindent Meanwhile, it also comes with corresponding drawbacks:

\begin{itemize}

    \item \textbf{High Cost and Low Efficiency.} Creating human generated datasets requires a substantial investment of manpower and time, making it less efficient compared to model constructed alternatives.
    
    \item \textbf{Subjectivity.} Human subjective judgment can introduce biases and inconsistencies into the datasets.

\end{itemize}

\noindent There are generally two ways to construct human generated datasets. The first way entails \textbf{direct creation of sets of instructional texts by company employees, volunteers, annotation platform personnel, etc., following given requirements and rules}. For instance, Databricks-dolly-15K \citep{bib60} is crafted by thousands of Databricks employees according to the instruction categories outlined in \citep{bib57}. Some instructions allow annotators to consult Wikipedia data as reference text. OASST1 \citep{bib61}, in contrast, is generated globally through crowdsourcing, with over 13.5K volunteers participating in the annotation process. OL-CC\footnote{\href{https://data.baai.ac.cn/details/OL-CC}{https://data.baai.ac.cn/details/OL-CC}} is the first open-source Chinese instruction dataset generated through crowdsourcing and manual efforts. On the open platform, 276 volunteers play the roles of both human users and AI assistants to create comprehensive text pairs. The Aya Dataset \citep{bib409}, as the largest manually annotated multilingual instruction dataset to date, is being collaboratively annotated by 2,997 contributors from 119 countries using the Aya Annotation Platform \citep{bib409}.

The second way entails \textbf{scraping human-generated real Q\&A data from webpages and standardizing them into instruction format}. The instructions in InstructionWild\_v2 \citep{bib62} are all collected from the web, covering social chat, code-related Q\&A, and more. LCCC \citep{bib63} is a Chinese conversation dataset primarily obtained by crawling user communication records on social media to capture authentic dialogues. Similarly, Zhihu-KOL\footnote{\href{https://github.com/wangrui6/Zhihu-KOL}{https://github.com/wangrui6/Zhihu-KOL}} is sourced from the well-known Chinese social media platform, Zhihu.

\subsubsection{Model Constructed Datasets}\label{subsubsec322}

The method of constructing the model involves leveraging a LLM, using various approaches to guide its generation of instructional data needed by humans. This approach has several advantages compared to human construction:

\begin{itemize}

    \item \textbf{Abundant Data.} LLMs can generate a vast amount of instructions, especially for content that occurs infrequently in real-world scenarios.
    
    \item \textbf{Cost-Effective and Efficient.} It reduces labor costs and time, enabling the acquisition of a large amount of data in a short period.

\end{itemize}

\noindent However, there are potential pitfalls in the content generated by the models, including:

\begin{itemize}

    \item \textbf{Variable Quality.} The quality of the generated content may not always be high. The model might produce hallucination, leading to inaccurate or inappropriate instructions. At the same time, the model itself may have inherent biases, and its output may not necessarily align with human values.
    
    \item \textbf{Post-Processing Required.} Generated samples need additional post-processing to ensure their quality and applicability before they can be used.
    
\end{itemize}

\noindent There are generally three methods for constructing datasets for model training. The first method involves \textbf{guiding a LLM to output instructions that meet expectations}. Typically, the LLM is given a certain identity (e.g., an expert question setter), along with requirements and examples for instruction generation. This allows the model to follow rules in answering questions or generating new instruction samples. Self-Instruct \citep{bib56} is a framework that sets initial instructions, automatically generates instruction samples, and iteratively filters them. The Self-Instruct dataset \citep{bib56} uses 175 manually written instructions as initial seeds and generates 52K instructions using this framework. Alpaca\_data \citep{bib64} improves on this framework, generating more diverse instruction data using the text-davinci-003. 

Other datasets, such as BELLE\_train\_0.5M\_CN \citep{bib58}, BELLE
\_train\_1M\_CN \citep{bib58}, InstructionWild\_v1 \citep{bib62}, and MOSS\_002\_sft\_data \citep{bib65}, also adopt this method for construction. Additionally, one can choose other well-performing models to build instruction datasets, like BELLE\_Generated\_Chat \citep{bib58}, BELLE\_Multiturn\_Chat \citep{bib58}, BELLE\_train\_2M\_CN \citep{bib58}, BELLE\_train\_3.5M\_CN \citep{bib58}, ChatGPT\_corpus\footnote{\href{https://github.com/PlexPt/chatgpt-corpus}{https://github.com/PlexPt/chatgpt-corpus}}, Unnatural Instructions \citep{bib73}, MOSS\_003\_sft\_plugin\_data \citep{bib65}, and others.

 To obtain higher-quality instructions, RedGPT-Dataset-V1-CN \citep{bib66} uses pre-existing LLMs to generate multi-turn dialogues. The pre-trained base model is fine-tuned, and the resulting RedGPT model \citep{bib66} is further used for instruction generation in an iterative manner to obtain a massive amount of high-quality data. WebGLM-QA \citep{bib67} generates data in three stages: Prompt Formulation, Instruction Inducting, and Few-shot In-context Learning. Wizard\_evol\_instruct\_196K \citep{bib68} and Wizard\_evol\_instruct\_70K \citep{bib68} use the Evol-Instruct method, subjecting 175 seed instructions to four evolution stages to enhance the complexity of generated instructions.

The second method involves \textbf{using real interactive conversations between humans and LLMs as instructional datasets}. ShareGPT\footnote{\href{https://sharegpt.com/}{https://sharegpt.com/}} can be used to share the dialogue outcomes between users and ChatGPT. ShareGPT90K\footnote{\href{https://huggingface.co/datasets/RyokoAI/ShareGPT52K}{https://huggingface.co/datasets/RyokoAI/ShareGPT52K}} and OpenChat \citep{bib61} have compiled tens of thousands of real conversations from ShareGPT. ShareGPT-Chinese-English-90k\footnote{\href{https://huggingface.co/datasets/shareAI/ShareGPT-Chinese-English-90k}{https://huggingface.co/datasets/shareAI/ShareGPT-Chinese-English-90k}} provides human-machine Q\&A data in parallel Chinese-English corpora. ShareChat\footnote{\href{https://paratranz.cn/projects/6725}{https://paratranz.cn/projects/6725}} translates all acquired ShareGPT data into Chinese. LMSYS-Chat-1M \citep{bib69} has gathered real conversation data from 25 LLMs between April and August 2023.

When constructing datasets, a combination of the first two methods can be employed. For instance, MOSS\_003\_sft\_data \citep{bib65} incorporates user data from MOSS-002 model \citep{bib65} and generated data from GPT-3.5-Turbo.

The third method involves \textbf{engaging in conversations using multiple LLM agents to obtain dialogue data}. CAMEL \citep{bib70} introduces a “role-playing” framework where LLMs generate metadata, creating 50 assistant roles and user roles for the “AI society.” UltraChat \citep{bib71} involves the interaction of multiple ChatGPT APIs in a dialogue. It employs an LSTM \citep{bib72} to process input and output for each round, simultaneously utilizing attention mechanisms to model contextual information.

\subsubsection{Collection and Improvement of Existing Datasets}\label{subsubsec323}

The Collection and Improvement of Existing Datasets is also a method for constructing instruction fine-tuning datasets. This method involves integrating and modifying several open-source datasets, ultimately consolidating them into a new dataset for LLM instruction fine-tuning. Such datasets can also be referred to as “Data Repositories.” It offers several advantages: 

\begin{itemize}

    \item \textbf{Diversity and Comprehensiveness.} The resulting datasets possess characteristics of rich data sources, diverse task types, and broad domain coverage.
    
    \item \textbf{Large Scale.} The more datasets selected, the larger the scale.
    
    \item \textbf{Time-saving.} It reduces the time required for dataset construction. 

\end{itemize}

\noindent However, it has its drawbacks: 

\begin{itemize}

    \item \textbf{Quality and Format Standardization.} It is necessary to comprehensively consider the quality of the source datasets and standardize the format of the data.
    
    \item \textbf{Dataset Licenses.} It is crucial to pay attention to the licenses of different source datasets to avoid privacy and regulatory issues.

\end{itemize}

\noindent A total of 16 datasets are compiled for this analysis. The source datasets for these “Data Repositories” primarily come from open-source traditional NLP datasets and other instruction fine-tuning datasets.

\textbf{CrossFit} \citep{bib74}. To investigate models’ few-shot learning capabilities across tasks, a collection of 269 NLP task datasets, known as CrossFit, has been assembled, covering 13 task types \citep{bib81}. In addition to being used for instruction fine-tuning, this dataset is employed for studying models’ cross-task generalization and transfer learning abilities.

\textbf{DialogStudio} \citep{bib75}. The DialogStudio dataset has gathered 87 open-source datasets, spanning six major task categories. The dataset integrates each sub-dataset while preserving the original information and is specifically designed for research on LLM instruction fine-tuning.

\textbf{Dynosaur} \citep{bib76}. The Dynosaur dataset is designed to study the dynamic expansion of instruction fine-tuning data. With a focus on minimizing maintenance costs, it incorporates approximately 802K data instances. During construction, metadata from existing NLP datasets is used to generate instructions for various NLP tasks, and the necessary data fields for building the dataset are identified. Furthermore, the dataset achieves dynamic growth by integrating new datasets from the Hugging Face\footnote{\href{https://huggingface.co/}{https://huggingface.co/}} data platform.

\textbf{Flan-mini} \citep{bib77}. The Flan-mini dataset is a subset selected from the Flan 2022 \citep{bib79}. It maintains a high level of task diversity while reducing the overall dataset size. The dataset includes specific tasks from the Flan 2022 and additional code-related datasets. Each instruction here has been processed, with the random addition of various prompt templates.

\textbf{Flan 2021} \citep{bib78}. The Flan 2021 dataset aggregates 62 existing NLP datasets, covering 12 tasks such as language understanding, generation, translation, and more. The collected datasets are predominantly in English.

\textbf{Flan 2022} \citep{bib79}. The Flan 2022 dataset consists of five parts, namely Flan 2021, T0 \citep{bib80}, SUPER-NATURAL INSTRUCTIONS \citep{bib81}, CoT datasets, and Dialog datasets. It encompasses as many as 1836 tasks. Each instruction provides four distinct instruction input templates, along with the incorporation of zero-shot, few-shot, CoT templates, as well as techniques like task mixing and input reversal.

\textbf{InstructDial} \citep{bib82}. The InstructDial dataset integrates 59 open-source dialogue datasets, covering 48 task types. Its goal is to enhance the models’ performance on dialogue-related tasks through instruction fine-tuning. Models fine-tuned on this dataset exhibit good performance in Out-of-Distribution (OOD) scenarios and few-shot learning.

\textbf{NATURAL INSTRUCTIONS} \citep{bib83}. The NATURAL INST-RUCTIONS dataset comprises 61 task datasets spanning 6 task types, totaling 193K instances. The dataset maps sub-datasets into a unified task pattern, exploring the cross-task generalization performance of models.

\textbf{OIG}\footnote{\href{https://huggingface.co/datasets/laion/OIG}{https://huggingface.co/datasets/laion/OIG}}. The OIG dataset, which stands for Open Instruction Generation, aims to create a collection that includes a large-scale of medium-quality instructions and a smaller scale of high-quality instructions. The dataset continues to incorporate new sub-datasets. As of February 2024, it contains 3.88M instructions, predominantly in English.

\textbf{Open-Platypus} \citep{bib84}. The Open-Platypus dataset aims to enhance the logical reasoning capabilities of models and is used to train the Platypus2 \citep{bib84}. By conducting keyword searches on other open-source datasets and using Sentence Transformers \citep{bib85}, questions with a similarity exceeding 80\% are filtered out. This process results in approximately 25K English instructions.

\textbf{OPT-IML Bench} \citep{bib86}. The OPT-IML Bench dataset comprises 2K NLP task datasets spanning 93 task types. The creators integrate and filter eight large data repositories, including the CrossFit, UnifiedSKG \citep{bib87}, PromptSource \citep{bib88}, and others. OPT-IML Bench is utilized to investigate the impact of a series of decisions in instruction fine-tuning on the downstream task performance.

\textbf{PromptSource} \citep{bib88}. The PromptSource dataset encompasses 176 NLP task datasets across 13 task types. Its strength lies in constructing a diverse set of prompts, offering ample resources for research areas such as instruction fine-tuning.

\textbf{SUPER-NATURAL INSTRUCTIONS} \citep{bib81}. The SUPER-NATURAL INSTRUCTIONS dataset comprises 1616 task datasets spanning 76 task types. It holds a linguistic advantage compared to other datasets, covering 55 languages. It is also suitable for studying the OOD capabilities of LLMs.

\textbf{T0} \citep{bib80}. The T0 dataset comprises 62 task datasets spanning 12 task types. Constructed by collecting NLP datasets and modifying prompts, it aims to test the zero-shot generalization capabilities of LLMs across many tasks.

\textbf{UnifiedSKG} \citep{bib87}. The UNIFIEDSKG framework proposed by \cite{bib87} integrates 21 structured knowledge grounding datasets into a text-to-text format, facilitating systematic SKG research. This dataset encompasses six task types, including semantic parsing and knowledge base Q\&A.

\textbf{xP3} \citep{bib89}. The xP3 dataset is a multilingual multitask dataset comprising 82 source datasets spanning 13 task types and 46 languages. The dataset is fine-tuned on multilingual pretrained models, resulting in variants of models such as BLOOMZ and mT0 \citep{bib89}. This exploration investigates performance on cross-lingual tasks.

\subsubsection{Datasets Created with Multiple Methods}\label{subsubsec324}

During the construction of certain general instruction fine-tuning datasets, multiple methods are concurrently employed to leverage the strengths of each, thereby enhancing the datasets’ qualities. The three methods are mentioned in previous sections, and through various combinations, four scenarios can be generated: \textbf{HG \& CI}, \textbf{HG \& MC}, \textbf{CI \& MC}, and \textbf{HG \& CI \& MC}. Here, “HG” stands for Human-Generated Datasets, “MC” for Model-Constructed Datasets, and “CI” for Collection and Improvement of Existing Datasets.

\textbf{HG \& CI.} \textbf{(1) While collecting data from other datasets, manual creation of data is concurrently undertaken to supplement missing task types}. Firefly \citep{bib59} gathers 23 common Chinese NLP tasks and constructs numerous tasks related to Chinese culture, such as couplets, poetry creation, and more. Each task is accompanied by manually written instruction templates to ensure high-quality and richness of the data. \textbf{(2) The collected data undergoes manual selection}. LIMA-sft \citep{bib90} includes 1330 instructions carefully chosen and prepared by human experts to validate the importance of high-quality instruction data.

\textbf{HG \& MC.} \textbf{Combine manually authored data with user-model dialogue data}. The InstructGPT-sft dataset \citep{bib57}, used in training the InstructGPT model \citep{bib57} by OpenAI, has two sources: one authored by annotators and the other consisting of instructions submitted via API to early models.

\textbf{CI \& MC.} \textbf{(1) Using other datasets as instruction inputs and selecting different models to generate responses}. Alpaca\_GPT4\_data \citep{bib91} employs instructions from the Alpaca\_data \citep{bib64} as input, generating responses using GPT-4 \citep{bib92}. Alpaca\_GPT4\_data\_zh \citep{bib91} and Wizard\_evol\_instruct\_zh dataset \citep{bib93} translate English instructions into Chinese before invoking models to generate Chinese responses. Bactrain-X \citep{bib94} utilizes a translation API to translate instruction inputs from the Alpaca\_data and databricks-dolly-15K into 51 languages, then inputs them into ChatGPT to obtain responses. GPT4All \citep{bib95} uses instructions from five public datasets as input and generates responses using GPT-3.5-Turbo. LogiCoT \citep{bib96} and OpenOrca \citep{bib97} follow similar methods. GuanacoDataset\footnote{\href{https://guanaco-model.github.io/}{https://guanaco-model.github.io/}} expands the language of instruction data from English to Chinese and Japanese. LaMini-LM \citep{bib98} uses the model to simultaneously generate synthetic instructions and responses corresponding to real instructions. These datasets reference existing instructions and are secondarily constructed with the assistance of models. \textbf{(2) Using open-source datasets as seed instructions to guide the generation of dialogues between models}. Baize \citep{bib99} samples “seeds” from specific datasets, allowing ChatGPT to engage in self-dialogue and batch generate high-quality multi-turn dialogue data. The dialogues cover both general and some vertical domains. \textbf{(3) Directly constructing input-output text pairs using existing data}. LongForm \citep{bib100} generates complete instructions for existing pre-trained corpus documents using LLMs, then expands them using structured corpus examples and task instances. Luotuo-QA-B \citep{bib101} instructs the model to generate five input-output text pairs for summaries or news content from three datasets.

\textbf{HG \& CI \& MC.} The six datasets combine the three construction methods mentioned in previous sections. The relevant information is as follows. \textbf{(1) COIG} \citep{bib102}. The COIG dataset consists of 191K Chinese instructions categorized into five types. Translated instructions are derived from open-source datasets, and the translation process involves three stages: automatic translation, manual verification, and manual correction. Exam instructions are primarily sourced from Chinese college entrance exams, high school entrance exams, and civil service exams. Human value alignment instructions consist of two series—one focusing on general human value alignment in Chinese regions, and the other on human value alignment specific to certain countries or regional cultures. Counterfactual correction multi-round chat are built based on the CN-DBpedia knowledge graph dataset \citep{bib103}, addressing hallucination issues in LLMs. Leetcode instructions gather programming-related prompts. \textbf{(2) HC3} \citep{bib104}. The HC3 dataset has both Chinese and English versions, totaling 37K Q\&A pairs. The dataset is designed to compare responses between human experts and ChatGPT across various domains. It can be used for research in areas such as instruction fine-tuning, human value alignment, model response characteristics, and more. \textbf{(3) Phoenix-sft-data-v1} \citep{bib105}. The 464K multilingual dialogue data in the Phoenix-sft-data-v1 dataset is primarily divided into two parts: single-turn instructions and multi-turn conversations. Single-turn instructions include Chinese and English instructions from Alpaca, translated multilingual instructions, and user-generated multilingual instructions. Multi-turn conversations mainly originate from ShareGPT and Discord\footnote{\href{https://discord.com/}{https://discord.com/}}. \textbf{(4) TigerBot\_sft\_en \& TigerBot\_sft\_zh} \citep{bib25}. These two datasets are fine-tuning data for the TigerBot \citep{bib25}, containing a large amount of collected open-source data and self-developed data. The construction of the dataset mainly follows five principles: annotating and summarizing 10 instruction categories and 120 sub-task types based on the distribution of instructions; generating instructions using the Self-Instruct method; organizing question and answer data based on manual question generation, web search, and other methods; converting and cleaning the format based on public datasets; the overall data distribution conforms to the natural distribution of instructions. \textbf{(5) Aya Collection} \citep{bib409}. The Aya Collection is a comprehensive and large corpus of datasets designed for training multilingual models, aimed at researchers worldwide. It comprises three primary sources of data: templated data, translated data, and the Aya Dataset \citep{bib409}. Templated data involves collaboration with fluent speakers to create templates for automatic dataset expansion into various languages. Translated data involves translating a subset of 19 datasets into 101 languages using the NLLB 3.3B machine translation model \citep{bib358}. The Aya Dataset is a human-annotated subset of the overall collection. 

\subsection{Domain-specific Instruction Fine-tuning Datasets}\label{subsec33}

\begin{figure}[h!]
\centering
\includegraphics[width=0.7\textwidth]{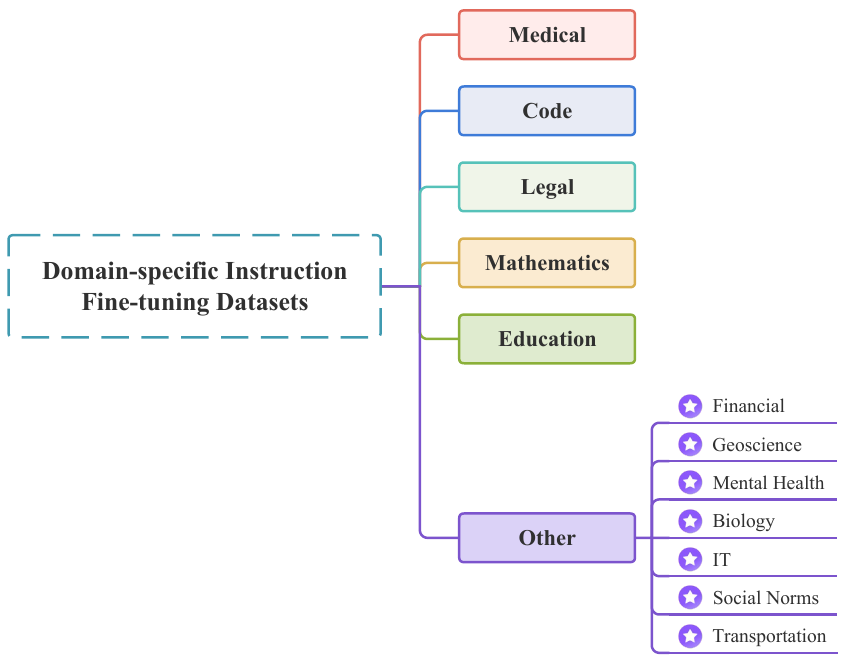}
\caption{Domain categories of the domain-specific instruction fine-tuning datasets}\label{fig13}
\end{figure}

The domain-specific instruction fine-tuning datasets are constructed for a particular domain by formulating instructions that encapsulate knowledge and task types closely related to that domain. After fine-tuning the pre-trained base model on the domain-specific instruction fine-tuning datasets, it can be applied to various scenario tasks within that domain, exhibiting outstanding performance. As shown in Figure~\ref{fig13}, the domain-specific instruction fine-tuning datasets are categorized into six major classes: medical, code, legal, mathematical, educational, and other domains. The collected and organized information from the domain-specific instruction fine-tuning datasets is presented in Table~\ref{tab7} and Table~\ref{tab8}.

\begin{table*}[h!]
        \captionsetup{singlelinecheck=off, justification=justified}
        \captionof{table}{Summary of \textbf{Domain-specific Instruction Fine-tuning Datasets} Information \textbf{Part I}. Release Time: “X” indicates unknown month. Public or Not: “All” indicates full open source; “Partial” indicates partially open source. “License” indicates the dataset follows a certain protocol. If the dataset is built upon other datasets, the licenses of the source datasets must also be adhered to}\label{tab7}
        \centering
        \resizebox{\textwidth}{!}{
            \begin{tabular}{llllll}
            \hline
            \textbf{Dataset} & \textbf{Publisher} & \textbf{Release Time} & \textbf{Size} & \textbf{Public or Not} & \textbf{License} \\ \hline
            BELLE\_School\_Math & BELLE & 2023-5 & 248481 instances & All & GPL-3.0 \\ 
            ChatDoctor & University of Texas Southwestern Medical Center et al. & 2023-3 & 115K instances & All & Apache-2.0 \\ 
            ChatMed\_Consult\_Dataset & michael-wzhu & 2023-5 & 549326 instances & All & CC-BY-NC-4.0 \\ 
            Child\_chat\_data & Harbin Institute of Technology et al. & 2023-8 & 5000 instances & All & - \\ 
            CMtMedQA & Zhengzhou University & 2023-8 & 68023 instances & All & MIT \\ 
            Code\_Alpaca\_20K & Sahil Chaudhary & 2023-3 & 20K instances & All & Apache-2.0 \\ 
            CodeContest & DeepMind & 2022-3 & 13610 instances & All & Apache-2.0 \\ 
            CommitPackFT & Bigcode & 2023-8 & 702062 instances & All & MIT \\ 
            DISC-Fin-SFT & Fudan University et al. & 2023-10 & 246K instances & Partial & Apache-2.0 \\ 
            DISC-Law-SFT & Fudan University et al. & 2023-9 & 403K instances & Partial & Apache-2.0 \\ 
            DISC-Med-SFT & Fudan University et al. & 2023-8 & 464898 instances & All & Apache-2.0 \\ 
            Educhat-sft-002-data-osm & East China Normal University et al. & 2023-7 & 4279419 instances & All & CC-BY-NC-4.0 \\ 
            GeoSignal & Shanghai Jiao Tong University et al. & 2023-6 & 22627272 instances & Partial & Apache-2.0 \\ 
            Goat & National University of Singapore & 2023-5 & 1746300 instances & All & Apache-2.0 \\ 
            HanFei 1.0 & Chinese Academy of Sciences et al. & 2023-5 & 255K instances & All & Apache-2.0 \\ 
            HuatuoGPT-sft-data-v1 & The Chinese University of Hong Kong et al. & 2023-5 & 226042 instances & All & Apache-2.0 \\ 
            Huatuo-26M & The Chinese University of Hong Kong et al. & 2023-5 & 26504088 instances & Partial & Apache-2.0 \\ 
            LawGPT\_zh & Shanghai Jiao Tong University & 2023-5 & 200K instances & Partial & - \\ 
            Lawyer LLaMA\_sft & Peking Universit & 2023-5 & 21476 instances & Partial & Apache-2.0 \\ 
            MeChat & Zhejiang University et al. & 2023-4 & 56K instances & All & CC0-1.0 \\ 
            MedDialog & UC San Diego & 2020-4 & 3.66M instances & All & - \\ 
            Medical Meadow & University Hospital Aachen et al. & 2023-4 & 160076 instances & All & GPL-3.0 \\ 
            Medical-sft & Ming Xu & 2023-5 & 2.07M instances & All & Apache-2.0 \\ 
            Mol-Instructions & Zhejiang University et al. & 2023-6 & 2043586 instances & All & CC-BY-4.0 \\ 
            MWP & Xihua University et al. & 2021-9 & 251598 instances & All & MIT \\ 
            OpenMathInstruct-1 & NVIDIA & 2024-2 & 1.8M instances & All & NVIDIA License \\
            Owl-Instruction & Beihang University et al. & 2023-9 & 17858 instances & All & - \\ 
            PROSOCIALDIALOG & Allenai & 2022-5 & 165681 instances & All & CC-BY-4.0 \\ 
            QiZhenGPT-sft-20k & Zhejiang University & 2023-5 & 20K instances & Partial & GPL-3.0 \\ 
            ShenNong\_TCM\_Dataset & michael-wzhu & 2023-6 & 112565 instances & All & Apache-2.0 \\ 
            TaoLi\_data & Beijing Language and Culture University et al. & 2023-X & 88080 instances & All & Apache-2.0 \\ 
            ToolAlpaca & Chinese Information Processing Laboratory et al. & 2023-6 & 3928 instances & All & Apache-2.0 \\ 
            ToolBench & Tsinghua University et al. & 2023-7 & 126486 instances & All & Apache-2.0 \\ 
            TransGPT-sft & Beijing Jiaotong University & 2023-7 & 58057 instances & All & Apache-2.0 \\ \hline
            \end{tabular}
        }
\end{table*}

\begin{table*}[h!]
        \captionsetup{singlelinecheck=off, justification=justified}
        \captionof{table}{Summary of \textbf{Domain-specific Instruction Fine-tuning Datasets} Information \textbf{Part II}. Language: “EN” indicates English, “ZH” indicates Chinese, “PL” indicates Programming Language, and the number in parentheses indicates the number of programming languages included. “CM” indicates Construction Methods, where “HG” indicates Human Generated Datasets, “MC” indicates Model Constructed Datasets, and “CI” indicates Collection and Improvement of Existing Datasets. “IC” indicates Instruction Category}\label{tab8}
        \centering
        \resizebox{\textwidth}{!}{
            \begin{tabular}{llllll}
            \hline
            \textbf{Dataset} & \textbf{Language} & \textbf{CM} & \textbf{Domain} & \textbf{IC} & \textbf{Source} \\ \hline
            BELLE\_School\_Math & ZH & MC & Math & Math & Generated by ChatGPT \\ 
            ChatDoctor & EN & HG \& MC & Medical & Multi & Real conversations between doctors and patients \& Generated by ChatGPT \\ 
            ChatMed\_Consult\_Dataset & ZH & MC & Medical & Multi & Generated by GPT-3.5-Turbo \\ 
            Child\_chat\_data & ZH & HG \& MC & Education & Multi & Real conversations \& Generated by GPT-3.5-Turbo \\ 
            CMtMedQA & ZH & HG & Medical & Multi & Real conversations between doctors and patients \\ 
            Code\_Alpaca\_20K & EN \& PL & MC & Code & Code & Generated by Text-Davinci-003 \\ 
            CodeContest & EN \& PL & CI & Code & Code & Collection and improvement of various datasets \\ 
            CommitPackFT & EN \& PL (277) & HG & Code & Code & GitHub Action dump \\ 
            DISC-Fin-SFT & ZH & HG \& CI \& MC & Financial & Multi & Open source datasets \& Manually collect financial data \& ChatGPT assistance \\ 
            DISC-Law-SFT & ZH & HG \& CI \& MC & Law & Multi & Open source datasets \& Legal-related Text Content \& Generated by GPT-3.5-Turbo \\ 
            DISC-Med-SFT & ZH & HG \& CI & Medical & Multi & Open source datasets \& Manually selected data \\ 
            Educhat-sft-002-data-osm & EN \& ZH & CI & Education & Multi & Collection and improvement of various datasets \\ 
            GeoSignal & EN & HG \& CI \& MC & Geoscience & Multi & Open source datasets \& Geoscience-related Text Content \& Generated by GPT-4 \\ 
            Goat & EN & HG & Math & Math & Artificially synthesized data \\ 
            HanFei 1.0 & ZH & - & Law & Multi & Filter legal-related data according to rules \\ 
            HuatuoGPT-sft-data-v1 & ZH & HG \& MC & Medical & Multi & Real conversations between doctors and patients \& Generated by ChatGPT \\ 
            Huatuo-26M & ZH & CI & Medical & Multi & Collection and improvement of various datasets \\ 
            LawGPT\_zh & ZH & CI \& MC & Law & Multi & Real conversations \& Generated by ChatGPT \\ 
            Lawyer LLaMA\_sft & ZH & CI \& MC & Law & Multi & Generated by ChatGPT with other datasets’ prompts \\ 
            MeChat & ZH & CI \& MC & Mental Health & Multi & Based on PsyQA dataset with the proposed SMILE method \\ 
            MedDialog & EN \& ZH & HG & Medical & Multi & Real conversations between doctors and patients \\ 
            Medical Meadow & EN & HG \& CI & Medical & Multi & Crawl data from the Internet \& Collection and improvement of various NLP datasets \\ 
            Medical-sft & EN \& ZH & CI & Medical & Multi & Collection and improvement of various NLP datasets \\ 
            Mol-Instructions & EN & HG \& CI \& MC & Biology & Multi & Molecule-oriented, Protein-oriented, Biomolecular text instructions \\ 
            MWP & EN \& ZH & CI & Math & Math & Collection and improvement of various datasets \\
            OpenMathInstruct-1 & EN & CI \& MC & Math & Math & GSM8K and MATH datasets (original questions); Generated using Mixtral-8×7B model \\
            Owl-Instruction & EN \& ZH & HG \& MC & IT & Multi & Generated by GPT-4 \& Manual verification \\ 
            PROSOCIALDIALOG & EN & HG \& MC & Social Norms & Social Norms & Generated by humans with GPT-3 created prompts \\ 
            QiZhenGPT-sft-20k & ZH & CI & Medical & Multi & Collection and improvement of various datasets \\ 
            ShenNong\_TCM\_Dataset & ZH & MC & Medical & Multi & Generated by ChatGPT \\ 
            TaoLi\_data & ZH & HG \& CI & Education & Multi & Collection and improvement of various datasets \& Manually extract dictionary data \\ 
            ToolAlpaca & EN \& PL & HG \& MC & Code & Code & Manually filter APIs \& Generated by ChatGPT \\ 
            ToolBench & EN \& PL & HG \& MC & Code & Code & Manually filter APIs \& Generated by ChatGPT \\ 
            TransGPT-sft & ZH & HG & Transportation & Multi & Manually collect traffic-related data \\ \hline
            \end{tabular}
        }
\end{table*}

\subsubsection{Medical Domain}\label{subsubsec331}

Currently, there are numerous open-source large-scale models for medical tasks in both Chinese and English. All of them have constructed instruction fine-tuning datasets in the medical domain for supervised fine-tuning, demonstrating excellent generalization capabilities. In some cases, the performance even close to that of professional doctors in specific scenarios. CMtMedQA \citep{bib106} and MedDialog \citep{bib107} exclusively utilize authentic doctor-patient multi-turn dialogues, where all instructions belong to real-world scenario data. In contrast, ChatMed\_Consult\_Dataset \citep{bib108} and ShenNong\_TCM\_Dataset \citep{bib109} adopt the Self-Instruct method, utilizing the model to generate medical Q\&A data. The former focuses on medical consultations, while the latter concentrates on traditional Chinese medicine knowledge Q\&A. 

Some datasets are collected and curated from open-source data such as knowledge bases and forums. For instance, Huatuo-26M \citep{bib110} has multiple sources, including medical encyclopedia Q\&A, medical knowledge graphs, and doctor-patient Q\&A. QiZhenGPT-sft-20k\footnote{\href{https://github.com/CMKRG/QiZhenGPT}{https://github.com/CMKRG/QiZhenGPT}} formulates instructions based on the content collected from the Qizhen medical knowledge base. Medical-sft\footnote{\href{https://github.com/shibing624/MedicalGPT}{https://github.com/shibing624/MedicalGPT}} merges several Chinese and English medical datasets, including the ChatDoctor \citep{bib111} and QiZhenGPT-sft-20k, among others.

In addition to the aforementioned, some datasets may comprise a combination of real and synthetic data or involve manual curation based on existing datasets. ChatDoctor and HuatuoGPT-sft-data-v1 \citep{bib112}, while collecting authentic doctor-patient dialogues, incorporate conversation data generated by ChatGPT and information from a disease database. DISC-Med-SFT \citep{bib113} and Medical Meadow \citep{bib114} meticulously select several data sources, undergoing a certain degree of reconstruction to enhance the overall quality of the datasets.

\subsubsection{Code Domain}\label{subsubsec332}

The purpose of the code instruction fine-tuning datasets is to enhance the capabilities of LLMs in tasks such as code generation and tool invocation. Some datasets focus on instructions tailored for code generation tasks. CommitPackFT \citep{bib210} extracts code files covering 350 programming languages, rigorously filtering and retaining code instruction data for 277 programming languages. Code\_Alpaca\_20K \citep{bib116} follows the construction method of the Alpaca\_data \citep{bib64}, generating 20K instructions for fine-tuning the Code Alpaca model \citep{bib116}. CodeContest \citep{bib117} merges data collected from Codeforces\footnote{\href{https://codeforces.com/blog/entry/89502}{https://codeforces.com/blog/entry/89502}}, Description2Code \citep{bib118}, and CodeNet \citep{bib119}. In addition, some datasets emphasize instructions for tool invocation tasks. ToolAlpaca \citep{bib120} creates a highly diverse tool usage dataset through the construction of a multi-agent simulation environment, fine-tuning the model with 3,928 instances of tool usage. The construction of the ToolBench \citep{bib121} involves three stages: API collection, instruction generation, and solution path annotation, aiming to fine-tune the model for tool usage instructions.

\subsubsection{Legal Domain}\label{subsubsec333}

Various LLMs in the legal domain have been introduced, but there is a relatively limited availability of open-source legal instruction datasets. Here, we compile information on four partially or fully open-source legal instruction datasets that can be utilized to enhance model capabilities in tasks such as legal Q\&A, judgment prediction, and case classification. DISC-Law-SFT \citep{bib122} is divided into two sub-datasets, each introducing legal reasoning abilities and the utilization of external knowledge to the model. Han Fei 1.0 \citep{bib123} merges general instructions with legal instructions, aiming to equip the model with legal knowledge while retaining its general capabilities. LawGPT\_zh \citep{bib124} includes scenario-based Q\&A with legal basis and single-turn legal Q\&A obtained through model cleaning. Lawyer LLaMA\_sft \citep{bib125} involves model-generated Chinese judicial exam Q\&A, legal consultation responses, and multi-turn dialogue data.

\subsubsection{Mathematics Domain}\label{subsubsec334}

The performance and future potential of LLMs in the field of mathematics have always been a focal point of attention. Mathematical problems assess various skills such as computation, reasoning, spatial thinking, making them inherently challenging. This often results in model performance on mathematical problems falling below expectations. Consequently, one common approach to improving models’ mathematical abilities is to perform supervised fine-tune using effective mathematical instruction datasets.

BELL\_School\_Math \citep{bib58} generates Chinese mathematical problems, including the solution process, through the model. However, the overall difficulty is low, and the answers have not undergone rigorous verification, potentially containing errors. Goat \citep{bib126} consists entirely of artificially synthesized data for arithmetic tasks, covering addition, subtraction, multiplication, and division operations, with difficulty levels not posing significant challenges for humans. MWP \citep{bib127} unifies eight mathematics-related NLP datasets into instruction format, offering both single-equation and multiple-equation forms. 
OpenMathInstruct-1 \citep{bib411} leverages the Mixtral-8x7B model \citep{bib412} to reason over questions from the GSM8K \citep{bib331} and MATH \citep{bib334} datasets, generating a plethora of question-solution text pairs. It significantly enhances the models’ mathematical capabilities.

Currently, there is a scarcity of high-difficulty mathematical instruction datasets, limited by factors such as high entry barriers, complex symbols, high costs, and non-open sourcing.

\subsubsection{Education Domain}\label{subsubsec335}

LLMs in the education domain focus on course guidance, emotional support, child companionship, knowledge learning, and other aspects, serving teachers, students, and parents. Their goal is to become new tools applied in the education industry. LLMs in the education domain undergo fine-tuning using specifically collected education-related instructions. Child\_chat\_data\footnote{\href{https://github.com/HIT-SCIR-SC/QiaoBan}{https://github.com/HIT-SCIR-SC/QiaoBan}} primarily revolves around the theme of emotional companionship for children, containing both real and synthetic Chinese dialogue data related to emotional companionship for children. Educhat-sft-002-data-osm \citep{bib128} is used for the development of the EduChat project and combines multiple Chinese and English educational instructions and dialogue data. It is used to train models that can provide open-ended questioning, emotional support, essay correction, and other functions in an educational setting. TaoLi\_data \citep{bib129} is constructed based on internationally circulated Chinese teaching materials, Hanyu Shuiping Kaoshi (HSK) exams\footnote{\url{https://www.chinesetest.cn/}}, Chinese dictionaries, and other resources. It includes various forms of instructions to enable the model to acquire knowledge related to international Chinese education.

\subsubsection{Other Domains}\label{subsubsec336}

Currently, other domain-specific fine-tuning datasets are gradually being open-sourced. The seven domains mentioned belows, although having fewer open resources for fine-tuning instructions, still hold significant meaning and value.

\textbf{Financial Domain.} DISC-Fin-SFT \citep{bib130} is a high-quality Chinese financial dataset. It is utilized for LoRA \citep{bib131} instruction fine-tuning on the Baichuan-13B-Chat model, ultimately resulting in the financial LLM DISC-FinLLM \citep{bib130}. The dataset comprises 246K instructions categorized into four subtypes: financial consultation, financial tasks, financial calculations, and retrieval enhancement. Sourced diversely from financial NLP datasets, manually curated Q\&A pairs, and model-generated dialogues, a portion of this dataset is currently open-sourced.

\textbf{Geoscience Domain.} GeoSignal \citep{bib132} is being used for the fine-tuning of instructions for K2 \citep{bib132}, the first LLM in the field of geoscience. The creators have collected extensive data from various databases and websites in the earth science domain. They have restructured this data into a unified sequence format suitable for tasks such as interpretation, named entity recognition, reasoning, text classification, and Q\&A. The original dataset size is 22.6M instances, but after cleaning, 40K data instances have been retained. A complete version is planned for future release.

\textbf{Mental Health Domain.} MeChat \citep{bib133} is Chinese psychological health dialogue data. Builders transform real psychological mutual assistance Q\&A into multi-turn dialogues using models. The dataset comprises 56K instructions, catering to extended conversational scenarios.

\textbf{Biology Domain.} Mol-Instructions \citep{bib134} consists of three main components: Molecule-oriented instructions, Protein-oriented instructions, and Biomolecular text instructions. Each part focuses on chemical reactions and molecular design, protein prediction, and bioinformatics in biochemistry, respectively. The dataset’s construction involves a combination of human-machine collaboration, database resource processing, and the transformation of biological data.

\textbf{IT Domain.} Owl-Instruction \citep{bib135} is utilized for the instruction fine-tuning of the Owl model \citep{bib135}. The instructions are specifically designed for handling IT-related tasks such as troubleshooting, log analysis, etc. The dataset construction involves four stages: data generation, GPT-4 filtering, manual verification, and supervised fine-tuning. It comprises 18K single-turn and multi-turn instructions.

\textbf{Social Norms Domain.} PROSOCIALDIALOG \citep{bib136} is a multi-turn English conversation dataset that instructs models to respond to problematic inputs according to human social norms. The dataset covers various unethical, problematic, biased, and harmful scenarios, created using a human-machine collaboration framework.

\textbf{Transportation Domain.} TransGPT-sft \citep{bib46} serves as the fine-tuning component for China’s pioneering open-source TransGPT traffic model \citep{bib46}. Adopting a dialogue-centric methodology, the dataset involves extracting content from documents in formats like PDFs and Doc files. LLMs are then employed to generate dialogues related to traffic based on the document content.

\subsection{Distribution Statistics of Instruction Fine-tuning Datasets}\label{subsec34}

Figure~\ref{fig14} provides statistics on 103 instruction fine-tuning datasets from six aspects: release time, license, data category, construction method, language, and domain. The following conclusion can be drawn:

\begin{figure}
\centering
\includegraphics[width=0.9\textwidth]{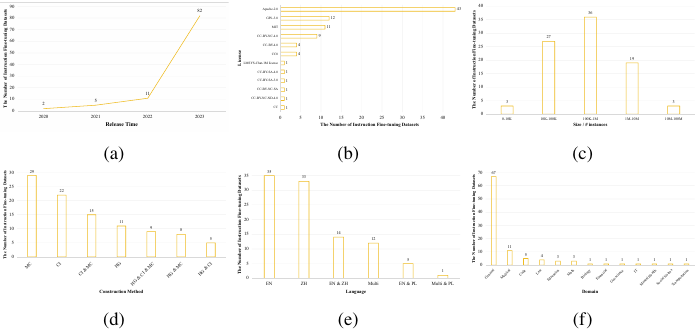}
\caption{Statistics distribution of instruction fine-tuning datasets. (a) illustrates the quantity trend over time. (b) depicts the quantity distribution under different licenses, considering only the datasets with listed licenses. (c) shows the quantity distribution across different data scales. (d) displays the quantity distribution for different construction methods. (e) represents the quantity distribution across different languages. (f) illustrates the quantity distribution across different domains. Zoom in for better view}\label{fig14}
\end{figure}

(1) The number of instruction fine-tuning datasets is showing a growing trend. The widespread attention to LLMs and the application of the instruction fine-tuning paradigm have greatly facilitated the construction and open-sourcing of instruction fine-tuning datasets. The demand for model fine-tuning and research interest in this area are rapidly expanding.

(2) Data licenses to some extent reflect the openness and accessibility of datasets. For instruction fine-tuning datasets, the Apache-2.0 license is the most commonly used, covering 43 datasets, followed by the GPL-3.0 license and the MIT license. This reflects the developers’ inclination towards open and shared data.

(3) The majority of instruction fine-tuning datasets are concentrated in the range of 10K to 1M, totaling 63 datasets. This indicates that, in practical applications, datasets of this scale are sufficient to meet the demand. However, there are relatively fewer small-scale and large-scale datasets, reflecting the challenges and scarcity at both extremes. Small-scale datasets emphasize quality but may lack category richness, while large-scale datasets offer diversity but may be constrained by computational resources and affected by data redundancy.

(4) The “utilizing model-constructed instructions” method is the most prevalent in constructing datasets, highlighting its potential in dataset creation. The quality of such datasets relies primarily on the models’ performance and the guidance provided during construction. The second most common method is “curating existing datasets and improving them,” indicating the active utilization of open-source data. The number of datasets manually generated is comparatively lower due to efficiency and cost considerations. There are 22 datasets that employ combinations of different methods to further enhance dataset quality, suggesting that this approach may become more mainstream in the future.

(5) Chinese and English instruction datasets hold a crucial position in research, garnering greater attention. Mixed Chinese and English, as well as multilingual datasets, show a considerable quantity, indicating that cross-language research is becoming a focus. There is a scarcity of open-source instruction datasets related to programming languages, primarily tailored for specific application scenarios.

(6) The number of general-domain datasets is 67, aligning with the widespread demand for instruction fine-tuning techniques in various application scenarios. Research and construction of instruction datasets for relevant LLMs have also been conducted in common fields such as healthcare, programming, law, etc. There are datasets available in other domains as well, indicating the potential applications of LLMs in diverse disciplines and industries. However, there are still instruction datasets for niche fields awaiting further research and exploration.

\section{Preference Datasets}\label{sec4}

\begin{figure}[h!]
\centering
\includegraphics[width=0.9\textwidth]{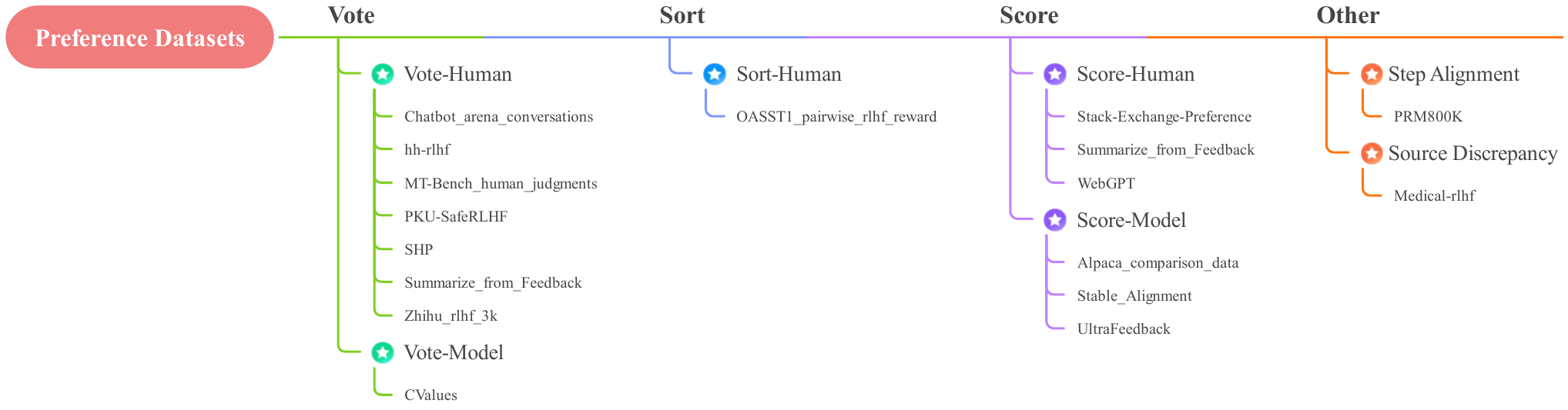}
\caption{Different preference datasets corresponding to various preference evaluation methods}\label{fig15}
\end{figure}

Preference datasets are collections of instructions that provide preference evaluations for multiple responses to the same instruction input. Typically, they consist of pairs of instructions with different responses, along with feedback from humans or other models. This setup reflects the relative preferences of humans or models for different responses within a given task or context. The feedback information in preference datasets is often manifested through voting, sorting, scoring, or other forms of comparison. Figure~\ref{fig15} categorizes various preference datasets based on the methods used for preference evaluation. The collected and organized information on preference datasets is presented in Table~\ref{tab9} and Table~\ref{tab10}.

Preference datasets are primarily utilized during the alignment phase of large models, aiming to assist in aligning the models’ outputs more closely with human preferences and expectations. The alignment with human preferences is manifested in three main aspects: \textbf{utility}, possessing the ability to follow instructions; \textbf{honesty}, avoiding fabrications; and \textbf{safety}, refraining from generating illegal or harmful information \citep{bib7}. Both RLHF \citep{bib139,bib140} and RLAIF (Reinforcement Learning from AI Feedback) \citep{bib142} employ reinforcement learning methods to optimize models using feedback signals. In addition to fine-tuning with instruction datasets, it is also possible to train reward models with preference datasets. Subsequently, the Proximal Policy Optimization (PPO) algorithm can be applied for further fine-tuning based on the feedback from the reward models \citep{bib141}.

\vspace{\baselineskip}

\begin{table*}[h!]
        \captionsetup{singlelinecheck=off, justification=justified}
        \captionof{table}{Summary of \textbf{Preference Datasets} Information \textbf{Part I}. Public or Not: “All” indicates full open source; “Partial” indicates partially open source. “License” indicates the dataset follows a certain protocol. If the dataset is built upon other datasets, the licenses of the source datasets must also be adhered to}\label{tab9}
        \centering
        \resizebox{\textwidth}{!}{
            \begin{tabular}{llllll}
            \hline
            \textbf{Dataset} & \textbf{Publisher} & \textbf{Release Time} & \textbf{Size} & \textbf{Public or Not} & \textbf{License} \\ \hline
            Alpaca\_comparison\_data & Stanford Alpaca & 2023-3 & 51K instances & All & Apache-2.0 \\ 
            Chatbot\_arena\_conversations & UC Berkeley et al. & 2023-6 & 33000 instances & All & CC-BY-4.0 \& CC-BY-NC-4.0 \\ 
            CValues & Alibaba Group & 2023-7 & 145K instances & All & Apache-2.0 \\ 
            hh-rlhf & Anthropic & 2022-4 & 169352 instances & All & MIT \\ 
            Medical-rlhf & Ming Xu & 2023-5 & 4K instances & All & Apache-2.0 \\ 
            MT-Bench\_human\_judgments & UC Berkeley et al. & 2023-6 & 3.3K instances & All & CC-BY-4.0 \\ 
            OASST1\_pairwise\_rlhf\_reward & Tasksource & 2023-5 & 18918 instances & All & Apache-2.0 \\ 
            PKU-SafeRLHF & Peking University & 2023-7 & 361903 instances & Partial & CC-BY-NC-4.0 \\ 
            PRM800K & OpenAI & 2023-5 & 800K instances & All & MIT \\ 
            SHP & Stanford University & 2021-10 & 385563 instances & All & - \\ 
            Stable\_Alignment & Dartmouth College et al. & 2023-5 & 169K instances & All & Apache-2.0 \\ 
            Stack-Exchange-Preferences & Anthropic & 2021-12 & 10807695 instances & All & CC-BY-SA-4.0 \\ 
            Summarize\_from\_Feedback & OpenAI & 2020-9 & 193841 instances & All & - \\ 
            UltraFeedback & Tsinghua University et al. & 2023-10 & 63967 instances & All & MIT \\ 
            WebGPT & OpenAI & 2021-12 & 19578 instances & All & - \\ 
            Zhihu\_rlhf\_3k & Liyucheng & 2023-4 & 3460 instances & All & CC-BY-2.0 \\ \hline
            \end{tabular}
        }
\end{table*}

\begin{table*}
        \captionsetup{singlelinecheck=off, justification=justified}
        \captionof{table}{Summary of \textbf{Preference Datasets} Information \textbf{Part II}. Language: “EN” indicates English, “ZH” indicates Chinese, “Multi” indicates Multilingual. “CM” indicates Construction Methods, where “HG” indicates Human Generated Datasets, “MC” indicates Model Constructed Datasets, and “CI” indicates Collection and Improvement of Existing Datasets. “IC” indicates Instruction Category. “PEM” indicates Preference Evaluation Method, where “VO” indicates Vote, “SO” indicates Sort, “SC” indicates Score, “-H” indicates Conducted by Humans, “-M” indicates Conducted by Models}\label{tab10}
        \centering
        \resizebox{\textwidth}{!}{
            \begin{tabular}{lllllll}
            \hline
                \textbf{Dataset} & \textbf{Language} & \textbf{CM} & \textbf{Domain} & \textbf{IC} & \textbf{PEM} & \textbf{Source} \\ \hline
                Alpaca\_comparison\_data & EN & MC & General & Multi & SC-M & Generated by three LLMs \& GPT-4 scoring \\ 
                Chatbot\_arena\_conversations & Multi & HG \& MC & General & Multi & VO-H & Generated by twenty LLMs \& Manual judgment \\ 
                CValues & ZH & MC & Social Norms & Social Norms & VO-M & Generated by LLMs \& Evaluation by the reward model \\ 
                hh-rlhf & EN & HG \& MC & General & Multi & VO-H & Generated by LLMs \& Manual judgment \\ 
                Medical-rlhf & ZH & CI \& MC & Medical & Multi & Other & Response\_chosen comes from the doctor's response \& Response\_rejected comes from the model's response \\ 
                MT-Bench\_human\_judgments & EN & HG \& MC & General & Multi & VO-H & Generated by LLMs \& Manual judgment \\ 
                OASST1\_pairwise\_rlhf\_reward & Multi & CI & General & Multi & SO-H & OASST1 datasets \& Manual sorting \\ 
                PKU-SafeRLHF & EN & HG \& CI \& MC & Social Norms & Social Norms & VO-H & Generated by LLMs \& Manual judgment \\ 
                PRM800K & EN & HG \& CI \& MC & Math & Math & Other & Generated by LLMs \& Mathematical reasoning steps are determined manually \\ 
                SHP & EN & HG & General & Multi & VO-H & Reddit data \& Manual judgment \\ 
                Stable\_Alignment & EN & MC & General & Multi & SC-M & Generated by LLMs \& Model scoring \\ 
                Stack-Exchange-Preferences & EN & HG & General & Multi & SC-H & Stackexchange data \& Manual scoring \\ 
                Summarize\_from\_Feedback & EN & HG \& CI & News & Multi & VO-H \& SC-H & Open source datasets \& Manual judgment and scoring \\ 
                UltraFeedback & EN & CI \& MC & General & Multi & SC-M & Generated by seventeen LLMs \& Model scoring \\ 
                WebGPT & EN & HG \& CI & General & Multi & SC-H & Open source datasets \& Manual scoring \\ 
                Zhihu\_rlhf\_3k & ZH & HG & General & Multi & VO-H & Zhihu data \& Manual judgment \\ \hline
            \end{tabular}
        }
\end{table*}

\subsection{Preference Evaluation Methods}\label{subsec41}

\begin{figure}[h!]
\centering
\includegraphics[width=0.8\textwidth]{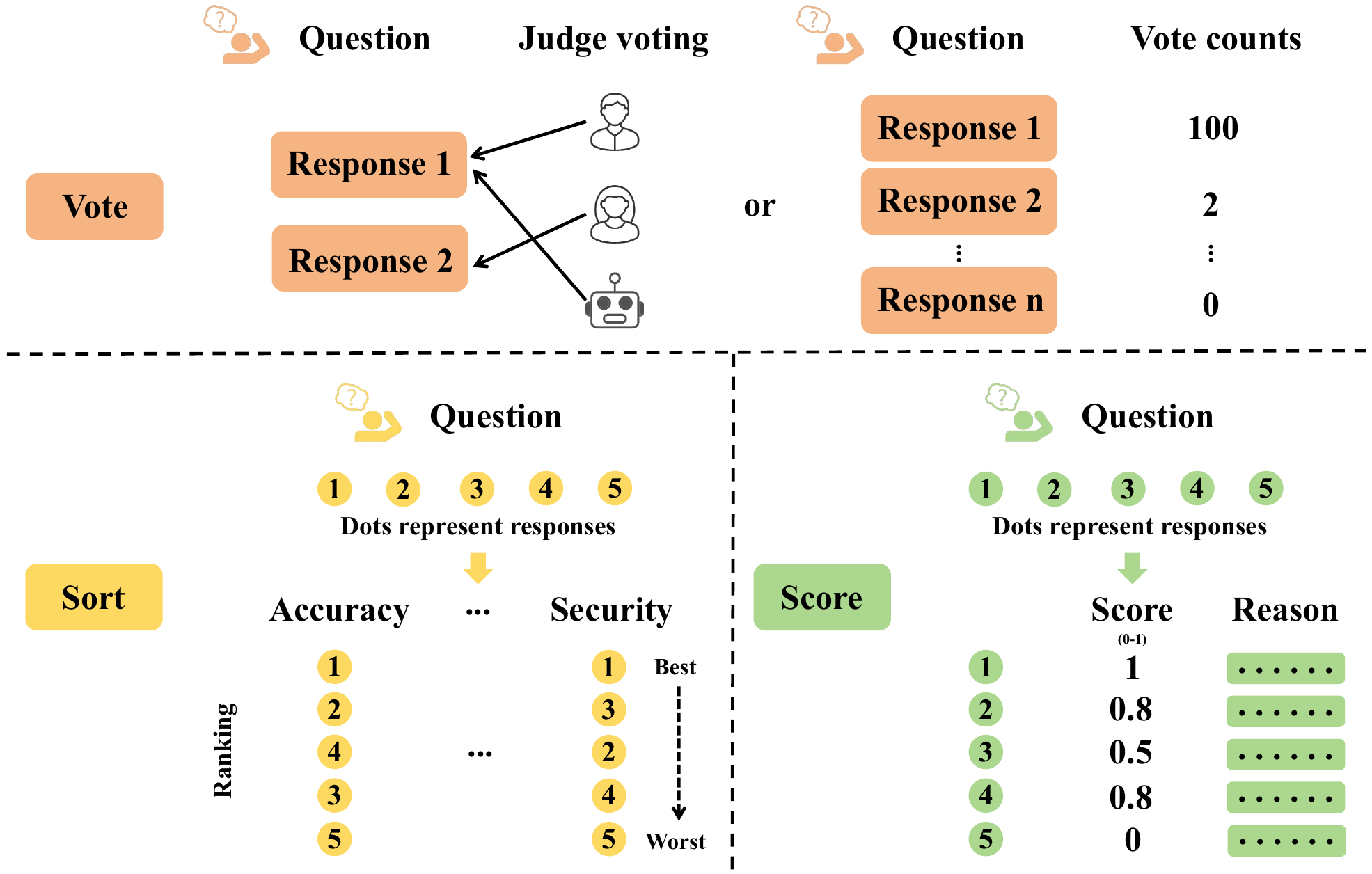}
\caption{Different preference evaluation methods}\label{fig16}
\end{figure}

The preference evaluation methods for preference datasets can be categorized into \textbf{voting}, \textbf{sorting}, \textbf{scoring}, and \textbf{other methods}. Each method can be conducted by humans or aligned high-quality LLMs. Human feedback provides preferences that are more aligned with real-world scenarios, capturing intuitive human cognition and language understanding. However, it may exhibit subjectivity and inconsistencies due to individual differences, requiring more time and cost for annotation. Model feedback can leverage learned human preference information and extensive knowledge from high-quality models, saving annotation time and cost. However, it may be influenced by inherent biases in the model, and the feedback information may be less authentic compared to human feedback. In general, a comprehensive approach that combines various forms and sources of preference data may be more advantageous. Figure~\ref{fig16} visually presents various preference evaluation methods.

\subsubsection{Vote}\label{subsubsec411}

The voting method typically involves selecting the better option from two answers or choosing several preferred options from multiple answers. The advantage is its simplicity and intuitiveness, making it easy to collect and reflecting the opinions of the group. However, the drawback is the lack of granularity in information.

Datasets using the “human vote” method are as follows. Chatbot\_arena\_con-versations \citep{bib143} includes examples with answers from two models to the same question and the selection made by a human judge. It comprises outputs from a total of 20 models in 96 languages. The dataset also annotates unsafe conversations for related research. The hh-rlhf dataset \citep{bib144,bib145} includes instances with accepted and rejected answers, where crowdworkers instruct the model to perform a task and choose the more useful and honest answer from two options. MT-Bench\_human\_judgments \citep{bib143} involves graduate students comparing pairwise preferences for 80 instructions generated by six models. PKU-SafeRLHF \citep{bib146} focuses on comparing performance and safety preferences. After evaluating the harmlessness of instructions, choices are made based on usefulness and harmlessness in the Q\&A format. Each entry in the final dataset includes two answers and feedback information. SHP \citep{bib147} is crawled from Reddit. Each post contains a question and a pair of answers, with one answer being more favored by Reddit users, constructing a preference dataset reflecting human preferences. Similarly, Zhihu\_rlhf\_3k\footnote{\href{https://huggingface.co/datasets/liyucheng/zhihu_rlhf_3k}{https://huggingface.co/datasets/liyucheng/zhihu\_rlhf\_3k}} is built in the same way using the Zhihu. Summarize\_from\_Feedback \citep{bib148} is primarily constructed to optimize summarization models. The dataset consists of two parts: one where annotators choose the better of two summaries, and the other where summaries are rated using a Likert scale. The dataset uses both human voting and human scoring.

A representative dataset for the “model vote” method is CValues \citep{bib149}. The CValues dataset encompasses three types of responses: safe and responsibility, safe, and unsafe, focusing on the domain of social norms. During construction, models assign different types to various responses, enabling a safety comparison between pairs of responses.

\subsubsection{Sort}\label{subsubsec412}

The sorting method involves arranging multiple responses to the same question in descending order according predefined criteria. The criteria for sorting are determined by specific requirements. This method provides more detailed information, reflecting the relative preference order, but it is cumbersome to collect and process the sorting information, and the sorting criteria need to be standardized. OASST1\_pairwise\_rlhf\_reward\footnote{\href{https://huggingface.co/datasets/tasksource/oasst1_pairwise_rlhf_reward}{https://huggingface.co/datasets/tasksource/oasst1\_pairwise\_rlhf\_reward}} is a representative dataset in this category. It undergoes post-processing on the OASST1 \citep{bib61}, generating data directly used for RLHF. The dialogues in the OASST1, constructed by humans and accompanied by quality ratings, allow for direct sorting of different responses based on annotations, reflecting human preferences.

\subsubsection{Score}\label{subsubsec413}

The scoring method involves assigning scores within a certain range to several responses to the same question. This method provides a continuous evaluation, offering a more flexible representation of preference intensity, allowing the model to understand human preferences in a more nuanced manner. However, it is important to note issues related to the uniformity of scoring criteria and subjective awareness in the scoring process.

Some datasets use human scoring to reflect preferences. Stack-Exchange-Preferences \citep{bib150} is derived from StackOverflow, where each answer is assigned a score defined by \cite{bib150}. This score is based on the number of likes the answer receives and whether it is accepted by the question asker. In Summarize\_from\_Feedback \citep{bib148}, a portion of it involves scoring the quality of different answers using the Likert scale. WebGPT \citep{bib151} includes examples with two model answers to a question along with relevant metadata. Each answer has a preference score assigned by humans to indicate which answer is better.

In addition to human scoring, models can also be used to replace humans in this process. Alpaca\_comparison\_data \citep{bib91} involves three models generating different responses, with GPT-4 scoring the quality of the responses. Each example contains one high-quality answer and one low-quality answer. Stable\_Alignment \citep{bib152} includes three types of alignment data from simulated social interactions, with multiple different model-generated responses and corresponding scores for each data point. UltraFeedback \citep{bib153} employs models to score four answers from four dimensions, providing detailed textual explanations for improving the answers, thereby enriching the dimensions of instructions, models, and preferences.

\subsubsection{Other}\label{subsubsec414}

In addition to the three methods mentioned earlier, a small portion of preference datasets employs alternative preference evaluation methods.

\textbf{Medical-rlhf} \citep{bib45}. The Medical-rlhf dataset is a Chinese dataset designed for aligning medical models. The dataset consists of 4K examples sampled from a Chinese medical dialogue dataset. Each example includes two responses, with the higher-quality response being authentic professional replies from real doctors and the lower-quality response being model-generated. Nevertheless, the dataset has a relatively small scale, and the categorization of high and low quality is too direct and absolute for the given questions.

\textbf{PRM800K} \citep{bib155}. The PRM800K dataset is used for supervised learning of the steps in the CoT process for mathematics. It contains 102K samples of mathematical solutions and 1M step-level labels, covering 12K mathematical problems. Human annotators have labeled each step of the model-generated solutions, providing an assessment of correctness. This supervision method can also be viewed as providing an alignment signal to the model.

\subsection{Distribution Statistics of Preference Datasets}\label{subsec42}

Figure~\ref{fig17} provides statistics on 16 preference datasets from six aspects: release time, license, preference evaluation method, construction method, language, and domain. The following conclusion can be drawn:

\begin{figure}
\centering
\includegraphics[width=0.9\textwidth]{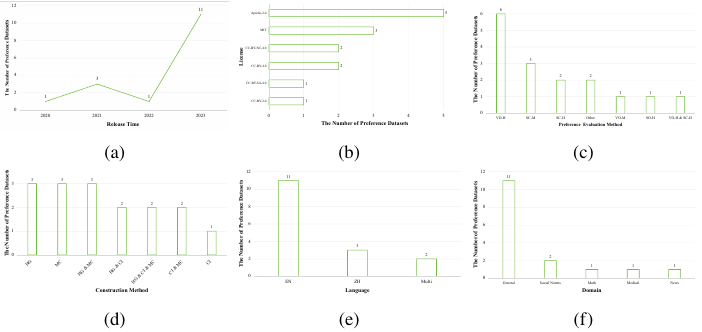}
\caption{Statistics distribution of preference datasets. (a) illustrates the quantity trend over time. (b) depicts the quantity distribution under different licenses, considering only the datasets with listed licenses. (c) shows the quantity distribution across different preference evaluation methods. (d) displays the quantity distribution for different construction methods. (e) represents the quantity distribution across different languages. (f) illustrates the quantity distribution across different domains. Zoom in for better view}\label{fig17}
\end{figure}

(1) The introduction of reinforcement learning and the in-depth research on LLMs alignment \citep{bib139,bib140,bib142} have spurred the development of preference datasets, showing a rapid growth trend in 2023. The alignment between models and humans has become an increasingly important aspect.

(2) The majority of preference datasets are available for commercial purposes, with Apache-2.0 license being predominant among them.

(3) Among all preference evaluation methods, human voting is the most commonly used. This method has a more convenient annotation process and reflects genuine human preferences. The next in popularity are human scoring and model scoring, which present preferences in a more intuitively distinguishable manner through scores. The sorting method and the combination of multiple evaluation methods are rarely used, constrained by the cumbersome process and inconsistencies in standards.

(4) From the perspective of dataset construction, the most common approach for preference datasets is human preference annotation and model-assisted generation of responses of varying quality, as these datasets require annotating feedback information based on different responses. The second approach involves scraping Q\&A from social platforms and using metrics like upvotes as a preference indicator.

(5) Preference datasets are predominantly in English, with a small portion in Chinese or a mixture of multiple languages. Overall, preference datasets in languages other than English are relatively scarce.

(6) Preference dataset examples mainly focus on general domains and social norm domains, especially in the realm of social norms. The primary goal is to ensure that LLMs align with human expectations across various general tasks and comprehensive safety considerations. Preference datasets specifically tailored for other vertical domains have not received significant attention at the moment.

\section{Evaluation Datasets}\label{sec5}

Evaluation datasets are a carefully curated and annotated set of data samples used to assess the performance of LLMs across various tasks. Different evaluation datasets focus on different evaluation aspects, providing an objective measure of different models. By solely adjusting the conditions of the training, including the pre-training corpora, instruction fine-tuning datasets, and preference datasets, the performance of LLMs on corresponding evaluation datasets can indirectly reflect the quality and effectiveness of the datasets. This, in turn, aids in the ongoing optimization of training data. The collected and organized information on representative existing evaluation datasets is presented in Table~\ref{tab11}, Table~\ref{tab12}, and Table~\ref{tab13}.

\begin{figure}[h!]
\centering
\includegraphics[width=0.9\textwidth]{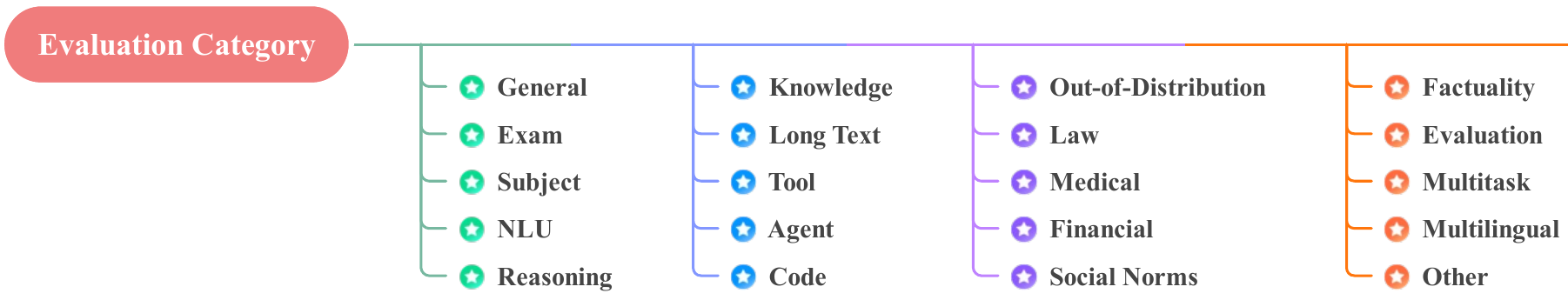}
\caption{Evaluation categories of the evaluation datasets}\label{fig18}
\end{figure}

\begin{table*}[h!]
        \captionsetup{singlelinecheck=off, justification=justified}
        \captionof{table}{Summary of \textbf{Evaluation Datasets} Information \textbf{Part I}. Public or Not: “All” indicates full open source; “Partial” indicates partially open source; “Not” indicates not open source. “License” indicates the dataset follows a certain protocol. If the dataset is built upon other datasets, the licenses of the source datasets must also be adhered to}\label{tab11}
        \centering
        \resizebox{\textwidth}{!}{
            \begin{tabular}{llllll}
            \hline
            \textbf{Dataset} & \textbf{Publisher} & \textbf{Release Time} & \textbf{Size} & \textbf{Public or Not} & \textbf{License} \\ \hline
            AgentBench & Tsinghua University et al. & 2023-8 & 1360 instances & All & - \\ 
            AGIEval & Microsoft & 2023-4 & 8062 instances & All & MIT \\ 
            ALCUNA & Peking University & 2023-10 & 84351 instances & All & MIT \\ 
            AlpacaEval & Stanford et al. & 2023-5 & 805 instances & All & Apache-2.0 \\ 
            API-Bank & Alibaba DAMO Academy et al. & 2023-4 & 264 dialogues & All & MIT \\ 
            APIBench & UC Berkeley et al. & 2023-5 & 16450 instances & All & Apache-2.0 \\ 
            APPS & UC Berkeley et al. & 2021-5 & 10000 instances & All & MIT \\ 
            ARB & DuckAI et al. & 2023-7 & 1207 instances & All & MIT \\ 
            BayLing-80 & Chinese Academy of Sciences & 2023-6 & 320 instances & All & GPL-3.0 \\ 
            BBF-CFLEB & Fudan University et al. & 2023-2 & 11327 instances & All & - \\ 
            BBH & Google Research et al. & 2022-10 & 6511 instances & All & MIT \\ 
            BELLE\_eval & BELLE & 2023-4 & 1000 instances & All & Apache-2.0 \\ 
            BIG-Bench & Google et al. & 2022-6 & - & All & Apache-2.0 \\ 
            BIRD & The University of Hong Kong et al. & 2023-5 & 12751 instances & All & CC-BY-NC-4.0 \\ 
            BOSS & Tsinghua University et al. & 2023-6 & - & All & MIT \\ 
            CBLUE & Zhejiang University et al. & 2022-5 & 195820 instances & All & Apache-2.0 \\ 
            C-CLUE & Tianjin University & 2021-8 & - & All & CC-BY-SA-4.0 \\ 
            CELLO & Fudan University et al. & 2023-9 & 523 instances & All & - \\ 
            C-Eval & Shanghai Jiao Tong University & 2023-5 & 13948 instances & All & CC-BY-NC-SA-4.0 \\ 
            CG-Eval & LanguageX AI Lab et al. & 2023-8 & 11000 instances & All & CC-BY-SA-4.0 \\ 
            Chain-of-Thought Hub & University of Edinburgh et al. & 2023-5 & - & All & MIT \\ 
            Choice-75 & University of Pittsburgh et al. & 2023-9 & 650 instances & All & - \\ 
            CLEVA & The Chinese University of Hong Kong et al. & 2023-8 & 370K instances & All & CC-BY-NC-ND-4.0 \\ 
            CLiB & jeinlee1991 & 2023-6 & 90 instances & All & - \\ 
            CLUE & CLUE team & 2020-12 & 9 datasets & All & - \\ 
            CMB & The Chinese University of Hong Kong et al. & 2023-8 & 281047 instances & All & Apache-2.0 \\ 
            CMMLU & MBZUAI & 2023-6 & 11528 instances & All & CC-BY-NC-4.0 \\ 
            CodeXGLUE & Peking University et al. & 2021-2 & 4.12M instances & All & C-UDA \\ 
            CrowS-Pairs & New York University & 2020-11 & 1508 instances & All & CC-SA-4.0 \\ 
            CUGE & Tsinghua University et al. & 2021-12 & 33.4M instances & All & - \\ 
            decaNLP & Salesforce Research & 2018-6 & 2010693 instances & All & BSD-3-Clause \\ 
            DS-1000 & The University of Hong Kong et al. & 2022-11 & 1000 instances & All & CC-BY-SA-4.0 \\ 
            EcomGPT\_eval & Alibaba & 2023-8 & 6000 instances & All & - \\ 
            EmotionBench & The Chinese University of Hong Kong et al. & 2023-8 & - & All & GPL-3.0 \\ 
            FACTOR & AI21 Labs & 2023-7 & 4030 instances & Partial & MIT \\ 
            FActScore & University of Washington et al. & 2023-5 & 500 instances & All & MIT \\ 
            FactualityPrompt & Hong Kong University of Science and Technology et al. & 2022-6 & 16000 instances & All & Apache-2.0 \\ 
            FairEval & Peking University et al. & 2023-5 & 80 instances & All & - \\ 
            FewCLUE & CLUE team & 2021-7 & 9 datasets & Partial & - \\ 
            FinancelQ & Du Xiaoman & 2023-9 & 7173 instances & All & CC-BY-NC-SA-4.0 \\ 
            FinBen & The Fin AI et al. & 2024-2 & 69805 instances & All & - \\
            FinEval & Shanghai University of Finance and Economics & 2023-8 & 4661 instances & All & CC-BY-NC-SA-4.0 \\ 
            FlagEval & BAAI et al. & 2023-6 & 84433 instances & Partial & - \\ 
            FLUE & Georgia Institute of Technology et al. & 2022-10 & 26292 instances & All & - \\ 
            FreshQA & Google et al. & 2023-10 & 600 instances & All & - \\ 
            GAOKAO-Bench & Fudan University et al. & 2023-5 & 2811 instances & All & Apache-2.0 \\ 
            GeoBench & Shanghai Jiao Tong University et al. & 2023-6 & 2517 instances & All & Apache-2.0 \\ 
            GLUE & New York University et al. & 2018-11 & 9 datasets & All & - \\ 
            GLUE-X & Westlake University et al. & 2022-11 & 6404940 instances & All & - \\ 

            \hline
            \end{tabular}
        }
\end{table*}

\begin{table*}[h!]
        \captionsetup{singlelinecheck=off, justification=justified}
        \captionof*{table}{\textbf{Table 11} (continued)}\label{tab112}
        \centering
        \resizebox{\textwidth}{!}{
            \begin{tabular}{llllll}
            \hline
            \textbf{Dataset} & \textbf{Publisher} & \textbf{Release Time} & \textbf{Size} & \textbf{Public or Not} & \textbf{License} \\ \hline
            HalluQA & Fudan University et al. & 2023-10 & 450 instances & All & - \\ 
            HaluEval & Renmin University of China et al. & 2023-5 & 35000 instances & All & MIT \\ 
            HELM & Stanford University et al. & 2022-11 & - & All & Apache-2.0 \\ 
            HuaTuo26M-test & The Chinese University of Hong Kong et al. & 2023-5 & 6000 instances & All & Apache-2.0 \\ 
            HumanEval & OpenAI et al. & 2021-7 & 164 instances & All & MIT \\ 
            HumanEvalPack & Bigcode & 2023-8 & 984 instances & All & MIT \\ 
            InfiniteBench & Tsinghua University et al. & 2023-11 & 3932 instances & All & Apache-2.0 \\ 
            KoLA & Tsinghua University & 2023-6 & 2138 instances & Partial & GPL-3.0 \\ 
            LAiW & Sichuan University et al. & 2023-10 & - & Partial & - \\ 
            LawBench & Nanjing University et al. & 2023-9 & - & All & Apache-2.0 \\ 
            LegalBench & Stanford University et al. & 2023-8 & 90417 instances & All & - \\ 
            L-Eval & Fudan University et al. & 2023-7 & 2043 instances & All & GPL-3.0 \\ 
            LexGLUE & University of Copenhagen et al. & 2021-10 & 237014 instances & All & - \\ 
            LEXTREME & University of Bern et al. & 2023-1 & 3508603 instances & All & - \\ 
            LILA & Arizona State Univeristy et al. & 2022-10 & 317262 instances & All & CC-BY-4.0 \\ 
            LLMEVAL-1 & Fudan University et al. & 2023-5 & 453 instances & All & - \\ 
            LLMEVAL-2 & Fudan University et al. & 2023-7 & 480 instances & All & - \\ 
            LLMEVAL-3 & Fudan University et al. & 2023-9 & 200K instances & Not & - \\ 
            $\mathrm{LLMEval}^2$ & Chinese Academy of Sciences et al. & 2023-8 & 2533 instances & All & MIT \\ 
            LMentry & Tel Aviv University et al. & 2023-7 & 110703 instances & All & - \\ 
            LMExamQA & Tsinghua University et al. & 2023-6 & 10090 instances & All & - \\ 
            LongBench & Tsinghua University et al. & 2023-8 & 4750 instances & All & MIT \\ 
            LongEval & LMSYS & 2023-6 & - & All & Apache-2.0 \\ 
            LooGLE & BIGAI et al. & 2023-11 & 6448 instances & All & CC-BY-SA-4.0 \\ 
            MCTS & Beijing Language and Culture University & 2023-6 & 723 instances & All & - \\ 
            miniF2F\_v1 & Ecole Polytechnique et al. & 2021-9 & 488 instances & All & - \\ 
            MINT & University of Illinois Urbana-Champaign et al. & 2023-9 & 586 instances & All & Apache-2.0 \\ 
            MMCU & LanguageX AI Lab & 2023-4 & 11845 instances & All & - \\ 
            MMLU & UC Berkeley et al. & 2020-9 & 15908 instances & All & MIT \\ 
            MT-Bench & UC Berkeley et al. & 2023-6 & 80 instances & All & Apache-2.0 \\ 
            MTPB & Salesforce Research & 2022-3 & 115 instances & All & Apache-2.0 \\ 
            MultiMedQA & Google Research et al. & 2022-12 & 212822 instances & All & - \\ 
            M3Exam & Alibaba Group et al. & 2023-6 & 12317 instances & All & - \\ 
            M3KE & Tianjin University et al. & 2023-5 & 20477 instances & All & Apache-2.0 \\ 
            NeuLR & Xi’an Jiaotong University et al. & 2023-6 & 3000 instances & All & - \\ 
            ODEX & Carnegie Mellon University et al. & 2022-12 & 945 instances & All & CC-BY-SA-4.0 \\ 
            Owl-Bench & Beihang University et al. & 2023-9 & 1317 instances & All & - \\ 
            PandaLM\_testset & Peking University et al. & 2023-4 & 999 instances & All & Apache-2.0 \\ 
            PromptBench & Microsoft Research et al. & 2023-6 & 583884 instances & All & MIT \\ 
            PromptCBLUE & East China Normal University et al. & 2023-4 & 20640 instances & All & - \\ 
            QiZhenGPT\_eval & Zhejiang University et al. & 2023-5 & 94 instances & All & GPL-3.0 \\ 
            RAFT & Ought et al. & 2021-9 & 28712 instances & All & - \\ 
            SafetyBench & Tsinghua University et al. & 2023-9 & 11435 instances & All & MIT \\ 
            Safety-Prompts & Tsinghua University et al. & 2023-4 & 100K instances & Partial & Apache-2.0 \\ 
            SCALE & University of Bern et al. & 2023-6 & 1.86M instances & All & CC-BY-SA \\ 
            SCIBENCH & University of California et al. & 2023-7 & 695 instances & All & MIT \\ 
            SentEval & Facebook Artificial Intelligence Research & 2018-5 & 28 datasets & All & BSD \\ 
            ScienceQA & University of California et al. & 2022-9 & 21208 instances & All & CC-BY-NC-SA-4.0 \\ 
            SocKET & University of Michigan et al. & 2023-5 & 2616342 instances & All & CC-BY-4.0 \\ 
            SuperCLUE & CLUE et al. & 2023-7 & 3754 instances & Not & - \\ 
            SuperCLUE-Agent & CLUEbenchmark & 2023-10 & - & Not & - \\ 
            SuperCLUE-Safety & CLUEbenchmark & 2023-9 & 4912 instances & Not & - \\ 
            SuperGLUE & New York University et al. & 2019-5 & 8 datasets & All & - \\ 
            TabMWP & University of California et al. & 2022-9 & 38431 instances & All & CC-BY-NC-SA-4.0 \\ 
            TheoremQA & University of Waterloo et al. & 2023-5 & 800 instances & All & MIT \\ 
            ToolBench & SambaNova Systems et al. & 2023-5 & 795 instances & All & Apache-2.0 \\ 
            TRUSTGPT & Sichuan University et al. & 2023-6 & 2000 instances & All & MIT \\ 
            TruthfulQA & University of Oxford et al. & 2022-5 & 817 instances & All & Apache-2.0 \\ 
            Vicuna Evaluation & LMSYS ORG & 2023-3 & 80 instances & All & Apache-2.0 \\ 
            XiezhiBenchmark & Fudan University et al. & 2023-6 & 249587 instances & All & CC-BY-NC-SA-4.0 \\ 
            XNLI & Facebook AI et al. & 2018-10 & 112500 instances & All & CC-BY-NC-4.0 \\ 
            XTREME & Carnegie Mellon University et al. & 2020-3 & - & All & Apache-2.0 \\ 
            ZeroSCROLLS & Tel Aviv University et al. & 2023-5 & 4378 instances & All & MIT \\ \hline
            \end{tabular}
        }
\end{table*}

\begin{table*}[h!]
        \captionsetup{singlelinecheck=off, justification=justified}
        \captionof{table}{Summary of \textbf{Evaluation Datasets} Information \textbf{Part II}. Language: “EN” indicates English, “ZH” indicates Chinese, “PL” indicates Programming Language, “Multi” indicates Multilingual, and the number in parentheses indicates the number of languages included. “CM” indicates Construction Methods, where “HG” indicates Human Generated Datasets, “MC” indicates Model Constructed Datasets, and “CI” indicates Collection and Improvement of Existing Datasets. “QT” indicates Question Types, where “SQ” indicates Subjective Questions, “OQ” indicates Objective Questions, and “Multi” indicates Multiple Question Types. “EM” indicates Evaluation Methods, where “CE” indicates Code Evaluation, “HE” indicates Human Evaluation, and “ME” indicates Model Evaluation}\label{tab12}
        \centering
        \resizebox{\textwidth}{!}{
            \begin{tabular}{lllllll}
            \hline
            \textbf{Dataset} & \textbf{Language} & \textbf{CM} & \textbf{QT} & \textbf{EM} & \textbf{Domain} & \textbf{Focus} \\ \hline
            AgentBench & EN & HG \& CI \& MC & SQ & CE & Agent & LLM-as-Agent’s reasoning and decision-making abilities \\ 
            AGIEval & EN \& ZH & HG \& CI & OQ & CE & Exam & Human-centric standardized exams \\ 
            ALCUNA & EN & HG & Multi & CE & Knowledge & Assess the ability of LLMs to respond to new knowledge \\ 
            AlpacaEval & EN & CI \& MC & SQ & ME & General & The performance on open-ended question answering \\ 
            API-Bank & EN \& PL & HG \& MC & SQ & HE \& CE & Tool & Plan step-by-step API calls, retrieve relevant APIs, and correctly execute API calls to meet human needs \\ 
            APIBench & EN \& PL & HG \& MC & SQ & CE & Tool & The reasoning ability for calling APIs \\ 
            APPS & EN \& PL & HG & SQ & CE & Code & The ability to take an arbitrary natural language specification and generate satisfactory Python code \\ 
            ARB & EN & CI & Multi & HE \& ME & Subject & Advanced reasoning problems in multiple fields \\ 
            BayLing-80 & EN \& ZH & HG \& CI & SQ & ME & General & Chinese-English language proficiency and multimodal interaction skills \\ 
            BBF-CFLEB & ZH & HG \& CI & SQ & CE & Financial & Language understanding and generation tasks in Chinese financial natural language processing \\ 
            BBH & EN & CI & Multi & CE & Multitask & Challenging tasks that have proven difficult for prior language model evaluations \\ 
            BELLE\_eval & ZH & HG \& MC & SQ & ME & General & The performance of Chinese language models in following instructions \\ 
            BIG-Bench & Multi & HG \& CI & Multi & CE & Multitask & The capabilities and limitations of language models \\ 
            BIRD & EN \& PL & HG \& CI \& MC & SQ & CE & Code & Text-to-SQL parsing \\ 
            BOSS & EN & CI & SQ & CE & OOD & Assess model performance under distribution shifts \\ 
            CBLUE & ZH & HG \& CI & SQ & CE & Medical & Chinese biomedical language understanding \\ 
            C-CLUE & ZH & HG & SQ & CE & Subject & Classical Chinese language understanding \\ 
            CELLO & EN & HG & SQ & CE & General & The ability of LLMs to understand complex instructions \\ 
            C-Eval & ZH & HG \& MC & OQ & CE & Subject & The advanced knowledge and reasoning abilities in a Chinese context \\ 
            CG-Eval & ZH & HG & SQ & CE & Subject & The generation capabilities of LLMs across various academic disciplines \\ 
            Chain-of-Thought Hub & EN & CI & SQ & CE & Reasoning & The multi-step reasoning capabilities \\ 
            Choice-75 & EN & HG \& CI \& MC & OQ & CE & Reasoning & Predict decisions based on descriptive scenarios \\ 
            CLEVA & ZH & HG \& CI & SQ & CE & Multitask & The performance of LLMs across various dimensions \\ 
            CLiB & ZH & - & SQ & HE & Multitask & Multidimensional capabilities \\ 
            CLUE & ZH & CI & SQ & CE & NLU & Natural language understanding capability \\ 
            CMB & ZH & HG & Multi & HE \& CE \& ME & Medical & The performance of LLMs in the field of medicine \\ 
            CMMLU & ZH & HG & OQ & CE & Subject & The knowledge and reasoning capabilities within the Chinese context \\ 
            CodeXGLUE & EN \& PL & CI & SQ & CE & Code & Program understanding and generation tasks \\ 
            CrowS-Pairs & EN & HG \& CI & SQ & CE & Social Norms & The presence of cultural biases and stereotypes in pretrained language models \\ 
            CUGE & EN \& ZH & CI & SQ & CE & NLU & Natural language understanding capability \\ 
            decaNLP & EN & CI & SQ & CE & Multitask & Multitask natural language processing capabilities \\ 
            DS-1000 & EN \& PL & HG & SQ & CE & Code & Code generation \\ 
            EcomGPT\_eval & EN \& ZH & CI & SQ & CE & E-commerce & E-commerce-related tasks \\ 
            EmotionBench & EN & HG \& MC & SQ & CE & Sentiment & The empathy ability \\ 
            FACTOR & EN & HG \& CI \& MC & OQ & CE & Factuality & The factuality of LLMs \\ 
            FActScore & EN & HG \& MC & SQ & HE \& ME & Factuality & The factuality of LLMs \\ 
            FactualityPrompt & EN & CI & SQ & CE & Factuality & The factuality of LLMs \\ 
            FairEval & EN & CI & SQ & CE & Evaluation & The performance in determining the quality of output content from different models \\ 
            FewCLUE & ZH & CI & SQ & CE &  Few-shot learning & Compare different few-shot learning methods \\ 
            FinancelQ & ZH & HG \& MC & OQ & CE & Financial & The knowledge and reasoning abilities in financial contexts \\ 
            FinBen & EN & CI & SQ & CE & Financial & NLP tasks in the financial domain \\
            FinEval & ZH & HG & OQ & CE & Financial & The performance in the financial domain knowledge \\ 
            FlagEval & EN \& ZH & HG \& CI & Multi & HE \& CE & Multitask & Multi-domain, multi-dimensional capabilities \\ 
            FLUE & EN & CI & SQ & CE & Financial & NLP tasks in the financial domain \\ 
            FreshQA & EN & HG & SQ & HE & Factuality & The factuality of LLMs \\ 
            GAOKAO-Bench & ZH & HG & Multi & HE \& CE & Exam & Chinese Gaokao examination \\ 
            GeoBench & EN & HG & Multi & HE \& CE \& ME & Geoscience & LLMs’ performance in understanding and utilizing geoscience knowledge \\ 
            GLUE & EN & CI & SQ & CE & NLU & Natural language understanding capability \\ 
            GLUE-X & EN & CI & SQ & CE & OOD & The out-of-distribution (OOD) robustness \\ 
            HalluQA & ZH & HG \& MC & SQ & ME & Factuality & The factuality of LLMs \\ 
            HaluEval & EN & HG \& CI \& MC & SQ & CE & Factuality & The factuality of LLMs \\ 
            HELM & EN & CI & SQ & HE \& CE & Multitask & Evaluate LLMs on a wide range of scenarios and metrics \\ 

            \hline
            \end{tabular}
        }
\end{table*}

\begin{table*}[h!]
        \captionsetup{singlelinecheck=off, justification=justified}
        \captionof*{table}{\textbf{Table 12} (continued)}\label{tab122}
        \centering
        \resizebox{\textwidth}{!}{
            \begin{tabular}{lllllll}
            \hline
            \textbf{Dataset} & \textbf{Language} & \textbf{CM} & \textbf{QT} & \textbf{EM} & \textbf{Domain} & \textbf{Focus} \\ \hline
            HuaTuo26M-test & ZH & CI & SQ & CE & Medical & Understand and generate complex medical language \\ 
            HumanEval & EN \& PL & HG & SQ & CE & Code & The correctness of problem-solving abilities in the context of program synthesis \\ 
            HumanEvalPack & EN \& PL & HG \& CI & SQ & CE & Code & The correctness of problem-solving abilities in the context of program synthesis \\ 
            InfiniteBench & EN \& ZH & HG \& CI \& MC & Multi & - & Long Text & Long text task capability \\ 
            KoLA & EN & HG \& CI & SQ & CE & Knowledge & The ability to grasp and utilize world knowledge \\ 
            LAiW & ZH & CI & SQ & CE & Law & Legal capabilities \\ 
            LawBench & ZH & HG \& CI & Multi & CE & Law & Legal capabilities \\ 
            LegalBench & EN & HG \& CI & SQ & HE \& CE & Law & Legal reasoning \\ 
            L-Eval & EN & HG \& CI & SQ & HE \& CE \& ME & Long Text & Long text task capability \\ 
            LexGLUE & EN & CI & SQ & CE & Law & Legal capabilities \\ 
            LEXTREME & Multi (24) & CI & SQ & CE & Law & Legal capabilities \\ 
            LILA & EN & CI & Multi & CE & Reasoning & Mathematical reasoning across diverse tasks \\ 
            LLMEVAL-1 & ZH & HG & SQ & HE \& ME & Multitask & Multidimensional capabilities \\ 
            LLMEVAL-2 & ZH & HG & Multi & HE \& ME & Knowledge & Knowledge capability \\ 
            LLMEVAL-3 & ZH & HG & SQ & ME & Subject & Subject-specific knowledge capability \\ 
            $\mathrm{LLMEval}^2$ & Multi & CI & SQ & CE & Evaluation & The performance in determining the quality of output content from different models \\ 
            LMentry & EN & HG & SQ & CE & Multitask & The performance on challenging tasks \\ 
            LMExamQA & EN & MC & SQ & ME & Knowledge & The performance on open-ended question answering \\ 
            LongBench & EN \& ZH & CI & SQ & CE & Long Text & Long text task capability \\ 
            LongEval & EN & HG & SQ & CE & Long Text & Long text task capability \\ 
            LooGLE & EN & HG \& CI \& MC & SQ & HE \& CE \& ME & Long Text & Long text task capability \\ 
            MCTS & ZH & HG & SQ & CE & NLU & Text simplification ability \\ 
            miniF2F\_v1 & EN & HG \& CI & SQ & CE & Reasoning & The performance on formal Olympiad-level mathematics problem statements \\ 
            MINT & EN & CI & SQ & CE & Multi-turn interactions & Solve complex tasks through multi-turn interactions using tools and leveraging natural language feedback \\ 
            MMCU & ZH & HG & OQ & CE & Subject & Multidisciplinary abilities \\ 
            MMLU & EN & HG & OQ & CE & Subject & Knowledge in academic and professional domains \\ 
            MT-Bench & EN & HG & SQ & ME & General & The performance on open-ended question answering \\ 
            MTPB & EN \& PL & HG & SQ & CE & Code & Multi-turn Programming \\ 
            MultiMedQA & EN & HG \& CI & Multi & HE \& CE & Medical & The performance in medical and clinical applications \\ 
            M3Exam & Multi (9) & HG & OQ & CE & Exam & The comprehensive abilities in a multilingual and multilevel context using real human exam questions \\ 
            M3KE & ZH & HG & OQ & CE & Subject & Multidisciplinary abilities \\ 
            NeuLR & EN & CI & SQ & CE & Reasoning & Logical reasoning capabilities \\ 
            ODEX & Multi \& PL   & HG \& CI & SQ & CE & Code & Natural language to Python code generation \\ 
            Owl-Bench & EN \& ZH & HG & Multi & ME & IT & The performance in IT-related tasks \\ 
            PandaLM\_testset & EN & HG \& MC & SQ & CE & Evaluation & The performance in determining the quality of output content from different models \\ 
            PromptBench & EN & CI & SQ & CE & Robustness & The models’ robustness \\ 
            PromptCBLUE & ZH & CI & SQ & CE & Medical & The performance in Chinese medical scenarios \\ 
            QiZhenGPT\_eval & ZH & HG & SQ & HE & Medical & Indications for use of drugs \\ 
            RAFT & EN & HG \& CI & SQ & CE & NLU & Text classification ability \\ 
            SafetyBench & EN \& ZH & HG \& CI \& MC & OQ & CE & Social Norms & The safety of LLMs \\ 
            Safety-Prompts & ZH & MC & SQ & HE \& ME & Social Norms & The safety of LLMs \\ 
            SCALE & Multi (5) & HG \& CI & SQ & CE & Law & Legal multidimensional abilities \\ 
            SCIBENCH & EN & HG & SQ & CE & Subject & The performance in university-level science and engineering domains \\ 
            SentEval & EN & CI & SQ & CE & NLU & The quality of universal sentence representations \\ 
            ScienceQA & EN & HG & OQ & CE & Subject & Science question-answering ability \\ 
            SocKET & EN & CI & SQ & CE & Knowledge & Mastery of social knowledge \\ 
            SuperCLUE & ZH & HG \& MC & Multi & HE \& CE & General & The performance in a Chinese context \\ 
            SuperCLUE-Agent & ZH & - & SQ & - & Agent & Agent capabilities of LLMs \\ 
            SuperCLUE-Safety & ZH & - & SQ & ME & Social Norms & The safety of LLMs \\ 
            SuperGLUE & EN & CI & SQ & CE & NLU & Natural language understanding capability \\ 
            TabMWP & EN & HG & Multi & CE & Reasoning & Mathematical reasoning ability involving both textual and tabular information \\ 
            TheoremQA & EN & HG & SQ & CE & Subject & Science subject question-answering ability \\ 
            ToolBench & EN & HG \& CI & SQ & CE & Tool & The enhancement in tool manipulation for real-world software tasks \\ 
            TRUSTGPT & EN & CI & SQ & CE & Social Norms & The performance in toxicity, bias, and value alignment \\ 
            TruthfulQA & EN & HG & SQ & CE \& ME & Factuality & The factuality of LLMs \\ 
            Vicuna Evaluation & EN & HG & SQ & ME & General & The performance on open-ended question answering \\ 
            XiezhiBenchmark & EN \& ZH & HG \& MC & OQ & CE & Subject & Multidisciplinary abilities \\ 
            XNLI & Multi (15) & HG & SQ & CE & Multilingual & Multilingual NLI \\ 
            XTREME & Multi (40) & CI & SQ & CE & Multilingual & The cross-lingual generalization capabilities \\ 
            ZeroSCROLLS & EN & HG \& CI & Multi & CE & Long Text & Long text task capability \\ 
            \hline
            \end{tabular}
        }
\end{table*}

\begin{table*}[h!]
        \captionsetup{singlelinecheck=off, justification=justified}
        \captionof{table}{Summary of \textbf{Evaluation Datasets} Information \textbf{Part III}. “NC” indicates Numbers of Evaluation Categories. “NS” indicates Numbers of Evaluation Subcategories. Zoom in for better view}\label{tab13}
        \centering
        \resizebox{\textwidth}{!}{
            \begin{tabular}{llll}
            \hline
            \textbf{Dataset} & \textbf{NC} & \textbf{NS} & \textbf{Category} \\ \hline
            AgentBench & 8 & - & Operating system, Database, Knowledge graph, Digital card game, Lateral thinking puzzles, House-holding, Web shopping, Web browsing \\ 
            AGIEval & 7 & 20 & Gaokao, SAT, JEC, LSAT, LogiQA, AQuA-RAT, Math \\ 
            ALCUNE & 3 & - & Knowledge understanding, Knowledge differentiation, Knowledge association \\ 
            AlpacaEval & 1 & - & Open-ended question answering \\ 
            API-Bank & 3 & - & Call, Retrieval+Call, Plan+Retrieval+Call \\ 
            APIBench & 1 & - & API call \\ 
            APPS & 1 & - & Code generation \\ 
            ARB & 5 & - & Mathematics, Physics, Law, MCAT(Reading), MCAT(Science) \\ 
            BayLing-80 & 9 & - & Writing, Roleplay, Common-sense, Fermi, Counterfactual, Coding, Math, Generic, Knowledge \\ 
            BBF-CFLEB & 6 & - & FinNL, FinNA, FinRE, FinFE, FinQA, FinNSP \\ 
            BBH & 23 & 27 & Boolean expressions, Causal judgement, Date understanding, Disambiguation QA, etc. \\ 
            BELLE\_eval & 9 & - & Extract, Closed qa, Rewrite, Summarization, Generation, Classification, Brainstorming, Open qa, Others \\ 
            BIG-Bench & 95 & 204 & Linguistics, Child development, Mathematics, Common sense reasoning, Biology, etc. \\ 
            BIRD & 1 & - & Text-SQL \\ 
            BOSS & 5 & 20 & Sentiment analysis, Toxicity detection, Natural language inference, Named entity recognition, Extractive Question answering \\ 
            CBLUE & 5 & 8 & Information extraction from the medical text, normalization of the medical term, medical text classification, medical sentence similarity estimation, medical QA \\ 
            C-CLUE & 2 & - & Named entity recognition, Relation extraction \\ 
            CELLO & 2 & 10 & Complex task description, Complex input \\ 
            C-Eval & 4 & 52 & STEM, Social Science, Humanity, Other \\ 
            CG-Eval & 6 & 55 & Science and engineering, Humanities and social sciences, Mathematical calculations, Medical practitioner qualification Examination, Judicial Examination, Certfied public accountant examination \\ 
            Chain-of-Thought Hub & 6 & 8 & Math, Science, Symbolic, Knowledge, Coding, Factual \\ 
            Choice-75 & 4 & - & Easy, Medium, Hard, N/A \\ 
            CLEVA & 2 & 31 & Ability, Application \\ 
            CLiB & 4 & - & Classification, Information extraction, Reading comprehension, Tabular question answering \\ 
            CLUE & 3 & 9 & Single-sentence tasks, Sentence pair tasks, Machine reading comprehension tasks \\ 
            CMB & 2 & 7 & CMB-Exam, CMB-Clin \\ 
            CMMLU & 5 & 67 & Social science, STEM, Humanities, China specific, Other \\ 
            CodeXGLUE & 4 & 10 & Code-Code, Text-Code, Code-Text, Text-to-Text \\ 
            CrowS-Pairs & 9 & - & Race, Gender, Sexual orientation, Religion, Age, Nationality, Disability, Physical appearance, Occupation \\ 
            CUGE & 7 & 18 & Language understanding (word-sentence or discourse level), Information acquisition and question answering, Language generation, Conversational interaction, Multilingual, Mathematical reasoning \\ 
            decaNLP & 10 & - & Question answering, Machine translaion, Summarization, Natural language inference, Sentiment analysis, Semantic role labeling, Zero-shot relation extraction, Goal-oriented dialogue, Semantic parsing, Pronoun resolution \\ 
            DS-1000 & 1 & - & Code generation \\ 
            EcomGPT\_eval & 4 & 12 & Classification, Generation, Extraction, Others \\ 
            EmotionBench & 8 & 36 & Anger, Anxiety, Depression, Frustration, Jealous, Guilt, Fear, Embarrassment \\ 
            FACTOR & 2 & - & Wiki, News \\ 
            FActScore & 7 & - & Single-sentence contradiction (words or beyond words), Page-level contradiction, Subjective, Fact is irrelevant, Wiki is inconsistent \& wrong, Annotation error \\ 
            FactualityPrompt & 2 & - & Factual prompts, Nonfactual prompts \\ 
            FairEval & 1 & - & Evaluate the quality of answers \\ 
            FewCLUE & 3 & 9 & Single sentence tasks, Sentence pair tasks, Reading comprehension \\ 
            FinancelQ & 10 & 36 & Bank, Fund, Securities, Futures and derivatives, CICE, Actuarial science, Financial planning, CPA, Taxation, Economics \\ 
            FinBen & 3 & 6 & Foundamental tasks, Advanced cognitive engagement, General intelligence \\
            FinEval & 4 & 34 & Finance, Economy, Accounting, Certificate \\ 
            FlagEval & 3 & 21 & Choice qa, Classification, Generation qa \\ 
            FLUE & 5 & 6 & Financial sentiment analysis, News headline classification, Named entity recognition, Structure boundary detection, Question answering \\ 
            FreshQA & 4 & - & Never-changing, Slow-changing, Fast-changing, False-premise \\ 
            GAOKAO-Bench & 10 & - & Chinese, Mathematics (2 categories), English, Physics, Chemistry, Biology, Politics, History, Geography \\ 
            GeoBench & 2 & - & NPEE, APTest \\ 
            GLUE & 3 & 9 & Single-sentence tasks, Similarity and paraphrase tasks, Inference tasks \\ 
            GLUE-X & 7 & 16 & Sentiment analysis, Linguistic acceptability, Textual similarity, Natural language inference, Question answering NLI, Textual entailment, Paraphrase \\ 
            HalluQA & 3 & - & Misleading, Misleading-hard, Knowledge \\ 
            HaluEval & 3 & - & QA, Dialogue, Summarization \\ 
            HELM & 73 & - & Question answering, Information retrieval, Sentiment analysis, Toxicity detection, Aspirational scenarios, etc. \\ 
            HuaTuo26M-test & 3 & - & Medical consultant records, Encyclopedias, Knowledge bases \\ 
            HumanEval & 1 & - & Code generation \\ 
            HumanEvalPack & 3 & - & HumanEvalFix, HumanEvalExplain, HumanEvalSynthesize \\ 
            InfiniteBench & 5 & 12 & Mathematics, Code, Dialogue, Books, Retrieval \\ 
            KoLA & 4 & 19 & Knowledge memorization, Knowledge understanding, Knowledge applying, Knowledge creating \\ 
            LAiW & 3 & 13 & Basic legal NLP, Basic legal application, Complex legal application \\ 
            LawBench & 3 & 20 & Legal knowledge memorization, Legal knowledge understanding, Legal knowledge applying \\ 
            LegalBench & 6 & 162 & Issue-spotting, Rule-recall, Rule-application, Rule-conclusion, Interpretation, Rhetorical-understanding \\ 
            L-Eval & 1 & 18 & Long text task \\ 
            LexGLUE & 3 & - & Multi-label classification, Multi-class classification, Multiple choice QA \\ 
            LEXTREME & 18 & - & Brazilian court decisions, German argument mining, Greek legal code, Swiss judgment prediction, etc. \\ 
            LILA & 4 & 23 & Math ability, Language, Knowledge, Format \\ 
            LLMEVAL-1 & 17 & - & Fact-based question answering, Reading comprehension, Framework generation, Paragraph rewriting, etc. \\ 
            LLMEVAL-2 & 12 & - & Computer science, Economics, Foreign languages, Law, Mathematics, Medicine, Optics, Physics, Social sciences, Chinese language and literature, Chemistry, Life sciences \\ 
            LLMEVAL-3 & 13 & - & Philosophy, Economics, Law, Education, Literature, History, Science, Engineering, Agriculture, Medicine, Military science, Management, Fine arts \\ 
            $\mathrm{LLMEval}^2$ & 1 & - & Evaluate the quality of answers \\ 
            LMentry & 25 & - & Sentence containing word, Sentence not containing word, Word containing letter, Word not containing letter, etc. \\ 
            LMExamQA & 3 & 25 & Knowledge memorization, Knowledge comprehension, Knowledge analysis \\ 
            LongBench & 6 & 21 & Single-doc QA, Multi-doc QA, Summarization, Few-shot learning, Synthetic tasks, Code completion \\ 
            LongEval & 2 & - & Coarse-grained topic retrieval, Fine-grained line retrieval \\ 
            LooGLE & 2 & 4 & Long dependency tasks, Short dependency tasks \\ 
            MCTS & 1 & - & Text simplification \\ 
            miniF2F\_v1 & 1 & - & Math \\ 
            MINT & 3 & - & Code generation, Decision making, Reasoning \\ 
            MMCU & 4 & 25 & Medicine, Law, Psychology, Education \\ 
            MMLU & 4 & 57 & Humanities, Social science, STEM, Other \\ 
            MT-Bench & 8 & - & Writing, Roleplay, Reasoning, Math, Coding, Extraction, STEM, Humanities \\ 
            MTPB & 1 & - & Code generation \\ 
            MultiMedQA & 1 & - & Medical question answering \\ 
            M3Exam & 3 & - & Low, Mid, High \\ 
            M3KE & 4 & 71 & Arts \& Humanities, Social sciences, Natural sciences, Other \\ 
            NeuLR & 3 & - & Deductive, Inductive, Abductive \\ 
            ODEX & 1 & - & Code generation \\ 
            Owl-Bench & 9 & - & Information security, Application, System architecture, Software architecture, Middleware, Network, Operating system, Infrastructure, Database \\ 
            PandaLM\_testset & 1 & - & Evaluate the quality of answers \\ 
            PromptBench & 10 & 15 & Sentiment analysis, Grammar correctness, Duplicate sentence detection, Natural language inference, etc. \\ 
            PromptCBLUE & 16 & - & Medical named entity recognition, Medical entity relation extraction, Medical event extraction, etc. \\ 
            QiZhenGPT\_eval & 1 & - & Drug indication question answering \\ 
            RAFT & 1 & 11 & Text classification \\ 
            SafetyBench & 7 & - & Offensiveness, Unfairness and bias, Physical health, Mental Health, Illegal activities, Ethics and morality, Privacy and Property \\ 
            Safety-Prompts & 2 & 13 & Typical security scenarios, Instruction attack \\ 
            SCALE & 4 & - & Processing long documents, Utilizing domain specific knowledge, Multilingual understanding, Multitasking \\ 
            SCIBENCH & 3 & 10 & Physics, Chemistry, Math \\ 
            SentEval & 1 & 21 & Universal sentence representations \\ 
            ScienceQA & 3 & 26 & Natural science, Social science, Language science \\ 
            SocKET & 4 & 58 & Classification, Regression, Pair-wise comparison, Span identification \\ 
            SuperCLUE & 2 & - & Open multi-turn open questions, OPT objective questions \\ 
            SuperCLUE-Agent & 3 & 10 & Tool utilization, Task planning, Long-term and short-term memory \\ 
            SuperCLUE-Safety & 3 & 20+ & Traditional security category, Responsible artificial intelligence, Instruction attacks \\ 
            SuperGLUE & 4 & 8 & Word sense disambiguation, Natural language inference, Coreference resolution, Question answering \\ 
            TabMWP & 1 & - & Mathematical reasoning and table QA \\ 
            TheoremQA & 4 & 39 & Mathematics, Physics, Finance, CS \& EE \\ 
            ToolBench & 8 & - & Open weather, The cat API, Home search, Trip booking, Google sheets, Virtual home, Web shop, Tabletop \\ 
            TRUSTGPT & 3 & - & Toxicity, Bias, Value-alignment \\ 
            TruthfulQA & 38 & - & Health, Law, Conspiracies, Fiction, Misconceptions, Paranormal, Economics, Biology, Language, Indexical etc. \\ 
            Vicuna Evaluation & 9 & - & Generic, Knowledge, Roleplay, Common-sense, Fermi, Counterfactual, Coding, Math, Writing \\ 
            XiezhiBenchmark & 13 & 516 & Medicine, Literature, Economics, Agronomy, Science, Jurisprudence, History, Art studies, Philosophy, Pedagogy, Military science, Management, Engineering \\ 
            XNLI & 1 & - & Multilingual natural language inference \\ 
            XTREME & 4 & 9 & Classification, Structured prediction, QA, Retrieval \\ 
            ZeroSCROLLS & 3 & 10 & Summarization, Question Answering, Aggregation \\ 
            \hline
            \end{tabular}
        }
\end{table*}

\subsection{Evaluation Domains}\label{subsec51}

\cite{bib10} categorizes LLM evaluations into five evaluation categories based on different dimensions: knowledge and capability evaluation, alignment evaluation, safety evaluation, specialized LLMs evaluation, and evaluation organization. As shown in Figure~\ref{fig18}, this paper focuses on the key evaluation domains of each evaluation dataset, finely categorizing 112 datasets into 20 evaluation domains, namely: \textbf{General}, \textbf{Exam}, \textbf{Subject}, \textbf{Natural Language Understanding (NLU)}, \textbf{Reasoning}, \textbf{Knowledge}, \textbf{Long Text}, \textbf{Tool}, \textbf{Agent}, \textbf{Code}, \textbf{OOD}, \textbf{Law}, \textbf{Medical}, \textbf{Financial}, \textbf{Social Norms}, \textbf{Factuality}, \textbf{Evaluation}, \textbf{Multitask}, \textbf{Multilingual}, and \textbf{Other}. The “Other” category includes seven sub-domains: E-commerce, Few-shot learning, Geoscience, IT, Multi-turn interactions, Robustness, and Sentiment. Each sub-domain has one representative dataset.

\subsubsection{General}\label{subsubsec511}

General evaluation datasets are typically of smaller scale and primarily focus on assessing how well LLMs perform in two aspects. The first aspect involves \textbf{evaluating their performance on general instructions across multiple categories and domains, mainly examining their versatility}. Vicuna Evaluation\footnote{\href{https://github.com/lm-sys/vicuna-blog-eval}{https://github.com/lm-sys/vicuna-blog-eval}}, for instance, assesses models in nine question categories, using GPT-4 to judge the quality of outputs and providing a preliminary evaluation of overall model quality. Building upon this, AlpacaEval \citep{bib156} includes instructions covering samples from various datasets, offering a broader evaluation of performance on different open-ended questions. BayLing-80 \citep{bib157} further expands Vicuna Evaluation, evaluating the general capabilities and conversational abilities of LLMs in both Chinese and English. BELLE\_eval \citep{bib158} and MT-Bench \citep{bib143} adopt similar evaluation methods. The former aims to assess the models’ general performance in Chinese scenarios across nine instruction types, while the latter focuses on evaluating their general performance in English scenarios across eight instruction types. The number of instructions in these datasets is all within 1K, with some limitations in comprehensiveness. SuperCLUE \citep{bib159} expands the scale of evaluation content. It serves as a comprehensive evaluation benchmark for Chinese general LLMs, designed to assess the effectiveness of current Chinese LLMs. The tasks include multi-turn open-ended Q\&A and objective multiple-choice Q\&A, with monthly updates and significant reference value.

The second aspect involves \textbf{assessing the ability of LLMs to follow instructions, especially when faced with complex directives}. Datasets like Vicuna Evaluation, AlpacaEval, and BayLing-80 incorporate various types of instructions, evaluating both generalization and the models’ capacities to comprehend the requirements of instructions. CELLO \citep{bib160} enhances the complexity of instructions by systematically evaluating the models’ capabilities to follow complex directives from two perspectives: complex task description and complex input.

\subsubsection{Exam}\label{subsubsec512}

Evaluation datasets within the examination domain are crafted with the specific purpose of formulating instructions derived from significant exam questions across diverse nations. In this scenario, LLMs take on the role of candidates, responding to queries in accordance with specified guidelines. The primary objective is to assess the proficiency of LLMs in comprehending the nuances of question intent and their reservoir of knowledge pertaining to examinations. GAOKAO-Bench \citep{bib161} employs Gaokao (China’s National College Entrance Examination) questions as the basis for evaluation, seeking to appraise the proficiency of LLMs across various subjects, encompassing a spectrum of 10 disciplines. AGIEval \citep{bib162} expands the ambit of inquiries by devising benchmarks centered on human-centric tests, featuring a selection of 20 official, public, and stringent entrance and qualification examinations, including Gaokao, the U.S. SAT, the bar exam, and the national civil service exam. M3Exam \citep{bib163} assembles an array of multi-modal, multi-lingual, and multi-tiered sets of multiple-choice questions, sourcing exam questions from primary, secondary, and high school exams in nine countries distinguished by diverse languages.

\subsubsection{Subject}\label{subsubsec513}

Evaluation datasets in academic domains thoroughly gauge the mastery of LLMs in diverse fields, including disciplines like mathematics, law, psychology, and more. C-CLUE\footnote{\href{https://github.com/jizijing/C-CLUE}{https://github.com/jizijing/C-CLUE}} stands as a benchmark for assessing classical Chinese language comprehension. It centers on tasks like NER and RE, all grounded in a historical knowledge graph. This dataset primarily scrutinizes proficiency within individual disciplines, yet it exhibits limited diversity. MMCU \citep{bib164} broadens the horizons by incorporating disciplines such as medicine, law, psychology, and education to measure Chinese semantic comprehension. In the realm of university-level science and engineering, SCIBENCH \citep{bib165} is tailor-made to evaluate LLMs’ capabilities, demanding the resolution of challenging subjective questions related to mathematics, physics, and chemistry. TheoremQA \citep{bib166} narrows its focus to 350 theorems from mathematics, physics, finance, and CS \& EE (Computer Science \& Electrical Engineering). Lastly, ARB \citep{bib167} introduces a more demanding examination, appraising LLMs’ prowess in text comprehension and domain-specific reasoning. The questions delve into profound knowledge across disciplines such as mathematics, physics, biology, chemistry, and law.

The aforementioned datasets focus on evaluating specific disciplines on a smaller scale. In contrast, some datasets aim to comprehensively assess disciplinary capabilities, encompassing a wide range of subjects. ScienceQA \citep{bib168} gathers multiple-choice questions from 26 subcourses in natural sciences, social sciences, and linguistics. C-Eval \citep{bib169} compiles 52 diverse subject questions, categorized into four difficulty levels, providing a holistic evaluation of models’ comprehensive subject proficiency in Chinese. Similarly, CG-Eval \citep{bib170} requires LLMs to accurately answer 55 sub-subject questions across six major categories for automatic scoring. LLMEVAL-3\footnote{\href{https://github.com/llmeval/llmeval-3}{https://github.com/llmeval/llmeval-3}} concentrates on evaluating proficiency in specialized knowledge, featuring generated questions from 13 academic categories outlined by China’s Ministry of Education and over 50 subcategories. It introduces a “question bank exam” mode. MMLU \citep{bib171} assesses subjects ranging from traditional fields like mathematics and history to professional areas such as law and ethics, covering 57 subjects with difficulty levels from elementary to professional. As the content of MMLU is in English, CMMLU \citep{bib172} is created as its Chinese counterpart for evaluating subject knowledge proficiency in a Chinese context, covering 67 subjects. M3KE \citep{bib173}, originating from the Chinese education system, collects multiple-choice questions from 71 subjects spanning from primary school to university. XiezhiBenchmark \citep{bib174}, covering a record 516 different subjects, attains a scale of approximately 250K questions. Overall, these subject evaluation datasets share a high degree of similarity in data sources, primarily sourced from online materials related to their respective subjects. Additionally, multiple-choice question formats, conducive to automated evaluation, are particularly favored.

\subsubsection{Natural Language Understanding}\label{subsubsec514}

This class of evaluation datasets aims to comprehensively evaluate the multifaceted abilities of LLMs in natural language understanding (NLU) tasks, covering fundamental comprehension of grammatical structures to advanced semantic reasoning and context handling. MCTS \citep{bib175} and RAFT \citep{bib176} serve as benchmarks for individual NLU tasks. The former stands as the most extensive evaluation dataset for Chinese text simplification, while the latter functions as a benchmark for text classification. Multiple NLU tasks are encompassed by most datasets. GLUE \citep{bib177} incorporates nine English NLU tasks, assessing LLMs in tasks such as sentiment analysis, semantic matching, and textual entailment. Building upon GLUE, SuperGLUE \citep{bib178} raises task difficulty, reflecting LLMs’ performance in a broader scope of language understanding. To evaluate the NLU capabilities of models in the Chinese context, CLUE \citep{bib179} is constructed with reference to GLUE. Comprising nine Chinese NLU tasks, the CLUE dataset evaluates LLMs in tasks like semantic matching, text classification, and reading comprehension. CUGE \citep{bib180} is organized hierarchically by language-task-dataset structure, using 21 sub-datasets to evaluate LLMs in language understanding, information retrieval, Q\&A, and language generation. SentEval \citep{bib181} aggregates NLU datasets for 21 sub-tasks.

\subsubsection{Reasoning}\label{subsubsec515}

Reasoning evaluation datasets are designed to gauge the proficiency of LLMs in tasks related to logical reasoning and inference. Chain-of-Thought Hub \citep{bib182} selects eight open-source datasets and evaluates LLMs’ multi-step reasoning performance by utilizing few-shot CoT prompting across domains like mathematics, science, and symbols. Choice-75 \citep{bib183} tasks LLMs with selecting an appropriate decision solution in various given scenarios, assessing their competence in decision reasoning. NeuLR \citep{bib184} assesses deductive reasoning, inductive reasoning, and abductive reasoning, emphasizing LLMs’ capabilities in these distinct reasoning directions. TabMWP \citep{bib185}, LILA \citep{bib186}, and miniF2F\_v1 \citep{bib187} all scrutinize LLMs’ reasoning prowess in mathematics. The TabMWP dataset requires LLMs to engage in table-based Q\&A and mathematical reasoning based on provided text and table data. The LILA dataset serves as a comprehensive mathematical reasoning benchmark, evaluating various mathematical skills, including basic proficiency, algebra, calculus, and more. The miniF2F\_v1 dataset is a compilation of Olympiad-level mathematical problems, posing a substantial challenge to the mathematical acumen of LLMs. In summary, reasoning evaluation datasets encompass diverse assessment directions, categorized into multi-step reasoning, decision reasoning, deductive reasoning, mathematical reasoning, and other forms of reasoning.

\subsubsection{Knowledge}\label{subsubsec516}

Datasets for evaluating knowledge not only gauge the knowledge retention capabilities of LLMs but also assess additional skills such as knowledge analysis, learning novel information, and knowledge induction. LLMEVAL-2 \citep{bib188}, derived from external databases, constructs a repository of knowledge questions across 12 domains. Curated by GPT-4, LMExamQA \citep{bib189} categorizes questions based on the requisite knowledge level, spanning memorization, comprehension, and analysis. KoLA \citep{bib190} predominantly examines LLMs’ proficiency in grasping and applying world knowledge, categorized into memory, comprehension, application, and creation according to the cognitive hierarchy of knowledge. Serving as an assessment benchmark for LLMs’ command of social knowledge, SocKET \citep{bib191} classifies knowledge into humor and satire, aggressiveness, emotion, credibility, and social facts. While previous datasets evaluate models from the perspective of existing knowledge, the challenge lies in appraising the models’ learning abilities with entirely unfamiliar new knowledge. Hence, \cite{bib192} employs the knowGen method to generate new knowledge, resulting in the inaugural benchmark dataset, ALCUNA \citep{bib192}, for evaluating and scrutinizing the models’ understanding, differentiation, and association capabilities regarding new knowledge.

\subsubsection{Long Text}\label{subsubsec517}

In recent times, numerous LLMs, including ChatGLM2\footnote{\url{https://github.com/THUDM/ChatGLM2-6B}} and Gemini 1.5\footnote{\url{https://deepmind.google/technologies/gemini/\#introduction}}, have sought to expand the context length of models to the scale of millions of tokens while maintaining performance \citep{bib193}. This has given rise to the development of long text evaluation datasets to better assess the capabilities of LLMs in processing and understanding extensive textual inputs. Notable datasets in this domain include ZeroSCROLLS \citep{bib194}, L-Eval \citep{bib195}, LongEval \citep{bib196}, and LooGLE \citep{bib197}, all focusing on the evaluation of lengthy English texts. ZeroSCROLLS standardizes datasets from diverse sources into a consistent input format with an average length of 10K words for assessment across 10 natural language tasks. L-Eval serves as a comprehensive evaluation suite for long-context language models, covering input lengths ranging from 4K to 60K words. It encompasses 18 multi-domain tasks involving inference, Q\&A, summarization, and more on long documents. LongEval introduces two tasks of varying difficulty, gauging LLM performance in fine-grained topic retrieval and line retrieval with input lengths between 5K and 16K tokens. LooGLE focuses on more challenging tasks with long dependencies, evaluating performance on tasks such as multiple information retrieval and timeline reorder with an average length of 20K words. In contrast, LongBench \citep{bib193} comprises a diverse set of 14 English tasks, 5 Chinese tasks, and 2 code tasks, with most tasks exhibiting an average length between 5K and 15K tokens. Despite claims of some models supporting 100K+ contexts, the previously mentioned datasets reveal limitations in evaluating such lengths. To address this, InfiniteBench \citep{bib198} increases the average length of evaluations in both Chinese and English to 200K tokens, introducing 10 new tasks among the set of 12 evaluation tasks to fill the void in assessing long texts exceeding 100K tokens.

\subsubsection{Tool}\label{subsubsec518}

The datasets for evaluating tools gauge the adeptness of LLMs in utilizing tools and invoking APIs. API-Bank \citep{bib199} replicates real-world scenarios, establishing an API library with 53 commonly used tools for LLMs to call upon. Tasks involving API invocation are designed to assess the models’ abilities to effectively use APIs in fulfilling user requirements within a given conversational context. APIBench \citep{bib200}, crafted for evaluation purposes, generates 16,450 instructions derived from 1,645 API documents. These instructions are formatted to suit LLM-friendly chat interactions and are accompanied by evaluation scripts. ToolBench \citep{bib201}, functioning as a benchmark for tool operations, encompasses a variety of software tools employed in real-world tasks. Tool invocations span single-step and multi-step action generation, covering eight subtasks, including open weather and webshop.

\subsubsection{Agent}\label{subsubsec519}

The research and application of LLMs as AI Agents, exemplified by entities like AutoGPT\footnote{\href{https://github.com/Significant-Gravitas/AutoGPT}{https://github.com/Significant-Gravitas/AutoGPT}} and AgentGPT\footnote{\href{https://github.com/reworkd/AgentGPT}{https://github.com/reworkd/AgentGPT}}, are continuously advancing. Agent evaluation datasets specifically concentrate on the capabilities of LLMs functioning as Agents. AgentBench \citep{bib202} undergoes assessment within English scenarios. It stands out as the inaugural benchmark designed to evaluate the performance of LLM-as-Agent across various environments, encompassing eight distinct settings and providing a thorough examination of LLMs’ competence as independent agents. SuperCLUE-Agent\footnote{\href{https://github.com/CLUEbenchmark/SuperCLUE-Agent}{https://github.com/CLUEbenchmark/SuperCLUE-Agent}} is subjected to evaluation within the Chinese context. This dataset gauges the Agent capabilities of LLMs in a Chinese context through three core abilities and ten foundational tasks, covering aspects such as tool usage, task planning, and both short-term and long-term memory.

\subsubsection{Code}\label{subsubsec5110}

The coding evaluation datasets aim to assess the capabilities of LLMs in handling programming-related tasks, including but not limited to code interpretation, code generation, code correction, and code optimization. These datasets are primarily categorized into two types. The first type is \textbf{single-task evaluation}. APPS \citep{bib203} serves as a benchmark for code generation, specifically evaluating the ability to generate Python code. Other datasets such as DS-1000 \citep{bib204}, HumanEval \citep{bib205}, MTPB \citep{bib32}, and ODEX \citep{bib206} investigate code generation abilities in different forms. DS-1000 introduces data science problems related to seven Python libraries. HumanEval assesses LLMs using manually written programming problems, mitigating data leakage concerns to some extent. MTPB tasks LLMs with synthesizing a subroutine at each step, requiring consideration of both the current task description and previous steps. ODEX extends the variety of natural languages, using English, Spanish, Japanese, and Russian to describe code intent, evaluating LLMs’ abilities to generate code under multilingual descriptions. Additionally, BIRD \citep{bib207} is a large-scale database benchmark for text-to-SQL (Structured Query Language) tasks that, compared to previous popular datasets like Spider \citep{bib208}, reduces the gap between academic research and practical applications, enhancing the level of difficulty. The second type is \textbf{multi-task evaluation}. CodeXGLUE \citep{bib209} categorizes code abilities into four types based on input-output pairs: code-code, text-code, code-text, and text-text. HumanEvalPack \citep{bib210} is an extension of the HumanEval, covering six programming languages and three code tasks, including code fixing, code comment generation, and code generation.

\subsubsection{Out-of-Distribution}\label{subsubsec5111}

The out-of-distribution (OOD) evaluation dataset is designed to gauge the capabilities of pre-trained base models after fine-tuning with instructions from a subset of tasks on previously unseen tasks. The emphasis is on scrutinizing the robustness of LLMs. \cite{bib211} conducted experiments on the BOSS dataset \citep{bib211}, encompassing 5 tasks and 20 sub-datasets, to scrutinize the OOD performance of LLMs. \cite{bib212} employed GLUE-X \citep{bib212} to assess the models’ OOD performance and offered insights into the measurement and enhancement of model OOD performance.

\subsubsection{Law}\label{subsubsec5112}

Legal evaluation datasets play a crucial role in the application of LLMs in the legal domain by providing standardized performance assessments and driving research and development in legal LLMs. The datasets can be categorized based on the linguistic environment they target. LAiW \citep{bib213} and LawBench \citep{bib214} are designed for the Chinese language environment. LAiW serves as a Chinese legal LLMs evaluation benchmark, focusing on 13 foundational tasks across three legal competencies. It compares LLMs in terms of NLP basic capabilities, fundamental application abilities, and complex application capabilities. LawBench, benchmarked on the Chinese legal system, evaluates LLMs’ legal abilities across 20 tasks simulating knowledge retention, understanding, and application, closely related to real-world applications.

In the English language environment, LegalBench \citep{bib215} and LexGLUE \citep{bib216} are relevant. LegalBench, constructed with the assistance of cross-disciplinary professionals, is a legal reasoning benchmark comprising six types of legal reasoning and 162 tasks. LexGLUE integrates open-source English legal datasets as an evaluation benchmark, examining legal Q\&A and classification tasks.

For a multilingual environment, LEXTREME \citep{bib217} and SCALE \citep{bib218} are applicable. LEXTREME divides 18 legal-related tasks from 11 open-source datasets, covering 24 languages. SCALE challenges current LLMs in four dimensions: handling long documents, applying legal knowledge, multilingual comprehension, and multitask processing. The benchmark is derived from the Swiss legal system, involving five languages.

\subsubsection{Medical}\label{subsubsec5113}

The medical evaluation datasets focus on examining the comprehensive capabilities of LLMs in medical tasks such as term explanation, disease diagnosis, and treatment recommendations. This enables a comparison of the proficiency gap between various medical models and professional doctors. MultiMedQA \citep{bib219} serves as an evaluation benchmark for medical Q\&A, blending multiple open-source datasets and proprietary datasets to assess LLMs’ abilities to address medical queries. QiZhenGPT-eval\footnote{\href{https://github.com/CMKRG/QiZhenGPT/tree/main/data/eval}{https://github.com/CMKRG/QiZhenGPT/tree/main/data/eval}} focuses on drug indication evaluation, tasking LLMs with identifying diseases for which a given drug is suitable. However, single-task datasets are overly restrictive in evaluation dimensions and may not reflect other medical competencies. Consequently, various integrated datasets have been gradually proposed.

CBLUE \citep{bib220} is an evaluation dataset for Chinese medical language understanding, presenting five medical tasks using authentic medical data. It assesses LLMs in medical text information extraction and medical Q\&A. The design of CMB \citep{bib221} is based on the Chinese language and cultural framework, evaluating LLMs from the perspective of Chinese-style medical exams and complex clinical diagnoses. HuaTuo26M-test \citep{bib110} is randomly sampled from various sources, including medical encyclopedias and knowledge graphs, offering diverse task types. PromptCBLUE\footnote{\href{https://github.com/michael-wzhu/PromptCBLUE}{https://github.com/michael-wzhu/PromptCBLUE}} transforms 16 different NLP tasks in medical scenarios into an evaluation format, forming the first systematic Chinese benchmark for medical scenarios.

\subsubsection{Financial}\label{subsubsec5114}

The financial evaluation dataset, akin to the legal and medical evaluation datasets mentioned in previous sections, focuses on knowledge related to the financial domain, assessing the performance of LLMs in handling financial texts and executing financial tasks. BBF-CFLEB \citep{bib43} encompasses six sub-datasets for financial tasks, strategically evaluating the language understanding and language generation capabilities of financial models from multiple perspectives. Both FinancelQ\footnote{\href{https://github.com/Duxiaoman-DI/XuanYuan/tree/main/FinanceIQ}{https://github.com/Duxiaoman-DI/XuanYuan/tree/main/FinanceIQ}} and FinEval \citep{bib223} emphasize knowledge and reasoning abilities in financial scenarios, incorporating multiple-choice questions on different financial topics to assess LLMs’ financial knowledge. While the preceding datasets target the Chinese environment, FLUE \citep{bib224} serves as an English-oriented testing benchmark, amalgamating six financial NLP datasets with a focus on NLU in the financial domain. FinBen \citep{bib413} is also an English benchmark dataset for evaluating the capabilities of LLMs in the financial domain. It gathers 35 existing datasets covering 23 financial tasks, categorized into three difficulty levels: foundamental tasks, advanced cognitive engagement, and general intelligence.

\subsubsection{Social Norms}\label{subsubsec5115}

The assessment dataset for societal norms evaluates LLMs across dimensions such as ethics, morality, prejudice, toxicity, and safety. It primarily investigates whether the models generate outputs that violate ethical and legal standards, display biased discrimination, or produce toxic and harmful content in response to unsafe instructions. Datasets of this nature hold significant importance and societal value in the safety scrutiny of LLMs. CrowS-Pairs \citep{bib225} assesses LLMs for biases and discrimination within the context of American culture, encompassing nine stereotypes related to prejudice, including race, religion, age, and more. SafetyBench \citep{bib226} stands as the inaugural benchmark for evaluating LLM safety through multiple-choice questions in both Chinese and English, covering seven distinct safety dimensions. Safety-Prompts \citep{bib227}, featuring 13 safety scenarios and prompt attack evaluation data generated by ChatGPT, enables a comprehensive evaluation of the models’ safety. However, constrained by ChatGPT’s performance, occasional errors may be present in questions or answers. TRUSTGPT \citep{bib228} evaluates LLMs in three crucial domains: toxicity, bias, and value consistency. Compared to previous mainstream safety benchmarks, SuperCLUE-Safety\footnote{\href{https://github.com/CLUEbenchmark/SuperCLUE-safety}{https://github.com/CLUEbenchmark/SuperCLUE-safety}} introduces heightened challenges by incorporating adversarial techniques and multi-turn interactions, thereby enhancing the identification of LLM safety protection capabilities under various adverse inputs.

\subsubsection{Factuality}\label{subsubsec5116}

The outputs produced by LLMs may exhibit deviations from the specified input criteria, preceding contextual information, or established facts and knowledge—a phenomenon commonly known as the hallucination of LLMs \citep{bib229}. Addressing this issue necessitates the use of datasets designed for factual evaluation to gauge the extent of hallucination in LLMs. There are three distinct forms of evaluating the factual accuracy of LLMs.

The first method entails \textbf{the presentation of various options, prompting LLMs to discern the factually correct choice among alternatives or to assess the factual alignment of the provided content}. In the FACTOR dataset \citep{bib230}, each instance comprises a prefix and four completions, with only one completion being factually accurate. LLMs are required to identify the accurate choice based on the given prefix and pertinent knowledge. HaluEval \citep{bib231} furnishes inputs and outputs for tasks like Q\&A, dialogue, and text summarization, challenging LLMs to recognize the existence of hallucination.

The second method entails \textbf{assessing the factual accuracy of open-ended content generated by LLMs}. FActScore \citep{bib232} employs information from biographies to create a factual evaluation dataset, incorporating novel evaluation techniques for appraising the factual precision of LLMs in producing extensive content. FactualityPrompt \citep{bib233} similarly evaluates the factual aspects of LLMs in open-text generation, demanding the generation of accurate content under genuine and non-genuine prompts.

The third method involves \textbf{interrogating LLMs to assess the prevalence of hallucinatory phenomena}. TruthfulQA \citep{bib234} meticulously devises English questions prone to generating erroneous answers due to potential misunderstandings, evaluating the veracity of LLMs’ responses. Taking cues from this, HalluQA \citep{bib235} formulates Chinese questions designed to mislead Chinese LLMs, evaluating the hallucinatory tendencies in Chinese LLMs. FreshQA \citep{bib236} acts as a dynamic benchmark for factual Q\&A, necessitating not only a mastery of rapidly evolving world knowledge but also the ability to refute incorrect factual premises.

\subsubsection{Evaluation}\label{subsubsec5117}

The rise of LLMs has ushered in a fresh paradigm for model evaluation, allowing proficient LLMs to act as evaluators in scoring the outputs of other models. However, the reliability of involving LLMs in assessments and the performance variability among different LLMs in appraising the quality of model responses prompt inquiries. Consequently, datasets falling under the evaluation category are specifically tailored to probe into the potential and competence of LLMs as evaluators. FairEval \citep{bib237} critically examines the model evaluation paradigm to explore the dependability of LLMs as assessors. It utilizes the Vicuna Evaluation dataset\footnote{\url{https://github.com/lm-sys/vicuna-blog-eval}} as instructions, generating responses from various models, and subsequently engages models such as ChatGPT, GPT-4, and others to evaluate diverse responses. PandaLM\_testset \citep{bib238}, enriched with human annotations, serves to validate the assessment capabilities of trained PandaLM \citep{bib238} when evaluating other LLMs. $\mathrm{LLMEval}^2$ \citep{bib239}, currently the largest and most diversified English benchmark for evaluating LLMs, spans 15 tasks and 8 abilities, employing innovative methods to gauge the quality of LLMs’ evaluation responses.

\subsubsection{Multitask}\label{subsubsec5118}

Multitask evaluation datasets present a thorough examination of LLMs’ comprehensive capabilities, characterized by a substantial task volume, extensive scale, broad domains, and diverse task types. In the realm of English, DecaNLP \citep{bib240} transforms 10 distinct task datasets into a Q\&A format, introducing the “Decathlon” multitask challenge within the natural language domain. LMentry \citep{bib241} provides a swift, automated “unit test,” assessing LLMs’ performance across 25 task types that are relatively simple for human understanding. However, these datasets still lack task type richness. BIG-Bench \citep{bib242} impressively includes 95 task types, totaling 204 tasks, covering a wide array of topics such as linguistics, common-sense reasoning, social biases, software development, and more. BBH \citep{bib243} carefully selects 23 challenging tasks from BIG-Bench, where previous language models have not surpassed average human performance, presenting a considerable challenge. HELM \citep{bib244} contemplates holistic model evaluation, establishing a comprehensive evaluation system for LLMs with 73 evaluation scenarios and 65 evaluation metrics, ensuring a thorough and rigorous assessment. 

In the Chinese domain, CLEVA \citep{bib245} stands as a comprehensive Chinese evaluation benchmark, featuring 11 application assessment tasks and 20 capability assessment tasks, with a scale reaching 370K. CLiB\footnote{\href{https://github.com/jeinlee1991/chinese-llm-benchmark}{https://github.com/jeinlee1991/chinese-llm-benchmark}} serves as a Chinese proficiency test list for LLMs, covering LLMs such as GPT-4, ERNIE Bot \citep{bib246}, QWen \citep{bib247}, and supporting multidimensional capability evaluations like classification and information extraction. LLMEVAL-1 \citep{bib248}, comprising 17 task categories, 5 scoring items, and various evaluation methods, systematically evaluates LLMs. Furthermore, FlagEval\footnote{\href{https://github.com/FlagOpen/FlagEval}{https://github.com/FlagOpen/FlagEval}} scrutinizes the models’ comprehensive performance in both Chinese and English environments, serving as an evaluation toolkit for AI base models capable of assessing over 600 sub-dimensions of base models.

\subsubsection{Multilingual}\label{subsubsec5119}

Multilingual evaluation datasets assess the performance of LLMs in cross-lingual tasks using data encompassing multiple languages, contributing to the exploration of LLMs’ capabilities across diverse linguistic challenges. XNLI \citep{bib249} is specialized for evaluating low-resource language transfer and cross-lingual sentence classification, incorporating 15 languages, including English, French, Spanish, Chinese, and German. Conversely, XTREME \citep{bib250} expands language coverage by translating content for four NLP tasks into 40 languages, crossing 12 language families. In essence, multilingual evaluation datasets typically build on traditional NLP tasks, extend language diversity, maintain a moderate task difficulty, and necessitate a wealth of language knowledge.

\subsubsection{Other}\label{subsubsec5120}

Apart from the aforementioned assessment datasets, there exist several datasets specifically dedicated to diverse domains, addressing deficiencies in the evaluation landscape. The subsequent section provides an overview of pivotal datasets within seven subdomains for reference.

\textbf{E-commerce Domain.} The EcomGPT\_eval dataset \citep{bib251} is designed to evaluate the efficacy of LLMs in tasks within the realm of e-commerce. It consists of 6K instances, with 500 instances sampled from each of the 12 held-out datasets tailored for e-commerce evaluation. Tasks in the e-commerce domain are classified into four categories: classification, generation, extraction, and miscellaneous. These tasks span coarse and fine-grained product classification, product title generation, attribute value detection, and e-commerce NER, among others.

\textbf{Few-shot Learning Domain.} The FewCLUE dataset \citep{bib252} has been created with a specific focus on assessing few-shot learning in the Chinese language. Its purpose is to leverage the generalization capabilities of pre-trained models and investigate the practicality of few-shot learning models applied to Chinese. The dataset is composed of nine sub-datasets, with some containing slightly over a hundred annotated samples, providing a means to evaluate model generalization under conditions of extremely limited labeled data.

\textbf{Geoscience Domain.} The GeoBench dataset \citep{bib132} serves as a means to evaluate the proficiency of language models in tackling questions related to geoscience, assessing their capacity to comprehend and apply knowledge in this domain. The dataset is bifurcated into two sections. The initial segment comprises questions from the Chinese graduate entrance examination in geology and geography, encompassing 182 multiple-choice questions, 150 fill-in-the-blank questions, 454 vocabulary explanation questions, and 335 essay questions. The subsequent segment includes 1,395 multiple-choice questions from advanced research examinations in the United States.

\textbf{IT Domain.} The Owl-Bench dataset \citep{bib135} serves as a bilingual evaluation benchmark tailored for IT operations and maintenance contexts. It encompasses 317 questions and answers, in addition to 1K multiple-choice questions. The tasks address numerous real-world industrial scenarios, spanning nine distinct sub-domains: information security, applications, system architecture, software architecture, middleware, networks, operating systems, infrastructure, and databases.

\textbf{Multi-turn Interaction Domain.} LLMs frequently interact with users across multiple turns, yet assessments typically focus on individual turns, overlooking their interactive capabilities. Thus, the MINT dataset \citep{bib253} is designed to evaluate LLMs in tasks involving multi-turn interactions, employing tools or utilizing natural language feedback. In this evaluation framework, the model being tested can access tools through the execution of Python code, receiving feedback simulated by GPT-4 to facilitate multi-turn interactive assessments.

\textbf{Robustness Domain.} The PromptBench dataset \citep{bib254} extensively explores the robustness of LLMs when confronted with seven distinct types of adversarial prompts. Simultaneously, it performs an analysis of the transferability of adversarial prompts generated by various models. The examination of robustness encompasses eight diverse NLP tasks across thirteen open-source datasets, encompassing domains like sentiment analysis, multi-task knowledge, reading comprehension, mathematics, and beyond.

\textbf{Sentiment Domain.} The EmotionBench dataset \citep{bib255} presents a pioneering benchmark for assessing the empathetic abilities of LLMs, examining how LLMs undergo emotional changes in response to particular situations. Encompassing more than 400 scenarios, the dataset generates eight distinct emotional categories: anger, anxiety, depression, frustration, jealousy, guilt, fear, and embarrassment.

\subsection{Evaluation Methods}\label{subsec52}

\begin{figure}
\centering
\includegraphics[width=0.9\textwidth]{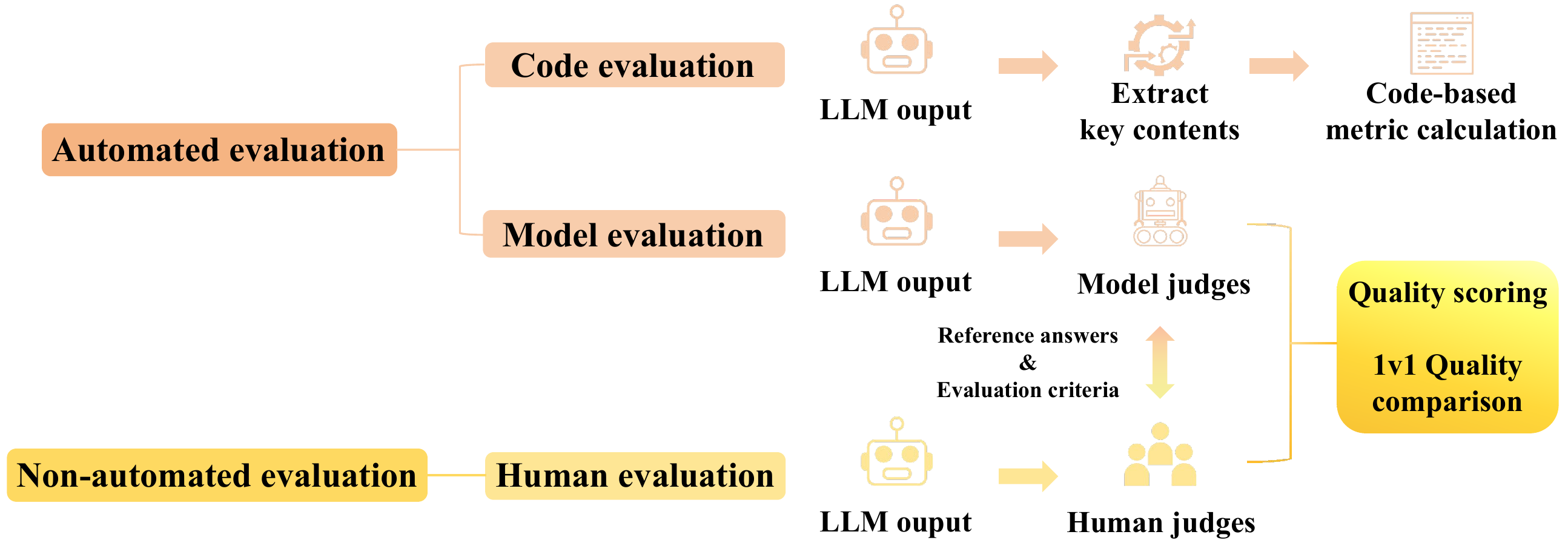}
\caption{Three evaluation methods}\label{fig19}
\end{figure}

In this section, evaluation methods are classified into three types: \textbf{code evaluation}, \textbf{human evaluation}, and \textbf{model evaluation}. Figure~\ref{fig19} illustrates these three evaluation methods. Notably, code evaluation and model evaluation operate with minimal human intervention, with evaluation results being automatically computed and generated through the pipeline. These two methods are categorized as automated evaluation. In contrast, human evaluation is characterized as a non-automated approach.

The approach of code evaluation entails the extraction of responses from LLMs, referencing authentic annotations, and utilizing code to statistically compute predefined evaluation metrics. The efficacy of LLMs is consequently gauged through the numerical values of these metrics. Prominent evaluation metrics include accuracy, F1 score, BLEU \citep{bib256}, ROUGE \citep{bib257}, Exact Match\footnote{\href{https://huggingface.co/spaces/evaluate-metric/exact_match}{https://huggingface.co/spaces/evaluate-metric/exact\_match}}, Pearson correlation coefficient\footnote{\href{https://libguides.library.kent.edu/SPSS/PearsonCorr}{https://libguides.library.kent.edu/SPSS/PearsonCorr}}, among others. For instance, accuracy can be employed in classification tasks to appraise the precision of LLMs’ classifications. In translation tasks, BLEU serves to assess the resemblance between LLMs’ translations and authentic annotations. Certain evaluation datasets not only provide custom calculation methods but also furnish pertinent code, facilitating direct application for the evaluation and analysis of LLMs’ performance. This evaluation methodology is commonly used for objective questions and straightforward subjective questions with predefined answers, such as basic knowledge queries and translation exercises. While its simplicity is beneficial, it may not be as effective for assessing open-ended subjective questions such as those involve generation and brainstorming.

The human evaluation approach, on the other hand, often involves the evaluation of LLM outputs by crowdsourced individuals, trained volunteers, students with relevant expertise, or expert panels. Evaluation methods include quality scoring (as seen in the QizhenGPT\_eval dataset\footnote{\url{https://github.com/CMKRG/QiZhenGPT/tree/main/data/eval}} and the CLiB dataset\footnote{\url{https://github.com/jeinlee1991/chinese-llm-benchmark}}), quality comparison assessment \citep{bib159}, and similar techniques. This manual evaluation method is versatile, suitable for various question types, especially open-ended subjective inquiries and complex problems lacking standard answers. Nevertheless, its limitation lies in the substantial costs, the need for extensive human resources, and a potential for subjective bias.

The method of evaluating models represents a novel paradigm in which questions, reference answers, evaluation criteria and standards, along with the responses of the tested models, are integrated into an optimal prompt. This combined information is then inputted to the model for evaluation \citep{bib158,bib143,bib157,bib156,bib235,bib189,bib135}. This evaluation approach emphasizes the selection of LLMs with currently high performance and provides suitable evaluation instructions. Its advantage lies in its capacity to substitute for a considerable amount of manual effort, resulting in a quicker evaluation process. Nevertheless, the limitation lies in the dependency on the LLMs’ performance and may not always correspond with human values and judgements.

It is increasingly common to employ a mix of multiple assessment methods \citep{bib195,bib161,bib227,bib167,bib199,bib221,bib232,bib132,bib244,bib215,bib248,bib188,bib219,bib159,bib234}, leveraging the strengths and mitigate the weaknesses of each method. This approach aims to achieve a comprehensive, rigorous, and standardized evaluation.

\subsection{Distribution Statistics of Evaluation Datasets}\label{subsec53}

Figure~\ref{fig20} provides statistics on 112 evaluation datasets from eight aspects: release time, license, size, construction method, language, domain, question type, and evaluation method. Based on these statistics, the following conclusions can be drawn:

\begin{figure}
\centering
\includegraphics[width=0.9\textwidth]{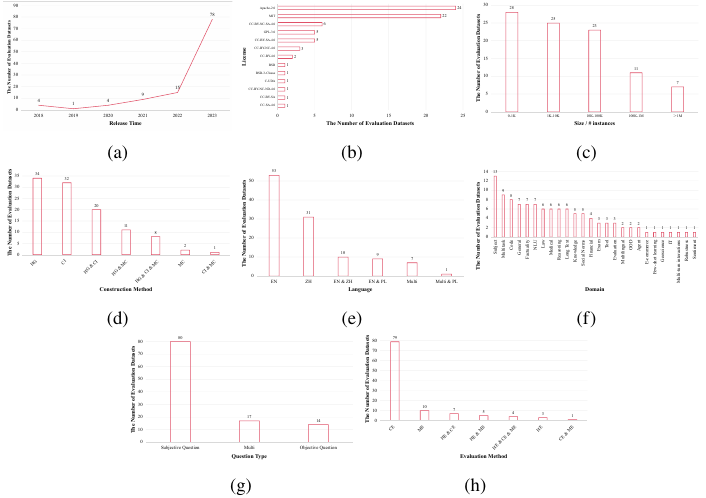}
\caption{Statistics distribution of evaluation datasets. (a) illustrates the quantity trend over time. (b) depicts the quantity distribution under different licenses, considering only the datasets with listed licenses. (c) shows the quantity distribution across different data scales. (d) displays the quantity distribution for different construction methods. (e) represents the quantity distribution across different languages. (f) illustrates the quantity distribution across different domains. (g) indicates the distribution based on various question types; and (h) outlines the distribution employing diverse evaluation methods. Zoom in for better view}\label{fig20}
\end{figure}

(1) There is a noticeable upward trend in the evaluation datasets. The ongoing maturation of technologies related to LLMs is driving the expansion of datasets tailored for LLMs evaluation. Specifically, in the year 2023, there has been a significant surge in the number of evaluation datasets, reflecting the need for diverse datasets to keep pace with the rapid iteration of LLMs and to improve model performance.

(2) The distribution of evaluation dataset licenses shows a preference for widely recognized licenses such as, Apache-2.0 and MIT. The overall pattern of distribution in these protocols underscores the delicate equilibrium sought within the LLMs data evaluation domain, balancing knowledge sharing and intellectual property protection. The flexibility provided by open licenses such as Apache-2.0 and MIT contributes to the widespread use and sharing of evaluation datasets, which is essential for advancing relative research.

(3) The majority of evaluation datasets fall within the 0-100K size range, with datasets containing fewer than 10K samples constituting 56.4\% of the total. This indicates that many tasks can be effectively assessed with relatively small datasets, which may be also due to cost considerations during dataset construction and evaluation. Nevertheless, a few datasets still surpass the 1M mark, mainly derived from web scraping or the consolidation of open-source datasets.

(4) Manual construction and the compilation of open-source datasets are the dominant methods for creating evaluation datasets. Manual construction is often preferred for its precision and relevance to specific domains, whereas the combination of open-source datasets creates common benchmarks for evaluation. The use of model-generated data for evaluation is less common due to concerns about question authenticity and answer accuracy, and it is generally used as a supplemental method.

(5) English language datasets are the most prevalent, with Chinese language datasets also being significant, reflecting the focus on evaluating LLM performance for tasks in these two languages. Although there are a limited number of datasets that cover evaluations in other languages, resources for low-resource minority languages are notably limited.

(6) Evaluation datasets including multiple disciplines and task types are prevalent, underscoring the increased focus on evaluating the holistic capabilities of LLMs. The research community is particularly concerned with the model’s general applicability and extensive knowledge. Various evaluation datasets cover conventional instructions, knowledge domains, social norms, and several prevalent vertical fields. Nevertheless, the distribution of domains within evaluation datasets continues to exhibit a long-tail pattern, with niche areas like e-commerce and earth sciences having limited evaluation resources. Notably, domains like ancient texts and cultures currently lack evaluation benchmarks.

(7) Subjective questions, especially those related to Natural Language Understanding (NLU), dominate the evaluation datasets. A minority of datasets encompasses objective questions, including multiple-choice and fill-in-the-blank formats. Regarding the methodologies employed for evaluation, the widespread use of code-based assessment is attributable to its applicability for objective questions and straightforward subjective tasks, manifesting advantages in efficiency and consistency. Conversely, manual evaluation is unsuitable for extensive tasks and objective questions due to cost considerations and is consequently infrequently utilized. It is crucial to highlight that model evaluation, to some degree, amalgamates the strengths of code-based and manual evaluations, potentially steering towards becoming the predominant evaluation methodology in the future. Naturally, the strategic combination of evaluation methods should consider practical aspects, including the scale and diversity  of questions.

\section{Traditional NLP Datasets}\label{sec6}

Diverging from instruction fine-tuning datasets, we categorize text datasets dedicated to natural language tasks before the widespread adoption of LLMs as traditional NLP datasets. These datasets, devoid of instructional formats, are specifically crafted for training, optimizing, and testing traditional NLP models. The resultant NLP models find application in diverse text processing tasks, including text classification, information extraction, text summarization, etc.

In contemporary LLMs projects, a plethora of traditional NLP datasets finds application. These datasets undergo dual roles: firstly, their format and content transform into instructional formats for the instruction-guided fine-tuning phase of LLMs, augmenting the models’ capacities to adhere to instructions and excel in such tasks; secondly, they serve as evaluation datasets for LLMs, enabling the comparison of diverse LLMs in natural language tasks. Notably, several LLMs instruction datasets and evaluation datasets emerge from the conversion of traditional NLP datasets. Consequently, this section succinctly summarizes classical traditional NLP datasets commonly integrated into existing LLMs and various LLMs evaluation platforms. The objective is to streamline and offer references for traditional NLP datasets, facilitating the dataset selection process for LLMs projects.

\begin{figure}
\centering
\includegraphics[width=0.9\textwidth]{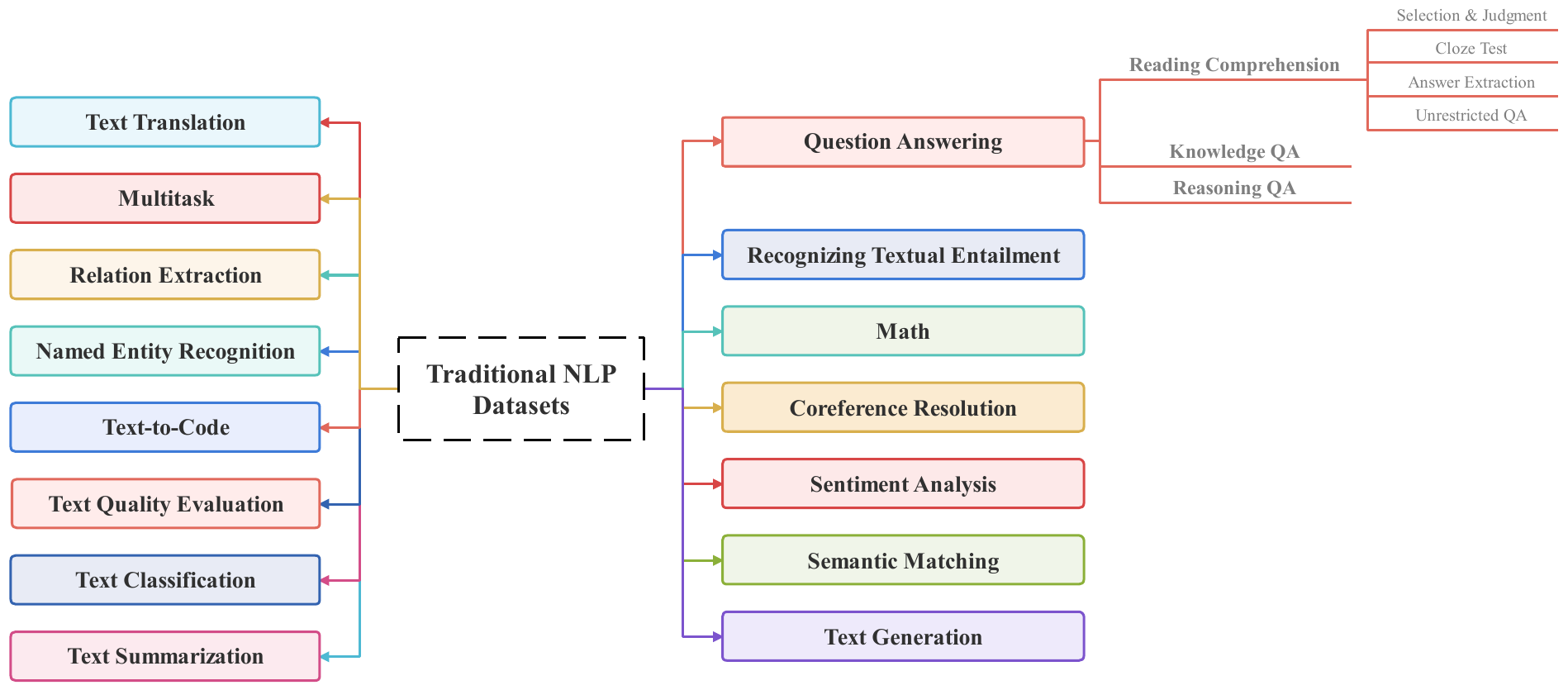}
\caption{Different NLP task categories of the traditional NLP datasets}\label{fig21}
\end{figure}

In this context, the compiled traditional NLP datasets are systematically classified into 15 distinct categories, aligning with various tasks. Figure~\ref{fig21} visually represents these categories, encompassing \textbf{question answering}, \textbf{recognizing textual entailment}, \textbf{math}, \textbf{coreference resolution}, \textbf{sentiment analysis}, \textbf{semantic matching}, \textbf{text generation}, \textbf{text translation}, \textbf{text summarization}, \textbf{text classification}, \textbf{text quality evaluation}, \textbf{text-to-code}, \textbf{named entity recognition}, \textbf{relation extraction}, and \textbf{multitask}. We will summarize various categories of NLP datasets in a straightforward manner using text and tables (Table~\ref{tab14} to Table~\ref{tab30}). Detailed information about the datasets is presented in the Appendix~\ref{secE}.

\subsection{Question Answering}\label{subsec61}

The task of question-answering requires the model to utilize its knowledge and reasoning capabilities to respond to queries based on provided text (which may be optional) and questions. This task often includes subcategories like \textbf{reading comprehension}, \textbf{knowledge QA}, and \textbf{reasoning QA}.

\subsubsection{Reading Comprehension}\label{subsubsec611}

The task of reading comprehension entails presenting a model with a designated text passage and associated questions, prompting the model to understand the text for the purpose of answering the questions. Based on the answering approach of the task, it can be roughly classified into four categories: \textbf{selection \& judgment}, \textbf{cloze test}, \textbf{answer extraction}, and \textbf{unrestricted QA}.

There are two modes for \textbf{selection \& judgment} tasks. \textbf{Mode one requires the model to select the most appropriate option from several answer options}. RACE \citep{bib265} and DREAM \citep{bib268} are specifically selected from English exams designed by human experts, requiring the model to answer multiple-choice questions about the content of given English articles. Similarly, $\mathrm{C}^3$ \citep{bib266} and ReClor \citep{bib267} are extracted from corresponding Chinese exams and graduate entrance exams, respectively, each containing relevant multiple-choice questions. \textbf{Mode two involves judging the correctness of a question using either “Yes” or “No.”} BoolQ \citep{bib260} requires the model to respond with “Yes” or “No” to complex inquiries and non-factual information. CondaQA \citep{bib262}, as the first English dataset to assess negation statements, tests the model’s understanding of negative assertions, with answers in the form of “Yes,” “No,” or “Don’t Know.” PubMedQA \citep{bib263}, focusing deeply on the biomedical field, presents higher professional knowledge requirements, necessitating judgment on the correctness of questions based on the abstracts of medical articles.

The \textbf{cloze task} requires the model to select a word or sentence to fill in the missing part of the text, making the text coherent and logical. Tasks are typically set at both the word and sentence levels. LAMBADA \citep{bib272} and CLOTH \citep{bib273} are English word-level cloze datasets. By perceiving the context, the model predicts the positions of missing words in the sentences. ChID \citep{bib271} requires the model to choose the correct idiom to fill in the blank, focusing on testing the model’s understanding of Chinese idioms. CMRC2019 \citep{bib274} is a sentence-level cloze-style dataset that requires the model to fill in several blank spaces in the article with candidate sentences. 

The \textbf{answer extraction} task involves the model pinpointing a continuous excerpt within the text as the answer to a given question. Fundamentally, the answers to the questions can be extracted or composed directly from the textual content, eliminating the necessity of generating supplementary open-ended content. SQuAD \citep{bib275} extracts text passages and answers to questions from Wikipedia articles for answer extraction tasks. SQuAD 2.0 \citep{bib276} extends the SQuAD dataset by adding unanswerable questions, testing the models’ ability to judge ambiguous questions. Adversarial QA \citep{bib284} expands upon the SQuAD dataset by creating more challenging questions using adversarial human annotations. Additionally, other datasets such as TriviaQA \citep{bib278}, Natural Questions \citep{bib279}, and CMRC2018 \citep{bib283} feature more complex, challenging, and realistic reading comprehension questions.

The \textbf{unrestricted QA} task exhibits greater openness when contrasted with answer extraction tasks. The task entails producing a fitting response by leveraging both textual content and a posed question. The answer, rather than being an exact extraction from the text, is openly generated by the models. Presently, this task category stands as a predominant focus in the evaluation of LLMs. DROP \citep{bib290} and QASPER \citep{bib292} assess models’ reasoning ability to generate open-ended answers. Answers cannot be directly extracted from the text but require models to search for clues from multiple sources and then perform certain operations. CoQA \citep{bib291} measures models’ ability to answer related questions, with answers being in free-form text. Compared to the previous datasets, DuReader 2.0 \citep{bib294} expands the scale of text and questions, conducting open-domain Q\&A at the document level.

\subsubsection{Knowledge QA}\label{subsubsec612}

In the knowledge QA task, models respond to questions by leveraging world knowledge, common sense, scientific insights, domain-specific information, and more. Unlike reading comprehension tasks, each instance does not come with a reference text. This task assesses the model’s depth of knowledge and its capacity to comprehend questions.

ARC \citep{bib295}, CommonsenseQA \citep{bib296}, and OpenBookQA \citep{bib297} evaluate models’ knowledge mastery and comprehension abilities based on scientific facts and human common sense. These datasets emphasize general knowledge known to the general public. However, some datasets place more emphasis on testing vertical domain knowledge. PIQA \citep{bib298} and SciQ \citep{bib302} examine knowledge of science, JEC-QA \citep{bib299} examines legal analysis, WebMedQA \citep{bib306} examines medical diagnosis, and PsyQA \citep{bib305} examines psychological counseling.

\subsubsection{Reasoning QA}\label{subsubsec613}

The focal point of reasoning QA tasks is the requirement for models to apply abilities such as logical reasoning, multi-step inference, and causal reasoning in answering questions. These types of questions typically necessitate models to grasp the logical connections within the text, deduce concealed information, and arrive at sensible conclusions.

HellaSwag \citep{bib309}, Social IQa \citep{bib311}, ROPES \citep{bib318}, and WIQA \citep{bib315} are grounded in contextual reasoning, aiming to enable models to infer the subsequent development direction based on given contexts. COPA \citep{bib308} specifically tests causal reasoning ability, selecting appropriate causal relationships based on premises. LogiQA \citep{bib312} extensively investigates logical reasoning, covering various deductive patterns. Thus, it is evident that datasets for reasoning question answering tasks involve different dimensions of reasoning.

\begin{table*}[h!]
        \captionsetup{singlelinecheck=off, justification=justified}
        \captionof{table}{Summary of \textbf{Reading Comprehension Datasets} Information. Release Time: “X” indicates unknown month. “Train Size,” “Dev Size,” “Test Size,” and “All Size” provide statistics on the respective question quantities in the dataset. Language: “EN” indicates English, “ZH” indicates Chinese, “Multi” indicates Multilingual, and the number in parentheses indicates the number of languages included}\label{tab14}
        \centering
        \resizebox{\textwidth}{!}{
            \begin{tabular}{lllllllll}
            \hline
            \textbf{Dataset} & \textbf{Publisher} & \textbf{Release Time} & \textbf{Train Size} & \textbf{Dev Size} & \textbf{Test Size} & \textbf{All Size} & \textbf{License} & \textbf{Language} \\ \hline
            \multicolumn{9}{c}{\textbf{Selection \& Judgment}} \\ \hline
            BoolQ & University of Washington et al. & 2019-5 & 9427 & 3270 & 3245 & 15942 & CC-SA-3.0 & EN \\ 
            CondaQA & Carnegie Mellon University et al. & 2022-11 & 5832 & 1110 & 7240 & 14182 & Apache-2.0 & EN \\ 
            CosmosQA & University of Illinois Urbana-Champaign et al. & 2019-9 & 25588 & 3000 & 7000 & 35588 & CC-BY-4.0 & EN \\ 
            $\mathrm{C}^3$ & Cornell University et al. & 2019-4 & 11869 & 3816 & 3892 & 19577 & - & ZH \\ 
            DREAM & Cornell University et al. & 2019-2 & 6116 & 2040 & 2041 & 10197 & - & EN \\ 
            Dureader Yes/No & Baidu Inc. et al. & 2019-12 & 75K & 5.5K & 11K & 91.5K & Apache-2.0 & ZH \\ 
            MCTest & Microsoft Research & 2013-10 & 1200 & 200 & 600 & 2000 & - & EN \\ 
            MultiRC & University of Pennsylvania et al. & 2018-6 & - & - & - & 9872 & MultiRC License & EN \\ 
            PubMedQA & University of Pittsburgh et al. & 2019-9 & - & - & - & 273.5K & MIT & EN \\ 
            QuAIL & University of Massachusetts Lowell & 2020-4 & 10346 & - & 2164 & 12510 & CC-NC-SA-4.0 & EN \\ 
            RACE & Carnegie Mellon University & 2017-4 & 87866 & 4887 & 4934 & 97687 & - & EN \\ 
            ReClor & National University of Singapore & 2020-2 & 4638 & 500 & 1000 & 6138 & - & EN \\ \hline
            \multicolumn{9}{c}{\textbf{Cloze Test}} \\ \hline
            ChID & Tsinghua University et al. & 2019-6 & 605K & 23.2K & 83.3K & 711.5K & Apache-2.0 & ZH \\ 
            CLOTH & Carnegie Melon University & 2018-10 & 76850 & 11067 & 11516 & 99433 & MIT & EN \\ 
            CMRC2019 & Harbin Institute of Technology et al. & 2020-12 & 100009 & 3053 & 5118 & 108180 & CC-BY-SA-4.0 & ZH \\ 
            LAMBADA & University of Trento et al. & 2016-6 & 2662 & 4869 & 5153 & 12684 & CC-BY-4.0 & EN \\ \hline
            \multicolumn{9}{c}{\textbf{Answer Extraction}} \\ \hline
            Adversarial QA & University College London & 2020-2 & 30000 & 3000 & 3000 & 36000 & MIT & EN \\ 
            CMRC2018 & Harbin Institute of Technology et al. & 2019-11 & 10321 & 3351 & 4895 & 18567 & CC-BY-SA-4.0 & ZH \\ 
            CUAD & UC Berkeley et al. & 2021-3 & 22450 & - & 4182 & 26632 & CC-BY-4.0 & EN \\ 
            Dureader Checklist & Baidu Inc. et al. & 2021-3 & 3K & 1.1K & 4.5K & 8.6K & Apache-2.0 & ZH \\ 
            Dureader Robust & Baidu Inc. et al. & 2020-3 & 15K & 1.4K & 4.8K & 21.2K & Apache-2.0 & ZH \\ 
            HOTPOTQA & Carnegie Mellon University et al. & 2018-9 & 90447 & 7405 & 7405 & 105257 & CC-BY-SA-4.0 & EN \\ 
            MLQA & Facebook AI Research et al. & 2020-7 & - & 4199 & 42246 & 46445 & CC-BY-SA-3.0 & Multi (7) \\ 
            MS MARCO & Microsoft AI \& Research & 2016-11 & 808731 & 101093 & 101092 & 1010916 & MIT & EN \\ 
            Natural Questions & Google Research & 2019-X & 307372 & 7830 & 7842 & 323044 & CC-BY-4.0 & EN \\ 
            QuAC & AI2 et al. & 2018-8 & 83568 & 7354 & 7353 & 98407 & CC-BY-SA-4.0 & EN \\
            Quoref & AI2 et al. & 2019-8 & 19399 & 2418 & 2537 & 24354 & CC-BY-4.0 & EN \\ 
            ReCoRD & Johns Hopkins University et al. & 2018-10 & 100730 & 10000 & 10000 & 120730 & - & EN \\ 
            SQuAD & Stanford University & 2016-11 & 87599 & 10570 & 9533 & 107702 & CC-BY-4.0 & EN \\ 
            SQuAD 2.0 & Stanford University & 2018-6 & 130319 & 11873 & 8862 & 151054 & CC-BY-SA-4.0 & EN \\ 
            TriviaQA & Univ. of Washington et al. & 2017-7 & - & - & - & 95000 & Apache-2.0 & EN \\
            TyDiQA & Google Research & 2020-3 & 116916 & 18670 & 18751 & 154337 & Apache-2.0 & Multi (11) \\ \hline
            \multicolumn{9}{c}{\textbf{Unrestricted QA}} \\ \hline
            CoQA & Stanford University & 2018-8 & - & - & - & 127K & - & EN \\ 
            DROP & University of California et al. & 2019-6 & 77409 & 9536 & 9622 & 96567 & CC-BY-4.0 & EN \\ 
            DuoRC & IBM Research et al. & 2018-7 & 130261 & 27914 & 27914 & 186089 & MIT & EN \\ 
            Dureader 2.0 & Baidu Inc. et al. & 2018-4 & - & - & - & 200K & Apache-2.0 & ZH \\ 
            QASPER & AI2 et al. & 2021-5 & - & - & - & 5049 & CC-BY-4.0 & EN \\ \hline
            \end{tabular}
        }
\end{table*}

\begin{table*}[h!]
        \captionsetup{singlelinecheck=off, justification=justified}
        \captionof{table}{Summary of \textbf{Knowledge QA Datasets} Information. Release Time: “X” indicates unknown month. “Train Size,” “Dev Size,” “Test Size,” and “All Size” provide statistics on the respective question quantities in the dataset. Language: “EN” indicates English, “ES” indicates Spanish, “ZH” indicates Chinese}\label{tab15}
        \centering
        \resizebox{\textwidth}{!}{
            \begin{tabular}{lllllllll}
            \hline
            \textbf{Dataset} & \textbf{Publisher} & \textbf{Release Time} & \textbf{Train Size} & \textbf{Dev Size} & \textbf{Test Size} & \textbf{All Size} & \textbf{License} & \textbf{Language} \\ \hline
            ARC & AI2 & 2018-3 & 3370 & 869 & 3548 & 7787 & CC-BY-SA & EN \\ 
            CMD & Toyhom & 2019-X & - & - & - & 792099 & MIT & ZH \\ 
            cMedQA2 & National University of Defense Technology & 2018-11 & 100000 & 4000 & 4000 & 108000 & GPL-3.0 & ZH \\ 
            CommonsenseQA & Tel-Aviv University et al. & 2018-11 & 9797 & 1225 & 1225 & 12247 & MIT & EN \\ 
            ECQA & IIT Delhi et al. & 2021-8 & 7598 & 1090 & 2194 & 10882 & CDLA-Sharing-1.0 & EN \\ 
            HEAD-QA & Universidade da Coruna & 2019-7 & 2657 & 1366 & 2742 & 13530 & MIT & EN \& ES \\ 
            JEC-QA & Tsinghua University et al. & 2019-11 & - & - & 26365 & 26365 & CC-NC-ND-4.0 & EN \\ 
            OpenBookQA & AI2 et al. & 2018-10 & 4957 & 500 & 500 & 5957 & Apache-2.0 & EN \\ 
            PIQA & AI2 et al. & 2019-11 & 16.1K & 1.84K & 3.08K & 21.02K & MIT & EN \\ 
            PsyQA & The CoAI group et al. & 2021-6 & - & - & - & 22346 & PsyQA User Agreement & ZH \\ 
            SciQ & University College London et al. & 2017-9 & 11679 & 1000 & 1000 & 13679 & CC-BY-NC-3.0 & EN \\ 
            WebMedQA & Chinese Academy of Sciences et al. & 2018-12 & 50610 & 6337 & 6337 & 63284 & Apache-2.0 & ZH \\ 
            WikiQA & Georgia Institute of Technology et al. & 2015-9 & 2118 & 296 & 633 & 3047 & Microsoft Research Data License & EN \\ \hline
            \end{tabular}
        }
\end{table*}

\begin{table*}[h!]
        \captionsetup{singlelinecheck=off, justification=justified}
        \captionof{table}{Summary of \textbf{Reasoning QA Datasets} Information. “Train Size,” “Dev Size,” “Test Size,” and “All Size” provide statistics on the respective question quantities in the dataset. Language: “EN” indicates English, “ZH” indicates Chinese}\label{tab16}
        \centering
        \resizebox{\textwidth}{!}{
            \begin{tabular}{lllllllll}
            \hline
            \textbf{Dataset} & \textbf{Publisher} & \textbf{Release Time} & \textbf{Train Size} & \textbf{Dev Size} & \textbf{Test Size} & \textbf{All Size} & \textbf{License} & \textbf{Language} \\ \hline
            COPA & Indiana University et al. & 2011-6 & - & 500 & 500 & 1000 & BSD 2-Clause & EN \\ 
            CREAK & The University of Texas at Austin & 2021-9 & 10176 & 1371 & 1371 & 13418 & MIT & EN \\ 
            HellaSwag & University of Washington et al. & 2019-7 & 39905 & 10042 & 10003 & 59950 & MIT & EN \\ 
            LogiQA & Fudan University et al. & 2020-7 & 7376 & 651 & 651 & 8678 & - & EN \& ZH \\ 
            PROST & University of Colorado Boulder & 2021-8 & - & - & 18736 & 18736 & Apache-2.0 & EN \\ 
            QASC & AI2 et al. & 2019-10 & 8134 & 926 & 920 & 9980 & CC-BY-4.0 & EN \\ 
            QuaRel & AI2 & 2018-11 & 1941 & 278 & 552 & 2771 & CC-BY-4.0 & EN \\ 
            QuaRTz & AI2 & 2019-11 & 2696 & 384 & 784 & 3864 & CC-BY-4.0 & EN \\ 
            ROPES & AI2 & 2019-8 & 10K & 1.6K & 1.7K & 13.3K & CC-BY-4.0 & EN \\ 
            Social IQa & AI2 & 2019-4 & 33410 & 1954 & - & 35364 & - & EN \\ 
            StoryCloze & University of Rochester et al. & 2016-6 & - & 1871 & 1871 & 3742 & - & EN \\ 
            STRATEGYQA & Tel Aviv University et al. & 2021-1 & 2290 & - & 490 & 2780 & MIT & EN \\ 
            WIQA & AI2 & 2019-9 & 29808 & 6894 & 3993 & 40695 & - & EN \\ \hline
            \end{tabular}
        }
\end{table*}

\subsection{Recognizing Textual Entailment}\label{subsec62}

The primary objective of tasks related to Recognizing Textual Entailment (RTE) is to assess whether information in one textual segment can be logically inferred from another. This is formally structured with a “premise” denoted as \textit{P} and a “hypothesis” denoted as \textit{H}, aimed at determining the relationship between \textit{P} and \textit{H}. If \textit{P} logically entails \textit{H}, it is categorized as “Entailment”; if \textit{P} and \textit{H} are logically contradictory, it is categorized as “Contradiction”; if there is no discernible logical connection or contradiction between \textit{P} and \textit{H}, it is categorized as “Neutral.” In some instances, the latter two scenarios are combined into “Non-Entailment.”

For example, RTE \citep{bib321,bib322,bib323,bib324} integrates a portion of the Recognizing Textual Entailment challenge datasets, comprising two types of relationships: “Entailment” and “Non-Entailment.” CommitmentBank \citep{bib327}, OCNLI \citep{bib330}, and CINLID\footnote{\url{https://www.luge.ai/\#/luge/dataDetail?id=39}} expand the judgment of relationships to three types. ANLI \citep{bib320} introduces adversarial samples, increasing the difficulty of textual relationship judgment and making it more challenging.

\begin{table*}[h!]
        \captionsetup{singlelinecheck=off, justification=justified}
        \captionof{table}{Summary of \textbf{Recognizing Textual Entailment Datasets} Information. Release Time: “X” indicates unknown month. “Train Size,” “Dev Size,” “Test Size,” and “All Size” provide statistics on the respective question quantities in the dataset. Language: “EN” indicates English, “ZH” indicates Chinese}\label{tab17}
        \centering
        \resizebox{\textwidth}{!}{
            \begin{tabular}{lllllllll}
            \hline
            \textbf{Dataset} & \textbf{Publisher} & \textbf{Release Time} & \textbf{Train Size} & \textbf{Dev Size} & \textbf{Test Size} & \textbf{All Size} & \textbf{License} & \textbf{Language} \\ \hline
            ANLI & UNC Chapel Hill et al. & 2019-10 & 162865 & 3200 & 3200 & 169265 & CC-NC-4.0 & EN \\ 
            CINLID & Gao et al. & 2021-4 & 80124 & - & 26708 & 106832 & - & ZH \\ 
            CMNLI & CLUE team & 2020-12 & 391783 & 12426 & 13880 & 418089 & - & ZH \\ 
            CommitmentBank & The Ohio State University et al. & 2019-X & - & - & - & 1200 & - & EN \\ 
            MedNLI & University of Massachusetts Lowell et al. & 2018-8 & 11232 & 1395 & 1422 & 14049 & - & EN \\ 
            MultiNLI & New York University & 2018-6 & 392702 & 19647 & - & 412349 & - & EN \\ 
            OCNLI & Indiana University et al. & 2020-10 & 50K & 3K & 3K & 56K & CC-BY-NC-2.0 & ZH \\ 
            RTE & The PASCAL Recognising Textual Entailment Challenge & - & 2.49K & 277 & 3K & 5.77K & CC-BY-4.0 & EN \\ 
            SNLI & Stanford Linguistics et al. & 2015-8 & 550152 & 10000 & 10000 & 570152 & CC-BY-SA-4.0 & EN \\ 
            WANLI & University of Washington et al. & 2022-1 & 102885 & - & 5000 & 107885 & CC-BY-4.0 & EN \\ \hline
            \end{tabular}
        }
\end{table*}

\subsection{Math}\label{subsec63}

Mathematical assignments commonly involve standard mathematical calculations, theorem validations, and mathematical reasoning tasks, among others. These tasks aim to investigate the latent capabilities of models within the field of mathematics.

Datasets related to mathematical tasks vary in difficulty. GSM8K \citep{bib331}, ASDiv \citep{bib333}, Math23K \citep{bib336}, and Ape210K \citep{bib335} only contain primary school mathematical calculations, which are relatively simple for humans. MATH \citep{bib334} targets mathematical competition problems, which are more challenging and also examine the models’ ability to follow thinking chains when solving problems. NaturalProofs \citep{bib339} involves mathematical proposition proofs, axiom inferences, and so on.

\begin{table*}[h!]
        \captionsetup{singlelinecheck=off, justification=justified}
        \captionof{table}{Summary of \textbf{Math Datasets} Information. “Train Size,” “Dev Size,” “Test Size,” and “All Size” provide statistics on the respective question quantities in the dataset. Language: “EN” indicates English, “ZH” indicates Chinese}\label{tab18}
        \centering
        \resizebox{\textwidth}{!}{
            \begin{tabular}{lllllllll}
            \hline
            \textbf{Dataset} & \textbf{Publisher} & \textbf{Release Time} & \textbf{Train Size} & \textbf{Dev Size} & \textbf{Test Size} & \textbf{All Size} & \textbf{License} & \textbf{Language} \\ \hline
            Ape210K & Yuanfudao AI Lab et al. & 2020-9 & 200488 & 5000 & 5000 & 210488 & - & ZH \\ 
            AQUA-RAT & DeepMind & 2017-7 & 100949 & 250 & 250 & 101499 & Apache-2.0 & EN \\ 
            ASDiv & Institute of Information Science & 2021-6 & - & - & - & 2305 & CC-BY-NC-4.0 & EN \\ 
            GSM8k & OpenAI & 2021-10 & 7.5K & - & 1K & 8.5K & MIT & EN \\ 
            MATH & UC Berkeley et al. & 2021-3 & 7500 & - & 5000 & 12500 & MIT & EN \\ 
            MathQA & University of Washington et al. & 2019-5 & 29837 & 4475 & 2985 & 37297 & Apache-2.0 & EN \\ 
            Math23K & Tencent AI Lab & 2017-9 & - & - & - & 23161 & MIT & ZH \\ 
            NaturalProofs & University of Washington et al. & 2021-4 & - & - & - & 80795 & MIT & EN \\ 
            SVAMP & Microsoft Research India & 2021-3 & - & - & - & 1000 & MIT & EN \\ \hline
            \end{tabular}
        }
\end{table*}

\subsection{Coreference Resolution}\label{subsec64}

The core objective of tasks related to coreference resolution is the identification of referential relationships within texts. Pronouns, noun phrases, or alternative expressions are occasionally employed in textual passages to refer to entities introduced earlier. This task entails the recognition of entities referred to by different segments of the text and is a fundamental research area in the field of NLP.

WiC \citep{bib342} and CLUEWSC2020 \citep{bib179} are coreference resolution datasets in the English and Chinese domains, respectively, used to determine whether words in different sentences have the same referential meaning. WSC \citep{bib19} does not require comparison but rather demands the specific content to which words refer. WinoGrande \citep{bib341} adjusts the WSC dataset by redesigning the task in a fill-in-the-blank format. WinoWhy \citep{bib343} extends the WSC dataset by introducing a new task of explaining referential relationships.

\subsection{Sentiment Analysis}\label{subsec65}

The sentiment analysis task, commonly known as emotion classification, seeks to analyze and deduce the emotional inclination of provided texts, commonly categorized as positive, negative, or neutral sentiments. This task finds practical utility in diverse domains, including social media monitoring, product review analysis, and market research.

Classic sentiment analysis datasets include IMDB \citep{bib344}, Sentiment140 \citep{bib345}, SST-2 \citep{bib346}, and EPRSTMT \citep{bib252}. The textual content of these datasets originates from real-life scenarios such as movie reviews, product reviews, and tweet content, hence possessing diversity and authenticity. Each sample is manually labeled as expressing either positive or negative sentiment based on the emotions conveyed in the text.

\begin{table*}[h!]
        \captionsetup{singlelinecheck=off, justification=justified}
        \captionof{table}{Summary of \textbf{Coreference Resolution Datasets} Information. Release Time: “X” indicates unknown month. “Train Size,” “Dev Size,” “Test Size,” and “All Size” provide statistics on the respective question quantities in the dataset. Language: “EN” indicates English, “ZH” indicates Chinese}\label{tab19}
        \centering
        \resizebox{\textwidth}{!}{
            \begin{tabular}{lllllllll}
            \hline
            \textbf{Dataset} & \textbf{Publisher} & \textbf{Release Time} & \textbf{Train Size} & \textbf{Dev Size} & \textbf{Test Size} & \textbf{All Size} & \textbf{License} & \textbf{Language} \\ \hline
            CLUEWSC2020 & CLUE team & 2020-12 & 1244 & 304 & 290 & 1838 & - & ZH \\ 
            DPR & University of Texas at Dallas & 2012-7 & 1322 & - & 564 & 1886 & - & EN \\ 
            WiC & University of Cambridge & 2018-8 & 5428 & 638 & 1400 & 7466 & CC-NC-4.0 & EN \\ 
            WinoGrande & AI2 et al. & 2019-7 & 63238 & 1267 & 1767 & 66272 & CC-BY & EN \\ 
            WinoWhy & HKUST & 2020-7 & - & - & - & 43972 & MIT & EN \\ 
            WSC & University of Toronto et al. & 2012-X & - & - & 285 & 285 & CC-BY-4.0 & EN \\ \hline
            \end{tabular}
        }
\end{table*}

\begin{table*}[h!]
        \captionsetup{singlelinecheck=off, justification=justified}
        \captionof{table}{Summary of \textbf{Sentiment Analysis Datasets} Information. Release Time: “X” indicates unknown month. “Train Size,” “Dev Size,” “Test Size,” and “All Size” provide statistics on the respective question quantities in the dataset. Language: “EN” indicates English, “ZH” indicates Chinese}\label{tab20}
        \centering
        \resizebox{\textwidth}{!}{
            \begin{tabular}{lllllllll}
            \hline
            \textbf{Dataset} & \textbf{Publisher} & \textbf{Release Time} & \textbf{Train Size} & \textbf{Dev Size} & \textbf{Test Size} & \textbf{All Size} & \textbf{License} & \textbf{Language} \\ \hline
            EPRSTMT & CLUE team & 2021-7 & 32 & 32 & 1363 & 20992 & - & ZH \\ 
            IMDB & Stanford University & 2011-6 & 25000 & - & 25000 & 50000 & - & EN \\ 
            Sentiment140 & Stanford University & 2009-X & 1600000 & - & 359 & 1600359 & - & EN \\ 
            SST-2 & Stanford University & 2013-10 & 67349 & 872 & 1821 & 70042 & - & EN \\ \hline
            \end{tabular}
        }
\end{table*}

\subsection{Semantic Matching}\label{subsec66}

The task of semantic matching entails evaluating the semantic similarity or degree of correspondence between two sequences of text. Models must grasp the semantic information within the text to perform tasks such as assessing text similarity, matching sentences, and determining semantic relationships. This task is widely applied in domains such as information retrieval and dialogue systems.

MRPC \citep{bib347}, QQP \citep{bib177}, and PAWS \citep{bib348} are commonly used English semantic matching datasets, used for determining semantic similarity at the sentence level. AFQMC \citep{bib179} and LCQMC \citep{bib351} are commonly used large-scale Chinese datasets. Specifically, the LCQMC dataset is more inclined towards matching the intent of questions rather than semantic matching. To address the lack of other languages, PAWS-X \citep{bib352} translates the PAWS dataset into 6 other languages. The most notable is the STSB dataset \citep{bib349}, which not only includes 10 languages but also employs continuous similarity scores as labels rather than simple binary labels.

\begin{table*}[h!]
        \captionsetup{singlelinecheck=off, justification=justified}
        \captionof{table}{Summary of \textbf{Semantic Matching Datasets} Information. Release Time: “X” indicates unknown month. “Train Size,” “Dev Size,” “Test Size,” and “All Size” provide statistics on the respective question quantities in the dataset. Language: “EN” indicates English, “ZH” indicates Chinese, “Multi” indicates Multilingual, and the number in parentheses indicates the number of languages included}\label{tab21}
        \centering
        \resizebox{\textwidth}{!}{
            \begin{tabular}{lllllllll}
            \hline
            \textbf{Dataset} & \textbf{Publisher} & \textbf{Release Time} & \textbf{Train Size} & \textbf{Dev Size} & \textbf{Test Size} & \textbf{All Size} & \textbf{License} & \textbf{Language} \\ \hline
            AFQMC & CLUE team & 2020-12 & 34.3K & 4.3K & 3.9K & 42.5K & - & ZH \\ 
            BQ & Harbin Institute of Technology et al. & 2018-10 & 100000 & 10000 & 10000 & 120000 & - & ZH \\ 
            BUSTM & CLUE team & 2021-7 & 32 & 32 & 3772 & 8087 & - & ZH \\ 
            DuQM & Baidu Inc. et al. & 2021-9 & - & - & - & 10121 & Apache-2.0 & ZH \\ 
            LCQMC & Harbin Institute of Technology et al. & 2018-8 & 238766 & 8802 & 12500 & 260068 & CC-BY-4.0 & ZH \\ 
            MRPC & Microsoft Research & 2005-X & 4076 & - & 1725 & 5801 & - & EN \\ 
            PAWS & Google AI Language & 2019-6 & 49401 & 8000 & 8000 & 65401 & - & EN \\ 
            PAWS-X & Google Research & 2019-8 & 296406 & 11815 & 11844 & 320065 & - & Multi (6) \\ 
            QQP & New York University et al. & 2018-11 & 364K & - & - & 364K & - & EN \\ 
            STSB & Google Research et al. & 2017-8 & 5749 & 1500 & 1379 & 8628 & - & Multi (10) \\ \hline
            \end{tabular}
        }
\end{table*}

\subsection{Text Generation}\label{subsec67}

The scope of text generation tasks is broad, encompassing the generation of content summaries or dialogues. In a specific context, we narrow down the definition of text generation tasks to differentiate them from tasks like text summarization and translation. The narrow definition of text generation tasks is bound by provided content and specific requirements. It involves utilizing benchmark data, such as descriptive terms and triplets, to generate corresponding textual descriptions.

The first form involves \textbf{generating sentences in a colloquial manner using specific words}. CommonGen \citep{bib354} and E2E \citep{bib356} task models with generating coherent sentences related to given vocabulary terms. The second form involves \textbf{mapping structured data to text}. DART \citep{bib355} and WebNLG \citep{bib357} input structured data as triples to the model to obtain relevant descriptive sentences.

\begin{table*}[h!]
        \captionsetup{singlelinecheck=off, justification=justified}
        \captionof{table}{Summary of \textbf{Text Generation Datasets} Information. “Train Size,” “Dev Size,” “Test Size,” and “All Size” provide statistics on the respective question quantities in the dataset. Language: “EN” indicates English, “RU” indicates Russian}\label{tab22}
        \centering
        \resizebox{\textwidth}{!}{
            \begin{tabular}{lllllllll}
            \hline
            \textbf{Dataset} & \textbf{Publisher} & \textbf{Release Time} & \textbf{Train Size} & \textbf{Dev Size} & \textbf{Test Size} & \textbf{All Size} & \textbf{License} & \textbf{Language} \\ \hline
            CommonGen & University of Southern California et al. & 2019-11 & 67389 & 4018 & 1497 & 72904 & MIT & EN \\ 
            DART & Yale University et al. & 2020-7 & 30526 & 2768 & 6959 & 40253 & MIT & EN \\ 
            E2E & Heriot-Watt University & 2017-6 & 42061 & 4672 & 4693 & 51426 & CC-BY-SA-3.0 & EN \\ 
            WebNLG & LORIA et al. & 2017-7 & 49665 & 6490 & 7930 & 64085 & CC-BY-NC-SA-4.0 & EN \& RU \\ \hline
            \end{tabular}
        }
\end{table*}

\subsection{Text Translation}\label{subsec68}

Text translation involves transforming text from one language to another. Models must adeptly grasp the meaning of the source language text and produce equivalent text that conforms to the grammar and context of the target language.

WMT\footnote{\url{https://www.statmt.org/wmt22/index.html}} is one of the most commonly used text translation datasets. It aggregates data from the Workshop on Statistical Machine Translation competition, with a large-scale dataset covering a wide range of languages. NLLB \citep{bib358} provides open-access to three text translation evaluation benchmarks, offering high-quality translations in over 200 languages, including many low-resource languages. IWSLT 2017 \citep{bib359} is also representative and commonly used for training and evaluation in translation tasks.

\begin{table*}[h!]
        \captionsetup{singlelinecheck=off, justification=justified}
        \captionof{table}{Summary of \textbf{Text Translation Datasets} Information. “Train Size,” “Dev Size,” “Test Size,” and “All Size” provide statistics on the respective question quantities in the dataset. Language: “Multi” indicates Multilingual, and the number in parentheses indicates the number of languages included}\label{tab23}
        \centering
        \resizebox{\textwidth}{!}{
            \begin{tabular}{lllllllll}
            \hline
            \textbf{Dataset} & \textbf{Publisher} & \textbf{Release Time} & \textbf{Train Size} & \textbf{Dev Size} & \textbf{Test Size} & \textbf{All Size} & \textbf{License} & \textbf{Language} \\ \hline
            IWSLT 2017 & FBK et al. & 2017-12 & 1108475 & 4442 & 41921 & 1154838 & CC-BY-NC-ND-4.0 & Multi (11) \\ 
            NLLB & NLLB Team et al. & 2022-7 & - & - & - & - & MIT & Multi \\ 
            WMT & ACL et al. & - & - & - & - & - & - & Multi \\ \hline
            \end{tabular}
        }
\end{table*}

\subsection{Text Summarization}\label{subsec69}

The task of text summarization pertains to the extraction or generation of a brief summary or headline from an extended text to encapsulate its primary content. Summaries are expected to retain the pivotal information from the original text, effectively conveying its fundamental ideas, while headlines demand brevity and inclusiveness.

News is the most common source for text summarization datasets. CNN-DM \citep{bib361} utilizes a large number of news articles to create tens of thousands of article-summary pairs. Compared to the CNN-DM dataset, XSum \citep{bib365} has shorter text content and richer vocabulary. In addition to obtaining data samples from various news sources, SAMSum \citep{bib364}, Opinion Abstracts \citep{bib366}, LCSTS \citep{bib368}, MediaSum \citep{bib372}, and AESLC \citep{bib360} respectively focus on real dialogues, movie reviews, social media texts, interview transcripts, and emails. This ensures that different text summarization datasets have diverse styles of content and do not become overly homogeneous.

\subsection{Text Classification}\label{subsec610}

Text classification tasks aim to assign various text instances to predefined categories, comprising text data and category labels as pivotal components. Sentiment analysis and semantic matching, previously mentioned, are encompassed within the domain of text classification. Due to the unique nature of these tasks and their frequent exploration as standalone subtasks by researchers, this paper provides separate summaries for sentiment analysis, semantic matching, and text classification.

AGNEWS \citep{bib373} and TNEWS \citep{bib179} evaluate models’ classification performance on English and Chinese news topics, respectively. They involve a relatively small number of categories, not exceeding 15. CSLDCP \citep{bib252} requires models to classify Chinese literature disciplines, expanding the categories to 67. IFLYTEK \citep{bib179} categorizes descriptive text based on app functionality for model classification, with an astonishing 119 categories.

\begin{table*}[h!]
        \captionsetup{singlelinecheck=off, justification=justified}
        \captionof{table}{Summary of \textbf{Text Summarization Datasets} Information. “Train Size,” “Dev Size,” “Test Size,” and “All Size” provide statistics on the respective question quantities in the dataset. Language: “EN” indicates English, “ZH” indicates Chinese, “Multi” indicates Multilingual, and the number in parentheses indicates the number of languages included}\label{tab24}
        \centering
        \resizebox{\textwidth}{!}{
            \begin{tabular}{lllllllll}
            \hline
            \textbf{Dataset} & \textbf{Publisher} & \textbf{Release Time} & \textbf{Train Size} & \textbf{Dev Size} & \textbf{Test Size} & \textbf{All Size} & \textbf{License} & \textbf{Language} \\ \hline
            AESLC & Yale University et al. & 2019-7 & 14436 & 1960 & 1906 & 18302 & CC-BY-NC-SA-4.0 & EN \\ 
            CNewSum & ByteDance & 2021-10 & 275596 & 14356 & 14355 & 304307 & Apache-2.0 & ZH \\ 
            CNN-DM & Stanford University et al. & 2017-4 & 287113 & 13368 & 11490 & 311971 & Apache-2.0 & EN \\ 
            Gigaword & Facebook AI Research et al. & 2015-9 & 3803957 & 189651 & 1951 & 3995559 & MIT & EN \\ 
            LCSTS & Harbin Institute of Technology & 2015-6 & 2400000 & 10000 & 1000 & 2411000 & CC-BY-4.0 & ZH \\ 
            MediaSum & Microsoft Cognitive Services Research Group & 2021-3 & 443596 & 10000 & 10000 & 463596 & - & EN \\ 
            MultiNews & Yale University & 2019-7 & 44972 & 5622 & 5622 & 56216 & - & EN \\ 
            Newsroom & Cornell University & 2018-6 & 995041 & 108837 & 108862 & 1212740 & - & EN \\ 
            Opinion Abstracts & Northeastern University et al. & 2016-6 & 5990 & - & - & 5990 & - & EN \\ 
            SAMSum & Samsung R\&D Institute Poland & 2019-11 & 14732 & 818 & 819 & 16369 & CC-BY-NC-ND-4.0 & EN \\ 
            WikiHow & University of California & 2018-10 & - & - & - & 230K & CC-BY-NC-SA & EN \\ 
            WikiLingua & Columbia University et al. & 2020-10 & - & - & - & 770087 & CC-BY-3.0 & Multi (18) \\ 
            XL-Sum & BUET et al. & 2021-8 & 1122857 & 114198 & 114198 & 1351253 & CC-BY-NC-SA-4.0 & Multi (45) \\ 
            XSum & University of Edinburgh & 2018-10 & 204045 & 11332 & 11334 & 226711 & MIT & EN \\ \hline
            \end{tabular}
        }
\end{table*}

\begin{table*}[h!]
        \captionsetup{singlelinecheck=off, justification=justified}
        \captionof{table}{Summary of \textbf{Text Classification Datasets} Information. Release Time: “X” indicates unknown month. “Train Size,” “Dev Size,” “Test Size,” and “All Size” provide statistics on the respective question quantities in the dataset. Language: “EN” indicates English, “ZH” indicates Chinese, “Multi” indicates Multilingual, and the number in parentheses indicates the number of languages included}\label{tab25}
        \centering
        \resizebox{\textwidth}{!}{
            \begin{tabular}{lllllllll}
            \hline
            \textbf{Dataset} & \textbf{Publisher} & \textbf{Release Time} & \textbf{Train Size} & \textbf{Dev Size} & \textbf{Test Size} & \textbf{All Size} & \textbf{License} & \textbf{Language} \\ \hline
            AGNEWS & New York University & 2015-9 & 120000 & - & 7600 & 127600 & - & EN \\ 
            CSLDCP & CLUE team & 2021-7 & 536 & 536 & 4783 & 23966 & - & ZH \\ 
            IFLYTEK & CLUE team & 2020-12 & 12.1K & 2.6K & 2.6K & 17.3K & - & ZH \\ 
            MARC & Amazon et al. & 2020-11 & 1200000 & 30000 & 30000 & 1260000 & - & Multi (6) \\ 
            THUCNews & Tsinghua University & 2016-X & - & - & - & 1672165 & MIT & ZH \\ 
            TNEWS & CLUE team & 2020-11 & 53.3K & 10K & 10K & 73.3K & - & ZH \\ \hline
            \end{tabular}
        }
\end{table*}

\subsection{Text Quality Evaluation}\label{subsec6-11}

The task of text quality evaluation, also referred to as text correction, involves the identification and correction of grammatical, spelling, or language usage errors in text. This task is akin to a teacher correcting writing errors made by students.

CoLA \citep{bib375} is used to evaluate models’ ability to judge the grammatical correctness of English sentences, which can be seen as a binary classification task. In contrast, SIGHAN \citep{bib376,bib377,bib378} and YACLC \citep{bib379} require models to proofread and correct Chinese spelling and grammar, presenting greater difficulty. Different from these two datasets, CSCD-IME \citep{bib380} is the first Chinese spelling correction dataset caused by errors in Pinyin input method, with different sources and distributions of errors.

\begin{table*}[h!]
        \captionsetup{singlelinecheck=off, justification=justified}
        \captionof{table}{Summary of \textbf{Text Quality Evaluation Datasets} Information. “Train Size,” “Dev Size,” “Test Size,” and “All Size” provide statistics on the respective question quantities in the dataset. Language: “EN” indicates English, “ZH” indicates Chinese}\label{tab26}
        \centering
        \resizebox{\textwidth}{!}{
            \begin{tabular}{lllllllll}
            \hline
            \textbf{Dataset} & \textbf{Publisher} & \textbf{Release Time} & \textbf{Train Size} & \textbf{Dev Size} & \textbf{Test Size} & \textbf{All Size} & \textbf{License} & \textbf{Language} \\ \hline
            CoLA & New York University & 2018-5 & 8511 & 1043 & - & 9554 & CC-BY-4.0 & EN \\ 
            CSCD-IME & Tencent Inc & 2022-11 & 30000 & 5000 & 5000 & 40000 & MIT & ZH \\ 
            SIGHAN & Chaoyang Univ. of Technology et al. & - & 6476 & - & 3162 & 9638 & - & ZH \\ 
            YACLC & Beijing Language and Culture University et al. & 2021-12 & 8000 & 1000 & 1000 & 10000 & - & ZH \\ \hline
            \end{tabular}
        }
\end{table*}

\subsection{Text-to-Code}\label{subsec6-12}

The Text-to-Code task involves models converting user-provided natural language descriptions into computer-executable code, thereby achieving the desired functionality or operation. Common subtasks include the generation of SQL query statements and generating code for different programming languages.

For example, MBPP \citep{bib381} serves as a benchmark comprising Python programming problems, assessing models’ proficiency in Python programming. On the other hand, DuSQL \citep{bib382}, CSpider \citep{bib383}, and Spider \citep{bib208} are applied in the Text-to-SQL task. They require models to generate corresponding SQL query statements from given databases based on questions.

\begin{table*}[h!]
        \captionsetup{singlelinecheck=off, justification=justified}
        \captionof{table}{Summary of \textbf{Text-to-Code Datasets} Information. “Train Size,” “Dev Size,” “Test Size,” and “All Size” provide statistics on the respective question quantities in the dataset. Language: “EN” indicates English, “ZH” indicates Chinese, “PL” indicates Programming Language}\label{tab27}
        \centering
        \resizebox{\textwidth}{!}{
            \begin{tabular}{lllllllll}
            \hline
            \textbf{Dataset} & \textbf{Publisher} & \textbf{Release Time} & \textbf{Train Size} & \textbf{Dev Size} & \textbf{Test Size} & \textbf{All Size} & \textbf{License} & \textbf{Language} \\ \hline
            CSpider & Westlake University & 2019-11 & - & - & - & 10181 & CC-BY-SA-4.0 & ZH \& PL \\ 
            DuSQL & Baidu Inc. et al. & 2020-11 & 18602 & 2039 & 3156 & 23797 & - & ZH \& PL \\ 
            MBPP & Google Research & 2021-8 & - & - & 974 & 974 & - & EN \& PL \\ 
            Spider & Yale University & 2018-9 & - & - & - & 10181 & CC-BY-SA-4.0 & EN \& PL \\ \hline
            \end{tabular}
        }
\end{table*}

\subsection{Named Entity Recognition}\label{subsec6-13}

The Named Entity Recognition (NER) task aims to discern and categorize named entities within a given text. Models are tasked with pinpointing entities, assigning them to predefined categories, and indicating their respective positions. These entities may include personal names, organizational names, geographic locations, dates, and other categories.

CoNLL2003 \citep{bib387} is a classic benchmark dataset in the field of NER. It categorizes entity types into 4 classes. OntoNotes 5.0 \citep{bib388} expands into an NER task dataset based on the corpus and provides 18 entity types. Subsequently, WUNT2017 \citep{bib385} focuses on models’ ability to recognize emerging named entities in new contexts within the NER task. Youku NER \citep{bib390}, Taobao NER \citep{bib390}, and Weibo NER \citep{bib391} are constructed for the entertainment, e-commerce, and social media domains, respectively, providing corresponding text-entity pairs.

\begin{table*}[h!]
        \captionsetup{singlelinecheck=off, justification=justified}
        \captionof{table}{Summary of \textbf{Named Entity Recognition Datasets} Information. “Train Size,” “Dev Size,” “Test Size,” and “All Size” provide statistics on the respective question quantities in the dataset. “NEC” indicates Number of Entity Categories. Language: “DE” indicates German, “EN” indicates English, “ZH” indicates Chinese, “Multi” indicates Multilingual, and the number in parentheses indicates the number of languages included}\label{tab28}
        \centering
        \resizebox{\textwidth}{!}{
            \begin{tabular}{llllllllll}
            \hline
            \textbf{Dataset} & \textbf{Publisher} & \textbf{Release Time} & \textbf{Train Size} & \textbf{Dev Size} & \textbf{Test Size} & \textbf{All Size} & \textbf{NEC} & \textbf{License} & \textbf{Language} \\ \hline
            CLUENER & CLUE Organization & 2020-1 & 10748 & 1343 & 1345 & 13436 & 10 & - & ZH \\ 
            CoNLL2003 & University of Antwerp & 2003-6 & 14041 & 3250 & 3453 & 20744 & 4 & - & EN \& DE \\ 
            Few-NERD & Tsinghua University et al. & 2021-5 & - & - & - & 188200 & 66 & CC-BY-SA-4.0 & EN \\ 
            MSRA & University of Chicago & 2006-7 & 46364 & - & 4365 & 50729 & 3 & CC-BY-4.0 & ZH \\ 
            OntoNotes 5.0 & Boston Childrens Hospital and Harvard Medical School et al. & 2013-10 & 59924 & 8528 & 8262 & 76714 & 18 & - & Multi (3) \\ 
            Resume & Singapore University of Technology and Design & 2018-7 & 3821 & 463 & 477 & 4761 & 8 & - & ZH \\ 
            Taobao NER & Singapore University of Technology and Design et al. & 2019-6 & 6000 & 998 & 1000 & 7998 & 9 & - & ZH \\ 
            Weibo NER & Johns Hopkins University & 2015-9 & 1350 & 269 & 270 & 1889 & 4 & CC-BY-SA-3.0 & ZH \\ 
            WUNT2017 & Johns Hopkins University et al. & 2017-9 & 3394 & 1009 & 1287 & 5690 & 6 & CC-BY-4.0 & EN \\ 
            Youku NER & Singapore University of Technology and Design et al. & 2019-6 & 8001 & 1000 & 1001 & 10002 & 9 & - & ZH \\ \hline
            \end{tabular}
        }
\end{table*}

\subsection{Relation Extraction}\label{subsec6-14}

The endeavor of Relation Extraction (RE) necessitates the identification of connections between entities within textual content. This process typically includes recognizing and labeling pertinent entities, followed by the determination of the specific types of relationships that exist among them. As an illustration, the Forbidden City (geographic location) is positioned in (type of relationship) Beijing (geographic location).

Dialogue RE \citep{bib394} is the first entirely human-annotated dataset for dialogue RE, comprising 36 types of relationship found in real dialogues. In contrast to sentence-level datasets, DocRED \citep{bib396} is constructed for RE tasks at the document level. Models are required to aggregate document information to infer relationships between entities. FewRel \citep{bib397} is the first to combine few-shot learning with relation extraction, and in its 2.0 version, it additionally evaluates models’ OOD capability.

\begin{table*}[h!]
        \captionsetup{singlelinecheck=off, justification=justified}
        \captionof{table}{Summary of \textbf{Relation Extraction Datasets} Information. “Train Size,” “Dev Size,” “Test Size,” and “All Size” provide statistics on the respective question quantities in the dataset. “NRC” indicates Number of Relationship Categories. Language: “EN” indicates English, “ZH” indicates Chinese}\label{tab29}
        \centering
        \resizebox{\textwidth}{!}{
            \begin{tabular}{llllllllll}
            \hline
            \textbf{Dataset} & \textbf{Publisher} & \textbf{Release Time} & \textbf{Train Size} & \textbf{Dev Size} & \textbf{Test Size} & \textbf{All Size} & \textbf{NRC} & \textbf{License} & \textbf{Language} \\ \hline
            Dialogue RE & Tencent AI Lab et al. & 2020-7 & 6100 & 2034 & 2034 & 10168 & 36 & - & EN \& ZH \\ 
            DocRED & Tsinghua University et al. & 2019-7 & 1546589 & 12332 & 12842 & 1571763 & 96 & MIT & EN \\ 
            FewRel & Tsinghua University & 2018-10 & - & - & - & 70000 & 100 & CC-BY-SA-4.0 & EN \\ 
            TACRED & Stanford University & 2017-9 & 68124 & 22631 & 15509 & 106264 & 42 & LDC & EN \\ \hline
            \end{tabular}
        }
\end{table*}

\subsection{Multitask}\label{subsec6-15}

Multitask datasets hold significance as they can be concurrently utilized for different categories of NLP tasks. Creators commonly manipulate the same batch of textual data through various configurations, transformations, and annotations to produce training or evaluation data for diverse NLP tasks, exemplifying the concept of “one dataset, multiple applications.”

For example, CSL \citep{bib398} contains a vast amount of information such as paper titles, abstracts, keywords, etc., which can be simultaneously applied to various NLP tasks such as title prediction, keyword generation, paper classification, and so on. QED \citep{bib399} extends the Natural Questions dataset \citep{bib279} by adding explanatory annotations and extends to different tasks such as sentence selection, equivalence recognition, etc. METS-CoV \citep{bib400} collects social media texts related to COVID-19, which are annotated by creators and used in NER and sentiment analysis tasks.

\begin{table*}[h!]
        \captionsetup{singlelinecheck=off, justification=justified}
        \captionof{table}{Summary of \textbf{Multitask Datasets} Information. “Train Size,” “Dev Size,” “Test Size,” and “All Size” provide statistics on the respective question quantities in the dataset. Language: “EN” indicates English, “ZH” indicates Chinese}\label{tab30}
        \centering
        \resizebox{\textwidth}{!}{
            \begin{tabular}{lllllllll}
            \hline
            \textbf{Dataset} & \textbf{Publisher} & \textbf{Release Time} & \textbf{Train Size} & \textbf{Dev Size} & \textbf{Test Size} & \textbf{All Size} & \textbf{License} & \textbf{Language} \\ \hline
            CSL & School of Information Engineering et al. & 2022-9 & - & - & - & 396209 & Apache-2.0 & ZH \\ 
            METS-CoV & Zhejiang University et al. & 2022-9 & - & - & - & - & Apache-2.0 & EN \\ 
            QED & Stanford University et al. & 2021-3 & 7638 & 1355 & - & 8993 & CC-BY-SA-3.0 \& GFDL & EN \\ \hline
            \end{tabular}
        }
\end{table*}

\section{Challenges and Future Directions}\label{sec7}

This section primarily elaborates on the existing challenges and future directions from four aspects: pre-training corpora, fine-tuning instruction datasets, preference datasets, and evaluation datasets.

\subsection{Pre-training Corpora}\label{subsec71}

The construction and open sourcing of pre-training corpora have experienced significant growth recently, with increasing emphasis on their quality by researchers. However, pre-training corpora still face challenges and shortcomings that not only impact the performance of models but also involve ethical and societal issues. Below, we briefly explore the challenges existing in current pre-training corpora and discuss future development directions.

\textbf{Data Selection.} Research indicates that the diversity of data is crucial, and a richer variety of domains is preferable \citep{bib54}. It is worth investigating how to make the content of pre-training corpora as diverse as possible. Currently, the majority of pre-training corpora are composed of web-scraped data, and the data types are not entirely comprehensive. There is a risk of excessive focus on popular content, resulting in category imbalance. This can lead to a severe lack of knowledge in certain domains, necessitating the subsequent collection of data for incremental pre-training. Moreover, the scale of English data is much larger than that of other languages, which can result in insufficient knowledge of other languages and poor performance of models in cross-language tasks. Therefore, data selection is a nuanced art. \textbf{First}, larger-scale, more diverse, and more broadly sourced pre-training corpora covering multiple languages and domains with better proportional representation will be a future trend. Therefore, choices and configurations regarding data scale, data sources, domain coverage, data proportions, and language distribution need to be carefully considered. \textbf{Secondly}, data will be subdivided into finer categories, similar to the further categorization of books in Figure~\ref{fig4}, to better measure the breadth of the corpora, facilitating improved data selection. \textbf{Thirdly}, there will be a gradual exploration of whether the addition of synthetic data is effective for the pre-training of models. \textbf{Fourthly}, many vertical domains lack open-source relevant data, such as in the fields of ancient texts or ethnic cultures.

\textbf{Timeliness.} Currently, the coverage time of most pre-training corpora is relatively outdated, lacking recent knowledge and making it challenging to achieve periodic updates. This results in inaccurate generation or outdated information and being unable to respond to recent content. Common Crawl, for instance, continually crawls the latest webpage data, but the majority is in English. Other types of data require reacquisition and preprocessing when updates are needed. In the future, \textbf{dynamic and automatic updates of pre-training corpora}, as well as \textbf{self-learning capabilities of LLMs regarding new knowledge}, will be crucial research directions.

\textbf{Quality Assessment.} \cite{bib54} conducts evaluations on The Pile \citep{bib30} and C4 \citep{bib12}, exploring potential features of the data using different data integration methods. \cite{bib55} designs the Task2Vec metric to measure the diversity of data. However, a systematic methodology for quality assessment has not yet been established. Most studies only assess specific aspects of the corpora. Questions about what makes a pre-training corpus of higher quality, how the quality of pre-training corpora should be compared, and what constitutes a more comprehensive quality evaluation remain largely unresolved.

\textbf{Data Preprocessing.} Each pre-training corpus has a unique preprocessing pipeline and methods, with some specific details yet to be disclosed. This gives rise to two issues. First, there is a lack of a unified framework and standardized processes for data preprocessing. The effectiveness of existing methods is sometimes challenging to assess. Second, \cite{bib54}, through experiments, demonstrated that the more harmful content is filtered out from pre-training data, the less harmful information the model generates, but its discrimination ability also weakens. Filtering out low-quality data too extensively reduces the diversity of the data. While enhancing discrimination ability, it may lead to the generation of more harmful information by the model. Whether a cleaner corpus is necessarily better and whether a small amount of harmful information and low-quality data can bring benefits are questions that need to be explored in the future. Determining the optimal extent of data cleaning is also a topic for future research.

\textbf{Building the Ecosystem of Pre-training Corpora.} Due to the rapid development of LLMs, a comprehensive ecosystem for pre-training corpora has not yet been established within the community. There is a lack of standards for data preprocessing, no systematic evaluation schemes for data, no established standards for the release of relevant data, and currently, there is no unified management and maintenance of data. Given these circumstances, there is still a long way to go in building the ecosystem for pre-training corpora.

\subsection{Instruction Fine-tuning Datasets}\label{subsec72}

During the instruction fine-tuning phase, creating high-quality datasets is crucial for driving model performance and expanding application domains. Several challenges currently pose tests to the future development of instruction fine-tuning datasets. Below, we briefly explore the challenges existing in current instruction fine-tuning datasets and look ahead to future directions.

\textbf{Subdivision of Instruction Categories.} In the majority of instruction fine-tuning datasets, instructions of various categories are mixed together without specifying the corresponding task types and associated domains for each instruction. For instance, in the classic Alpaca\_data dataset \citep{bib64}, each instruction consists of “instruction,” “input,” and “output” parts without category annotations. This makes it challenging to adjust the distribution of categories in the instruction fine-tuning dataset to enhance the performance of specific tasks or to add and simplify instructions. Additionally, while datasets like Firefly \citep{bib59} and BELLE\_train\_3.5M\_CN \citep{bib58} have added a field for instruction categories, they suffer from issues such as incomplete or overly broad categories. Taking the “code” category as an example, instructions could be further subdivided into more granular categories like “code correction,” “code generation” and “code improvement.” Therefore, in the future, \textbf{a more fine-grained category subdivision in datasets should become a standard, allowing users to better understand the overall composition and facilitating dataset optimization}. Of course, this may introduce challenges such as difficulty in standardizing category subdivisions and increased annotation costs and time.

\textbf{Domain Scarcity.} The majority of datasets focus on general domains, with datasets in vertical domains mostly concentrated in common areas such as healthcare, finance, and law. This results in a scarcity of instruction datasets for low-resource and niche domains, potentially limiting the performance improvement of models in certain specialized fields. For instance, in fields like traditional Chinese classics, antiques, or niche areas such as paleobiology, funeral studies, and minority languages. \textbf{Constructing corresponding datasets for these domains not only systematically integrates knowledge but also allows the application of trained LLMs in specific fields, serving as auxiliary tools with societal significance and value}.

\textbf{Quality Evaluation.} The quality evaluation of instruction fine-tuning datasets is a complex and subjective issue, and currently, there are no clear, universal standards or methods. In practice, quality evaluation may involve multiple aspects, including but not limited to: \textbf{(1) Model Performance Evaluation}. Assessing the performance of the fine-tuned model on evaluation datasets. The selected evaluation datasets should be diverse and reasonable to avoid evaluation contamination \citep{bib137}. \textbf{(2) Annotation Consistency and Rationality}. Evaluating the consistency among different annotators regarding instructions and the rationality and correctness of instruction input and answer output. \textbf{(3) Bias Analysis}. Assessing biases and harmful content in the dataset to ensure the model is not adversely affected. \textbf{(4) Timeliness Detection}. Regularly checking whether the content of instructions in the dataset has become outdated or inaccurate. \textbf{(5) Subjective Evaluation}. Manually conducting subjective scoring and inspection. In conclusion, future efforts may involve establishing more explicit evaluation standards and metrics, creating a unified evaluation framework to make it more scientifically objective.

\textbf{Legal and Ethical Risks.} \cite{bib138} research on instruction fine-tuning datasets has revealed that an increasing number of datasets are treated as wholes rather than a series of sources, undergoing multiple repackagings and reauthorizations without sufficient labeling of data sources and copyright information. This leads to issues such as data leakage and biased behavior, posing legal and ethical risks. Therefore, \textbf{there is a current need to enhance the transparency of datasets, improve quality and ethical compliance, and reduce potential problems}. \cite{bib138} provides a dataset audit and data provenance explorer tool to address this. In the future, establishing standards for dataset usage is a focal point of concern.

\subsection{Preference Datasets}\label{subsec73}

The significance of preference datasets lies in providing crucial training data for the models’ output decisions. Below, we briefly discuss the challenges currently faced by preference datasets and look forward to future directions.

\textbf{Limited Availability of Resources.} RLHF has been widely researched and applied by leading industry companies such as OpenAI, Anthropic, Google, etc. However, due to the lack of high-quality, publicly available preference datasets, the open-source community is still lagging in the research and practice of RLHF \citep{bib153}. Currently, there are not many open-source preference datasets, and the majority are in English. Non-English and domain-specific preference datasets are extremely scarce. One reason for the scarcity of resources is the relatively cumbersome annotation process and the high cost involved. Therefore, exploring weakly supervised learning methods, using simple labels such as user clicks, support amounts, instead of manual annotation, or leveraging high-quality models like GPT-4 to assist in voting and scoring, could be attempted. On the other hand, there is lower attention to preference datasets in other languages and vertical domains, leading to fewer related efforts.

\textbf{Preference Evaluation Method Settings.} The most commonly used preference evaluation method is still the voting method, but many preference datasets lack strict and uniform evaluation standards, providing feedback information only from a single dimension. Human preferences in the real world are diverse, and to more comprehensively and high-quality reflect them, corresponding standards need to be established to reduce subjective differences and conduct fine-grained evaluations from multiple dimensions \citep{bib153}. Employing various evaluation methods for comprehensive assessments is recommended. Defining these standards is a complex issue. Additionally, preference datasets often do not provide explicit reasons for why some answers are more favored by humans, introducing uncertainty into the model learning process. Therefore, it is advisable to include textual explanations in preference evaluations, stating the reasons for the assessment and providing suggestions for improving the responses. The construction of UltraFeedback \citep{bib153} is relatively more scientifically standardized, playing a positive role in fostering future developments.
 
\subsection{Evaluation Datasets}\label{subsec74}

Evaluation datasets play a crucial role in ensuring the reliability, practicality, and safety of LLMs. They provide researchers and practitioners with insights into the strengths and weaknesses of LLMs, facilitating continuous improvements and optimizations. The following discussion highlights the challenges within current evaluation datasets and suggests potential directions for future development. 

\textbf{Establishment of Evaluation Datasets.} When creating an evaluation dataset for a particular domain, several essential factors must be considered. \textbf{(1) Data sources.} There is a growing emphasis on evaluating the fairness and reliability of datasets \citep{bib258}, with particular attention to the risk of data pollution or leakage during assessments \citep{bib137}. \cite{bib137} has identified instances where LLMs unintentionally learned from evaluation data during pre-training or prompt fine-tuning, resulting in inflated evaluation scores and diminished generalization ability. To mitigate this, dataset providers should disclose training data compositions and provide detailed information about data sources to prevent contamination. Consequently, beyond publicly disclosing the composition of training data to avoid inappropriate selection of evaluation datasets, providers of evaluation datasets must furnish detailed data source information and assess the risks of data contamination. Whenever possible, data sources should consist of artificially generated or non-public data to ensure fair evaluations. The challenge of minimizing data pollution or leakage remains an open problem. \textbf{(2) Question design.} Various factors, including scale, question types, and topic distribution, should be considered when developing evaluation datasets. Achieving overall enhancement requires extensive research and practical application. Initially, the scale of the evaluation dataset should be determined based on specific evaluation content, emphasizing high-quality questions, diverse question types, and an evenly distributed array of topics before gradually expanding and regularly updating the evaluation dataset. This approach resembles Chinese Gaokao, where refined questions assess the mastery of comprehensive knowledge. Additionally, setting a reasonable difficulty level is crucial. Evaluation tasks should largely surpass the current capabilities of LLMs, establishing an appropriate upper and lower limit. Without a good design of evaluation benchmarks, many models achieving scores above 95\% are relatively unhelpful for advancing LLMs \citep{bib167}.

\textbf{Addressing Evaluation Gaps.} Persistent gaps in the evaluation landscape require researchers' attention to refine the evaluation framework. \textbf{(1) Evaluating in low-resource domains.} Evaluative datasets in certain domains are in nascent stages of development, such as the e-commerce domain \citep{bib251}, and the geoscience domain \citep{bib132}; while certain domains lack pertinent evaluation benchmarks temporarily, including the domain of ancient literature, cultural artifacts, tea culture, etc. \textbf{(2) Evaluating in other languages.} Beyond the predominantly featured English and Chinese  datasets, resources for evaluations in other languages are limited. \textbf{(3) Multi-turn evaluations.} The focus on single-turn assessments overlooks LLMs’ capabilities in multi-turn interactions and contextual understanding. \textbf{(4) Dynamic evaluations.} Many evaluative datasets employ static evaluation methods, introducing two drawbacks. On one hand, the evaluation data is utilized for training to enhance ranking on leaderboards; on the other hand, the initial evaluation content may gradually become inadequate for meeting the capabilities of LLMs, and the evaluated knowledge may become obsolete or erroneous \citep{bib10}.

\textbf{Choosing and Improving Evaluation Approaches.} The limitations of code evaluation, especially for open-ended questions, require addressing. Manual evaluations, while in-depth, can be costly and subject to human bias. Thus, model-based scoring is emerging as a promising alternative, striving for scientific reliability and the goal of fully automated evaluation processes.

\textbf{Comprehensive Evaluation Framework.} The complexity of selecting from numerous datasets, the lack of standardized data formats, and the diversity in evaluation methodologies pose significant challenges. A comprehensive evaluation framework could simplify the process by providing a central repository and an efficient, standardized API for model invocation. This framework should fulfill three criteria: simplicity, centralization, and efficiency. Firstly, the evaluation steps should be straightforward, requiring only the provision of an API for model invocation. Secondly, a unified repository should be available for selecting datasets spanning diverse domains and tasks. Lastly, the evaluation process should be efficient, covering a broad range of dimensions to yield rapid results. Achieving this goal poses various challenges, with familiar frameworks like the HELM evaluation framework \citep{bib244} and the OpenCompass evaluation platform \citep{bib259} evolving in this direction.

\section{Conclusion}\label{sec8}

In the vast landscape of AI, Large Language Models (LLMs) stand out as rapidly growing, prominent features—akin to towering trees in a dense forest. The datasets that feed their growth and development can be compared to the vital root system of these trees, providing the sustenance that is essential for their performance. Regrettably, the current landscape of LLM-related datasets is extensive, with a lack of cohesive synthesis across the various types of datasets. Understanding the current state and future trends of the LLM datasets presents a formidable challenge. Therefore, this survey offers a comprehensive analysis of LLMs datasets, categorizing and summarizing datasets associated with LLMs across five dimensions: pre-training corpora, fine-tuning instruction datasets, preference datasets, evaluation datasets, and traditional NLP datasets. Alongside this categorization, we identify the current challenges and outline potential directions for future dataset development in four key areas: pre-training, fine-tuning instruction, reinforcement learning, and model evaluation. It is our hope that this survey will serve as a valuable point of reference for researchers both in academia and industry, as well as newcomers and proficient practitioners engaged with LLMs. Our ultimate objective is to continually refine LLMs datasets, to foster a robust and standardized dataset ecosystem, as well as to support the progressive advancement of LLMs.

\newpage

\begin{appendices}

\section{Pre-training Corpus Information}\label{secA}

Appendix~\ref{secA} provides detailed information on each pre-training corpus mentioned in the main text.

\subsection{General Pre-training Corpora}\label{subsecA1}

\subsubsection{Webpages}\label{subsubsecA11}

\begin{itemize}

    \item \textbf{CC-Stories} \citep{bib17}. The CC-Stories corpus is approximately 31GB in size. It is a subset extracted from Common Crawl. The selected text aligns with the style of Winograd Schema stories, providing knowledge for models in commonsense reasoning and language modeling.

    \item \textbf{CC100} \citep{bib14}. The CC100 corpus includes monolingual data from 100 languages. Its construction process involves processing URL and paragraph indices based on the CC-Net repository, utilizing snapshots from Common Crawl spanning from January to December 2018.

    \item \textbf{CLUECorpus2020} \citep{bib20}. The CLUECorpus2020 corpus is a large-scale Chinese corpus released by the CLUE organization, comprising 100GB of raw text and 35 billion Chinese characters. It is derived from the processing of Chinese data in Common Crawl from July to December 2019.

    \item \textbf{Common Crawl}\footnote{\href{https://commoncrawl.org/}{https://commoncrawl.org/}}. The Common Crawl corpus is an extensive, unstructured, multilingual dataset of webpages, encompassing over 8 years of web crawler data. The data is available in web archive, web archive transformation, and web extracted text formats. Many pre-training corpora are obtained through data preprocessing based on this corpus.

    \item \textbf{CulturaX} \citep{bib21}. The CulturaX corpus is a multilingual corpus developed for LLMs, covering 167 languages with a total of 6.3T tokens. It underwent comprehensive cleaning and deduplication processes based on mC4 and OACAR.

    \item \textbf{C4} \citep{bib12}. The C4 corpus is constructed by obtaining snapshots of Common Crawl in April 2019 and extracting pure English text using multiple filters. C4 has a total of 5 variants, namely en, en.noclean, en.noblocklist, realnewslike, and multilingual.

    \item \textbf{mC4} \citep{bib13}. The mC4 corpus consists of natural text in 108 languages, serving as a multilingual extension of C4. The data is sourced from multiple monthly web data snapshots from Common Crawl, providing a more diverse linguistic range.
    
    \item \textbf{OSCAR 22.01} \citep{bib15}. OSCAR is an open-source project aimed at providing web-based multilingual resources. The project continuously develops high-performance data processing pipelines to build multilingual corpora. Currently, there are four versions, including OSCAR 2019, OSCAR 21.09, OSCAR 22.01, and OSCAR 23.01. This paper only lists OSCAR 22.01 as a representative example.

    \item \textbf{RealNews} \citep{bib18}. The RealNews corpus is a large-scale corpus of news articles sourced from data in Common Crawl. The corpus focuses on content from the news domain indexed by Google News, with a time coverage spanning from December 2016 to April 2019.

    \item \textbf{RedPajama-V2} \citep{bib16}. The RedPajama-V2 corpus comprises over 100 billion text documents from 84 Common Crawl snapshots and has undergone processing using the CC-Net pipeline. Among them, 30 billion texts have been annotated with high-quality labels.

    \item \textbf{RefinedWeb} \citep{bib11}. The RefinedWeb corpus is the English pre-training dataset for the Falcon model. The full version of this corpus contains 5TB tokens and has undergone rigorous filtering and extensive removal of duplicate data on Common Crawl.

    \item \textbf{WuDaoCorpora-Text} \citep{bib22}. The WuDaoCorpora-Text corpus has a pure text size of approximately 5TB, comprising over 50 industry data labels such as education and technology. The corpus has crawled a rich set of Chinese webpage data. Currently, 200GB of texts have been released as open source.

\end{itemize}

\subsubsection{Languages Texts}\label{subsubsecA12}

\begin{itemize}

    \item \textbf{ANC}\footnote{\href{https://anc.org/}{https://anc.org/}}. The ANC corpus includes textual records of various written and spoken materials in the United States since 1990. It is divided into the OANC (Open American National Corpus) and the MASC (Manually Annotated Sub-Corpus).

    \item \textbf{BNC}\footnote{\href{http://www.natcorp.ox.ac.uk/}{http://www.natcorp.ox.ac.uk/}}. The BNC corpus is jointly developed and established by institutions including Oxford University Press. It consists of 4124 representative texts of a wide range of modern British English, with a vocabulary exceeding 100M words. Written language accounts for 90\%, while spoken language makes up 10\% of the corpus.

    \item \textbf{News-crawl}\footnote{\href{https://data.statmt.org/news-crawl/}{https://data.statmt.org/news-crawl/}}. The News-crawl corpus comprises news texts in 59 different languages. The texts are crawled from online newspaper resources. The corpus is utilized for the Workshop on Machine Translation (WMT) series of shared tasks.

\end{itemize}

\subsubsection{Books}\label{subsubsecA13}

\begin{itemize}

    \item \textbf{Anna’s Archive}\footnote{\href{https://annas-archive.org/datasets}{https://annas-archive.org/datasets}}. The Anna’s Archive corpus claims to be the world’s largest open-source and open-data library. It has currently gathered resources from Libgen, Sci-Hub, Z-Library, and Internet Archive Controlled Digital Lending.

    \item \textbf{BookCorpusOpen} \citep{bib29}. The BookCorpusOpen corpus is a variant of Toronto Book Corpus. It comprises 17,868 book entries, with each entry containing a title and text. The titles represent the names of the books, while the text consists of the unprocessed content of the respective books.

    \item \textbf{PG-19} \citep{bib28}. The PG-19 corpus selects 28,752 books from Project Gutenberg published before 1919, totaling 11.74GB. The reason is to avoid being affected by international copyright issues.

    \item \textbf{Project Gutenberg}\footnote{\href{https://www.gutenberg.org/}{https://www.gutenberg.org/}}. The Project Gutenberg corpus was established in 1971, making it the earliest digital library. The majority of the books within it are original works of public domain literature, preserved over the long term through digital archiving. As of July 2018, the collection comprised over 57K books.

    \item \textbf{Smashwords}\footnote{\href{https://www.smashwords.com/}{https://www.smashwords.com/}}. The Smashwords corpus is a platform for publishing e-books, and it has been in operation since 2008. It offers diverse book formats. Many corpora source their book resources from Smashwords.

    \item \textbf{Toronto Book Corpus} \citep{bib27}. The Toronto Book Corpus is a large-scale corpus of book texts compiled by crawling and organizing content from e-book websites. It comprises a total of 11,038 e-books. The resources have not been made publicly available at present.

\end{itemize}

\subsubsection{Academic Materials}\label{subsubsecA14}

\begin{itemize}

    \item \textbf{arXiv}\footnote{\href{https://arxiv.org/}{https://arxiv.org/}}. The arXiv corpus is a website that compiles preprints of papers spanning physics, mathematics, computer science, biology, and quantitative economics. Operational since 1991, this resource features papers written in LATEX. Numerous pre-training corpora source their academic material data from this repository.

    \item \textbf{S2ORC} \citep{bib53}. The S2ORC corpus stands as an extensive academic literature corpus, encompassing 81M English-language academic papers across diverse academic disciplines. It features abundant metadata, paper abstracts, and meticulously resolved bibliographic references, offering structured full text for 8.1M open-access papers. Each full-text document is meticulously annotated, incorporating automatically detected inline citations, figures, and tables.

\end{itemize}

\subsubsection{Code}\label{subsubsecA15}

\begin{itemize}

    \item \textbf{BIGQUERY} \citep{bib32}. The BIGQUERY corpus is a subset of BigQuery, comprising code from six programming languages (all under open-source licenses), including C, C++, Go, Java, JavaScript, and Python.

    \item \textbf{Github}\footnote{\href{https://github.com/}{https://github.com/}}. The Github corpus is a hosting platform that offers features such as code repository management and code snippet sharing. It houses numerous well-known open-source projects.

    \item \textbf{phi-1} \citep{bib34}. The phi-1 corpus is employed to train models capable of generating Python functions and corresponding docstrings. The corpus comprises a curated code-language dataset, around 6 billion tokens, and a Python textbook and exercise dataset synthesized by GPT-3.5.

    \item \textbf{The Stack} \citep{bib31}. The Stack consists of over 6TB of open-source code files spanning 358 programming languages, all of which are licensed under permissive licenses. It serves as the pre-training corpus for Code LLMs.

\end{itemize}

\subsubsection{Parallel Corpus}\label{subsubsecA16}

\begin{itemize}

    \item \textbf{MTP}\footnote{\href{https://data.baai.ac.cn/details/BAAI-MTP}{https://data.baai.ac.cn/details/BAAI-MTP}}. The full name of MTP is Massive Text Pairs, comprising a total of 300M aligned Chinese-English text pairs. It serves as a crucial foundation for training Chinese-English semantic vector models.

    \item \textbf{MultiUN} \citep{bib37}. The MultiUN corpus is sourced from files within the United Nations Official Document System. These files cover the six official languages, namely Arabic, Chinese, English, French, Russian, and Spanish. Some of the files also provide versions in German.

    \item \textbf{ParaCrawl} \citep{bib36}. ParaCrawl utilizes open-source software to crawl webpages, creating a publicly available parallel corpus. The corpus of version 5.0 includes 223M filtered sentence pairs from approximately 150K websites, encompassing 42 languages.

    \item \textbf{UNCorpus v1.0} \citep{bib38}. The UNCorpus v1.0 corpus consists of text content written and manually translated from the years 1990 to 2014. These contents comprise public domain United Nations official records and other conference documents, totaling 799,276 files. The majority of these files cover the six official languages.

\end{itemize}

\subsubsection{Social Media}\label{subsubsecA17}

\begin{itemize}

    \item \textbf{OpenWebText} \citep{bib40}. The OpenWebText corpus is a reproduction of WebText. It extracts post URLs from Reddit, undergoes a series of filtering, deduplication, and tokenization operations, ultimately resulting in 8,013,769 documents.

    \item \textbf{Pushshift Reddit} \citep{bib41}. The Pushshift Reddit corpus is a platform for collecting, analyzing, and archiving social media data. It has been collecting data from Reddit since 2015 and receives regular updates.

    \item \textbf{Reddit}\footnote{\href{https://www.reddit.com/}{https://www.reddit.com/}}. The Reddit corpus is an entertainment, social, and news website where users can post texts or links and vote on posts. The site covers a variety of topics including news, gaming, music, and more. Many pre-trained language models source their social media data from here.

    \item \textbf{StackExchange}\footnote{\href{https://stackexchange.com/}{https://stackexchange.com/}}. The StackExchange corpus is a Q\&A website that stores questions and their corresponding answers posed by users on the Internet. It is one of the largest publicly available resources of Q\&A pairs. One of its prominent sub-sites is StacOverflow, which caters to programmers and developers.

    \item \textbf{WebText} \citep{bib39}. The WebText corpus is an internal dataset of OpenAI. It comprises a collection of text gathered from 45M links, totaling over 8M documents. All documents related to Wikipedia have been removed from this corpus.

    \item \textbf{Zhihu}\footnote{\href{https://www.zhihu.com/}{https://www.zhihu.com/}}. The Zhihu corpus is a Chinese knowledge-sharing social platform.It enables users to ask questions, provide answers, and share knowledge, maintaining a high level of quality. Simultaneously, the platform encourages users to build social connections through interactions such as following, upvoting, commenting, and more. Many Chinese social media datasets are derived from this platform.

\end{itemize}

\subsubsection{Encyclopedia}\label{subsubsecA18}

\begin{itemize}

    \item \textbf{Baidu baike}\footnote{\href{https://baike.baidu.com/}{https://baike.baidu.com/}}. The Baidu baike corpus is an open online encyclopedia launched by Baidu, Inc. The primary language is Chinese, and it was released in 2008. As of April 2023, it has accumulated more than 27M entries.

    \item \textbf{TigerBot-wiki} \citep{bib25}. The TigerBot-wiki corpus is specifically dedicated to collecting Chinese encyclopedia-related data. This constitutes the raw external brain data used during the rethinking process of the TigerBot model, with a scale of 205MB.

    \item \textbf{Wikipedia}\footnote{\href{https://www.wikipedia.org/}{https://www.wikipedia.org/}}. The Wikipedia corpus is an online encyclopedia written in multiple languages, freely open-sourced to users. Due to its rigorous content spanning various languages and domains, people often crawl relevant data, clean it, and use it for training large-scale models. Wikipedia is widely used across various applications.

\end{itemize}

\subsubsection{Multi-category Corpora}\label{subsubsecA19}

\begin{itemize}

    \item \textbf{ArabicText 2022}\footnote{\href{https://data.baai.ac.cn/details/ArabicText-2022}{https://data.baai.ac.cn/details/ArabicText-2022}}. The ArabicText 2022 corpus is the world’s largest open-source pretraining dataset for Arabic, specifically designed for training Arabic LLMs. The creators curate, expand, and clean existing Arabic web text data, resulting in a dataset of 201.9GB. Text and knowledge-related data constitute over 65\% of the corpus.

    \item \textbf{Dolma} \citep{bib408}. The Dolma corpus is a vast English-language corpus comprising 3T tokens. It encompasses six main data types: webpages, scholarly papers, code, books, social media, and encyclopedia. For each data type, specific design principles and processing details are openly disclosed. This corpus has been instrumental in training the OLMo model. Notably, its creators have transparently disclosed the selection of data sources and provided a detailed overview of the data curation process.

    \item \textbf{MNBVC} \citep{bib23}. The MNBVC corpus is an extremely large-scale Chinese corpus with the goal of matching the 40TB data capacity used in training ChatGPT. It includes all forms of pure-text Chinese data. The corpus is continuously being cleaned and updated. Until November 2023, the scale has reached 20,811GB.

    \item \textbf{RedPajama-V1}\footnote{\href{https://huggingface.co/datasets/togethercomputer/RedPajama-Data-1T}{https://huggingface.co/datasets/togethercomputer/RedPajama-Data-1T}}. The RedPajama-V1 corpus replicates the pre-training corpora used according to report on LLaMA. The data scale is 1.2TB, encompassing five languages and six data types.

    \item \textbf{ROOTS} \citep{bib42}. ROOTS stands for Responsible Open-science Open-collaboration Text Sources. It is composed of datasets from HuggingFace, Github repositories, OSCAR, etc. The corpus has a scale of 1.6TB and includes 46 natural languages and 13 programming languages.

    \item \textbf{The Pile} \citep{bib30}. The Pile is a large-scale, diverse language modeling dataset consisting of 22 data subsets. The goal is to capture text in as many forms as possible and cover a wide range of textual content. The corpus includes academic papers, code, legal materials, patents, subtitles, chat content, parallel corpora, etc.

    \item \textbf{TigerBot\_pretrain\_en \& TigerBot\_pretrain\_zh} \citep{bib25}. These two corpora are the Chinese and English corpora used in the pre-training of TigerBot. The corpus design is based on the pre-training data distribution of GPT-3. The creators filter the collected 20TB data down to 2TB while maintaining the proportional distribution of languages and categories. Finanlly, 100GB of data is randomly sampled for open-sourcing.

    \item \textbf{WanJuanText-1.0} \citep{bib24}. The data source of WanJuanText-1.0 includes patents, textbooks, exam questions, books and other materials. The dataset comprises over 500M Chinese and English documents, totaling 1,094GB. It standardizes data from many formats into the jsonl format and undergoes thorough cleaning, deduplication, and value alignment.

\end{itemize}

\subsection{Domain-specific Pre-training Corpora}\label{subsecA2}

\subsubsection{Financial Domain}\label{subsubsecA21}

\begin{itemize}

    \item \textbf{BBT-FinCorpus} \citep{bib43}. BBT-FinCorpus is a Chinese corpus in the financial domain. The text is primarily focused on financial news, company announcements, research reports, and social media. The data is sourced from several well-known financial websites and platforms on the Chinese Internet. The corpus has a scale of approximately 256GB and is utilized for training BBT-FinT5.

    \item \textbf{FinCorpus} \citep{bib44}. FinCorpus includes text types such as company announcements, financial information and news, and financial exam questions. The data is obtained through web crawling, with a scale of approximately 60GB. It is used for training XuanYuan.

    \item \textbf{FinGLM} \citep{bib26}. FinGLM incorporates 11,588 PDF files, all of which are annual reports from listed companies for the years 2019 to 2021. The corpus also includes corresponding TXT and HTML files.

    \item \textbf{TigerBot-earning \& TigerBot-research} \citep{bib25}. These represent the raw external brain data utilized during the rethinking phase of TigerBot. The former encapsulates 2.5K financial reports, while the latter encompasses 20K financial research reports. Data is stored on a paragraph-level granularity.

\end{itemize}

\subsubsection{Medical Domain}\label{subsubsecA22}

\begin{itemize}

    \item \textbf{Medical-pt} \citep{bib45}. Medical-pt is a Chinese-language corpus in the medical field. Approximately 360K entries are derived from medical encyclopedias, and 8,475 entries are from medical textbooks. It is primarily used for incremental pre-training of models in medical knowledge.

    \item \textbf{PubMed Central}\footnote{\href{https://www.ncbi.nlm.nih.gov/pmc/}{https://www.ncbi.nlm.nih.gov/pmc/}}. PubMed Central is an open-access repository of biomedical literature, offering free resources in the field of biomedicine. It comprises approximately 5M articles. The corpus is regularly updated, providing a wealth of medical knowledge.

\end{itemize}

\subsubsection{Other Domains}\label{subsubsecA23}

See Section~\ref{subsubsec223} for details.

\section{Instruction Fine-tuning Dataset Information}\label{secB}

Appendix~\ref{secB} provides detailed information on each instruction fine-tuning dataset mentioned in the main text.

\subsection{General Instruction Fine-tuning Datasets}\label{subsecB1}

\subsubsection{Human Generated Datasets}\label{subsubsecB11}

\begin{itemize}

    \item \textbf{Aya Dataset} \citep{bib409}. The Aya Dataset is the largest human-annotated multilingual instruction fine-tuning dataset to date, comprising over 204K instances across 65 languages. On the Aya Annotation Platform, contributors engage in three tasks: creating new examples from scratch (original annotations), enhancing existing examples for improved quality and comprehensiveness (re-annotations), and providing feedback on the quality of existing contributions (annotation feedback), following the find-fix-verify paradigm.
    
    \item \textbf{databricks-dolly-15K} \citep{bib60}. The databricks-dolly-15K dataset was constructed by Databricks employees in March and April 2023, comprising 15,011 high-quality English instruction pairs. The dataset encompasses eight instruction categories and is suitable for commercial applications. The data sources include manually generated data and selected text from Wikipedia.

    \item \textbf{InstructionWild\_v2} \citep{bib62}. The InstructionWild\_v2 dataset comprises approximately 110K instructions gathered from sources such as social media and code repositories. It provides additional information, including instruction types and special labels. The dataset is suitable for non-commercial research purposes.

    \item \textbf{LCCC} \citep{bib63}. The dataset is named Large-scale Cleaned Chinese Conversation, comprising two versions: LCCC-base and LCCC-large, with 6.8M and 12M dialogues, respectively. The instructions are meticulously cleaned from 79M original dialogue data. The construction process involves acquiring user communication records from social media.

    \item \textbf{OASST1} \citep{bib61}. The OASST1 dataset is designed to advance research in model instruction fine-tuning and alignment. It consists of 161K assistant-style dialogue messages covering 35 languages. Moreover, the dataset includes quality ratings, forming over 10K fully annotated dialogue trees.

    \item \textbf{OL-CC}\footnote{\href{https://data.baai.ac.cn/details/OL-CC}{https://data.baai.ac.cn/details/OL-CC}}. The dataset is named OpenLabel-Chinese Conversations, and it is a Chinese conversational instruction dataset. The creators utilize crowdsourcing, collecting data on an open platform, resulting in 10,006 “instruction-response” text pairs with answers and 1,649 without answers. The instruction types encompass many tasks such as Q\&A, text composition, brainstorming, mathematics, and more. The dataset is completed by 276 volunteers.

    \item \textbf{Zhihu-KOL}\footnote{\href{https://github.com/wangrui6/Zhihu-KOL}{https://github.com/wangrui6/Zhihu-KOL}}. The Zhihu-KOL dataset, a Chinese conversation dataset, was constructed in March 2023 by scraping the Zhihu website. The dataset construction employed a hierarchical Q\&A categorization method, involving three separate scraping processes targeting different levels of types.

\end{itemize}

\subsubsection{Model Constructed Datasets}\label{subsubsecB12}

\begin{itemize}

    \item \textbf{Alpaca\_data} \citep{bib64}. The Alpaca\_data dataset consists of 52K instructional data points used for fine-tuning the Alpaca model. Each data instance is presented in json format, including instruction descriptions, task inputs, and the answers generated by the model.

    \item \textbf{BELLE\_Generated\_Chat} \citep{bib58}. The BELLE\_Generated\_C-hat dataset comprises approximately 400K instances of personalized character dialogues generated by the BELLE project, along with introductions for each character. These data are constructed by ChatGPT without rigorous validation and may contain errors. The category of all instructions is generation.

    \item \textbf{BELLE\_Multiturn\_Chat} \citep{bib58}. The BELLE\_Multiturn\_Chat dataset comprises approximately 800K instances of multi-turn dialogues between users and assistants, generated by the BELLE project. These data are constructed by ChatGPT without rigorous validation and may contain errors.

    \item \textbf{BELLE\_train\_0.5M\_CN} \citep{bib58}. The BELLE\_train\_0.5M\_CN dataset is a subset of the Chinese training data for the BELLE model, consisting of approximately 520K Chinese instructions. All instructions are generated by the model. The entire dataset is currently open source.

    \item \textbf{BELLE\_train\_1M\_CN} \citep{bib58}. The BELLE\_train\_1M\_CN dataset is part of the Chinese training data for the BELLE model, comprising around 917K Chinese instructions. It shares the same construction method as the BELLE\_train\_0.5M\_CN, but it undergoes post-processing to remove low-quality data.

    \item \textbf{BELLE\_train\_2M\_CN \& BELLE\_train\_3.5M\_CN} \citep{bib58}. Th-ese two datasets are Chinese instructions datasets generated by the BELLE project, comprising around 2M and 3.5M diverse task data, respectively. In comparison to previous datasets, they offer a more extensive range of training data. BELLE\_train\_3.5M\_CN expands the fields of instruction categories, covering 13 types such as generation, extraction, role-playing, and others.

    \item \textbf{CAMEL} \citep{bib70}. The CAMEL dataset features an extensive collection of around 584K instructions, among which 107K have been translated into multiple languages. This dataset provides a wide array of dialogue resources covering both multilingual and code domains. The datasets introduces a communication agent framework called “role-playing” generated through three types of prompts and involving two contexts: “AI Society” and “Programming.”

    \item \textbf{Chatgpt\_corpus}\footnote{\href{https://github.com/PlexPt/chatgpt-corpus}{https://github.com/PlexPt/chatgpt-corpus}}. The Chatgpt\_corpus dataset contains 3.27M instances of the model engaging in self-conversation. This dataset offers Chinese dialogue resources, with each instruction accompanied by a label indicating the associated domain.

    \item \textbf{InstructionWild\_v1} \citep{bib62}. The InstructionWild\_v1 dataset furnishes 52K instructions in both Chinese and English. Constructed using a model-generated approach, the dataset involves providing five example prompts to the model, which then generates new instructions along with corresponding responses. The dataset is intended for non-commercial research purposes.

    \item \textbf{LMSYS-Chat-1M} \citep{bib69}. The LMSYS-Chat-1M dataset comprises 1M instances of authentic dialogue data, collected from various models responding to questions on a website. To ensure the secure release of the data, the creators remove conversations containing personal identifying information. However, they retain unsafe dialogues for the purpose of studying robustness and security.

    \item \textbf{MOSS\_002\_sft\_data} \citep{bib65}. The MOSS\_002\_sft\_data dataset is a collection of multi-turn dialogue data utilized by the MOSS-002 project. It comprises 570K English instructions and 590K Chinese instructions. The dataset encompasses three aspects: utility, fidelity, and harmlessness, all generated by the model.

    \item \textbf{MOSS\_003\_sft\_data} \citep{bib65}. The MOSS\_003\_sft\_data dataset compiles 100K user data instances from the beta testing phase of the MOSS-002 model and generated data from GPT-3.5-Turbo. In comparison to the MOSS\_002\_sft\_data, this collection aligns more closely with the distribution of real user intents. Furthermore, it features more detailed category labels, a broader range of harmless data, and longer dialogue sequences.

    \item \textbf{MOSS\_003\_sft\_plugin\_data} \citep{bib65}. The MOSS\_003\_sft\_plugin\_da-ta dataset is an augmented version of MOSS\_003\_sft\_data, comprising around 300K multi-turn dialogue instances. It is generated by four plugins: search engine, diagram generator, calculator, and equation solver.

    \item \textbf{OpenChat} \citep{bib61}. The OpenChat dataset consists of 70K user dialogues sourced from ShareGPT, comprising 6K instances generated by GPT-4 and the rest by GPT-3.5-Turbo. This dataset provides rich information for English dialogues.

    \item \textbf{RedGPT-Dataset-V1-CN} \citep{bib66}. The RedGPT-Dataset-V1-CN dataset is a Chinese instruction dataset generated by RedGPT. The dataset is divided into two parts: RedGPT-Fact, providing instructions related to factual knowledge, and RedGPT-Code, offering dialogues related to programming. The construction process involves: (1) generating multi-turn dialogues using open-source models, (2) utilizing the dialogues for model fine-tuning to obtain the RedGPT model, and (3) employing this model to generate the final instruction data.

    \item \textbf{Self-Instruct} \citep{bib56}. The Self-Instruct dataset comprises approximately 52K English instructions obtained through the model, covering a variety of task categories. The specific construction details involve the expansion of the dataset using seed instructions. The Self-Instruct construction framework used in the dataset has been widely applied.

    \item \textbf{ShareChat}\footnote{\href{https://paratranz.cn/projects/6725}{https://paratranz.cn/projects/6725}}. The ShareChat dataset comprises approximately 90K instructions, all sourced from dialogue data on ShareGPT. In terms of language distribution, there are 68K instructions in English, 11K in Chinese, and the remaining in other languages. The aim of this dataset is to translate all instructions in other languages into Chinese, contributing to the resources of Chinese instructions. All data undergoes manual inspection and verification.

    \item \textbf{ShareGPT-Chinese-English-90k}\footnote{\href{https://huggingface.co/datasets/shareAI/ShareGPT-Chinese-English-90k}{https://huggingface.co/datasets/shareAI/ShareGPT-Chinese-English-90k}}. The ShareGPT-Chinese-English-90k dataset is a parallel bilingual Q\&A database in Chinese and English. Unlike other Q\&A content generated through repeated calls to API interfaces, this dataset has a more robust instruction distribution, making it suitable for training bilingual dialogue models. All questions are spontaneously generated by users, and most conversations with relatively poor subjective experiences have been filtered out.

    \item \textbf{ShareGPT90K}\footnote{\href{https://huggingface.co/datasets/RyokoAI/ShareGPT52K}{https://huggingface.co/datasets/RyokoAI/ShareGPT52K}}. The ShareGPT90K dataset comprises approximately 90K dialogues from ShareGPT. Primarily in English, this dataset represents authentic data reflecting interactions between users and the model.

    \item \textbf{UltraChat} \citep{bib71}. The UltraChat dataset comprises 1.47M multi-turn dialogues. The data predominantly covers three main topics: questions about the world, writing and creativity, and assistance in paraphrasing existing materials. Two independent models are employed in the construction of this dataset for dialogue generation.

    \item \textbf{Unnatural Instructions} \citep{bib73}. The collection process of the Unnatural Instructions dataset involves minimal manual labor. The creators use seed instructions to prompt the model to generate 64K examples, and then instruct the model to rephrase each instruction to further expand the dataset. In the end, approximately 240K instructions are obtained.

    \item \textbf{WebGLM-QA} \citep{bib67}. The WebGLM-QA dataset is designed for training the WebGLM generation module and comprises approximately 43K high-quality samples. Constructed using a context-guided approach, the process involves prompt formulation, guided instructions, and few-shot context learning. All instructions belong to the category of Open QA.

    \item \textbf{Wizard\_evol\_instruct\_196K \& Wizard\_evol\_instruct\_70K} \citep{bib68}. Both of these datasets consist of English instructions, with approximately 196K and 70K data instances respectively. The construction methodology is based on the Evol-Instruct approach, involving four evolutionary stages for 175 human-created seed instructions, aimed at increasing the difficulty and complexity of the instructions.

\end{itemize}

\subsubsection{Collection and Improvement of Existing Datasets}\label{subsubsecB13}

See Section~\ref{subsubsec323} for details.

\subsubsection{Datasets Created with Multiple Methods}\label{subsubsecB14}

(1) HG \& CI

\begin{itemize}

    \item \textbf{Firefly} \citep{bib59}. The Firefly dataset is a large-scale collection encompassing 23 Chinese NLP tasks. It includes tasks related to Chinese culture such as couplet creation, poetry composition, Jin Yong’s novels, prose, and more. Each task is meticulously curated with a variety of manually crafted instruction templates, complemented by additional category labels. The dataset boasts a substantial scale, amounting to 1.6M instances.

    \item \textbf{LIMA-sft} \citep{bib90}. The LIMA-sft dataset comprises 1,330 meticulously curated human-selected instructions. Employing LIMA-sft for fine-tuning on LLaMA-65B, it investigates the significance of data quality during the large-scale model instruction fine-tuning phase, demonstrating that a limited dataset size is sufficient to instruct the model in generating high-quality outputs.

\end{itemize}

\noindent (2) HG \& MC

\begin{itemize}

    \item \textbf{InstructGPT-sft} \citep{bib57}. The InstructGPT-sft dataset, used for fine-tuning the InstructGPT model, comprises 14K instructions. Part of the dataset comes from user data on the platform, while the other portion is authored by 40 trained annotators through a process involving creating simple tasks, providing few-shot tasks, and writing instructions. Currently, the dataset is not open-source.

\end{itemize}

\noindent (3) CI \& MC

\begin{itemize}

    \item \textbf{Alpaca\_GPT4\_data} \citep{bib91}. The Alpaca\_GPT4\_data dataset utilizes instruction inputs from the Alpaca\_data and generates responses using GPT-4. The dataset comprises 52K English instructions. The format of the dataset is identical to that of the Alpaca\_data, with higher-quality generated answers.

    \item \textbf{Alpaca\_GPT4\_data\_zh} \citep{bib91}. The Alpaca\_GPT4\_data\_zh dataset translates the instruction inputs from the Alpaca\_data into Chinese and then generates responses using GPT-4. The dataset consists of 52K Chinese instructions. Alpaca\_GPT4\_data\_zh is the Chinese response version of Alpaca\_GPT4\_data, but it may exhibit semantic shifts during the translation process.

    \item \textbf{Bactrain-X} \citep{bib94}. The Bactrain-X dataset comprises 3.5M instructions, spanning 52 languages. During construction, 67K English instructions from the Alpaca\_data and databricks-dolly-15K are translated into 51 other languages using a translation API, followed by models generating responses. While this dataset encompasses a diverse array of languages, its quality is contingent upon the accuracy of translation and the models’ responses.

    \item \textbf{Baize} \citep{bib99}. The Baize dataset is employed to train the Baize model. Questions are drawn from datasets such as Quora, StackOverflow, MedQuAD, and others. The dataset comprises a total of 210K English dialogue samples generated through self-dialogue using ChatGPT. It encompasses not only general domain conversations but also includes dialogue data from the medical domain.

    \item \textbf{GPT4All} \citep{bib95}. The GPT4All dataset is utilized to train the GPT4All model, comprising approximately 740K English instructions. The construction process involves gathering questions from diverse domains through public datasets, invoking the model for responses, and subsequently performing operations such as semantic similarity-based deduplication of instructions and filtering out rejected model outputs.

    \item \textbf{GuanacoDataset}\footnote{\href{https://guanaco-model.github.io/}{https://guanaco-model.github.io/}}. The GuanacoDataset, short for Generative Universal Assistant for Natural-language Adaptive Context-aware Omnilingual outputs Datasets, comprises approximately 534K instructions, spanning various languages such as English, Simplified Chinese, Traditional Chinese (Taiwan), Traditional Chinese (Hong Kong), Japanese, German, and more. It provides different language versions based on 175 seed instructions.

    \item \textbf{LaMini-LM} \citep{bib98}. The LaMini-LM dataset is employed to train the LaMini model series. It comprises 2.58M English instructions, providing the advantages of a large scale and broad topic coverage. The creators utilize various instruction inputs from existing resources, including Self-Instruct, Flan 2022, and others, invoking the model to generate responses. The generation of instructions primarily follows guided strategies based on examples and themes.

    \item \textbf{LogiCoT} \citep{bib96}. The LogiCoT dataset is primarily designed to enhance the logical reasoning abilities of models, focusing on instructions falling under the category of reasoning. The dataset comprises 605K instructions in both Chinese and English, serving as an extension to four existing open-source NLP reasoning datasets.

    \item \textbf{LongForm} \citep{bib100}. The LongForm dataset is designed to enhance models’ long-text generation capabilities, featuring approximately 28K English instructions covering tasks such as Q\&A, email composition, grammar error correction, story and poetry generation, and text summarization. The dataset is constructed based on manually created documents from C4 and Wikipedia, where different documents are selected, and model-generated instructions are derived.

    \item \textbf{Luotuo-QA-B} \citep{bib101}. The 157K Chinese-English instructions in the Luotuo-QA-B dataset are generated based on CSL, CNN-DM, and arXiv. The model generates five corresponding instruction-text pairs for each abstract or news article in the source datasets.

    \item \textbf{OpenOrca} \citep{bib97}. The OpenOrca dataset is constructed based on the Flan 2022. It comprises 1M instructions generated by GPT-4 and 3.2M instructions generated by GPT-3.5-Turbo. The dataset holds a significant advantage in terms of scale.

    \item \textbf{Wizard\_evol\_instruct\_zh} \citep{bib93}. The Wizard\_evol\_instru-ct\_zh dataset translates the instructions from Wizard\_evol\_instruct\_70K into Chinese and then invokes the model to generate responses, resulting in 70K Chinese instructions. However, the dataset may contain translation errors.

\end{itemize}

\noindent (4) HG \& CI \& MC

See Section~\ref{subsubsec324} for details.

\subsection{Domain-specific Instruction Fine-tuning Datasets}\label{subsecB2}

\subsubsection{Medical Domain}\label{subsubsecB21}

\begin{itemize}

    \item \textbf{ChatDoctor} \citep{bib111}. The release of the ChatDoctor dataset primarily addresses the limitations of existing LLMs in the field of medical knowledge. The dataset comprises 115K English dialogue samples, including authentic conversations between real patients and doctors sourced from websites, as well as model-generated dialogues and disease database information. Fine-tuning with this dataset significantly enhances the models’ abilities to understand patient needs and provide recommendations.

    \item \textbf{ChatMed\_Consult\_Dataset} \citep{bib108}. The ChatMed\_Consu-lt\_Dataset is a Chinese medical online consultation dataset comprising 549K instructions. These instructions are divided into real internet medical consultation questions and dialogues generated by the model. The dataset aims to reflect the medical consultation needs of different patients. Subsequently, the creators will filter and curate the Q\&A pairs.

    \item \textbf{CMtMedQA} \citep{bib106}. The CMtMedQA dataset is a Chinese medical multi-turn dialogue dataset consisting of 68K authentic doctor-patient conversations, featuring a substantial number of actively inquiring statements. The dataset is utilized for training the Zhongjing model, enhancing the complexity and proactive inquiry capabilities of medical dialogues.

    \item \textbf{DISC-Med-SFT} \citep{bib113}. The DISC-Med-SFT dataset is a Chinese medical instruction dataset designed for training the DISC-MedLLM model. The dataset consists of 465K samples, covering various scenarios such as single-turn medical Q\&A, multi-turn medical consultations, and multiple-choice medical Q\&A. The construction process involves the use of a target-oriented strategy, selecting high-quality open-source datasets and restructuring them.

    \item \textbf{HuatuoGPT-sft-data-v1} \citep{bib112}. The HuatuoGPT-sft-data-v1 dataset is a Chinese medical instruction dataset designed for the instruction fine-tuning phase of the HuatuoGPT model. The dataset combines refined data generated by the model and authentic dialogue data provided by real doctors, totaling 226K instructions.

    \item \textbf{Huatuo-26M} \citep{bib110}. The Huatuo-26M dataset is a Chinese medical Q\&A dataset comprising 26M high-quality medical Q\&A pairs. The medical topics covered include diseases, symptoms, treatment methods, drug information, and more. The data is sourced from various channels, including online medical encyclopedias, medical knowledge graphs, and records from online medical consultations, ensuring data diversity. Currently, part of this dataset is open source.

    \item \textbf{MedDialog} \citep{bib107}. The MedDialog dataset is a collection of medical dialogue data in both Chinese and English. The Chinese dataset comprises 3.4M doctor-patient dialogues, covering 172 disease specialties, while the English dataset includes 0.26M doctor-patient dialogues, spanning 96 disease specialties. All data in the dataset are authentic inquiries from real interactions.

    \item \textbf{Medical Meadow} \citep{bib114}. The Medical Meadow dataset is an English medical instruction dataset, consisting of a total of 160K records. It has two primary sources: firstly, open-source medical NLP task datasets that have been standardized into instruction fine-tuning format, and secondly, web scraping of medical resources from the internet. The dataset encompasses a diverse range of medical domains, including biomedicine, health, bioinformatics, and more.

    \item \textbf{Medical-sft}\footnote{\href{https://github.com/shibing624/MedicalGPT}{https://github.com/shibing624/MedicalGPT}}. The Medical-sft dataset is a bilingual medical dataset containing two parts. The first part consists of Chinese data, including 1.95M records from consultations in six medical departments, online medical encyclopedia, and Q\&A from medical knowledge graphs. The second part comprises English medical inquiry dialogue data and NLP datasets, totaling 110K records.

    \item \textbf{QiZhenGPT-sft-20k}\footnote{\href{https://github.com/CMKRG/QiZhenGPT}{https://github.com/CMKRG/QiZhenGPT}}. The QiZhenGPT-sft-20k dataset is a collection of 20K Chinese medical instructions. The data is sourced from the Qizhen medical knowledge base and includes real doctor-patient knowledge Q\&A data, as well as instructions constructed from text knowledge based on drugs and diseases. It is primarily used to enhance the models’ accuracy in medical knowledge Q\&A and alleviate hallucination phenomena.

    \item \textbf{ShenNong\_TCM\_Dataset} \citep{bib109}. The ShenNong\_TC-M\_Dataset is a Chinese medical dataset. Based on an open-source traditional Chinese medicine knowledge graph, the dataset utilizes the Self-Instruct method to construct instruction data centered around traditional Chinese medicine. In total, it comprises 112K records. The dataset represents a promising resource in the field of traditional Chinese medicine.

\end{itemize}

\subsubsection{Code Domain}\label{subsubsecB22}

\begin{itemize}

    \item \textbf{Code\_Alpaca\_20K} \citep{bib116}. The Code\_Alpaca\_20K dataset is designed for fine-tuning the Code Alpaca model. The construction of this dataset follows the method used in the Alpaca\_data, resulting in 20K instructions. Its strength lies in contributing a dataset of code-related instructions.

    \item \textbf{CodeContest} \citep{bib117}. The CodeContest dataset is a collection of data related to programming contests, featuring 13.6K code competition examples. The data is sourced from Codeforces, Description2Code, and CodeNet. The dataset is characterized by its rich set of code instructions.

    \item \textbf{CommitPackFT} \citep{bib210}. The CommitPackFT dataset undergoes filtering based on the original dataset. The original dataset covers 350 programming languages, totaling 4TB. After filtering, it retains 702K instructions, supporting 277 programming languages. Multiple quality filters are applied during data processing to preserve content with commercially friendly licenses.

    \item \textbf{ToolAlpaca} \citep{bib120}. The ToolAlpaca dataset aims to enhance models’ abilities to use common tools, comprising a total of 3,928 instances and over 400 tools. During construction, 500 randomly selected APIs from a public API repository serve as a starting point. The models are then employed to generate more comprehensive documentation, resulting in the creation of a diverse collection of tools.

    \item \textbf{ToolBench} \citep{bib121}. The ToolBench dataset is a tool usage dataset created automatically by a model. The construction process primarily involves three stages: firstly, the collection of 16,464 real tool APIs covering 49 categories; secondly, the use of the model to generate various instructions for these APIs, including single-tool and multi-tool scenarios; and finally, the use of the model to search for effective solution paths for each instruction. The dataset comprises a total of 126K instances, providing a rich resource for tool invocation.

\end{itemize}

\subsubsection{Legal Domain}\label{subsubsecB23}

\begin{itemize}

    \item \textbf{DISC-Law-SFT} \citep{bib122}. The DISC-Law-SFT dataset is a Chinese legal instruction dataset that covers various judicial application scenarios, including legal information extraction, judgment prediction, document summarization, and legal Q\&A. The dataset comprises a total of 403K instructions and is divided into two subsets: DISC-Law-SFT-Pair and DISC-Law-SFT-Triplet. The former introduces legal reasoning capabilities, while the latter enhances the models’ abilities to utilize external knowledge. The data is sourced from three components: NLP judicial task public datasets, legal original texts, and general domain data. The creators utilize three approaches—behavior shaping, knowledge expansion, and mindset cultivation—to reconstruct the instruction data and improve data quality.

    \item \textbf{HanFei 1.0} \citep{bib123}. The HanFei 1.0 dataset is a Chinese legal instruction dataset that includes both general instructions and legal instructions. The total scale of the dataset is 255K instructions, with 147K specifically related to legal content. The dataset is constructed using rule-based filtering, and future versions will incorporate manual curation.

    \item \textbf{LawGPT\_zh} \citep{bib124}. The LawGPT\_zh dataset is a Chinese legal instruction dataset, primarily divided into two parts: scenario dialogues and legal knowledge Q\&A. The scenario dialogues consist of 200K authentic conversations between lawyers and users. After reprocessing the Q\&A using the model, 52K single-turn Q\&A and 92K scenario Q\&A with legal basis are obtained. The other part involves generating legal knowledge-related Q\&A pairs through a self-built legal professional knowledge database, which is currently not yet open source.

    \item \textbf{Lawyer LLaMA\_sft} \citep{bib125}. The Lawyer LLaMA\_sft dataset is a Chinese legal instruction dataset, totaling 21.5K records. The primary sources include model-generated answers to Chinese judicial exam questions, responses to legal consultations, and multi-turn legal consultation dialogues generated based on legal provisions. Currently, only a portion of the dataset is open source.

\end{itemize}

\subsubsection{Mathematics Domain}\label{subsubsecB24}

\begin{itemize}

    \item \textbf{BELLE\_School\_Math} \citep{bib58}. The BELLE\_School\_Math data-set is a Chinese mathematical question dataset released as part of the BELLE project, comprising approximately 248K mathematical questions along with their solution processes. All answers to the questions are generated by the model and have not undergone rigorous verification, thus potential errors may exist in both the questions and the solution processes.

    \item \textbf{Goat} \citep{bib126}. The Goat dataset is an instruction-synthesized dataset in the field of mathematics, consisting of 1.74M synthetic data instances for mathematical arithmetic tasks. Each instance includes instructions for an arithmetic expression, a randomly generated arithmetic expression in code, and the target output. However, the dataset is limited to arithmetic tasks involving addition, subtraction, multiplication, and division in the field of mathematics.

    \item \textbf{MWP} \citep{bib127}. The MWP dataset is focused on tasks related to solving mathematical word problems (MWP). It integrates eight popular MWP datasets, categorizing them into single-equation and multiple-equation types. The dataset comprises approximately 252K problems, providing a diverse corpus for studying the resolution of mathematical problems.

    \item \textbf{OpenMathInstruct-1} \citep{bib411}. OpenMathInstruct-1, a comprehensive math instruction tuning dataset, features 1.8M pairs generated by the Mixtral-8x7B model.   It encompasses subsets from GSM8K and MATH, offering synthetically generated solutions. The dataset is thoughtfully divided into train and validation subsets to cover the entirety of the training sets. It is constructed using the methods of prompting novelty and brute-force scaling.

\end{itemize}

\subsubsection{Education Domain}\label{subsubsecB25}

\begin{itemize}

    \item \textbf{Child\_chat\_data}\footnote{\href{https://github.com/HIT-SCIR-SC/QiaoBan}{https://github.com/HIT-SCIR-SC/QiaoBan}}. The Child\_chat\_data dataset, comprising 5K instances of Chinese children’s emotional companionship dialogue, serves as the training data for the QiaoBan model. The construction process unfolds through two key phases: (1) Sampling from real-life scenarios, volunteers curate high-quality emotional companionship dialogue data based on topic lists derived from genuine children’s conversations. Expert scholars actively participate, offering insights and recommendations to enhance the dataset’s quality. (2) Model-generated dialogue data is produced, catering to different topics within the dataset.

    \item \textbf{Educhat-sft-002-data-osm} \citep{bib128}. The Educhat-sft-002-data-osm dataset, consisting of 4.28M dialogues in both Chinese and English, is employed to train the EduChat model. This dataset amalgamates diverse educational data, enabling the model to possess functionalities such as question generation, homework grading, emotional support, and course guidance.

    \item \textbf{TaoLi\_data} \citep{bib129}. \cite{bib129} is building an international Chinese education resource library that includes over 500 international Chinese textbooks, HSK exam questions, Chinese dictionaries, and other resources. Based on this resource library, TaoLi\_data is being constructed. The task types for instructions involve grammar correction, meaning generation, text simplification, and controlled text generation, totaling 88K instances. Some of the data is generated by the model, and errors may occur.

\end{itemize}

\subsubsection{Other Domains}\label{subsubsecB26}

See Section~\ref{subsubsec336} for details.

\section{Preference Dataset Information}\label{secC}

Appendix~\ref{secC} provides detailed information on each preference dataset mentioned in the main text.

\subsection{Vote}\label{subsecC1}

\begin{itemize}

    \item \textbf{Chatbot\_arena\_conversations} \citep{bib143}. The Chatbot\_arena\_con-versations dataset collects 33K examples from Chatbot Arena, spanning from April to June 2023. Each example includes a question ID, the names and responses of two models, the choice of a human judge, language labels, toxic labels, and more. After analysis, a total of 20 models’ outputs and 96 languages are identified. Personal information is removed, and unsafe conversations are labeled and retained.

    \item \textbf{CValues} \citep{bib149}. The CValues dataset, also known as the CValues-Comparison dataset, consists of 145K aligned value samples. These samples are Chinese data in the domain of social norms. The dataset encompasses three types of responses: Safe and Responsibility, Safe, and Unsafe, ranked in descending order of safety. Through processes such as expanding seed instructions, model responses, categorizing positive and negative samples, and model rewrites, different responses are assigned types, creating safety comparisons between pairs.

    \item \textbf{hh-rlhf} \citep{bib144}. The hh-rlhf dataset consists of approximately 170K examples. Each line in each jsonl file of the dataset represents a pair of selected and rejected responses. The construction process involves crowdsourced workers choosing one response to continue the conversation based on the replies of two models. The collection process primarily includes basic model extraction, rejection sampling, and online iterative sampling. During annotation, creators encourage individuals to make selections based on their own criteria to maintain diversity in the data, although this approach may introduce subjectivity issues.

    \item \textbf{MT-Bench\_human\_judgments} \citep{bib143}. The MT-Bench\_huma-n\_judgments dataset is obtained through pairwise preference comparisons conducted by graduate students for 80 instructions generated separately by six models. The six models include GPT-4, GPT-3.5, Claude-v1, Vicuna-13B, Alpaca-13B, and LLaMA-13B. The data is in English, and the dataset is relatively small, comprising only 3.3K examples.

    \item \textbf{PKU-SafeRLHF} \citep{bib146}. The PKU-SafeRLHF dataset contains 362K human-annotated English data. The construction involves a two-stage annotation process. In the first stage, instructions are evaluated for harmlessness across 14 harmful categories. In the second stage, preferences are selected based on both usefulness and harmlessness. Each open-sourced example includes two responses along with preference information.

    \item \textbf{SHP} \citep{bib147}. The SHP dataset consists of 385K examples covering 18 topics. Each example includes a question and a pair of responses from Reddit posts, with one response being more favored by users. In contrast to the hh-rlhf dataset, the questions and answers in SHP are manually crafted rather than generated by models, enhancing authenticity.

    \item \textbf{Summarize\_from\_Feedback} \citep{bib148}. The purpose of creating the Summarize\_from\_Feedback dataset is to optimize summary generation models through human feedback. The dataset is divided into two parts: Comparisons and Axis. The former involves annotators selecting the better summary from two alternatives, while the latter includes annotators rating the quality of summaries using the Likert scale. In total, the dataset comprises approximately 194K examples focused on the news domain.
    
    \item \textbf{Zhihu\_rlhf\_3k}\footnote{\href{https://huggingface.co/datasets/liyucheng/zhihu_rlhf_3k}{https://huggingface.co/datasets/liyucheng/zhihu\_rlhf\_3k}}. The Zhihu\_rlhf\_3k dataset comprises 3,460 examples from Zhihu. Similar to SHP, each example consists of two responses, with the more popular answer determined by user votes, reflecting genuine user preferences. This dataset provides valuable Chinese preference instruction resources, which are relatively scarce.

\end{itemize}

\subsection{Sort}\label{subsecC2}

\begin{itemize}

    \item \textbf{OASST1\_pairwise\_rlhf\_reward}\footnote{\href{https://huggingface.co/datasets/tasksource/oasst1_pairwise_rlhf_reward}{https://huggingface.co/datasets/tasksource/oasst1\_pairwise\_rlhf\_reward}}. The OASST1\_pairwise\_rlhf\_reward dataset consists of 19K examples obtained through post-processing on the OASST1 dataset. The source dataset itself includes human quality ratings for different responses, allowing for a direct transformation into the form of preference data based on annotations, reflecting human preferences in a sorted manner.

\end{itemize}

\subsection{Score}\label{subsecC3}

\begin{itemize}

    \item \textbf{Alpaca\_comparison\_data} \citep{bib91}. The Alpaca\_comparison\_data dataset consists of 51K examples comparing three models. The results of the comparisons serve as a form of preference feedback. The preference evaluation method involves using GPT-4 to score the quality of responses, thus creating preference samples. Each example includes a prompt input, a high-quality answer, and a low-quality answer.

    \item \textbf{Stable\_Alignment} \citep{bib152}. The Stable\_Alignment dataset is used to train social intelligence agents to better align their responses. Examples are categorized into three types from simulated social interactions: imitation, self-critic, and realignment, totaling 168K examples. These agents learn to adjust their responses based on social value through simulated social interactions. Each example includes multiple different responses generated by the model and their corresponding scores.
    
    \item \textbf{Stack-Exchange-Preferences} \citep{bib150}. The Stack-Exchange-Preferences dataset comprises 10.8M examples sourced from Q\&A interactions on StackOverflow. Each answer is assigned a score based on two factors: the number of upvotes and whether it was accepted by the questioner. The score reflects the preference, with higher scores indicating stronger preference.

    \item \textbf{UltraFeedback} \citep{bib153}. The UltraFeedback dataset is a large-scale, diverse, and fine-grained preference dataset, consisting of approximately 64K English examples. Each example includes responses from four different models, model ratings for the responses, and detailed textual explanations for the ratings. The models assess the responses from four dimensions: instruction-following, truthfulness, honesty, and helpfulness. Instructions are sourced from various publicly available datasets, and the models randomly choose four out of 17 for response generation.
    
    \item \textbf{WebGPT} \citep{bib151}. The WebGPT dataset consists of approximately 19.6K examples. Each example includes answers from two model responses to a given question, along with relevant metadata. The answers are manually rated, and each final answer is assigned a preference score to determine its quality. The entire construction process involves collecting questions from the general domain.

\end{itemize}

\subsection{Other}\label{subsecC4}

See Section~\ref{subsubsec414} for details.

\section{Evaluation Dataset Information}\label{secD}

Appendix~\ref{secD} provides detailed information on each evaluation dataset mentioned in the main text.

\subsection{General}\label{subsecD1}

\begin{itemize}

    \item \textbf{AlpacaEval} \citep{bib156}. The 805 English instructions in the AlpacaEval dataset are sourced from various datasets, including Self-Instruct, Vicuna Evaluation, and others. The dataset primarily assesses the performance of LLMs on a variety of subjective open-ended questions in the general domain, employing models such as GPT-4 to score the outputs.

    \item \textbf{BayLing-80} \citep{bib157}. The BayLing-80 dataset comprises 320 single-turn and multi-turn instructions in both Chinese and English. Starting with the translation of 80 English instructions from Vicuna Evaluation into Chinese, a second round of instructions was manually expanded to create both single-turn and multi-turn instructions in both languages. The dataset primarily evaluates the cross-lingual and conversational capabilities of LLMs, covering nine tasks, including writing, roleplay, common-sense, fermi, counterfactual, coding, math, generic, and knowledge. GPT-4 is used for scoring in the evaluation process.

    \item \textbf{BELLE\_eval} \citep{bib158}. The BELLE\_eval dataset comprises 1K Chinese instructions created by the BELLE project. The dataset primarily assesses the general capabilities of LLMs in a Chinese context, covering nine tasks: extract, closed QA, rewrite, summarization, generation, classification, brainstorming, open QA, and others. The “others” category mainly focuses on tasks related to mathematics and coding. The evaluation is conducted using ChatGPT for scoring.

    \item \textbf{CELLO} \citep{bib160}. The CELLO dataset comprises 523 English directives, all derived from data manually curated in real-world situations. Its principal objective is to gauge the proficiency of LLMs in comprehending intricate instructions. The evaluation encompasses ten subtasks, addressing aspects related to both complex task description and complex input. The evaluation methodology employs code assessment.

    \item \textbf{MT-Bench} \citep{bib143}. The MT-Bench dataset encompasses 80 English instructions, all meticulously forged by human artisans. The principal objective of this dataset is to appraise the comprehensive competency of LLMs within the English milieu. It spans eight varied tasks, covering realms such as writing, roleplay, reasoning, mathematics, programming, information extraction, STEM (Science, Technology, Engineering, and Mathematics), and humanities. The assessment approach entails leveraging GPT-4 for the scoring process.

    \item \textbf{SuperCLUE} \citep{bib159}. The SuperCLUE dataset functions as an extensive benchmark designed to appraise the proficiency of large-scale Chinese models. It encompasses SuperCLUE-OPEN, targeting multi-turn open-ended questions, and SuperCLUE-OPT, focused on objective questions that test three primary capabilities. This dataset predominantly scrutinizes the models’ prowess in handling Chinese language tasks, spanning a spectrum of over a hundred subtasks. It undergoes monthly updates, with 3458 questions in September and 3754 questions in October. The evaluation combines manual assessment and code-based evaluation.

    \item \textbf{Vicuna Evaluation}\footnote{\href{https://github.com/lm-sys/vicuna-blog-eval}{https://github.com/lm-sys/vicuna-blog-eval}}. The Vicuna Evaluation dataset encompasses 80 instructions in the English language, meticulously composed by human creators. Its principal aim is to conduct a preliminary evaluation of the overall proficiencies of LLMs, addressing nine diverse tasks such as writing, roleplay, common-sense, fermi, counterfactual, coding, math, generic, and knowledge. The assessment approach employs GPT-4 to compare two responses.

\end{itemize}

\subsection{Exam}\label{subsecD2}

\begin{itemize}

    \item \textbf{AGIEval} \citep{bib162}. The AGIEval dataset encompasses 8,062 directives presented in both Chinese and English, featuring a combination of segments extracted from publicly available datasets and segments meticulously devised through manual efforts. The dataset’s principal objective revolves around appraising the competencies of models when engaged in tasks associated with human cognition and problemsolving, achieved through the scrutiny of 20 meticulously designed entrance and qualification examinations. The evaluative content is sourced from diverse domains, spanning general university admissions assessments (GRE, Gaokao, SAT), specialized entry evaluations (LSAT, GMAT), challenges derived from high school mathematical competitions (AMC, AIME), China’s civil service entrance examinations, and legal licensure tests. The question formats encompass multiple-choice queries and the completion of blanks, with the evaluation methodology grounded in code-based scrutiny.

    \item \textbf{GAOKAO-Bench} \citep{bib161}. The GAOKAO-Bench dataset incorporates 2,811 directives in Chinese, sourced exclusively from meticulously crafted authentic questions found in Gaokao. Comprising a spectrum of 10 subjects—ranging from the Chinese to science and liberal arts mathematics, English, physics, chemistry, biology, geography, politics, and history—the dataset serves as a means to assess the holistic capacities of LLMs. These capacities encompass language understanding and logical deduction, as manifested in their responses to Gaokao queries. The evaluative content spans questions spanning the period from 2010 to 2022, encompassing 1,781 objective questions and 1,030 subjective questions derived from GaoKao. The evaluative methodology entails automated scrutiny for objective questions and expert-assigned scores for subjective questions.

    \item \textbf{M3Exam} \citep{bib163}. The dataset named M3Exam encompasses 12,313 multiple-choice questions extracted from exams at primary, middle, and high school levels across nine countries and utilizing nine distinct languages. Approximately 23\% of the evaluation tasks include visual elements, testing the models’ capabilities from various linguistic, modal, and hierarchical perspectives.

\end{itemize}

\subsection{Subject}\label{subsecD3}

\begin{itemize}

    \item \textbf{ARB} \citep{bib167}. The ARB dataset contains 1,207 instructions in English, featuring sophisticated reasoning challenges spanning mathematics, physics, biology, chemistry, and law, delving into more intricate layers of knowledge. The questions encompass multiple-choice, brief-response, and open-answer formats, utilizing a blended assessment methodology involving code, human evaluation, and model analysis. The initiators introduce a rule-driven evaluation approach, enabling GPT-4 to assign scores to intermediary reasoning steps.

    \item \textbf{C-CLUE}\footnote{\href{https://github.com/jizijing/C-CLUE}{https://github.com/jizijing/C-CLUE}}. Derived from a crowdsourced annotation system, the C-CLUE dataset stands as a benchmark for evaluating classical Chinese language comprehension. It consists of 19,150 entities and 4,365 relation pairs. The primary focus of this dataset is to gauge LLMs’ proficiency in tasks related to NER and RE within the field of classical Chinese language studies. The assessment methodology is grounded in code-based evaluations.

    \item \textbf{C-Eval} \citep{bib169}. The C-Eval dataset comprises 13,948 Chinese multiple-choice questions, spanning 52 different academic disciplines and categorized into four difficulty levels. The subject categories are primarily divided into STEM, social science, humanity, and other. Some of the data is derived from freely available simulated and past-year exam questions on the Internet, while the remaining data, not freely accessible to the public, has been obtained with the appropriate authorization. The evaluation methodology involves code-based assessments.

    \item \textbf{CG-Eval} \citep{bib170}. Within the CG-Eval dataset, there exist 11K Chinese questions, encompassing a spectrum of six major categories: science and engineering, humanities and social sciences, mathematical computation, medical qualification exams, judicial exams, and certified public accountant exams, further segmented into 55 subtopics. This dataset serves as a counterpart to MMCU, with a focused emphasis on appraising the prowess of Chinese text generation within the academic realm. The evaluation employs a comprehensive scoring system, summing diverse criteria for non-computational questions and amalgamating computed results and problem-solving processes for computational queries.

    \item \textbf{CMMLU} \citep{bib172}. The CMMLU dataset functions as an all-encompassing Chinese assessment standard, covering a total of 67 academic fields, spanning from fundamental subjects to advanced professional domains. These disciplines include not only the natural sciences that demand computational reasoning but also the humanities and social sciences that require knowledge. Additionally, there are region-specific categories like Chinese driving rules and dietary culture. Given the presence of China-specific answers in numerous tasks, it stands as a thoroughly Sinicized evaluation benchmark.
    
    \item \textbf{LLMEVAL-3}\footnote{\href{https://github.com/llmeval/llmeval-3}{https://github.com/llmeval/llmeval-3}}. The LLMEVAL-3 dataset encompasses around 200K questions designed in a free-response format, classified into 13 overarching academic domains and spanning more than 50 specific sub-disciplines, systematically probing into the depth of expertise in specialized knowledge. The questions are predominantly curated from undergraduate assignments, examinations, and graduate entrance assessments. Meticulous efforts are made by the creators to procure evaluation content from sources beyond the internet sphere. Throughout the evaluation process, models are presented with a randomized subset of 1K questions drawn from the question bank, with their responses subjected to assessment through the GPT-4 scoring methodology.

    \item \textbf{MMCU} \citep{bib164}. The MMCU dataset incorporates 11,845 Chinese multiple-choice questions, spanning 25 subtasks across disciplines such as medicine, law, psychology, and education. These questions are curated with precision by experts who manually gather them from freely accessible online repositories, encompassing materials like legal qualification exams, psychological counselor certification tests, and Gaokao. The evaluation methodology applied involves the utilization of code-based assessment.

    \item \textbf{MMLU} \citep{bib171}. The MMLU dataset encompasses 15,908 multiple-choice questions in English, providing a benchmark to evaluate model knowledge proficiency through both zero-shot and few-shot assessments. Covering 57 subjects, including STEM, humanities, and social sciences, the benchmark spans difficulty levels from elementary to advanced. Students manually gathered questions from various online free resources, incorporating exercises from different subjects. The evaluation employs a code-based assessment methodology.

    \item \textbf{M3KE} \citep{bib173}. Comprising 20,477 multiple-choice questions in Chinese, the M3KE dataset spans 71 academic disciplines, ranging from primary education to university levels. It is broadly classified into arts \& humanities, social sciences, natural sciences, and other categories. The assessment methodology involves employing code-based evaluations.

    \item \textbf{SCIBENCH} \citep{bib165}. The SCIBENCH dataset includes 695 English questions derived from educational materials, functioning as an evaluative standard for university-level STEM disciplines such as mathematics, physics, and chemistry. Its primary focus lies in assessing the models’ intricate reasoning capabilities, knowledge proficiency, and computational skills. The questions are manually formulated, and the evaluation methodology encompasses code-based assessments.

    \item \textbf{ScienceQA} \citep{bib168}. The ScienceQA dataset consists of 21,208 multimodal English multiple-choice questions originating from primary and secondary school science courses. Within this set, 16,864 questions incorporate images, while 10,220 questions incorporate textual context. The dataset evaluates the scientific literacy of LLMs through a methodology based on code assessments.

    \item \textbf{TheoremQA} \citep{bib166}. The TheoremQA dataset includes 800 English questions formulated from a set of 350 theorems spanning mathematics, physics, finance, and CS \& EE. Highly specialized human experts meticulously curate the data, guaranteeing elevated quality and a moderate level of complexity.

    \item \textbf{XiezhiBenchmark} \citep{bib174}. The XiezhiBenchmark dataset encompasses 249,587 dual-language multiple-choice questions, representing 516 academic disciplines across 13 categories. These questions are predominantly sourced from two channels: approximately 170K questions gathered from six distinct examinations and roughly 80K questions autonomously generated through an automatic updating framework. The assessment methodology involves code-driven evaluations.

\end{itemize}

\subsection{Natural Language Understanding}\label{subsecD4}

\begin{itemize}

    \item \textbf{CLUE} \citep{bib17}. The CLUE dataset consists of 9 NLU Chinese datasets, namely TNEWS, IFLYTEK, CLUEWSC2020, AFQMC, CSL, OCNLI, CMRC 2018, ChID, and C3. The evaluated NLU tasks include text classification, coreference resolution, semantic matching, reading comprehension, and textual entailment.

    \item \textbf{CUGE} \citep{bib179}. The CUGE dataset comprises 21 NLU datasets, encompassing 7 language abilities and 18 predominant NLP tasks. These 7 abilities span language comprehension at the word and sentence levels, comprehension at the document level, information retrieval and question answering, language generation, conversational interaction, multilingualism, and mathematical reasoning.

    \item \textbf{GLUE} \citep{bib177}. The GLUE dataset consists of 9 NLU English datasets, namely CoLA, SST-2, MRPC, STS-B, QQP, MNLI, QNLI, RTE, and WNLI. The evaluated NLU tasks include grammaticality judgment, sentiment analysis, semantic matching, textual entailment, reading comprehension, and coreference resolution.

    \item \textbf{MCTS} \citep{bib178}. Containing 723 Chinese test samples, the MCTS dataset is the most extensive and widely cited evaluation dataset for tasks related to simplifying Chinese text. Originating from complex structures extracted from news corpora, each original sentence corresponds to multiple manually simplified versions. The main focus is on evaluating the understanding and rewriting capabilities of LLMs when dealing with intricate Chinese texts.

    \item \textbf{RAFT} \citep{bib176}. The RAFT dataset comprises 28,712 English test samples, serving as a real-world few-shot text classification benchmark. The 11 sub-datasets within it are all binary or multi-classification tasks, covering text content from various domains such as healthcare, customer interactions, Twitter, and more.

    \item \textbf{SentEval} \citep{bib181}. The SentEval dataset comprises 11 downstream tasks and 10 probing tasks, making a total of 21 subtasks. It functions as an evaluation toolkit for universal sentence representations, covering a range of tasks including binary classification, multi-classification, natural language inference (NLI), and semantic matching.
    
    \item \textbf{SuperGLUE} \citep{bib17}. An advanced iteration of GLUE, the SuperGLUE dataset serves as an expanded and enhanced benchmark for assessing NLU. Consisting of 8 NLU English datasets—BoolQ, ReCoRD, CB, WiC, WSC, RTE, COPA, and MultiRC—this dataset raises the bar for evaluating reading comprehension, textual entailment, semantic matching, and coreference resolution.

\end{itemize}

\subsection{Reasoning}\label{subsecD5}

\begin{itemize}

    \item \textbf{Chain-of-Thought Hub} \citep{bib182}. The Chain-of-Thought Hub dataset encompasses eight open-source datasets, establishing a comprehensive inventory of intricate reasoning tasks encompassing mathematics (GSM8K), science (MATH, TheoremQA), symbols (BBH), knowledge (MMLU, C-Eval), encoding (HumanEval), and facts (SummEdits). The evaluation employs the approach of few-shot CoT prompting.

    \item \textbf{Choice-75} \citep{bib183}. The Choice-75 dataset consists of 650 multiple-choice questions in English, establishing the inaugural benchmark to evaluate the decision reasoning prowess of LLMs within descriptive scenarios. The dataset is composed of 75 scripts and over 600 scenarios, classified based on different difficulty levels. Models are tasked with selecting the optimal option from two alternatives within predefined scenarios.

    \item \textbf{LILA} \citep{bib186}. The LILA dataset evaluates LLMs’ mathematical reasoning skills through 23 tasks across four dimensions. It scrutinizes fundamental mathematical skills, algebra, probability theory, calculus, and other pertinent mathematical knowledge. These four dimensions encompass mathematical proficiency, language format, language diversity, and external knowledge.

    \item \textbf{MiniF2F\_v1} \citep{bib187}. The MiniF2F\_v1 dataset consists of 488 statements presenting Olympiad-level mathematical problems, aiming to evaluate the neural mathematical reasoning capabilities. The data is gathered from diverse mathematical competitions, including AIME, AMC, IMO, along with materials from high school and undergraduate mathematics courses.
    
    \item \textbf{NeuLR} \citep{bib184}. The NeuLR dataset comprises 3K reasoning questions, representing an improvement over 15 standard logical reasoning datasets. Its primary focus is on assessing three distinct reasoning capabilities: deductive reasoning, inductive reasoning, and abductive reasoning. The evaluation system for logical reasoning capabilities is ultimately formed by six dimensions: accuracy, precision, self-awareness, activeness, orientation, and absence of hallucination.

    \item \textbf{TabMWP} \citep{bib185}. The TabMWP dataset includes 38,431 questions, with a primary focus on evaluating the mathematical reasoning prowess of LLMs in handling both textual and tabular data. This dataset serves as a benchmark for addressing more intricate challenges, particularly those related to models processing heterogeneous information. The questions are divided into two categories: question-answering and multiple-choice, and each is annotated with golden answers, offering a detailed understanding of the multi-step reasoning process.

\end{itemize}

\subsection{Knowledge}\label{subsecD6}

\begin{itemize}

    \item \textbf{ALCUNA} \citep{bib192}. The ALCUNA dataset includes 84,351 English queries and 3,554 independently created entities. Each entity, on average, encompasses 11.75 sets of attribute triples and 25.39 linked entities. Employing the innovative knowGen technique, adjustments are made to the attributes and relationships of existing entities to produce fresh knowledge. This process forms the foundation of ALCUNA, designed to evaluate LLMs’ proficiency in comprehending, distinguishing, and associating with novel knowledge.

    \item \textbf{KoLA} \citep{bib190}. The KoLA dataset encompasses 2,138 English test instances, assessing LLMs’ capacity to rationally deduce and generate knowledge at four hierarchical levels: memory, comprehension, application, and innovation. To ensure impartiality, the evaluation employs standardized scores and a comparative assessment system, with quarterly dataset updates implemented to mitigate the potential for model impropriety.

    \item \textbf{LLMEVAL-2} \citep{bib188}. The LLMEVAL-2 dataset comprises 480 Chinese questions that assess knowledge across different domains. The questions, sourced from external databases spanning 12 domains, encompass both multiple-choice and open-ended formats. Evaluation is conducted through a combination of manual and automated assessment methods.
    
    \item \textbf{LMExamQA} \citep{bib189}. The LMExamQA dataset comprises 10,090 English test instances, classified into knowledge recall, understanding, and analysis categories based on the complexity of the questions. It encompasses 25 different domains. The dataset introduces an innovative framework, treating language models as knowledgeable examiners who generate questions based on their understanding and evaluate responses without external references.
    
    \item \textbf{SocKET} \citep{bib191}. The SocKET dataset encompasses around 2.6M English test samples drawn from 58 NLP datasets designed to assess social knowledge. It functions as a benchmark for evaluating LLMs’ proficiency in comprehending various aspects of social knowledge, categorized into humor and sarcasm, aggression, emotion, credibility, and social facts. The task types encompass classification, regression, pair-wise comparison, and span identification.

\end{itemize}

\subsection{Long Text}\label{subsecD7}

\begin{itemize}

    \item \textbf{InfiniteBench} \citep{bib198}. The InfiniteBench dataset fills the void in assessing long texts beyond 100K, elevating the input length from the previous 10K to over 100K and reaching a maximum of 2M. Originating from five domains, namely mathematics, code, dialogue, books, and retrieval, the dataset incorporates diverse task formats, including Q\&A, multiple-choice, and summarization. With the exception of key information retrieval tasks, all others constitute novel evaluation tasks.
    
    \item \textbf{L-Eval} \citep{bib195}. The L-Eval dataset comprises 411 lengthy documents and 2,043 English prompts, assessing the capabilities of LLMs across diverse tasks when exposed to extensive text inputs. These tasks necessitate reasoning on prolonged textual content, encompassing activities such as text summarization, Q\&A, context assimilation, topic retrieval, and aiding in academic paper composition. The dataset is meticulously annotated and scrutinized, drawn from a diverse array of platforms and origins. The evaluation approach integrates code-based, human-based, and model-based assessments. The input length ranges from 4K to 60K.

    \item \textbf{LongBench} \citep{bib193}. The LongBench dataset encompasses 4,750 test instances, presented in both Chinese and English, and stands as the inaugural benchmark for a thorough assessment of LLMs’ abilities in cross-lingual, multitask, and comprehensive contextual comprehension. The dataset spans six pivotal application scenarios for handling long texts, encompassing single-document QA, multi-document QA, text summarization, few-shot learning, synthetic tasks, and code completion. The input length ranges from 1K to 22K.

    \item \textbf{LongEval} \citep{bib196}. The LongEval dataset supports the assessment of models relying on extensive text contexts. Tasks come in two different difficulty levels: one focusing on broad-topic retrieval, and the other emphasizing detailed passage retrieval. The evaluation data utilizes code synthesis to meet task specifications, providing a straightforward and rapid approach to gauge and compare models’ effectiveness with long texts. The input length ranges from 5K to 16K.

    \item \textbf{LooGLE} \citep{bib197}. The LooGLE dataset exhibits an average input length of 20K words. It is predominantly structured around two principal task categories: short dependency tasks and long dependency tasks. Short Q\&A is generated from Wikipedia articles and scripts for the former, while the latter involves the design of four tasks, including multiple information retrieval, timeline reorder, computation, and understanding with inference, specifically tailored to arXiv papers and extensive documents.

    \item \textbf{ZeroSCROLLS} \citep{bib194}. The ZeroSCROLLS dataset consists of 10 subsets of data, which are automatically transformed into a standardized input format with an average length of 10K words. Functioning as a zero-shot testing benchmark, it requires LLMs to engage in inference on diverse types of long texts across tasks like text summarization, Q\&A, aggregated sentiment classification, and information reordering.

\end{itemize}

\subsection{Tool}\label{subsecD8}

\begin{itemize}

    \item \textbf{API-Bank} \citep{bib199}. The API-Bank dataset encompasses APIs representing 53 frequently utilized tools, along with 264 dialogues subjected to manual curation, and 568 tasks involving API invocation. The tasks are stratified into three tiers: Tier 1 gauges the LLMs’ accuracy in invoking APIs, Tier 2 delves deeper into assessing their aptitude for retrieving APIs, and Tier 3 scrutinizes their competence in orchestrating multiple API invocations.

    \item \textbf{APIBench} \citep{bib200}. Derived from a collection of 1,645 API calls using the Self-Instruct technique, the APIBench dataset produces 16,450 English instructions. These instructions are versatile, serving both as guidance for fine-tuning LLMs and as a benchmark for evaluating the models’ proficiency in executing API-related instructions.

    \item \textbf{ToolBench} \citep{bib201}. The ToolBench dataset encompasses 795 instructions in English, designed to evaluate the proficiency of LLMs in manipulating various tools. The dataset is compiled from five pre-existing datasets and three recently acquired datasets. Evaluation tasks span diverse domains, including open weather, the cat API, home search, trip booking, Google Sheets, virtual home, webshop, and tabletop.

\end{itemize}

\subsection{Agent}\label{subsecD9}

\begin{itemize}

    \item \textbf{AgentBench} \citep{bib202}. The AgentBench dataset contains 1,360 English test samples and stands as the first benchmark to evaluate the performance of LLMs functioning as AI Agents across various environments. The dataset encompasses eight distinct environments, comprising five newly established domains: operating system, database, knowledge graph, digital card game, and lateral thinking puzzles, in addition to three domains adapted from public datasets: house-holding, web shopping, and web browsing.

    \item \textbf{SuperCLUE-Agent}\footnote{\href{https://github.com/CLUEbenchmark/SuperCLUE-Agent}{https://github.com/CLUEbenchmark/SuperCLUE-Agent}}. The SuperCLUE-Agent dataset addresses the gap in evaluating LLMs’ Agent capabilities in Chinese tasks and scenarios. The evaluation encompasses 10 tasks, distributed across three core abilities. Task planning capabilities include task decomposition, self-reflection, and CoT tasks; tool usage capabilities encompass API invocation, API retrieval, API planning, and general tool utilization tasks; long-term and short-term memory capabilities cover tasks such as few-shot learning, long-term dialogue, and multi-document Q\&A.

\end{itemize}

\subsection{Code}\label{subsecD10}

\begin{itemize}

    \item \textbf{APPS} \citep{bib203}. The APPS dataset contains 10K English programming questions, along with 131,777 test cases to verify solutions and 232,421 genuine solutions crafted by human authors. This dataset is primarily designed to evaluate the code generation prowess of LLMs, categorized into three difficulty levels: basic, interview, and competition. The data is collected from diverse open-access programming platforms and meticulously curated.

    \item \textbf{BIRD} \citep{bib207}. The BIRD dataset consists of 12,751 text-SQL pairs and 95 databases, spanning 37 professional domains. Its primary objective is to evaluate LLMs’ comprehension of database values and the external knowledge linkage between natural language queries and database values. The dataset is curated using techniques like web scraping and synthetic data generation.

    \item \textbf{CodeXGLUE} \citep{bib209}. The CodeXGLUE dataset is designed for the evaluation of coding abilities and consolidates existing datasets related to code. It categorizes tasks into four types based on input-output relationships, encompassing a total of nine tasks. Type 1: code-code includes tasks such as clone detection, defect detection, fill-in-the-blank tests, code completion, and code translation. Type 2: text-code involves tasks like code search and text-to-code generation. Type 3: code-text focuses on code summarization. Type 4: text-text consists of tasks like code documentation translation.

    \item \textbf{DS-1000} \citep{bib204}. The DS-1000 dataset includes 1K English coding queries associated with 7 Python libraries, designed to evaluate the proficiency in code generation. These queries are drawn from StackOverflow, ensuring a diverse and authentic representation. Moreover, each query has an automated assessment mechanism.

    \item \textbf{HumanEval} \citep{bib205}. The HumanEval dataset consists of 164 programming questions meticulously crafted by human experts. Beyond evaluating the ability to generate code, the dataset necessitates models to exhibit proficiency in language comprehension, algorithmic understanding, and mathematical knowledge. Each question encompasses functional descriptions, input-output examples, function definitions, and more, challenging the model to produce Python functions based on the given information and pass the provided test cases.

    \item \textbf{HumanEvalPack} \citep{bib210}. The HumanEvalPack dataset en-compasses 984 English-coded questions, spanning programming languages such as Python, JavaScript, Java, Go, C++, and Rust. While Python questions closely resemble those in HumanEval, questions pertaining to other programming languages are meticulously constructed by human experts. The primary assessment centers around gauging the proficiency of LLMs in rectifying code, generating code comments, and producing code.

    \item \textbf{MTPB} \citep{bib32}. The MTPB dataset comprises 115 code problems expertly crafted, each representing a multi-turn code generation task. In each problem, LLMs are tasked with synthesizing subprograms at each step, integrating the current task description with preceding steps. This multi-turn decomposed evaluation process serves to enhance the models’ understanding of user intent and its capabilities in code generation.

    \item \textbf{ODEX} \citep{bib206}. The ODEX dataset contains 945 pairs of natural language queries and associated code snippets, accompanied by 1,707 meticulously designed test cases. The task instructions for the queries span across four languages: English, Spanish, Japanese, and Russian, serving as an evaluation benchmark for the proficiency of LLMs in cross-lingual coding tasks.

\end{itemize}

\subsection{Out-of-Distribution}\label{subsecD11}

\begin{itemize}

    \item \textbf{BOSS} \citep{bib211}. The BOSS dataset is dedicated to the investigation of the OOD performance across different LLMs, presenting challenges within the framework of distributional transfer settings. Comprising 20 sub-datasets, the BOSS dataset spans tasks including sentiment analysis, toxicity detection, NLI, NER, and extractive Q\&A.

    \item \textbf{GLUE-X} \citep{bib212}. The GLUE-X dataset is comprised of 8 training datasets and 16 evaluation datasets, with all training datasets sourced from GLUE. It assesses the OOD performance of models across classic NLP tasks, encompassing syntactic judgment, sentiment analysis, semantic matching, textual entailment, and reading comprehension.

\end{itemize}

\subsection{Law}\label{subsecD12}

\begin{itemize}

    \item \textbf{LAiW} \citep{bib213}. The LAiW dataset systematically arranges pre-existing legal datasets, dividing assessment tasks into three primary legal competencies, comprising a total of 13 fundamental assignments. Fundamental tasks in NLP encompass the retrieval of statutes, recognition of elements, identification of named entities, summarization of judicial key points, and the identification of legal cases. Basic applications involve the mining of disputed focal points, matching cases, predicting criminal judgments, predicting civil judgments, and answering legal questions. Advanced applications include the generation of judicial reasoning, comprehension of case details, and the provision of legal consultation.

    \item \textbf{LawBench} \citep{bib214}. The LawBench dataset functions as an assessment benchmark designed for the Chinese legal framework. It evaluates legal capabilities through 20 tasks, such as legal entity recognition, reading comprehension, and crime amount calculation, originating from three judicial cognitive dimensions: legal knowledge retention, understanding, and application. A noteworthy strength of the dataset is its tasks being more closely aligned with real-world applications.

    \item \textbf{LegalBench} \citep{bib215}. The LegalBench dataset consists of 162 diverse legal tasks, covering six types of legal reasoning: issue-spotting, rule-recall, rule-application, rule-conclusion, interpretation, and rhetorical-understanding. Professionals from legal and computer science backgrounds collaborate in the development of the dataset.

    \item \textbf{LexGLUE} \citep{bib216}. Comprising seven open-source English legal datasets, the LexGLUE dataset involves tasks categorized into multi-label classification, multi-class classification, and multiple choice Q\&A.

    \item \textbf{LEXTREME} \citep{bib217}. Comprising 11 evaluation sub-datasets spanning 24 languages, the LEXTREME dataset categorizes all evaluation content into 18 tasks, such as Brazilian court decisions, German argument mining, Greek legal code, Swiss judgment prediction, among others.

    \item \textbf{SCALE} \citep{bib218}. The evaluation content of the SCALE dataset originates from diverse legal NLP datasets within the Swiss legal system and additional datasets, exploring research aspects related to five languages and the federal legal framework. The assessment of LLMs is conducted across four dimensions, namely handling lengthy documents of up to 50K tokens, applying legal knowledge, comprehending multiple languages, and processing multiple tasks. The multitasking component encompasses information retrieval, court view generation, decision summarization, citation extraction, and text classification.

\end{itemize}

\subsection{Medical}\label{subsecD13}

\begin{itemize}

    \item \textbf{CBLUE} \citep{bib220}. The CBLUE dataset encompasses five medical task categories and eight sub-datasets tailored for assessing Chinese medical language comprehension. These tasks involve extracting medical text information, normalizing medical terms, classifying medical texts, determining relationships between medical sentences, and answering medical questions. The dataset is sourced from authentic contexts like clinical trials, electronic health records, and textbooks, annotated by domain experts.

    \item \textbf{CMB} \citep{bib221}. The CMB dataset functions as an inclusive benchmark in the field of Chinese medicine, addressing both medical examination tasks and intricate clinical diagnosis challenges. The dataset consists of 281K questions, spanning five categories of medical exams: physician, nursing, pharmacist, disciplinary, and graduate entrance exams. The questions encompass various formats, including multiple-choice, open-ended, and multi-turn dialogues. The evaluation methodology incorporates assessments from code, experts, and models.

    \item \textbf{HuaTuo26M-test} \citep{bib110}. The testing set employed in the study \citep{bib110} is the HuaTuo26M-test dataset, randomly sampled from the Huatuo-26M dataset. The dataset comprises data collected from authentic sources like online medical encyclopedias, medical knowledge graphs, and medical Q\&A forums. It serves as a benchmark for evaluating current medical practices.

    \item \textbf{MultiMedQA} \citep{bib219}. The MultiMedQA dataset functions as an assessment benchmark for LLMs in the realm of medical Q\&A. It encompasses six publicly available medical datasets and an in-house Q\&A dataset, all expressed in the English language. The questions are structured in both multiple-choice and open-ended formats.

    \item \textbf{PromptCBLUE}\footnote{\href{https://github.com/michael-wzhu/PromptCBLUE}{https://github.com/michael-wzhu/PromptCBLUE}}. The PromptCBLUE dataset represents the first evaluation benchmark designed for LLMs in the realm of Chinese medical scenarios. This dataset integrates 16 pre-existing NLP tasks related to medical scenarios, converting them into language generation tasks based on prompts. The prompts within the dataset are structured using 94 diverse templates, demonstrating a high level of richness.

    \item \textbf{QiZhenGPT\_eval}\footnote{\href{https://github.com/CMKRG/QiZhenGPT/tree/main/data/eval}{https://github.com/CMKRG/QiZhenGPT/tree/main/data/eval}}. The QiZhenGPT\_eval dataset is utilized for evaluating LLMs’ abilities to answer questions regarding drug indications. With a dataset comprising 94 questions, the model is tasked with identifying the diseases for which a specific drug is applicable. The evaluation process involves comparing model responses with standard answers, conducted by medical experts who assign scores accordingly.

\end{itemize}

\subsection{Financial}\label{subsecD14}

\begin{itemize}

    \item \textbf{BBF-CFLEB} \citep{bib43}. The BBF-CFLEB dataset comprises six financial datasets: FinNA, FinQA, FinNL, FinRE, FinFE, and FinNSP. Each dataset is designed for specific financial tasks, including financial news summarization, event-based Q\&A, news classification, news relation extraction, sentiment analysis, and identification of negative news and subjects. The primary focus is on assessing language understanding and language generation proficiency.

    \item \textbf{FinancelQ}\footnote{\href{https://github.com/Duxiaoman-DI/XuanYuan/tree/main/FinanceIQ}{https://github.com/Duxiaoman-DI/XuanYuan/tree/main/FinanceIQ}}. The FinancelQ dataset focuses on the Chinese financial domain, specifically evaluating financial knowledge and reasoning abilities. It covers 10 major financial categories and 36 subcategories, comprising a total of 7173 multiple-choice questions. The dataset undergoes steps such as question selection and rewriting during construction to mitigate the potential impact of data leakage.

    \item \textbf{FinBen} \citep{bib413}. The FinBen dataset provides a thorough and comprehensive assessment of LLMs’ financial capabilities. It integrates 35 existing datasets covering 23 financial tasks. The creators have categorized the tasks into three difficulty levels: foundamental tasks, advanced cognitive engagement, and general intelligence. FinBen extends financial evaluation to a wide range of tasks including quantification, understanding, forecasting, and introduces the direct trading task for the first time.

    \item \textbf{FinEval} \citep{bib223}. The FinEval dataset serves as a benchmark for evaluating Chinese financial knowledge, encompassing 4,661 high-quality multiple-choice questions. The dataset spans four domains: finance, economy, accounting, and certificates, covering a total of 34 distinct academic subjects. The majority of the data is sourced from simulated exams and practice questions available through public channels, while some certificate-related questions are derived from printed papers.

    \item \textbf{FLUE} \citep{bib224}. The FLUE dataset aggregates six English NLP datasets pertaining to finance, establishing a standardized benchmark for financial evaluation. It primarily evaluates the proficiency in NLU, covering tasks like financial sentiment analysis, news headline classification, NER, structure boundary detection, and Q\&A.

\end{itemize}

\subsection{Social Norms}\label{subsecD15}

\begin{itemize}

    \item \textbf{CrowS-Pairs} \citep{bib225}. The utilization of the CrowS-Pairs dataset aims to assess the social biases exhibited by LLMs towards marginalized communities in the United States. Through a crowdsourcing approach, a benchmark of contrasting stereotypes is established. The dataset consists of 1,508 instances, each presenting sentences with varying degrees of stereotypical content. In its entirety, the material covers nine categories of stereotypes, encompassing race, gender, sexual orientation, religion, age, nationality, disability, physical appearance, and occupation.

    \item \textbf{SafetyBench} \citep{bib226}. The SafetyBench dataset encompasses 11,435 dual-language multiple-choice questions, delivering a thorough assessment of the safety aspects of LLMs. Evaluation is conducted across seven distinct safety dimensions, encompassing offensiveness, unfairness and bias, physical health, mental health, illegal activities, ethics and morality, as well as privacy and property.

    \item \textbf{Safety-Prompts} \citep{bib227}. The Safety-Prompts dataset contains 100K Chinese prompts depicting safety scenarios and paired responses from ChatGPT. It serves as a resource for assessing the safety performance of large-scale models and aligning with human safety values. The questions are categorized into typical safety scenarios and instruction attacks. Typical safety scenarios involve insult, unfairness and discrimination, crimes and illegal activities, physical harm, mental health, privacy and property, and ethics and morality. Instruction attacks include goal hijacking, prompt leaking, role play instruction, unsafe instruction topic, inquiry with unsafe opinion, and reverse exposure.

    \item \textbf{SuperCLUE-Safety}\footnote{\href{https://github.com/CLUEbenchmark/SuperCLUE-safety}{https://github.com/CLUEbenchmark/SuperCLUE-safety}}. The SuperCLUE-Safety dataset functions as a safety benchmark tailored for large-scale Chinese models engaged in multi-turn adversarial conversations. Comprising 2,456 test instances, each featuring a safety question and a subsequent inquiry, the dataset integrates adversarial techniques to heighten the complexity of the questions. It effectively simulates real-world user scenarios through multi-turn dialogues. This benchmark serves as an evaluation for three key capabilities: conventional safety, responsible AI, and instruction attacks.

    \item \textbf{TRUSTGPT} \citep{bib228}. Assessing toxicity, bias, and value alignment involves the evaluation of the TRUSTGPT dataset using 2K English test samples. The dataset undergoes scrutiny for toxicity using prompts aligned with social norms. Following this, bias levels of LLMs are quantified by measuring toxicity values across different groups. Ultimately, value alignment is appraised by gauging LLMs’ rejection of content that contradicts human values.

\end{itemize}

\subsection{Factuality}\label{subsecD16}

\begin{itemize}

    \item \textbf{FACTOR} \citep{bib230}. The FACTOR dataset is divided into Wiki-FACTOR and News-FACTOR, distinguished by their respective data sources. The dataset consists of 4,030 English samples, with 2,994 in Wiki-FACTOR and 1,036 in News-FACTOR. For each sample, LLMs are tasked with selecting the singular option that is factually accurate from among four completions, guided by provided prefixes and relevant knowledge. This evaluation seeks to gauge the proficiency of LLMs in factual accuracy.

    \item \textbf{FActScore} \citep{bib232}. The evaluation of LLMs’ factual accuracy in generating extensive content is conducted using the FActScore dataset. Consisting of 500 English evaluation samples, the dataset draws from biographical information found on Wikipedia. A novel approach is employed, dissecting the generated text into elemental facts and computing scores for the factual components endorsed by the knowledge source.

    \item \textbf{FactualityPrompt} \citep{bib233}. The FactualityPrompt dataset evaluates the factual accuracy of textual outputs generated by LLMs. With a dataset size of 16K English samples, evenly split into 8K authentic prompts and 8K fabricated prompts, it is an adaptation derived from the FEVER dataset. The study delves into the influence of two types of prompts on the factual accuracy of LLMs, replicating real-world situations with input inaccuracies.

    \item \textbf{FreshQA} \citep{bib236}. The FreshQA dataset functions as a dynamic QA benchmark, encompassing 600 English evaluation samples. Questions presented to LLMs are classified into four categories based on the characteristics of the answers: answers that remain constant, answers that undergo gradual changes, answers that undergo rapid changes, and answers with incorrect premises. This evaluation scrutinizes whether LLMs manifest hallucinatory phenomena in responding to questions and their ability to refute erroneous factual assumptions without succumbing to misdirection.

    \item \textbf{HalluQA} \citep{bib235}. The HalluQA dataset encompasses 450 Chinese questions specifically crafted to test the hallucinatory behaviors of Chinese LLMs. These questions are classified into three types: misleading questions, highly misleading questions, and long-tail knowledge questions. During the assessment, GPT-4 is utilized to discern whether the models’ responses demonstrate hallucination.

    \item \textbf{HaluEval} \citep{bib231}. The HaluEval dataset functions as a benchmark for assessing hallucination in LLMs, containing 35K English evaluation samples. The evaluation data is crafted through a combination of manual and automated methods using existing datasets. The assessment data involves inputs and outputs in scenarios like Q\&A, dialogue, text summarization, and human-computer interaction, challenging LLMs to identify the potential existence of hallucination.

    \item \textbf{TruthfulQA} \citep{bib234}. The TruthfulQA dataset is a standard for evaluating the authenticity of answers produced by LLMs, featuring 817 English questions across 38 domains. These questions are carefully designed and selected through manual curation.

\end{itemize}

\subsection{Evaluation}\label{subsecD17}

\begin{itemize}

    \item \textbf{FairEval} \citep{bib237}. All 80 instructions in the FairEval dataset originate from the Vicuna Evaluation dataset. The creators generate responses across various models, including ChatGPT, Vicuna-13B, Alpaca-13B. Subsequently, models such as ChatGPT, GPT-4, are employed to assess different responses. The evaluation underscores the importance of exercising caution when employing LLMs as evaluators, given the outcomes obtained.

    \item \textbf{$\mathbf{LLMEval}^2$} \citep{bib239}. The $\mathrm{LLMEval}^2$ dataset is presently the most extensive and diverse English benchmark for appraising the evaluation capabilities of LLMs, comprising a total of 2,553 samples. It incorporates 15 task types, spanning 8 abilities, designed for utilization by LLMs assessors. \cite{bib239} delves into the inquiry of whether a more profound and expansive network contributes to a fairer assessment.

    \item \textbf{PandaLM\_testset} \citep{bib238}. The PandaLM\_testset dataset contains 1K varied English samples, with human annotations for both context and labels. Its purpose is to evaluate the PandaLM model’s proficiency in comparing responses among different LLMs, demonstrating the dependability of PandaLM. The dataset also enables an investigation into the assessment capabilities of alternative LLMs.

\end{itemize}

\subsection{Multitask}\label{subsecD18}

\begin{itemize}

    \item \textbf{BBH} \citep{bib243}. The BBH dataset encompasses 23 tasks, consisting of 6,511 English test samples. These tasks are drawn from BIG-Bench and represent types where LLMs exhibit performance below the average human level. Included in these tasks are causal judgment, date understanding, disambiguation QA, among others.

    \item \textbf{BIG-Bench} \citep{bib242}. The BIG-Bench dataset stands as a comprehensive, intricate, and varied benchmark, honing in on tasks that pose challenges for contemporary language models. It serves as an evaluation platform for the behaviors exhibited by classical models. With a total of 95 task types encompassing 204 tasks, the dataset spans a broad spectrum of topics. Furthermore, there exists a task subset known as “BIG-Bench Lite,” featuring tasks that are representative, compact, and facilitate swifter evaluations.

    \item \textbf{CLEVA} \citep{bib245}. The CLEVA dataset functions as a benchmark for multi-task evaluation in Chinese, consisting of 370K test samples. A notable portion, approximately 33.98\%, is newly generated, addressing concerns associated with data leakage. The dataset covers 11 tasks for application evaluation and 20 tasks for capability assessment, with consistent preprocessing and standardized Chinese prompt templates applied to all data.

    \item \textbf{CLiB}\footnote{\href{https://github.com/jeinlee1991/chinese-llm-benchmark}{https://github.com/jeinlee1991/chinese-llm-benchmark}}. The CLiB dataset serves as an evaluation benchmark for assessing the Chinese language capabilities of LLMs. It conducts evaluations on 48 LLMs, whether commercial or open-source, across various dimensions, including classification, information extraction, reading comprehension, and table-based Q\&A. The dataset consistently releases new evaluation rankings and offers the original output results of the models.

    \item \textbf{decaNLP} \citep{bib240}. The decaNLP dataset spans ten distinct tasks, covering areas such as Q\&A, translation, text summarization, NLI, sentiment analysis, RE, semantic role labeling, goal-oriented dialogue, semantic parsing, and commonsense reasoning. Characterized by a substantial dataset, it evaluates the English task processing proficiency of LLMs.

    \item \textbf{FlagEval}\footnote{\href{https://github.com/FlagOpen/FlagEval}{https://github.com/FlagOpen/FlagEval}}. The FlagEval dataset currently encompasses 22 evaluation sets, featuring a collective of 84,433 questions. It introduces a nuanced evaluation framework based on the dimensions of “capability-task-metric,” offering detailed insights into the cognitive boundaries of models. This assessment explores over 600 sub-dimensions across 30+ capabilities, 5 primary tasks, and 4 key metrics.

    \item \textbf{HELM} \citep{bib244}. The HELM dataset has the objective of constructing a holistic evaluation framework to comprehensively appraise the capabilities of LLMs. Adopting a top-down classification methodology for domain coverage, it precisely delineates evaluation scenarios and metrics, facilitating a systematic selection process. Presently, there are a total of 73 evaluation scenarios. In terms of metric evaluation, the dataset emphasizes the reflection of diverse dimensions of capabilities, achieving metric diversification, with a current tally of 65 evaluation metrics.

    \item \textbf{LLMEVAL-1} \citep{bib248}. The LLMEVAL-1 dataset encompasses 453 questions in Chinese, covering 17 broad task categories, including tasks like providing factual answers, generating frameworks, and creating poetry. Simultaneously, five assessment criteria are defined, covering accuracy, fluency, informativeness, logic, and harmlessness. Evaluation is conducted through methods such as crowdsourced comparative assessment, public comparative assessment, and GPT-4 scoring.

    \item \textbf{LMentry} \citep{bib241}. The LMentry dataset evaluates how LLMs perform on tasks that humans consider simple. In contrast to benchmarks for more intricate tasks, LMentry provides rapid and interpretable insights into the fundamental capabilities and robustness of LLMs. The dataset consists of around 110K English samples, encompassing 25 task categories, including word selection and sentence composition.

\end{itemize}

\subsection{Multilingual}\label{subsecD19}

\begin{itemize}

    \item \textbf{XNLI} \citep{bib249}. The XNLI dataset evaluates the transfer of low-resource languages and cross-lingual sentence classification, featuring a total of 15 languages, including English, French, Spanish, German, Bulgarian, Russian, Turkish, Arabic, Vietnamese, Thai, Chinese, Hindi, Swahili, and Urdu. There are 7.5K evaluation samples for each language, with data for non-English languages derived from translation.

    \item \textbf{XTREME} \citep{bib250}. The XTREME dataset assesses LLMs through four NLP tasks conducted in a variety of languages, scrutinizing the linguistic competence of LLMs. Task categories encompass classification, structured prediction, Q\&A, and retrieval. The dataset encompasses 40 languages, representing 20 language families.

\end{itemize}

\subsection{Other}\label{subsecD20}

See Section~\ref{subsubsec5120} for details.

\section{Traditional NLP Dataset Information}\label{secE}

Appendix~\ref{secE} provides detailed information on each traditional NLP dataset mentioned in the main text.

\subsection{Question Answering}\label{subsecE1}

\subsubsection{Reading Comprehension}\label{subsubsecE11}

(1) Selection \& Judgment

\begin{itemize}

    \item \textbf{BoolQ} \citep{bib260}. The BoolQ dataset is crafted in an environment devoid of prompts and constraints, yielding 15,942 yes/no queries that delve into intricate inquiries and non-factual details, serving as a litmus test for the models’ reading comprehension and inferential prowess. Each instance comprises a question, a paragraph, and an answer, necessitating the model to provide a response using either yes or no.
    
    \item \textbf{CondaQA} \citep{bib262}. The CondaQA dataset represents a pioneering effort in English reading comprehension datasets dedicated to inferencing the implications of negated statements in textual content. Annotators, responding to text with negations, formulate queries assessing meaning comprehension. The text undergoes three types of modifications, involving rephrasing, changing the scope of negation, and inverting negations. Responses to queries are provided in three formats: “Yes,” “No,” and “Don't Know.”
    
    \item \textbf{CosmosQA} \citep{bib261}. The CosmosQA dataset necessitates models to undertake reading comprehension tasks by leveraging common sense, structured in a multiple-choice format. Utilizing everyday stories as textual input, it presents inquiries about the origins and repercussions of events. Models are expected not merely to provide surface-level answers but also to grasp the implicit common knowledge and logical connections embedded in the text.

    \item \textbf{$\mathbf{C}^3$} \citep{bib266}. The $\mathrm{C}^3$ dataset serves as an evaluation measure for the Chinese reading comprehension capabilities of models, encompassing 13,369 dialogues or texts of mixed genres and 19,577 multiple-choice questions. These inquiries are drawn from Chinese language exams intricately designed by educational specialists, resembling the structure of questions in RACE and DREAM. $\mathrm{C}^3$ has been included in CLUE, establishing itself as an assessment benchmark for Chinese NLU tasks.

    \item \textbf{DREAM} \citep{bib268}. The DREAM dataset comprises a dialogue-based multiple-choice reading comprehension exam with 10,197 questions and 6,444 dialogues. The dialogues are collected from English exams designed by human experts. 84\% of the answers are non-extractive, 85\% of the questions require reasoning across multiple sentences, and 34\% of the questions involve common-sense knowledge.

    \item \textbf{DuReader Yes/No}\footnote{\href{https://github.com/baidu/DuReader}{https://github.com/baidu/DuReader}}. Given the challenges in using metrics like F1 to gauge a model’s genuine comprehension of textual meaning in opinion-based questions, this dataset employs opinion polarity judgment as the focus of reading comprehension. The task necessitates the model to discern the polarity of answers from the provided question, text, and answer summary, encompassing positive, negative, and indeterminate polarities.

    \item \textbf{MCTest} \citep{bib270}. The MCTest dataset necessitates models to respond to multiple-choice questions related to imaginary narratives. Given the entirely fictional nature of the text, there is a scarcity of included world knowledge. The primary focus lies on evaluating the models’ proficiency in understanding story content and extracting relevant answers.

    \item \textbf{MultiRC} \citep{bib264}. The MultiRC dataset mandates that models incorporate information from several sentences in the text to address questions involving the selection of accurate options. The number of correct answer options varies for each question, thus requiring the model to evaluate the accuracy of each option. Furthermore, the dataset is sourced from diverse materials such as news articles, novels, historical texts, and seven other domains.
    
    \item \textbf{PubMedQA} \citep{bib263}. The PubMedQA dataset serves as a reading comprehension resource specifically designed for biomedical questions. It derives its content from abstracts within the PubMed Central. The assigned task requires models to respond to questions based on the article abstracts, with potential answers categorized as “Yes,” “No,” or “Maybe.” The dataset encompasses 1K meticulously annotated samples, along with an additional 61.2K unlabeled samples and 211.3K synthetically generated samples.

    \item \textbf{QuAIL} \citep{bib269}. The QuAIL dataset combines question types based on text, world knowledge, and unanswerable scenarios, totaling 15K multiple-choice questions spanning four domains. Notably, the dataset includes annotations for nine reasoning types, encompassing aspects such as time, causality, factual information, coreference, role attributes, belief states, entity states, event duration, and questions deemed unanswerable.

    \item \textbf{RACE} \citep{bib265}. The RACE dataset serves as a resource for evaluating proficiency in English reading comprehension, encompassing more than 28K articles and close to 100K inquiries. Derived from reading comprehension questions within Chinese English exams, all questions are structured in a multiple-choice format. The dataset is stratified by complexity, offering the “RACE-M” subset for middle school students and the “RACE-H” subset for high school students.

    \item \textbf{ReClor} \citep{bib267}. The ReClor dataset originates from standardized graduate entrance examinations, aiming to heighten the complexity of reading comprehension and introduce novel challenges to the logical reasoning capabilities of models. In order to mitigate the risk of models achieving elevated performance without a true comprehension of the text through the exploitation of inherent biases in the data, the dataset bias has been partitioned by the creators into easy and hard subsets.

\end{itemize}

\noindent (2) Cloze Test

\begin{itemize}

    \item \textbf{ChID} \citep{bib271}. The ChID dataset serves as a platform for evaluating models in the context of Chinese idiomatic expression reading comprehension. In this task, models are tasked with filling in the blanks by choosing the appropriate idiom based on the provided context. The dataset specifically targets the models’ comprehension of Chinese idioms and has been incorporated into the CLUE benchmark for assessing Chinese NLU capabilities.

    \item \textbf{CLOTH} \citep{bib273}. The CLOTH dataset stands as the pioneer in cloze-type reading comprehension datasets crafted manually. Sourced from English exam questions for Chinese middle and high school levels, the missing words and candidate options are meticulously curated by subject experts. The objective is for models to comprehensively grasp the meaning of the entire text and choose fitting English words to fill the gaps.

    \item \textbf{CMRC2019} \citep{bib274}. The CMRC2019 dataset stands as a sentence-level cloze-style reading comprehension benchmark. The objective is for models to intelligently insert sentences from a set of candidates into the blanks within a given article (featuring multiple blanks), ensuring the coherence and completeness of the text. This task critically assesses the models’ capacities for discerning logical relationships in context.

    \item \textbf{LAMBADA} \citep{bib272}. The LAMBADA dataset serves to evaluate model reading comprehension abilities by employing a word prediction task. Extracted from books, it includes 10K passages and over 100K English sentences. Each sentence concludes with a blank space, challenging the model to predict the missing word based on a comprehensive understanding of the context, thereby assessing its contextual awareness.

\end{itemize}

\noindent (3) Answer Extraction

\begin{itemize}

    \item \textbf{Adversarial QA} \citep{bib284}. \cite{bib284} delves into the exploration of model-driven cyclic adversarial annotations, leveraging SQuAD as its underpinning. Employing the paradigm of adversarial artificial annotations, queries are systematically generated until they render the adversarial model incapable of delivering correct responses. Consequently, this methodology is harnessed to formulate the Adversarial QA dataset characterized by its inherent challenges.

    \item \textbf{CMRC2018} \citep{bib283}. The CMRC2018 dataset is composed of approximately 20K Chinese reading comprehension questions, each representing genuine queries annotated by human experts on Wikipedia. Additionally, a challenging subset is presented, necessitating extensive comprehension and multi-sentence reasoning within context for model-derived answers. This dataset has been integrated into the CLUE dataset, serving as an assessment benchmark for Chinese NLU tasks.

    \item \textbf{CUAD} \citep{bib288}. The CUAD dataset concentrates on the realm of understanding legal contracts, encompassing 510 legal agreements and 41 distinct categories of crucial clauses. The task mandates models to comprehend the textual content of contracts and extract answers to queries pertaining to the contracts.

    \item \textbf{DuReader Checklist}\footnote{\href{https://github.com/baidu/DuReader}{https://github.com/baidu/DuReader}}. The DuReader Checklist dataset utilizes extractive reading comprehension queries and institute a comprehensive Checklist evaluation framework to methodically appraise models’ multidimensional and nuanced proficiency in reading comprehension. The evaluative aspects include lexical understanding, phrase comprehension, semantic role comprehension, and reasoning capabilities, among other dimensions.

    \item \textbf{DuReader Robust} \citep{bib287}. The DuReader Robust dataset represents the pioneering Chinese robust reading comprehension dataset, crafted to gauge the robustness of models by employing data instances from authentic real-world scenarios. Its objective is to appraise the models for their sensitivity, excessive stability, and generalization.

    \item \textbf{HOTPOTQA} \citep{bib277}. The HOTPOTQA dataset incorporates text sourced from Wikipedia, tasking models with deducing answers to questions from diverse document contents. Characterized by multi-document reasoning, absence of predefined knowledge base constraints, and provision of sentence-level supporting facts, it facilitates the exploration of multi-step reasoning involving information from multiple sources.

    \item \textbf{MLQA} \citep{bib286}. The MLQA dataset serves as a benchmark for assessing the multilingual Q\&A proficiency of models through the utilization of extractive reading comprehension prompts. Instances within this dataset span across seven languages, encompassing English, Arabic, German, Spanish, Hindi, Vietnamese, and Simplified Chinese.

    \item \textbf{MS MARCO} \citep{bib289}. The queries within the MS MARCO dataset originate from the Bing search engine. Each query is paired with manually crafted responses, and web documents retrieved from Bing searches serve as contextual information. The creators have consequently proposed three tasks with different levels of difficulty: “assessing answerability,” “generating answers,” and “ranking retrieval content.” Due to the dataset’s content being drawn from genuine user search histories, it possesses substantial scale, practical relevance, and thus, considerable reference merit.

    \item \textbf{Natural Questions} \citep{bib279}. The Natural Questions dataset mandates models to peruse and grasp complete Wikipedia articles, discerning if the articles encompass answers to posed questions. In the affirmative, the model must articulate the precise details of the response. The questions are derived from authentic user inquiries, enhancing the dataset's realism and complexity.

    \item \textbf{QuAC} \citep{bib281}. The QuAC dataset comprises 14K dialogue pairs and 100K questions designed for conversational reading comprehension. Annotators engage in a two-person dialogue, where one formulates a set of open-ended questions to unveil concealed information from Wikipedia text, and the other extracts concise excerpts from the text to respond to these questions. Notably, the questions in this dataset exhibit a greater degree of openness, with some questions finding significance only within the contextual framework of the dialogue.

    \item \textbf{Quoref} \citep{bib285}. The Quoref dataset serves as an assessment tool for models’ proficiency in co-reference reasoning within the domain of reading comprehension. Models, in order to address posed queries, are mandated to dissect intricate co-reference relationships embedded in the supplied textual content. The dataset encompasses a plethora of more than 47K paragraphs sourced from Wikipedia.

    \item \textbf{ReCoRD} \citep{bib280}. Included in SuperGLUE, the ReCoRD dataset serves as an assessment benchmark for English NLU tasks. The objective is for models to extract answers from provided news text given a set of questions. This task places a notable emphasis on evaluating the models’ capacities for common-sense reasoning during the comprehension process.
    
    \item \textbf{SQuAD} \citep{bib275}. The SQuAD dataset is constructed with over 100K samples through crowdsourcing. Annotators generate questions based on Wikipedia articles, and the answers are derived from corresponding passages in the text.

    \item \textbf{SQuAD 2.0} \citep{bib276}. Built upon the SQuAD dataset, SQuAD 2.0 introduces an additional 53,775 unanswerable questions crafted through crowdsourced reverse engineering. Responding to these questions necessitates models to decline providing an answer as the information cannot be located in the given text.

    \item \textbf{TriviaQA} \citep{bib278}. The TriviaQA dataset covers 95K Q\&A pairs, with an average of six associated evidence documents per question, constituting over 650K question-answer-evidence triplets. The questions are relatively intricate, demanding cross-sentence reasoning for answer identification, providing a closer representation of real-world scenarios.

    \item \textbf{TyDiQA} \citep{bib282}. TyDiQA, a Q\&A dataset, encompasses 11 distinct languages and consists of 204K Q\&A pairs, deliberately addressing language intricacies absent in conventional English-centric datasets. The dataset’s questions are authored by individuals genuinely seeking answers to inquiries they lack knowledge of. Answers are extracted directly from Wikipedia texts in the corresponding languages, eschewing the use of translation tools. Models are assigned the tasks of paragraph selection and determining minimal answer spans based on the given text and questions.

\end{itemize}

\noindent (4) Unrestricted QA

\begin{itemize}

    \item \textbf{CoQA} \citep{bib291}. The CoQA dataset encompasses in excess of 8K dialogues and over 127K Q\&A pairs, serving as a metric for assessing models’ adeptness in understanding text and responding to interconnected queries. Each dialogue originates from a conversation between two annotators, derived from the provided sets of questions and answers. A distinguishing feature of CoQA lies in the fact that responses can manifest as free-form textual expressions, with the pertinent context for the answers embedded within the text.
    
    \item \textbf{DROP} \citep{bib290}. The objective of the DROP dataset is to evaluate the pluralistic reasoning capabilities of models when dealing with textual information. Generated through crowdsourcing, it encompasses 96K interrogations. Models are compelled to explore diverse avenues for unraveling questions, sometimes engaging in computations, sorting, and other operations grounded in the textual data to derive answers. The task mandates models to cultivate a more profound comprehension of the text, given that answers might not be readily apparent within the provided textual context.

    \item \textbf{DuoRC} \citep{bib293}. The DuoRC dataset consists of 186,089 questions derived from 7,680 pairs of movie plots. Each plot pair includes two distinct portrayals of the same movie—one extracted from Wikipedia and the other from the IMDB website. Annotators generate questions based on one portrayal, and answers are then constructed using the alternate portrayal. As a result, certain questions do not share vocabulary with the provided text, requiring models to autonomously formulate language for responses.

    \item \textbf{DuReader 2.0} \citep{bib294}. The DuReader 2.0 dataset constitutes an expansive, authentic, and manually curated collection of Chinese reading comprehension data. Focused on open-domain Q\&A, this dataset comprises 200K questions, 420K answers, and 1M documents, all derived from real-world scenarios and extensively annotated. Models are tasked with deriving answers through summarization from several documents.

    \item \textbf{QASPER} \citep{bib292}. The QASPER dataset covers 1,585 NLP papers and 5,049 related questions, designed to facilitate understanding and reasoning across diverse sections of research papers. Each question is formulated by NLP professionals after perusing only the paper’s title and abstract. Following this, a separate group of practitioners responds to the questions and furnishes supporting evidence for their responses. Extracting answers directly from the text is not viable; instead, a degree of summarization and synthesis is necessary.

\end{itemize}

\subsubsection{Knowledge QA}\label{subsubsecE12}

\begin{itemize}

    \item \textbf{ARC} \citep{bib295}. The ARC dataset consists of 7,787 real elementary-level science knowledge questions, classified into a challenging subset (2,590 questions) and an easy subset (5,197 questions) based on question difficulty. The task mandates models to choose the optimal option through scientific knowledge and reasoning.

    \item \textbf{CMD}\footnote{\href{https://github.com/Toyhom/Chinese-medical-dialogue-data}{https://github.com/Toyhom/Chinese-medical-dialogue-data}}. The CMD dataset represents a Chinese medical Q\&A dataset aimed at evaluating the knowledge Q\&A capabilities of models within the medical domain. The dataset encompasses a total of 792,099 Q\&A pairs, classified into six sub-domains: andrology, internal medicine, obstetrics and gynecology, oncology, pediatrics, and surgery.

    \item \textbf{cMedQA2} \citep{bib300}. The cMedQA2 dataset represents an expanded and enhanced version of the cMedQA dataset. The initiators gather authentic doctor-patient dialogues from an online Chinese medical Q\&A forum as inquiries, wherein medical professionals respond to medical queries posed by patients. The primary focus is to assess the models’ abilities to answer questions within real scenarios where patients seek medical information.
    
    \item \textbf{CommonsenseQA} \citep{bib296}. The CommonsenseQA dataset consists of 12,102 multiple-choice questions demanding diverse forms of common-sense knowledge for accurate answer selection. Extracting various target concepts with semantic relations akin to the source concepts from CONCEPTNET, creators task annotators with crafting multiple-choice questions that discriminate between different target concepts. The objective of the task is to evaluate the models’ proficiency in common-sense knowledge.

    \item \textbf{ECQA} \citep{bib304}. The ECQA dataset, an abbreviation for Explanation CommonsenseQA, originates from CommonsenseQA. Following manual annotation, it encompasses positive and negative attributes, along with English explanations, for 11K QA pairs extracted from CommonsenseQA. Its objective is to furnish explanations for the knowledge-based question-answering task within CommonsenseQA, providing an in-depth comprehension of the general attributes linked to various options.

    \item \textbf{HEAD-QA} \citep{bib301}. The HEAD-QA dataset encompasses a variety of multiple-choice questions and answers, spanning disciplines such as medicine, pharmacology, psychology, nursing, biology, and chemistry. The questions are sourced from professional position exams within the Spanish healthcare system, adding a level of complexity. The dataset is available in both English and Spanish versions, covering a range of technical and societal knowledge.

    \item \textbf{JEC-QA} \citep{bib299}. The JEC-QA dataset consists of 26,365 multiple-choice questions, sourced exclusively from the Chinese National Judicial Examination. The primary objective is to evaluate the knowledge Q\&A capabilities of models within the legal domain. Questions can be classified into two categories: knowledge-driven, emphasizing legal concepts, and case analysis, necessitating an analysis of practical legal scenarios.
    
    \item \textbf{OpenBookQA} \citep{bib297}. The OpenBookQA dataset replicates the structure of open-book exams aimed at evaluating human comprehension across diverse subjects. Each sample comprises a question, four options along with their respective answers, and supplementary scientific facts and common-sense information. Models need to exhibit proficiency in multi-step reasoning, application of common-sense knowledge, and comprehension of textual content.

    \item \textbf{PIQA} \citep{bib298}. The PIQA dataset centers on the physics interaction Q\&A task, evaluating the models’ capacities to effectively respond to questions pertaining to physics common sense. The task mandates the model to apply physics common sense in selecting the most plausible solution from two presented alternatives based on a provided real-world scenario.

    \item \textbf{PsyQA} \citep{bib305}. The PsyQA dataset presents a collection of Chinese mental health data in a Q\&A format. Derived from a Chinese platform for mental health services, it encompasses 22K questions and 56K answers. The dataset’s knowledge-based Q\&A relies on psychological counseling theory, evaluating the models’ abilities to produce text related to mental health counseling. This assessment aims to improve the smoothness and utility of the generated answers.

    \item \textbf{SciQ} \citep{bib302}. The SciQ dataset consists of 13,679 science examination questions acquired through crowdsourcing, spanning disciplines like physics, chemistry, and biology. These questions are structured in a multiple-choice format, offering four answer options. Additional paragraphs and materials supporting the correct answers are included for the majority of questions.

    \item \textbf{WebMedQA} \citep{bib306}. The WebMedQA dataset represents a Chinese medical Q\&A dataset, akin to cMedQA2. Each instance is sourced from specialized health advisory websites, comprising questions, answers, adoption status, and categorized labels. Specifically, there are 23 distinct categories, encompassing a broad range of prevalent clinical departments, with internal medicine and surgery having the highest representation.
    
    \item \textbf{WikiQA} \citep{bib303}. The WikiQA dataset explores models for open-domain Q\&A. The origin of questions is sourced from Bing query logs, and answers are derived from the content available on Wikipedia. A collective total of 3,047 questions has been gathered through crowdsourcing.

\end{itemize}

\subsubsection{Reasoning QA}\label{subsubsecE13}

\begin{itemize}

    \item \textbf{COPA} \citep{bib308}. The COPA dataset is explicitly crafted for the common-sense causal reasoning task. Models are tasked with choosing the correct causal relationship based on provided premises. Incorporated into SuperGLUE, COPA serves as an assessment benchmark for English NLU tasks.

    \item \textbf{CREAK} \citep{bib319}. For the exploration of models’ abilities to amalgamate entity knowledge with common-sense reasoning, the CREAK dataset is introduced. It establishes a connection between factual details about entities (e.g., wizards like Harry Potter, proficient in broomstick flying) and common-sense reasoning principles (e.g., having expertise in a skill allows one to instruct others). This process results in the formulation of reasoning queries (e.g., is Harry Potter capable of instructing broomstick flying).

    \item \textbf{HellaSwag} \citep{bib309}. The HellaSwag dataset is curated for evaluating common-sense natural language reasoning. Each query includes a scenario and four conceivable outcomes, tasking models with deducing the most reasonable conclusion. Human-validated incorrect responses aim to mislead the model.

    \item \textbf{LogiQA} \citep{bib312}. For comprehensive exploration of logical reasoning, \cite{bib312} has engaged human experts to develop the LogiQA dataset, aimed at evaluating questions pertaining to human logical reasoning. It encompasses more than 8K Q\&A pairs, covering diverse types of deductive reasoning, including categorical reasoning, sufficient conditional reasoning, necessary conditional reasoning, disjunctive reasoning, and conjunctive reasoning.

    \item \textbf{PROST} \citep{bib313}. The PROST dataset, officially known as Physical Reasoning about Objects Through Space and Time, serves as a test for assessing physical reasoning capabilities. It consists of 18,736 multiple-choice questions created through 14 manually designed templates. The questions cover 10 concepts related to physical reasoning, encompassing direction, mass, height, circumference, stackable, rollable, graspable, breakable, slideable, and bounceable.

    \item \textbf{QASC} \citep{bib316}. The QASC dataset evaluates the multi-hop reasoning abilities of models. It involves retrieving pertinent facts from an extensive corpus and employing effective multi-hop reasoning methods to integrate these facts. Ultimately, the correct answer is selected from a pool of eight options.
    
    \item \textbf{QuaRel} \citep{bib317}. The QuaRel dataset is developed with the aim of fostering models’ comprehension and resolution of problems related to qualitative relationship inference. The dataset encompasses 2,771 narrative-based multiple-choice questions, exemplified by “Jenny notices a discrepancy in the speed of the robotic vacuum cleaner between the living room and bedroom carpets. Which carpet exhibits greater friction?”. The logical form of the questions is also provided.

    \item \textbf{QuaRTz} \citep{bib314}. The QuaRTz dataset presents a novel task involving the qualitative analysis of textual relationships, where common qualitative statements are paired with contextually generated questions through crowdsourcing. For example, the qualitative statement “Sunscreen with a higher SPF protects the skin for a longer time” is paired with the contextual question “Billy applies sunscreen with an SPF lower than Lucy’s. Who will receive better sun protection?”. Models must exhibit robust abilities in both reasoning transfer and analogical reasoning to effectively address these inquiries.

    \item \textbf{ROPES} \citep{bib318}. The ROPES dataset is primarily designed to evaluate the reasoning abilities of models within specific contexts. Models are presented with background articles containing pertinent knowledge, newly constructed scenarios, and questions. Its task is to employ background knowledge for reasoning through the questions within the provided context. These background articles are derived from scientific textbooks and Wikipedia, with scenarios, questions, and answers curated by annotators.

    \item \textbf{Social IQa} \citep{bib311}. The Social IQa dataset functions as benchmarks for commonsense reasoning within social contexts, incorporating questions that revolve around social interactions. The task necessitates models to choose the most reasonable option from three potential subsequent behaviors, all within a provided scenario. This introduces content pertaining to the reasoning of temporal relationships while evaluating fundamental common knowledge.

    \item \textbf{StoryCloze} \citep{bib310}. The StoryCloze dataset is devised to assess the causal reasoning capabilities of models within the realms of story comprehension, story generation, and script learning. Analogous to HellaSwag, the objective is for models to choose an accurate conclusion from four sentences portraying a story scenario. The dataset encapsulates intricate causal and temporal contextual associations prevalent in everyday occurrences.
    
    \item \textbf{STRATEGYQA} \citep{bib307}. The STRATEGYQA dataset acts as a benchmark for reasoning-based Q\&A. The necessary steps for models to respond are implicitly stated within the questions, and inference is carried out through the application of diverse strategies. It encompasses 2,790 samples, each consisting of a question focused on strategy, a breakdown of steps, and a paragraph providing evidence.
    
    \item \textbf{WIQA} \citep{bib315}. In particular, WIQA stands out as the inaugural dataset tailored for “What if...” queries pertaining to procedural reasoning. Models are tasked with deducing the repercussions of a disturbance occurring in a described process, utilizing knowledge embedded in the textual depiction of the process. For example, when presented with text detailing beach erosion, the objective is to predict the effects of a stormy weather event on the erosion level.

\end{itemize}

\subsection{Recognizing Textual Entailment}\label{subsecE2}

\begin{itemize}

    \item \textbf{ANLI} \citep{bib320}. The Adversarial Natural Language Inference (ANLI) dataset, in its entirety, focuses on evaluating the performance of models in inference scenarios with heightened challenges. A notable aspect is the incorporation of adversarial samples, modifications applied to annotated training samples, posing increased difficulty for models to accurately classify text entailment relationships.

    \item \textbf{CINLID}\footnote{\url{https://www.luge.ai/\#/luge/dataDetail?id=39}}. Comprising 106K pairs of manually generated Chinese idioms, the CINLID dataset serves as a semantic reasoning dataset. This collection includes a minor proportion of concise texts, such as riddles and colloquial expressions. Each pair presents two idioms, employed as the premise and hypothesis, prompting the assessment of their semantic relationship as either approximate, unrelated, or opposing.

    \item \textbf{CMNLI}\footnote{\href{https://github.com/CLUEbenchmark/CLUE}{https://github.com/CLUEbenchmark/CLUE}}. The Chinese version of the CMNLI dataset, employed for RTE tasks, is derived by translating the English segments from both MultiNLI and XNLI. Within CLUE, this dataset has been replaced by OCNLI.

    \item \textbf{CommitmentBank} \citep{bib327}. The CommitmentBank dataset leverages naturally unfolding discourse to explore whether assertions made by speakers entail commitments to forthcoming actions. SuperGLUE has designated CommitmentBank as the assessment standard for the English RTE task, categorizing the veracity between the initial dataset pairs as “Entailment,” “Neutral,” and “Contradiction.”

    \item \textbf{MedNLI} \citep{bib326}. The MedNLI dataset comprises RTE task data within the medical domain, annotated by expert physicians. During its development, transfer learning is applied, leveraging pre-existing open-source NLI datasets. Additionally, domain knowledge from external medical data and specialized medical terminology is integrated.

    \item \textbf{MultiNLI} \citep{bib328}. The MultiNLI dataset, denoted as Multi-Genre Natural Language Inference, is crafted by incorporating English textual and spoken content from ten distinct genres for the development of the RTE task dataset. This facilitates the assessment of generalization across different genres.

    \item \textbf{OCNLI} \citep{bib330}. The OCNLI dataset stands as the pioneer among non-translated Chinese RTE task datasets, generated exclusively from native Chinese sources. With a dataset size of 56K text pairs, it has been integrated into CLUE as the evaluation benchmark for Chinese NLU tasks.
    
    \item \textbf{RTE} \citep{bib321,bib322,bib323,bib324}. The RTE dataset is dedicated to the task of recognizing textual entailment. It is an amalgamation of datasets from various annual recognizing textual entailment challenges. Prominent RTE datasets encompass RTE1, RTE2, RTE3, and RTE5, necessitating the discernment of relationships categorized as either ‘Entailment’ or “Non-Entailment.” The RTE dataset has been included in GLUE and SuperGLUE, serving as an evaluation benchmark for English NLU tasks.

    \item \textbf{SNLI} \citep{bib329}. Human annotators have labeled the premises and hypotheses in the SNLI dataset by relying on image captions, resulting in a dataset of 570K text pairs. This dataset currently holds the record as the largest in scale for RTE.
    
    \item \textbf{WANLI} \citep{bib325}. The WANLI dataset encompasses 108K pairs of English textual samples. The dataset’s construction employed a hybrid approach involving both human and model contributions. Initially, a set of challenging samples was identified on MultiNLI. Following this, GPT-3 generated new instances using a comparable approach, and after automated filtration, annotated personnel conducted the labeling and refinement process.

\end{itemize}

\subsection{Math}\label{subsecE3}

\begin{itemize}

    \item \textbf{Ape210K} \citep{bib335}. The Ape210K dataset consists of 210K mathematical problems designed for the elementary school level in China, exhibiting a considerable scale in comparison to alternative datasets. Each problem is equipped with an optimal solution, the corresponding equation for obtaining the answer, and is enriched with a variety of templates. Tackling challenges within Ape210K necessitates multifaceted capabilities, encompassing natural language comprehension, mathematical reasoning, and common knowledge.

    \item \textbf{AQUA-RAT} \citep{bib338}. The AQUA-RAT dataset consists of around 100K algebraic problems. Each problem’s solution process is methodically elucidated through a step-by-step explanation in natural language, facilitating the training of models in CoT abilities within the realm of mathematics.

    \item \textbf{ASDiv} \citep{bib333}. The ASDiv dataset serves as repositories of mathematical application problems in the English language, employed to assess the proficiency of models in solving such problems. The dataset encompasses 2,305 questions, spanning diverse text patterns and encompassing most problem types encountered in elementary school mathematics. Each sample is annotated with its respective problem type and grade level.
    
    \item \textbf{GSM8K} \citep{bib331}. The GSM8K dataset encompasses 8.5K meticulously crafted elementary school mathematical problems. These mathematical computations, deemed facile for human comprehension, entail solution procedures spanning 2 to 8 steps. The primary operations involved are consecutive calculations using addition, subtraction, multiplication, and division.

    \item \textbf{MATH} \citep{bib334}. The MATH dataset encompasses 12.5K competitive mathematical problems, presenting a high level of difficulty. Each problem is accompanied by a complete step-by-step solution, providing a means to evaluate models’ CoT abilities in solving mathematical problems or allowing the models to learn the deductive process and explanation for generating answers.

    \item \textbf{MathQA} \citep{bib337}. Prior to the development of the MathQA dataset, available datasets in the realm of mathematics are either limited in scale or lack precise operational annotations for a diverse range of questions. In response, MathQA introduces a new representation language tailored to articulate the accurate operational procedures associated with mathematical problems. The overarching aim is to augment both the performance and interpretability of models.

    \item \textbf{Math23K} \citep{bib336}. The Math23K dataset is curated explicitly for tasks related to mathematical problem-solving, encompassing 23,161 math problems that include equation templates and answer labels. All the presented problems focus on linear algebra and involve a singular variable. Derived from several online educational platforms, these questions represent authentic problem sets designed for elementary school students.

    \item \textbf{NaturalProofs} \citep{bib339}. The focus of the NaturalProofs dataset is on mathematical propositions and proof-related tasks, exploring mathematical reasoning expressed in natural language. The problem content encompasses statements and proofs of theorems, mathematical definitions, inferences based on axioms, etc., sourced from real-world materials like compilations of mathematical proofs and textbooks.
    
    \item \textbf{SVAMP} \citep{bib332}. In addressing elementary applied mathematical problems, models are observed to predominantly depend on shallow heuristics rather than engage in deep reasoning. Consequently, a more challenging and reliably assessed SVAMP dataset is introduced. This dataset adapts examples from pre-existing datasets to evaluate the models’ sensitivity to problem-solving and reasoning abilities in the realm of mathematical problems, with difficulty maintained at a level equivalent to that of a fourth-grade elementary school.

\end{itemize}

\subsection{Coreference Resolution}\label{subsecE4}

\begin{itemize}

    \item \textbf{CLUEWSC2020} \citep{bib179}. The CLUEWSC2020 dataset serves as a Chinese rendition of the coreference resolution task, demanding models to assess the co-reference relationships within sentences involving pronouns or noun phrases. The sentences in the samples are meticulously chosen from 36 contemporary literary works and annotated by linguistic experts. This dataset is integrated into CLUE as a benchmark for evaluating Chinese NLU tasks.

    \item \textbf{DPR} \citep{bib340}. The primary objective of the DPR dataset is to address the referential connections involving target pronouns within sentences. The chosen sentences are sourced extensively, covering topics such as real events, movie plots, and purely fictional content. Each sample comprises textual content, a target pronoun, two candidate antecedents, and the correct answer.

    \item \textbf{WiC} \citep{bib342}. The WiC dataset functions as a lexical disambiguation task, posing a binary classification challenge in the context of sentence pairs. Tasking the model with evaluating two text segments and a word occurring in both sentences, the objective is to discern whether the word holds identical meanings in the given contexts.

    \item \textbf{WinoGrande} \citep{bib341}. The concept behind the WinoGrande dataset is rooted in WSC, with modifications undertaken to amplify data volume and enhance bias robustness. The dataset reconfigures the pronoun disambiguation task into a fill-in-the-blank structure, wherein the target pronoun is substituted with a blank space requiring selection from two candidate nouns that match the sentence’s meaning.

    \item \textbf{WinoWhy} \citep{bib343}. The WinoWhy dataset presents a novel task of elucidating pronoun reference connections, tasking models with choosing the accurate rationale from provided options for a pronoun that refers to a particular noun. Regarded as an extension of WSC, WinoWhy comprises the original WSC dataset’s data and an additional 4,095 constructed pronoun reference reasons.
    
    \item \textbf{WSC} \citep{bib19}. The WSC dataset is utilized for tasks related to pronoun disambiguation, necessitating models to infer the referent noun of the annotated pronoun within the given context. The presented texts commonly include pairs of nearly identical sentences, distinguished by only a few words. In situations where pronoun reference.

\end{itemize}

\subsection{Sentiment Analysis}\label{subsecE5}

\begin{itemize}

    \item \textbf{EPRSTMT} \citep{bib252}. The sentiment analysis data within the EPRSTMT dataset originates from product reviews on an e-commerce platform. Samples are categorized with either positive or negative sentiments. This dataset has been included in FewCLUE.
    
    \item \textbf{IMDB} \citep{bib344}. Derived from movie reviews on the IMDB website, the IMDB dataset comprises evaluations categorized as positive or negative sentiments. Each review is evenly distributed between positive and negative samples. The authenticity and diversity of these reviews stem from real user contributions on the movie website, enhancing the datasets’ representativeness.

    \item \textbf{Sentiment140} \citep{bib345}. Derived from tweet contents on Twitter, the Sentiment140 dataset consists of tweets labeled with positive or negative sentiment. The data is curated by the creator through API calls, filtering tweets from diverse domains such as consumer products, companies, individuals, and others, based on their content.

    \item \textbf{SST-2} \citep{bib346}. The SST-2 dataset encompasses thoroughly annotated sentiment parse tree corpora. Extracted from movie reviews and parsed using the Stanford parser, the annotations are conducted at the sentence level by three annotators. The reviews are categorized into positive and negative sentiments. The dataset is included in GLUE as an evaluation benchmark for English NLU tasks.

\end{itemize}

\subsection{Semantic Matching}\label{subsecE6}

\begin{itemize}

    \item \textbf{AFQMC} \citep{bib179}. The AFQMC dataset originates from the Ant Technology Exploration Conference Developer Competition, serving as a valuable resource for Chinese semantic similarity tasks. The textual content is extracted from data within the Ant Financial platform, with a specific focus on the financial domain. This dataset has been incorporated into CLUE as an assessment benchmark for Chinese NLU tasks.

    \item \textbf{BQ} \citep{bib350}. The BQ dataset serves as a corpus for recognizing semantic equivalence in Chinese sentences within the banking domain. Consisting of 120K question pairs extracted from a year’s worth of online banking customer service logs, the dataset employs a clustering-based annotation approach to form positive and negative pairs by combining questions with similar and dissimilar intents.

    \item \textbf{BUSTM} \citep{bib252}. The BUSTM dataset focuses on the intent matching task for short dialog texts. All textual content is sourced from the spoken language text generated by OPPO’s Xiao Bu Assistant. The objective is to assess whether the intent of the content is consistent across short spoken texts.

    \item \textbf{DuQM} \citep{bib353}. The DuQM dataset serves as a Chinese robust dataset for question matching, encompassing natural questions embedded with linguistic perturbations to assess the robustness of models in this particular task. DuQM comprises three overarching categories and thirteen subcategories of linguistic perturbation types, facilitating a comprehensive evaluation of diverse model performances.

    \item \textbf{LCQMC} \citep{bib351}. The LCQMC dataset constitutes an extensive Chinese corpus designed for the matching of questions, with a distinct focus on aligning the intentions behind questions rather than achieving paraphrastic alignment of sentences. The dataset is curated by the creators through the utilization of a search engine to gather question pairs related to high-frequency words across diverse domains, followed by a meticulous filtering process for validation.
    
    \item \textbf{MRPC} \citep{bib347}. The MRPC dataset serves as a prevalent benchmark for semantic matching tasks at the sentence level. Its primary purpose is to assess the semantic similarity or synonymy between two sentences. The textual content is derived from news articles on the internet. This dataset has been incorporated into GLUE as an assessment benchmark for English NLU tasks.

    \item \textbf{PAWS} \citep{bib348}. Researchers observed a deficiency in current semantic matching datasets, specifically in the absence of sentence pairs exhibiting both extensive lexical overlap and distinct semantic similarity. This observation led to the introduction of the PAWS dataset, where all pairs of sentences display substantial lexical commonality but may not align semantically, creating a potential source of confusion for models.

    \item \textbf{PAWS-X} \citep{bib352}. In order to compensate for the lack of semantic matching datasets in various languages, the PAWS-X dataset has been introduced. English sentence pairs from the original PAWS dataset underwent manual translation into six additional languages, specifically: French, Spanish, German, Chinese, Japanese, and Korean.
    
    \item \textbf{QQP} \citep{bib177}. Similar to MRPC, the QQP dataset is designed for semantic matching tasks at the sentence level. It derives its data from the Quora Q\&A community, an online platform dedicated to interactive Q\&A. Included in GLUE, this dataset serves as an evaluation benchmark for English NLU tasks.
    
    \item \textbf{STSB} \citep{bib349}. In comparison to other semantic matching datasets, the STSB dataset exhibits several noteworthy features. Firstly, its textual content is drawn from diverse domains, encompassing realms such as news and social media. Secondly, diverging from the binary labels commonly used in semantic matching datasets (0 and 1 to denote similarity or dissimilarity), STSB employs continuous similarity scores, rated on a scale of 0 to 5, where higher scores correlate with increased similarity. Lastly, the dataset incorporates text in a total of 10 distinct languages. It has been incorporated into GLUE as an assessment benchmark for English NLU tasks.

\end{itemize}

\subsection{Text Generation}\label{subsecE7}

\begin{itemize}

    \item \textbf{CommonGen} \citep{bib354}. The CommonGen dataset serves the purpose of a delimited text generation task, linked with benchmark datasets, designed to explicitly evaluate models’ commonsense reasoning and text narrative capabilities. When presented with a group of concepts or common words, the model produces a cohesive sentence describing an everyday scenario. This task resembles exercises in exams that involve constructing sentences using provided words.

    \item \textbf{DART} \citep{bib355}. The DART dataset is utilized for the generation task of transforming structured data records into text in an open-domain context. The model is given structured data records in the form of sets of entity-relation triplets, aiming to produce a textual description that encompasses all the elements of the triplets.

    \item \textbf{E2E} \citep{bib356}. The E2E dataset serves as a training resource for natural language generation systems tailored to the restaurant domain. Inputting data pertaining to restaurants enables the generation of sentences that articulate diverse information about the restaurant. The textual content within the dataset is meticulously composed, showcasing an extensive vocabulary and syntactic variety.

    \item \textbf{WebNLG} \citep{bib357}. Much like DART, the WebNLG dataset serves the purpose of mapping data to text. Extracted from DBpedia, the dataset consists of triplets, and the corresponding text represents the linguistic expressions of these triplets. Models are tasked with generating a detailed and seamlessly coherent textual description informed by the information encapsulated in the triplets.

\end{itemize}

\subsection{Text Translation}\label{subsecE8}

\begin{itemize}

    \item \textbf{IWSLT 2017} \citep{bib359}. The International Workshop on Spoken Language Translation (IWSLT) stands as a highly impactful competition, annually unveiling pertinent translation tasks and datasets. Notably, the IWSLT 2017 dataset is recurrently utilized for both training and evaluation in translation tasks, possessing a noteworthy level of representativeness. This dataset spans languages including English, French, German, and Arabic.

    \item \textbf{NLLB} \citep{bib358}. The No Language Left Behind (NLLB) initiative stands as a text translation project, unveiling three open-sourced benchmarks for text translation evaluation: FLORES-200, NLLB-MD, and Toxicity-200. Leveraging open-source models, the project enables the provision of high-quality translations among a diverse set of over 200 languages, encompassing even those with limited linguistic resources like Luganda and Urdu. As a result, its datasets for text translation offer substantial points of reference.
    
    \item \textbf{WMT}\footnote{\url{https://www.statmt.org/wmt22/index.html}}. The WMT dataset consolidates translation competition datasets publicly disclosed by the Workshop on Statistical Machine Translation across multiple years. It incorporates diverse sources, including news commentaries and parliamentary records. The datasets within the WMT series are characterized by their extensive data scale and encompassment of a wide range of languages.

\end{itemize}

\subsection{Text Summarization}\label{subsecE9}

\begin{itemize}

    \item \textbf{AESLC} \citep{bib360}. The AESLC dataset is formed by aggregating email messages from employees at Enron Corporation. The objective is to generate concise summaries for the textual content found in the email subjects. The creators argue that, in contrast to news articles where the initial and concluding sentences typically offer a summarizing overview of the article, the email domain presents a more challenging context.

    \item \textbf{CNewSum} \citep{bib369}. The CNewSum dataset caters to the requirements of Chinese news summarization endeavors. The creators curate a dataset comprising 304K extensive documents accompanied by manually generated news summaries. Two distinctive attributes characterize CNewSum: firstly, it facilitates model comprehension and summarization at the document level; secondly, the test set incorporates comprehensive and inferential annotations on the summaries, offering researchers a means to scrutinize and identify potential performance constraints of the models.
    
    \item \textbf{CNN-DM} \citep{bib361}. Utilizing a corpus exceeding 300K news articles from CNN and The Daily Mail, the CNN-DM dataset has been curated. Each instance comprises an article paired with its corresponding summary, facilitating the training and evaluation of models for text summarization. The most recent iteration accommodates both extractive and generative summarization techniques.

    \item \textbf{Gigaword} \citep{bib401}. The Gigaword dataset is an English text summarization task dataset, comprising approximately 4M samples. The content of the dataset is derived from global news over the past two decades. The creators have pruned and filtered the data based on heuristic filters. Each final sample includes the textual content and a summary headline.

    \item \textbf{LCSTS} \citep{bib368}. The LCSTS dataset represents a compilation of Chinese short-text summaries sourced from Sina Weibo, a widely used Chinese microblogging platform. With a voluminous scale exceeding 2.4M entries, each instance originates from genuine short texts composed by users of Sina Weibo, each supplemented with a succinct summary. Social media textual content exhibits traits such as brevity, a broad spectrum of language styles, and heightened levels of noise.

    \item \textbf{MediaSum} \citep{bib372}. In contrast to other text summarization datasets, which rely on news articles, the MediaSum dataset pivots towards the realm of media interviews. The creators have curated interview transcripts sourced from NPR and CNN, utilizing summaries and topic descriptions as abstracts. The content encompasses intricate and multifaceted dialogues among multiple parties.
    
    \item \textbf{MultiNews} \citep{bib362}. Derived from articles on news websites and summaries meticulously curated by seasoned editors, the MultiNews dataset boasts a diverse array of news sources, spanning across more than 1.5K unique sites.

    \item \textbf{Newsroom} \citep{bib363}. Constructed from 1.3M articles and their associated summaries, the Newsroom dataset is a compilation from 38 leading news publishers. The selected articles span the timeframe from 1998 to 2017, and the abstracts undergo preprocessing employing a diverse array of extractive and abstractive strategies.

    \item \textbf{Opinion Abstracts} \citep{bib366}. The creators of the Opinion Abstracts dataset gather data on movie reviews and debates for text summarization tasks. On one hand, they construct a consensus comment for each movie based on expert opinions in the reviews, summarizing the content and tendencies. On the other hand, they collect points for and against from debate discussions. The central ideas of the debates are summarized in a single sentence based on the debate topic and relevant arguments.
    
    \item \textbf{SAMSum} \citep{bib364}. Within the SAMSum dataset, one can find around 16K dialogues designed to emulate real-time messaging conversations, accompanied by corresponding summaries. Proficient linguists, well-versed in English, meticulously crafted and recorded these dialogues, infusing them with varied styles and language elements, including slang, emoticons, and occasional errors, thereby presenting fresh challenges for text summarization tasks.

    \item \textbf{WikiHow} \citep{bib371}. The majority of existing datasets for text summarization originate from news articles, characterized by a distinct writing style. To address the scarcity of text in alternative genres and styles, the WikiHow dataset has been introduced. Comprising over 230K pairs of articles and summaries, the dataset is sourced from a diverse range of authors contributing to an online knowledge repository.

    \item \textbf{WikiLingua} \citep{bib367}. The WikiLingua dataset serves the purpose of assessing cross-lingual abstract summarization tasks. Approximately 770K pairs of articles and summaries are extracted by the creators from the WikiHow website, encompassing 18 diverse languages. WikiHow constitutes a repository of multi-themed procedural guides, composed by human contributors. These guides typically feature instructive visuals, succinct summaries, and in-depth details. The information from the details and summaries under the same guide is amalgamated to yield article-summary pairs.

    \item \textbf{XL-Sum} \citep{bib370}. The XL-Sum dataset encompasses 1.35M pairs of professionally annotated articles and summaries, extracted from BBC through heuristic approaches, exclusively tailored for text summarization tasks. Encompassing 45 languages, over two-thirds of which qualify as low-resource languages, the dataset is designed to facilitate research in multilingual summarization.
    
    \item \textbf{XSum} \citep{bib365}. Functioning as a dataset tailored for single-document summarization tasks, XSum draws its content from online articles curated by the British Broadcasting Corporation. Spanning the timeframe from 2010 to 2017, this dataset explores diverse domains, including family, science, and weather. Notably, in contrast to CNN-DM, both the textual content and summaries within XSum are more concise, while simultaneously showcasing a more extensive lexicon.

\end{itemize}

\subsection{Text Classification}\label{subsecE10}

\begin{itemize}

    \item \textbf{AGNEWS} \citep{bib373}. The AGNEWS dataset encompasses 497K news articles sourced from a diverse array of over 2K news outlets. It functions as an evaluative benchmark for gauging the efficacy of models in the realm of news article topic classification. The news topics are broadly classified into four categories: world, sports, business, and science \& technology.

    \item \textbf{CSLDCP} \citep{bib252}. The CSLDCP dataset constitutes a subject classification dataset for Chinese scientific literature, encompassing 67 categories that span a spectrum from social sciences to natural sciences. Examples of these categories include, but are not limited to, “horticulture” and “mechanical engineering.” The content to be classified comprises excerpts from the abstracts of Chinese literature.

    \item \textbf{IFLYTEK} \citep{bib179}. Utilized for the Chinese long-text classification task, the IFLYTEK dataset encompasses more than 17K extensive texts focusing on app application descriptions. These texts are systematically organized into 119 categories based on the functional themes of the respective apps, including but not limited to “ride-hailing,” “map navigation,” and “payment.” The extensive variety of categories poses a considerable challenge for classification. As part of CLUE, IFLYTEK serves as an evaluative benchmark for tasks related to Chinese NLU.

    \item \textbf{MARC} \citep{bib374}. The MARC dataset comprises a multilingual assemblage designed for the categorization of Amazon product reviews. Product reviews are presented in multiple languages, including English, Japanese, German, French, Chinese, and Spanish. Each specific instance includes a review, star rating, and broad product category, covering classifications like “books” and “home appliances.”

    \item \textbf{THUCNews}\footnote{\url{https://github.com/thunlp/THUCTC}}. The THUCNews dataset originates from the curated historical data of Sina Weibo’s subscription channels between 2005 and 2011. Following a rigorous screening and refinement process, a corpus of 740K pertinent documents has been meticulously restructured and classified into 14 distinct thematic categories within the Sina News classification framework. These encompass finance, lottery, real estate, stocks, home decor, education, technology, society, fashion, current affairs, sports, astrology, gaming, and entertainment.
    
    \item \textbf{TNEWS} \citep{bib179}. The TNEWS dataset serves the purpose of news headline classification, comprising Chinese news headlines sourced from the Toutiao platform by Bytedance, up until May 2018. In its entirety, TNEWS encompasses 73.3K headlines, systematically categorized into 15 sections corresponding to different news genres, namely story, culture, entertainment, sports, finance, house, car, education, technology, military, travel, world, stock, agriculture, and game. Selected as an assessment benchmark for Chinese NLU tasks, this dataset has been incorporated into CLUE.

\end{itemize}

\subsection{Text Quality Evaluation}\label{subsecE11}

\begin{itemize}

    \item \textbf{CoLA} \citep{bib375}. The CoLA dataset explores models’ proficiency in evaluating the grammatical accuracy of sentences. Comprising 10K English sentences, the dataset includes both grammatically correct and erroneous sentences. The task doesn’t mandate the model to identify specific error locations or undertake corrections; rather, it focuses on determining correctness, presenting itself as a binary classification task. This dataset has been incorporated into GLUE as a benchmark for evaluating English NLU tasks.

    \item \textbf{CSCD-IME} \citep{bib380}. The CSCD-IME dataset marks a pioneering effort to address errors induced by Chinese Pinyin input methods in the context of Chinese spelling correction. The sentences targeted for correction originate from posts on Sina Weibo. The spelling errors introduced by Pinyin input methods manifest specific distributions at both the Pinyin and semantic levels, presenting a considerable level of complexity. Notably, this dataset currently represents the most extensive collection for Chinese spelling correction tasks.
    
    \item \textbf{SIGHAN} \citep{bib376,bib377,bib378}. The SIGHAN dataset, made publicly available by scholars, serves as a resource for Chinese text correction. Presently, it encompasses three editions: SIGHAN2013, SIGHAN2014, and SIGHAN2015. The objective of the task is to evaluate the proficiency of models in Chinese spell checking, involving distinct subtasks such as detecting error positions and performing error corrections.

    \item \textbf{YACLC} \citep{bib379}. Multiple universities collaborate to create the YACLC dataset, featuring Chinese text samples. Graduate students specializing in pertinent fields are enlisted to assess the acceptability of Chinese sentences, contributing annotations for both correction and fluency. Corrections involve grammatical adjustments to align sentences with Chinese grammar standards, whereas fluency annotations focus on refining sentences for improved smoothness and authenticity, in accordance with prevalent Chinese communication norms. YACLC finds practical utility in tasks like grammar correction and text proofreading.

\end{itemize}

\subsection{Text-to-Code}\label{subsecE12}

\begin{itemize}

    \item \textbf{CSpider} \citep{bib383}. The CSpider dataset represents a Chinese variant of the Text-to-SQL dataset, translated by researchers from the original English Spider dataset. In pursuit of diversity, sentences conveying similar meanings are translated into distinct expressions in Chinese to uphold richness. Concerning specifics, the databases’ table and column names remain unaltered in English, with the exception of localized treatment for certain personal and geographical names.

    \item \textbf{DuSQL} \citep{bib382}. The DuSQL dataset functions as a Chinese dataset designed for the cross-domain Text-to-SQL task, encompassing 200 databases, 813 tables, and 23,979 question-SQL pairs. The primary focus of the task lies in practical applications, spanning a breadth of 164 domains. The questions manifest in common formats, including matching, computation, and inference, thereby closely resembling scenarios encountered in real-world applications.

    \item \textbf{MBPP} \citep{bib381}. The MBPP dataset serves as a benchmark for code generation, comprising 974 crowdsourced Python programming questions. These programming questions cover fundamental programming knowledge, standard library functionalities, and more. Each question includes a task description, a code solution, and three automated test cases.
    
    \item \textbf{Spider} \citep{bib208}. The Spider dataset encompasses Text-to-SQL dataset in English, annotated by a student cohort, totaling 10,181 questions, 5,693 SQL queries, and 200 databases. The inclusion of varied and intricate SQL queries and databases across both the training and test sets presents a formidable challenge.

\end{itemize}

\subsection{Named Entity Recognition}\label{subsecE13}

\begin{itemize}

    \item \textbf{CLUENER} \citep{bib392}. The CLUENER dataset originates from a subset of the THUCNews text classification dataset, carefully selected to facilitate detailed annotation of named entities. There exist 10 distinct entity categories, encompassing address, book, company, game, government, movie, name, organization, position, and scene.

    \item \textbf{CoNLL2003} \citep{bib387}. The CoNLL2003 dataset is introduced during the CoNLL-2003 shared task, establishing itself as a benchmark within the NER domain. Entity categories within the dataset include personal names, organizational names, geographical locations, among others. The dataset is presented in both English and German variants.

    \item \textbf{Few-NERD} \citep{bib386}. The Few-NERD dataset constitutes an extensive and finely annotated resource for NER tasks. The dataset encompasses 188,200 sentences, 491,711 entities, and 4,601,223 labels. The entities are classified into 8 broader categories and 66 more specific categories. The creators have established three benchmark tasks, involving one supervised task and two tasks with limited samples.

    \item \textbf{MSRA} \citep{bib389}. The MSRA dataset is utilized in the NER task of the Third International Chinese LanguageProcessing Bakeoff. The competition furnishes two corpora in simplified Chinese and one in traditional Chinese for both training and testing purposes. The entities cover locations, personal names, and organizational names.

    \item \textbf{OntoNotes 5.0} \citep{bib388}. The OntoNotes dataset has evolved to its final version, denoted as Version 5.0. It stands as a multigenre and multilingual corpus, meticulously annotated with syntactic, semantic, and discourse information. The dataset has been extended to serve as a NER task dataset for the CoNLL-2012 shared task, featuring three languages—English, Chinese, and Arabic—and spanning across 18 entity categories.

    \item \textbf{Resume} \citep{bib393}. The Resume dataset is compiled using several resume profiles from Sina Finance. The creators conduct manual annotations for eight distinct categories of named entities, encompassing nationality, educational history, geographic locations, individual names, organizational titles, field of study, ethnicity, and professional designations.

    \item \textbf{Taobao NER} \citep{bib390}. The Taobao NER dataset serves as openly available resources for NER in the e-commerce sector, crafted from the e-commerce data of Taobao. The entity categories are categorized into four broader types (pattren, product, brand, misc) and nine more specific types (model Type, product description, core product, brand description, core brand, location, person, literature, product specification).

    \item \textbf{Weibo NER} \citep{bib391}. The Weibo NER dataset acts as openly available resources for NER in the realm of social media, compiled from Weibo information. The entity categories include geopolitical entities, geographical locations, institutional names, and personal names, offering a more nuanced perspective compared to MSRA.
    
    \item \textbf{WUNT2017} \citep{bib385}. The central emphasis of the WUNT2017 dataset is on recognizing unconventional and hitherto unencountered entities within a new context. It assesses the capacities of models to detect and categorize emerging named entities amidst noisy textual data. The entity categories encompass corporations, creative works, groups, locations, persons, and products.

    \item \textbf{Youku NER} \citep{bib390}. The Youku NER dataset functions as an openly accessible resource for NER within the entertainment domain. Derived from titles linked to Youku videos, the entity categories are delineated into three overarching types (figure, program, misc) and nine more specific types (figure, variety show, movie, animation, TV drama, character, number, location, song).

\end{itemize}

\subsection{Relation Extraction}\label{subsecE14}

\begin{itemize}

    \item \textbf{Dialogue RE} \citep{bib394}. The Dialogue RE dataset stands as the initial manually annotated dataset for relation extraction based on dialogues. It originates from 1,788 dialogues extracted from the American sitcom “Friends.” Annotators have meticulously labeled instances of 36 relationship types within the dialogues, offering versions in both Chinese and English.

    \item \textbf{DocRED} \citep{bib396}. Functioning as a dataset for document-level RE, DocRED draws its textual content from Wikipedia and Wikidata. With a composition of 132,375 entities, 56,354 relationship facts, and 5,053 documents, the dataset challenges models to engage with multiple sentences within a document for entity recognition and relationship inference through the synthesis of document-level information. This significantly diverges from dataset focused on RE at the sentence level.

    \item \textbf{FewRel} \citep{bib397}. The FewRel dataset comes in two iterations, denoted as versions 1.0 and 2.0. The inaugural version, 1.0, represents the pioneering integration of few-shot learning with RE. The training set incorporates 64 distinct relationships, and the test set comprises 16 relationships. Version 2.0 introduces challenges in domain adaptation and the detection of categories not covered above, evaluating the models’ transferability and OOD generalization capabilities.

    \item \textbf{TACRED} \citep{bib395}. The TACRED dataset encompasses 106,264 instances designed for relation extraction tasks. These instances are drawn from news articles and online texts utilized in the annual Text Analysis Conference Knowledge Base Population (TACKBP). In total, the dataset encompasses 41 distinct relationship types among diverse entities or denotes the absence of a relationship.
    
\end{itemize}

\subsection{Multitask}\label{subsecE15}

\begin{itemize}

    \item \textbf{CSL} \citep{bib398}. The CSL dataset represents a sizable Chinese scientific literature database, incorporating titles, abstracts, keywords, and academic domain details from 396K papers. Beyond serving as a pretraining corpus, it can be configured into distinct NLP task datasets. The creators have employed it in tasks like predicting titles, generating keywords, and classifying papers.

    \item \textbf{METS-CoV} \citep{bib400}. The METS-CoV dataset provides medical annotations for COVID-19-related social media texts, facilitating tasks in NER and sentiment analysis. A collection of 10K tweets is manually annotated, encompassing four medical entity categories (disease, drug, symptom, vaccine) and three general entity categories (person, location, organization). In exploring sentiment attitudes toward specific entities, sentiment polarity labels are additionally applied to individuals, organizations, drugs, and vaccines.
    
    \item \textbf{QED} \citep{bib399}. Derived from a scalable framework that furnishes explanations in Q\&A scenarios, the QED dataset delineates explanations for answers on Natural Questions as discrete, human-understandable step combinations. Each instance is sourced from samples in Natural Questions and is accompanied by QED-style explanatory annotations. This dataset is applicable to tasks like single-sentence selection, answer selection, equality recognition, and the extraction of inference patterns.

\end{itemize}

\end{appendices}












\bibliography{sn-bibliography}

\end{document}